\documentclass[fleqn,12pt]{wlscirep}
\usepackage[utf8]{inputenc}
\usepackage[T1]{fontenc}
\usepackage{setspace}

\usepackage{array}
\newcolumntype{L}[1]{>{\raggedright\let\newline\\\arraybackslash\hspace{0pt}}m{#1}}
\newcolumntype{C}[1]{>{\centering\let\newline\\\arraybackslash\hspace{0pt}}m{#1}}
\newcolumntype{R}[1]{>{\raggedleft\let\newline\\\arraybackslash\hspace{0pt}}m{#1}}

\usepackage{caption}
\usepackage{subcaption}
\usepackage{url}

\graphicspath{{images/}}
\usepackage{bm}
\usepackage{color}
\usepackage{siunitx}
\usepackage{pifont}

\definecolor{Gray}{gray}{0.935}
\newcommand{\disclaimer}[1]{\textcolor{red}{#1~}}
\definecolor{lightblue}{rgb}{.90,.95,1}
\newcommand{\chosen}[1]{\colorbox{lightblue}{#1}}

\usepackage{multirow}
\usepackage{makecell}
\usepackage{longtable}
\usepackage{graphicx}

\newcommand{\rone}[1]{{#1}}

\newcommand{\editnew}[1]{{#1}}
\newcommand{\editsnew}[1]{{#1}}

\newcommand{\showedits}[1]{{#1}}
\newcommand{\resubmission}[1]{{#1}}

\newcommand{\xhdr}[1]{\vspace{1.6mm}\noindent{{\bf #1.}}}

\newcommand*{\framework}{\textsc{Bolt}}

\definecolor{valbest}{HTML}{d9ead3}

\definecolor{valgood}{HTML}{d0e0e3}

\definecolor{valmid}{HTML}{fce5cd}

\definecolor{valbad}{HTML}{ead1dc}

\definecolor{greenplot}{HTML}{66C2A5}
\definecolor{redplot}{HTML}{e64340}
\definecolor{blueplot}{HTML}{91BFDB}
\definecolor{purpleplot}{HTML}{DDCBE2}
\definecolor{orangeplot}{HTML}{FC8D59}

\title{A Computational Framework for Behavioral Assessment of LLM Therapists}

\author[1,*]{Yu Ying Chiu}
\author[2,*]{Ashish Sharma}
\author[2]{Inna W. Lin}
\author[2,+]{Tim Althoff}
\affil[1]{Department of Linguistics, University of Washington, Seattle, WA, USA}
\affil[2]{Paul G. Allen School of Computer Science and Engineering, University of Washington, Seattle, WA, USA}
\affil[*]{Equal Contribution}
\affil[+]{Corresponding Author: althoff@cs.washington.edu}

\begin{document}

\flushbottom
\maketitle

\section*{Abstract}

\editnew{The emergence of large language models (LLMs) like ChatGPT has increased interest in their use as therapists to address mental health challenges and the widespread lack of access to care. However, \showedits{experts have emphasized the critical need for systematic evaluation of LLM-based mental health interventions to accurately assess their capabilities and limitations.} Here, we propose \framework, \showedits{a proof-of-concept} computational framework to systematically assess the conversational behavior of LLM therapists. We quantitatively measure LLM behavior across 13 psychotherapeutic approaches \showedits{with in-context learning methods}. \showedits{Then,} we compare the behavior of LLMs against high- and low-quality human therapy. Our analysis based on Motivational Interviewing therapy reveals that \textit{LLMs often resemble behaviors more commonly exhibited in low-quality therapy rather than high-quality therapy,} such as offering a higher degree of problem-solving advice when clients share emotions. However, unlike low-quality therapy, LLMs reflect significantly more upon clients' needs and strengths. \textit{Our findings caution that LLM therapists still require further research for consistent, high-quality care.}}

\newpage

\disclaimer{Important: This paper does \textit{NOT advocate} for the use of large language models  (LLMs) in therapeutic settings, \textit{NOR establish their readiness.} Instead, our objective is to enable \textit{systematic characterization and assessment} of the behavior of current LLMs when they are used for therapy to inform this assessment. Readiness assessments further need to consider patient safety, which is outside the scope of this study.}

Ever since Eliza, a mock psychotherapist, was introduced in the 1960s\cite{weizenbaum1966eliza}, the prospect of therapy chatbots that can offer mental health support has intrigued clinicians, researchers, as well as the general public. The recent advent of ChatGPT and the rapid progress in large language models (LLMs)\cite{brown2020language,OpenAI2023GPT4TR,touvron2023llama,bubeck2023sparks} have amplified the excitement. \showedits{This growing excitement is driven by a stark reality: Hundreds of millions of people do not or cannot receive mental health care for various reasons, including a shortage of professionals, lack of insurance coverage, high costs, and stigma surrounding therapy\cite{hosny2019artificial,world2022mental,olfson2016building,sickel2014mental}. As a result, individuals and researchers have started using LLMs as a potential solution to bridge this critical access gap\cite{hamilton-linkedin,fastcompany-therapy,stade2023artificial,mental-health-gpts,sharma2023naturemi,sharma-etal-2023-cognitive,Sharma2020-dx,Sharma2023FacilitatingSM,lin2024imbue}. However, mental health experts have raised concerns and caution about using LLMs in therapeutic contexts without appropriate evaluation\cite{stade2023artificial,choudhury2023llms,li2020developing,tate2023chatgpt,van2024adapted}.}

\showedits{LLMs have the potential to augment and support both therapists and clients. 
Emerging LLM techniques can train therapists\cite{Tanana2019-bj}, assess treatment adherence\cite{flemotomos2022automated}, and streamlining administrative tasks like note-taking \cite{krishna2021generating}. Traditionally, such processes can be slow, expensive, labor-intensive, and require extensive expert instruction\cite{Tanana2019-bj,moyers2005assessing}. By training LLMs on thousands of therapy conversations, LLMs can simulate diverse virtual clients, allowing therapists to practice skills in a safe environment. 
Moreover, LLMs may provide feedback on key therapy skills, potentially enhancing the efficiency and effectiveness of therapy. }


\showedits{Simultaneously, LLM-based self-guided interventions for mental health are increasingly becoming available, offering innovative tools and techniques for individuals seeking support. Such LLM-augmented ``Do-It-Yourself'' tools can help people learn therapeutic skills\cite{Sharma2023FacilitatingSM} and improve mental health ``in-the-moment'' of crisis. Furthermore, specialized therapy chatbots like Woebot (\href{woebothealth.com}{woebothealth.com}), Wysa (\href{wysa.com}{wysa.com}), and Sonia (\href{soniahealth.com}{soniahealth.com}) are being developed to provide personalized support at scale. Interestingly, many individuals are also using general purpose LLMs like ChatGPT for therapeutic purposes, even when not explicitly designed for it, sharing anecdotal LLM examples that resemble therapist behaviors\cite{openai-employee,reddit-examples,reddit-examples-2,twitter-examples}.}

\showedits{However, mental health experts have cautioned about the potential risks of using LLMs in therapeutic contexts. Both general-purpose and mental health-specific LLMs have been scrutinized for generating biased, harmful, or factually incorrect information\cite{blodgett2020language,lin-etal-2022-gendered,ganguli2022red}, potentially undermining their effectiveness in supporting both therapists and clients. Moreover, exposing vulnerable individuals to untested interventions, such as under-development chatbots, raises potential risks and ethical concerns. Therefore, evaluating and addressing the potential risks and ethical concerns is critical to ensure the safe and effective integration of these LLMs into existing and new mental health interventions. To achieve this, experts have called for developing mechanisms for continuous monitoring and evaluation of LLM systems used within a high-stakes domain like mental health to ensure responsible usage and adherence to established interventions\cite{stade2023artificial,choudhury2023llms,li2020developing,tate2023chatgpt,van2024adapted}.}

\showedits{Currently, there is a significant lack of research and evidence demonstrating the effectiveness, safety, and feasibility of LLMs in providing therapeutic interventions. Established interventions like Cognitive Behavioral Therapy (CBT) and Motivational Interviewing (MI), even in computer-assisted formats, are widely recognized as evidence-based practices for a variety of mental health conditions due to extensive research and evaluation\cite{beck1976cognitive,miller2012motivational,wright2019computer}. While high-quality LLM-delivered CBT or MI \textit{could} potentially be effective, we currently lack methods to evaluate this technology. There is a risk of LLMs exhibiting clinically contraindicated behaviors (e.g., offering solutions before reflecting on client's feelings and experiences \cite{cochran2015heart}). Without systematic large-scale evaluation frameworks, we cannot determine the presence or frequency of such undesirable behaviors. Understanding the behavior of LLMs is particularly important in high-stakes settings such as mental health where undesirable behavior could have severe consequences on vulnerable clients. Moreover, identifying the desirable and undesirable behaviors can help identify the potential benefits and risks, which can inform adoption, iterative development, and continuous monitoring.}


\showedits{Also, LLMs can intentionally or unintentionally generate harmful advice or inaccurate information\cite{ganguli2022red,augenstein2023factuality}. Due to the vulnerability of individuals seeking mental health support, such LLM responses, even if detrimental, may be perceived as helpful, leading to ``pathological helpfulness''\cite{oakley2011pathological} by LLMs that could exacerbate existing mental health conditions or even trigger new ones. If an LLM offers weight loss advice to someone with an eating disorder, it may worsen their condition. They might see the LLM's response as support for their harmful behaviors which could increase distress and put their health at risk. However, similar weight loss advice given to someone without an eating disorder might not have the same negative impact. }


Such contexts highlight the critical need to evaluate whether LLM-based interventions are feasible, effective, and safe in mental healthcare, how they compare to established interventions, and whether they would ultimately be helpful to patients and clinicians. \resubmission{However, this evaluation necessitates making advances across various fields including artificial intelligence, natural language processing, human-computer interaction, psychology and mental health, posing significant ethical and technical challenges. Bridging these diverse disciplines presents a formidable challenge due to inherent differences in their research methodologies and a limited understanding of what's feasible at their intersection.} The rapid pace at which LLMs are being developed necessitates exploring \textit{automated} and \textit{general} evaluation methods to keep up with advancements. However, \resubmission{in the mental health field}, evaluation frameworks like the Cognitive Therapy Rating Scale (CTRS)\cite{vallis1986cognitive} have been designed for manual assessment of human therapists and are difficult to adapt to automated, large-scale LLM evaluation, due to their complexity. Therefore, efforts to translate these guidelines into actionable metrics for automatic, large-scale evaluations have been limited, with existing work focusing only on specific therapy types\cite{flemotomos2022automated}. Also, the evaluation necessitates a large-scale assessment of different types of open-ended texts that LLMs may generate, in response to a diverse range of clients, for different types of support, and at different stages of support. \resubmission{In the fields of artificial intelligence and natural language processing, such a behavioral assessment is significantly challenging due to its complex and multi-faceted nature.}


In this paper, we conduct a \showedits{proof-of-concept study proposing a framework for evaluating LLMs in mental health support.} We develop \framework, a novel, general computational framework to systematically study the \textbf{B}ehavior \textbf{O}f \textbf{L}LMs in the context of their application in \textbf{T}herapy. Given an LLM (e.g., GPT-4) that is being used as a therapist \resubmission{by an end user (e.g., by asking a chatbot to behave like a therapist}; see Methods section), we aim to quantitatively assess the ways in which it responds when supporting clients with mental health issues (see Approach section). Subsequently, we aim to compare the behavior of \textit{LLM therapists} against that of high- and low-quality human therapists, and study whether their behavior can be modulated into more desirable behaviors (see Approach section).

\begin{figure*}
\includegraphics[width=\textwidth]{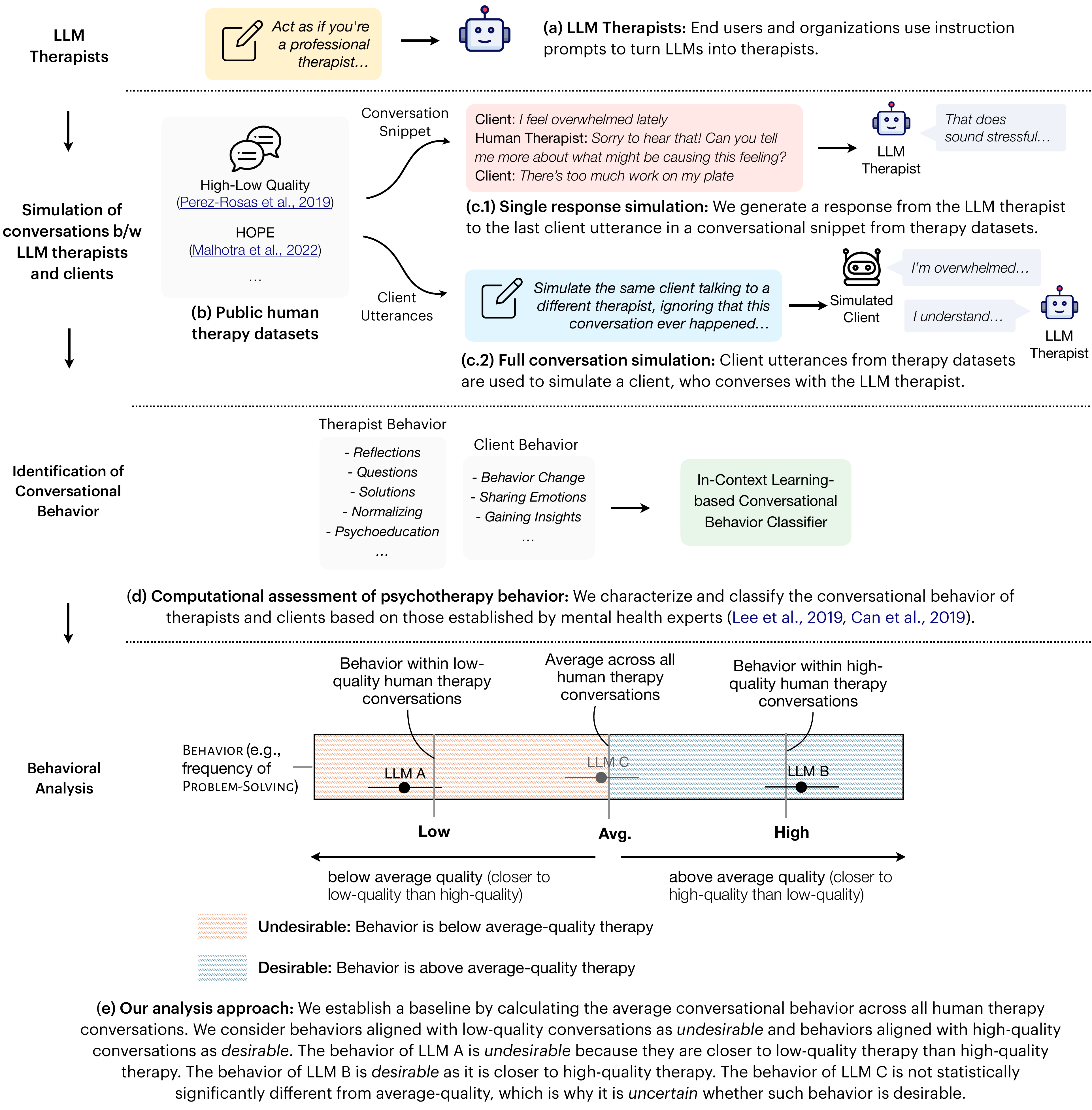}
\centering
\caption{\showedits{Overview of \framework, a computational framework that enables systematic assessment of the behavior of LLM therapists and compares them to high- and low-quality human therapy.}}
\label{fig:intro-figure}
\end{figure*}


To analyze the behavior of LLM therapists, we need conversations between LLM therapists and clients. \showedits{\framework~simulates conversations between simulated clients and LLMs, leveraging client-human therapist conversations from existing public datasets of therapy conversations to ensure high fidelity (see Approach section). Next, to assess the conversational behavior of therapists and clients in these conversations, \framework~leverages a prompting-based classifier that uses psychotherapy-based definitions and in-context examples to identify the psychotherapy techniques (e.g, \textsc{Problem-Solving}) underlying all utterances (see Approach section). }


\showedits{Furthermore, \framework~establishes behaviors that are representative of high-quality therapy and low-quality therapy sessions based on established principles of psychotherapy. Based on this, \framework~analyzes these conversations using the prompting-based behavior classifier to identify specific therapist and client behaviors associated with high-quality and low-quality therapy (see Approach section).} The analysis is based on the \textit{frequency} of their behavior, the \textit{temporal order} in which the behavior is expressed in conversations, and their \textit{adaptability} to different behaviors from clients. \showedits{Finally, we study whether variations in prompts may help in modulating the behavior of LLM therapists toward high-quality therapy.}


\showedits{To illustrate the utility of \framework, we develop a case study applying it to Motivational Interviewing (MI), a popular, well-established therapeutic approach\cite{miller2012motivational} and} evaluating four popular, state-of-the-art LLMs -- GPT-4\cite{OpenAI2023GPT4TR}, GPT-3.5-turbo\cite{brown2020language}, Llama2-70b, and Llama2-13b\cite{touvron2023llama}. As different therapy frameworks (e.g., MI vs. Problem-Solving Therapy\cite{haley1992problem}) might have varying nuances regarding what is considered high-quality therapy, an assessment within a specific therapeutic approach is required to interpret the findings in a principled manner and at high granularity. \showedits{However, note that \framework~is intentionally designed to be generalizable, with behavioral techniques that are broadly applicable in psychotherapy. Such a design \resubmission{builds upon what is feasible in AI and NLP, has the potential to adapt to future LLM advancements, and} could potentially generalize to other psychotherapy contexts and varied therapeutic frameworks. We demonstrate this through a secondary experiment with other therapeutic approaches (see Methods section). }

\showedits{While anecdotal examples have demonstrated striking similarity to plausible therapist statements, our systematic approach reveals that LLM therapists often more closely resemble the behavior of human therapists in low-quality rather than high-quality sessions.}

\resubmission{With \framework, we develop a general conversation behavior assessment that builds upon what is currently feasible at the intersection of AI and mental health fields. This offers a meaningful first step for conducting such assessments and provides a vital foundation for developing more comprehensive and higher-fidelity assessments. Our evaluation is based on high-quality therapy behaviors and does not incorporate patient outcomes in its assessment, which are challenging to obtain and difficult to simulate. Also, our behavioral assessment focuses on ``what'' the LLM therapists do (e.g., offer reflections) and ``when'' they do that (e.g., early on in the sessions or when client behave in a certain manner). However, we do not assess ``how'' skillfully they execute these behaviors (e.g., the skillfullness of their reflections). Nevertheless, we our proposed methodology identified crucial differences between LLM and human therapists already in terms of ``what'' behaviors are expressed, offering a critical first step for evaluation and improvement in the AI for mental health field.} 

\section*{Approach}
\label{sec:results}

\subsection*{Simulating Conversations between LLM Therapists and Clients}

\showedits{To characterize and analyze the behavior of LLM therapists, we need conversation data between LLM therapists and clients. \framework~uses public therapy datasets to simulate these conversations (see Methods section). We develop two simulation strategies representing common LLM therapist use cases --}


\xhdr{(1) LLM Single Response Simulation}
\showedits{We use LLMs to write a single response within an ongoing conversation session between a human therapist and a client (see Methods section). The LLM therapist here responds to a prior, real-world, authentic interaction between a human client and a human therapist. This is helpful in not biasing the LLM based on simulated clients or therapists.}



\xhdr{(2) LLM Full Conversation Simulation} 
\showedits{In addition to single-response generation, we also simulate a {full, multi-turn} conversation session between LLMs and clients through a novel method based on highly consistent simulated clients (see Methods section). Such a simulation is helpful in assessing the behavior of LLM therapists across multiple turns of support to the same simulated client. Importantly, we evaluate the consistency of these simulated conversations and find that the simulated conversations have a {near-perfect} mean consistency of 2.95 (out of 3; std = 0.13; see Methods section). }

\subsection*{Behavioral Techniques in Psychotherapy}
\showedits{To assess the conversational behavior exhibited by therapists and clients during therapy sessions, \framework~characterizes the psychotherapeutic approaches underlying utterances using clinically relevant dialogue acts\cite{stolcke-etal-2000-dialogue} established by psychotherapy experts\cite{lee-etal-2019-identifying, cao2019observing}.}


\showedits{For characterizing the conversational behavior of therapists, we focus on 13 major psychotherapeutic approaches organized into five semantically related categories: (a) \textsc{Reflections} on -- (1) \textsc{Needs}, (2) \textsc{Emotions}, (3) \textsc{Values}, (4) \textsc{Consequences}, (5) \textsc{Conflicts}, and (6) \textsc{Strengths}; (b) \textsc{Questions} on -- (7) \textsc{Experiences}, (8) \textsc{Perspectives}, and (9) \textsc{Emotions}; (c) \textsc{Solutions} -- (10) \textsc{Problem-Solving}, and (11) \textsc{Planning}; (d) (12) \textsc{Normalizing}; (e) (13) \textsc{Psychoeducation}. }

\showedits{Moreover, for characterizing the conversational behavior of clients, we use six types of client expressions organized into three semantically related categories: (a) \textsc{Behavior Change} -- (1) \textsc{Changing Unhealthy Behavior} and (2) \textsc{Sustaining Unhealthy Behavior}; (b) \textsc{Self-Disclosure of Affect or Experiences} -- (3) \textsc{Sharing Positive Emotions}, (4) \textsc{Sharing Negative Emotions}, and (5) \textsc{Sharing Experiences}; (c) (6) \textsc{Gaining Insights}.}

\showedits{Next, \framework~automatically identifies the conversational behavior of LLM therapists using GPT-based\cite{OpenAI2023GPT4TR} methods (see Methods section). We find that the prompting-based methods that use definitions and examples achieve the best performance. For therapist behavior, we achieve $57.7\%$ macro-F1 (std $=2.5\%$) which is $43.6\%$ greater than a uniform random baseline, which attained $14.1\%$ macro-F1 (std $=1.3\%$). For client behavior, we achieve $76.7\%$ macro-F1 (std $=6.3\%$) which is $50.8\%$ greater than a uniform random baseline which attained $25.9\%$ macro-F1 (std $=6.2\%$).  }

\subsection*{Associating Conversational Behavior with High-Quality and Low-Quality Therapy}
\showedits{Now that we can identify behaviors in conversations, we need to understand which behaviors are high-quality or low-quality. This differentiation is crucial, as it helps us determine desirable and undesirable behaviors. Here, we leverage the High-Low Quality dataset from Perez-Rosas et al.\cite{perez-rosas-etal-2019-makes} containing 259 Motivational Interviewing (MI) conversations annotated as either high-quality or low-quality. We analyze these conversations using our behavior classifier to identify specific therapist and client behaviors associated with high-quality and low-quality MI. Our assessment focuses on the \textit{frequency} of behavior, the \textit{temporal order} in which it's expressed, and its \textit{adaptability} to different client behaviors.}



\showedits{What differences between high- and low-quality human-delivered MI therapy does this dataset reveal? We find that therapists from {high-quality human therapy} conversations more frequently express \textsc{Reflections on Needs}, \textsc{Reflections on Emotions}, \textsc{Reflection on Values}, \textsc{Reflections on Conflicts}, \textsc{Reflections on Strengths}, \textsc{Questions on Experiences}, and  \textsc{Questions on Emotions}by 5.6\%, 8.3\%, 3.0\%, 3.3\%, 6.6\%, 9.9\%, and 3.2\%, as compared to {low-quality human therapy}. Moreover, therapists from high-quality human therapy conversations less frequently express \textsc{Reflections on Consequences}, \textsc{Problem Solving}, \textsc{Normalizing}, and \textsc{Psychoeducation} by 7.7\%, 9.2\%, 5.4\%, and 4.6\% than low-quality human therapy. }


\showedits{Moreover, previous psychotherapy research has demonstrated that focusing on solving client problems \textit{before} empathizing may negatively affect relationship forming between therapists and clients \cite{cochran2015heart}. Here, we find that therapists from high-quality human therapy conversations respond with \textsc{Reflections On Needs} 8.8 turns earlier, \textsc{Reflections on Emotions} 5.3 turns earlier, \textsc{Reflections on Consequences} 3.8 turns earlier, \textsc{Reflections on Conflicts} 8.4 turns earlier, and \textsc{Reflections on Strengths} 8.2 turns earlier than low-quality human therapy.}


\showedits{Also, it is important to appropriately respond and adapt to the client's behavior \cite{bordin1979generalizability,horvath1989development,miller2012motivational}. When analyzing adaptability to client behaviors, if clients express \textsc{Gained Insights}, therapists from high-quality human therapy respond with 15.0\% lower \textsc{Psychoeducation} compared to low-quality human therapy. If clients express \textsc{Changing Unhealthy Behavior}, therapists from high-quality human therapy respond with 9.6\% higher \textsc{Reflections on Strengths} than low-quality human therapy, likely to reinforce the positive change and promote self-efficacy in clients. Also, if clients \textsc{Share Negative Emotions} therapists from high-quality human therapy respond with 15.9\% higher \textsc{Reflections on Emotions} than low-quality human therapy, potentially to foster empathy.}

\showedits{These differences between high-quality and low-quality conversations help us analyze LLMs. We assess how LLM therapists compare against human therapists, determining whether their conversational behavior is closer to high-quality or low-quality human therapists. To establish a baseline, we calculate the average conversational behavior across all human therapy conversations. We consider behaviors aligned with low-quality conversations as below average quality and behaviors aligned with high-quality conversations as above average quality. We then assess the LLM therapists' behavior in relation to this average. This allows us to assess whether the behavior of LLM therapists falls below (\textit{undesirable}) or above (\textit{desirable}) this average, indicating closer alignment with low-quality or high-quality human therapists, respectively. Figure~\ref{fig:intro-figure}e provides an overview with example LLM behaviors. }

\showedits{Our null hypothesis is that the behavior of an LLM therapist is similar to that of average-quality human therapy. To test this, we conduct a two-sided Student’s t-test. The results reported below are statistically significant at $p = \frac{0.05}{m}$, following Bonferroni correction (m: number of intents tested). We report descriptive statistics in Supplementary Tables~\ref{tab:dist_new_hl}-\ref{tab:adapt_new_hl_p2}.}







\section*{Results}

\subsection*{Behavior Frequency -- How frequently do LLM Therapists exhibit specific conversational behaviors relative to human therapists?} \label{sec:frequency-analysis}



\begin{figure*}
\includegraphics[width=0.9\textwidth]{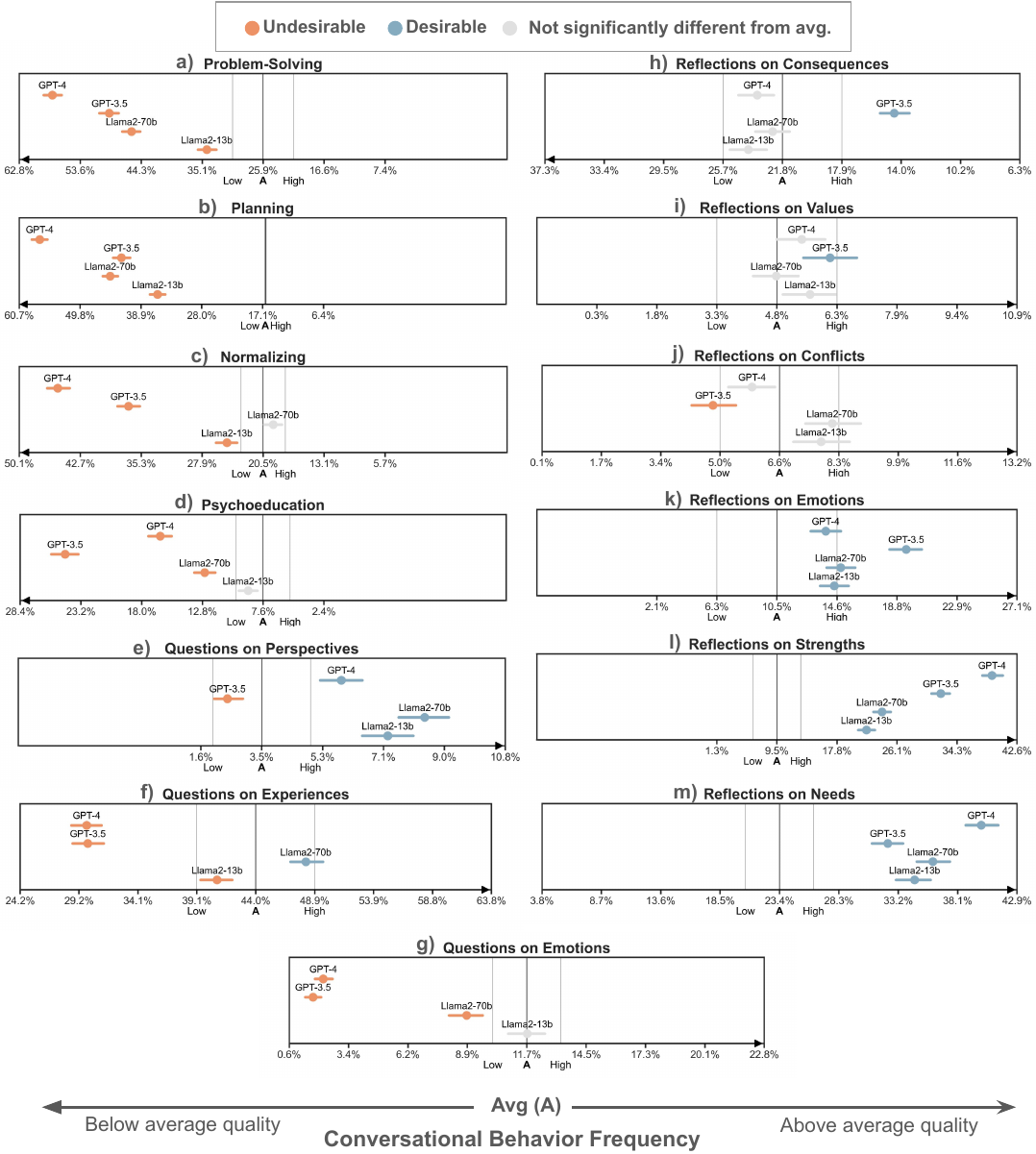}
\centering
\caption{\showedits{Difference in the frequency of conversational behaviors exhibited by LLM therapists (GPT-4, GPT-3.5-turbo, Llama2-70b, Llama2-13b), relative to average-, low-, and high-quality human therapy. \textbf{A}: average-quality, Low: low-quality, and High: high-quality therapy. The direction of the arrow on the x-axis indicates the direction in which the frequency is increasing (we flip the axis if low-quality is more frequent than high-quality, such that low-quality is visualized below the average quality marker). Values colored in \textcolor{blueplot}{blue} indicate desirable behaviors (significantly closer to high-quality than low-quality) whereas values colored in \textcolor{orangeplot}{orange} indicate undesirable behaviors (significantly closer to low-quality than high-quality). Values in \textcolor{gray}{gray} are not statistically significantly different from average-quality at p = 0.05 at $p = \frac{0.05}{m}$ using Two-sided Student’s t-test, following Bonferroni correction (m: number of intents tested = 13). Error bars indicate 95\% bootstrapped confidence intervals. A key insight we find is that LLMs respond with significantly higher \textsc{Problem-Solving} (subfigure (a)), similar to low-quality human therapy. On the other hand, LLMs respond with significantly higher \textsc{Reflections on Strengths} (subfigure (l)), similar to high-quality therapy, but with a frequency that significantly exceeds high-quality therapy.}}
\label{fig:distribution-avg-new}
\end{figure*}

\xhdr{Summary} \showedits{Based on behavior frequency, LLM therapists providing MI support often resemble therapists in \textit{low-quality} MI sessions (Figure~\ref{fig:distribution-avg-new}; Supplementary Table~\ref{tab:dist_new_hl}). Both tend to offer more \textsc{Solutions} compared to average-quality human therapy. However, unlike low-quality human therapy, LLMs tend to emphasize \textsc{Reflections}.}

\showedits{\xhdr{LLM therapists respond with significantly higher degrees of \textsc{Solutions}, \textit{similar to} low-quality human therapy conversations} Specifically, we find that GPT-4, GPT-3.5, Llama2-70b, and Llama2-13b are more frequent in exhibiting \textsc{Problem-Solving} by 31.9\%, 23.1\%, 19.9\%, and 8.5\% respectively than average human therapists (Figure~\ref{fig:distribution-avg-new}a). This more closely resembles the higher degrees of \textsc{Problem-Solving} behavior observed in low-quality therapy, suggesting that such behavior may be undesirable. In fact, GPT-4, GPT-3.5, Llama2-70b, and Llama
2-13b even exceeds the frequency of low-quality therapy by 27.3\%, 18.5\%, 15.3\%, and 3.9\% respectively, raising concerns about their potentially overly aggressive approach to problem-solving. To corroborate this finding, we also evaluate the word usage of LLM therapists using LIWC -- the gold-standard lexicon-based method for analyzing psychologically relevant constructs in text  \cite{boyd2022development} and find a similar trend significantly more Behavioral Activation \cite{kanter2010behavioral} words related to identifying goals, problem-solving, and planning (Supplementary Table~\ref{tab:liwc_new_hl}).}

\showedits{A likely, plausible reason for this problem-solving prominent behavior could be the Reinforcement Learning with Human Feedback (RLHF) alignment that these LLMs incorporate. A key focus of RLHF alignment in LLMs is helping users solve their tasks and provide advice \cite{bai2022training,ouyang2022training}. This so-called alignment may lead to LLMs overly focusing on offering solutions to client's problems even when LLMs are used as therapists.}

\showedits{\xhdr{LLM therapists respond with {higher} \textsc{Normalizing} and \textsc{Psychoeducation}, \textit{similar to} low-quality human therapy} Also, we we find that GPT-4, GPT-3.5, and Llama2-13b are more frequent in exhibiting \textsc{Normalizing} by 24.9\%, 13.5\%, and 4.4\% respectively than average human therapists (Figure~\ref{fig:distribution-avg-new}c). GPT-4, GPT-3.5, and Llama2-70b are also more frequent in exhibiting \textsc{Psychoeducation} by 8.8\%, 20.1\%, and 5.0\% respectively than average human therapists (Figure~\ref{fig:distribution-avg-new}d). This more closely resembles the higher degrees of \textsc{Normalizing} and \textsc{Psychoeducation} observed in low-quality therapy, suggesting that such behavior may be undesirable. Such behavior aligns with previous studies on RLHF data, which have demonstrated that ``empathetic'' and ``informative'' LLM generations are considered more helpful and more preferred based on human feedback \cite{sharma2023towards}.}

\showedits{\xhdr{Most LLM therapists respond with significantly fewer \textsc{Questions on Experiences} and \textsc{Questions on Emotions}, \textit{similar to} low-quality human therapy} Among LLM therapists, we observe that GPT-4, GPT-3.5, and Llama2-13b are less frequent in exhibiting \textsc{Questions on Experiences} by 14.2\%, 12.9\%, and 3.2\% than average human therapists (Figure~\ref{fig:distribution-avg-new}f). Furthermore,  GPT-4, GPT-3.5, and Llama2-70b are less frequent in exhibiting \textsc{Questions on Emotions} by 9.5\%, 9.8\%, and 2.8\% than average human therapists (Figure~\ref{fig:distribution-avg-new}g). This more closely resembles the lower degrees of \textsc{Questions on Experiences} and \textsc{Questions on Emotions} observed in low-quality therapy, suggesting that such behavior may be undesirable.}




\showedits{\xhdr{LLM therapists respond with more \textsc{Reflections}, \textit{unlike} low-quality human therapy} Moreover, all LLM therapists respond with more \textsc{Reflections} than average-quality human therapy conversations in general (Figures~\ref{fig:distribution-avg-new}h-m). We find that GPT-4, GPT-3.5, Llama2-70b, and Llama2-13b are more frequent in exhibiting \textsc{Reflections on Emotions} by 3.4\%, 7.9\%, 4.4\%, and 4.0\% respectively than average-quality human therapy. This resembles high-quality human therapy suggesting that such a behavior may be desirable. All LLM therapists are also above average-quality therapy in exhibiting \textsc{Reflections on Strengths} and \textsc{Reflections on Needs}. However, their frequency significantly exceeds high-quality therapy as well, raising concerns about the desirability of such behavior.}





\subsection*{Temporal Order of Behavior -- How do LLM therapists structure their conversation compared to human therapists?} 

\label{sec:first-and-relative-occur-analysis}

\begin{figure*}
\includegraphics[width=0.9\textwidth]{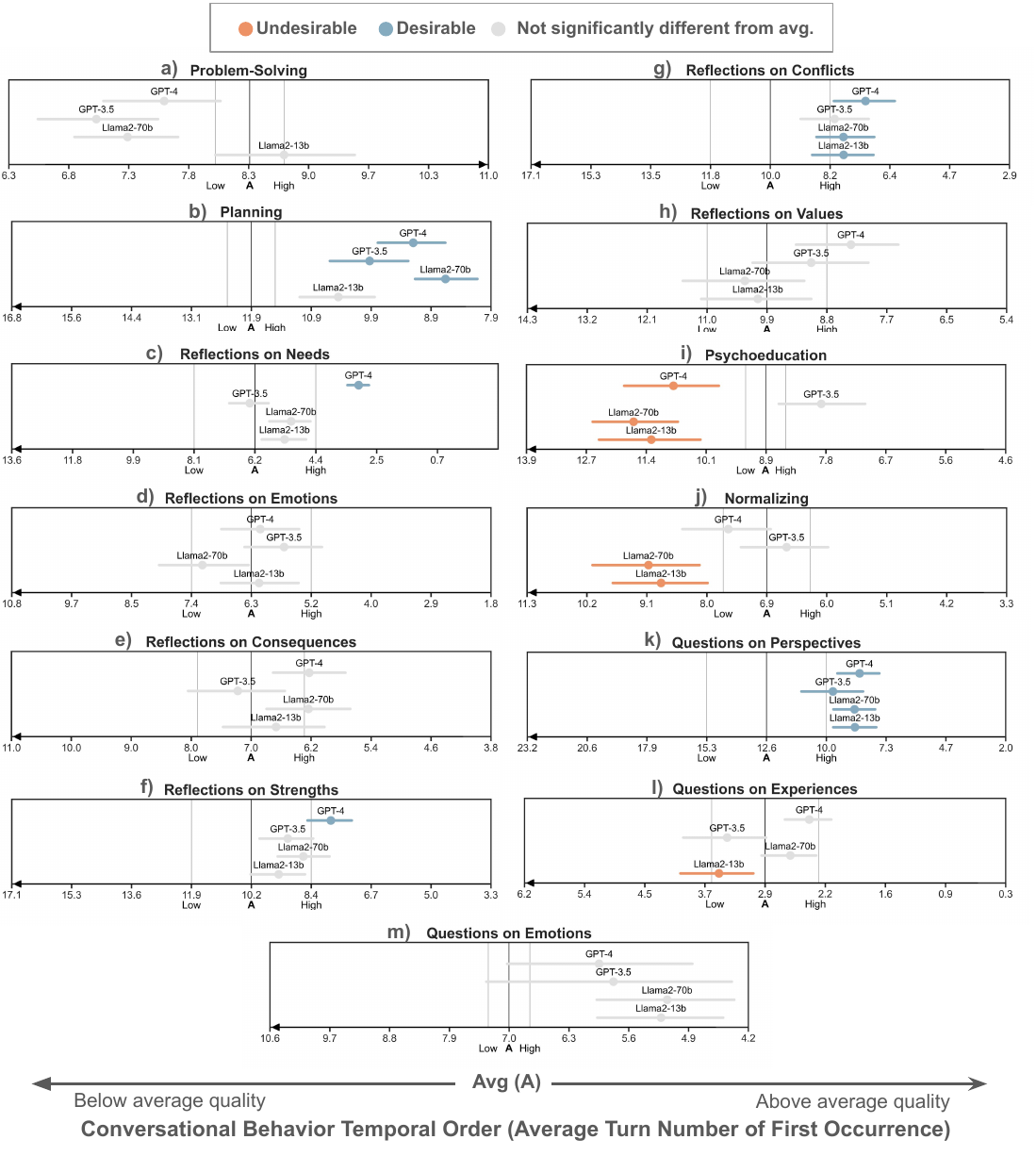}
\centering
\caption{\showedits{Difference in the \textit{temporal order} of conversational behaviors, operationalized as the turn numbers in which behaviors are first exhibited in a conversation by LLM therapists (GPT-4, GPT-3.5-turbo, Llama2-70b, Llama2-13b), relative to average-, low-, and high-quality human therapy. \textbf{A}: average-quality, Low: low-quality, and High: high-quality therapy. The direction of the arrow on the x-axis indicates the direction in which the order is increasing (we flip the axis if low-quality is exhibited later than high-quality, such that low-quality is visualized below the average quality marker). Values colored in \textcolor{blueplot}{blue} indicate desirable behaviors (significantly closer to high-quality than low-quality) whereas values colored in \textcolor{orangeplot}{orange} indicate undesirable behaviors (significantly closer to low-quality than high-quality). Values in \textcolor{gray}{gray} are not statistically significantly different from average-quality at $p = \frac{0.05}{m}$ using Two-sided Student’s t-test, following Bonferroni correction (m: number of intents tested = 13). Error bars indicate 95\% bootstrapped confidence intervals. Most LLM therapists start providing \textsc{Planning} (subfigure (b)) earlier in the conversations but provide \textsc{Normalizing} (subfigure (j)) later against common recommendations \cite{cochran2015heart}. }}



\label{fig:temporal-order-avg-new}
\end{figure*}



\showedits{\xhdr{Summary} \editsnew{LLM therapists providing MI support exhibit a temporal ordering of behaviors that often aligns with low-quality human MI therapy (Figure~\ref{fig:temporal-order-avg-new}; Supplementary Table~\ref{tab:order_new_hl}).} For instance, both tend to offer \textsc{Solutions} before \textsc{Normalizing}. }



\showedits{\xhdr{Most LLM therapists do not prioritize \textsc{Normalizing} before \textsc{Solutions}, \textit{similar to} low-quality human therapy} 
We compare the order between \textsc{Solutions} and \textsc{Normalizing} exhibited by LLM therapists (Figure~\ref{fig:temporal-order-avg-new}a-b, Figure~\ref{fig:temporal-order-avg-new}j). We find that GPT-4 and Llama2-70b provide \textsc{Problem-Solving} 0.5 and 0.9 turns earlier and GPT-4, GPT-3.5, and Llama2-70b provide \textsc{Planning} 2.6, 1.8, and 3.2 turns earlier than average-quality human therapy (Figure~\ref{fig:temporal-order-avg-new}a-b). In addition, GPT-4, Llama2-70b, and Llama2-13b exhibit \textsc{Normalizing} 1.0, 2.1, and 2.3 turns later than average-quality human therapy. This more closely resembles the earlier expression of \textsc{Solutions} and later expressions of \textsc{Normalizing} in low-quality therapy than high-quality therapy. \editnew{Importantly, previous psychotherapy studies have shown that the earlier focus on providing \textsc{Solutions} rather than \textsc{Normalizing} may negatively impact the therapeutic relationship both in MI as well as general psychotherapy \cite{cochran2015heart,hill2009helping}. This suggests that LLM therapists often exhibit an {undesirable} temporal ordering of \textsc{Solutions} and \textsc{Normalizing}.}}

\showedits{\xhdr{Most LLM therapists provide \textsc{Psychoeducation} later, \textit{similar to} low-quality human therapy}
\textsc{Psychoeducation} acts as an effective supplement approach to help clients understand therapy, and be better equipped with coping skills \cite{bhattacharjee2011psychoeducation}. We find that GPT-4, Llama2-70b, and Llama2-13b first respond with \textsc{Psychoeducation} 2.2, 2.6, and 3.0 turns later than therapists from average-quality human therapy respectively (Figure~\ref{fig:temporal-order-avg-new}i). Interestingly, we do not find significant differences for GPT-3.5. This is more similar to the later expressions of \textsc{Psychoeducation} in low-quality therapy than high-quality therapy suggesting that such a behavior may be {undesirable}.}

\showedits{Psychotherapists commonly begin sessions by imparting psychoeducation to set the objectives and to outline client expectations (e.g., ``\textit{In this session, we will focus on cognitive restructuring, which is a key technique to help with our negative thinking patterns...}''). However, our qualitative analysis suggests that LLM therapists frequently do not start sessions with such psychoeducation.}

\subsection*{Adaptability to Client Behaviors -- In what ways do LLM Therapists adapt to different client behaviors compared to human therapists?}
\label{sec:client-therapist-vs-analysis}

\begin{figure*}
\vspace{-20pt}
\includegraphics[width=\textwidth]{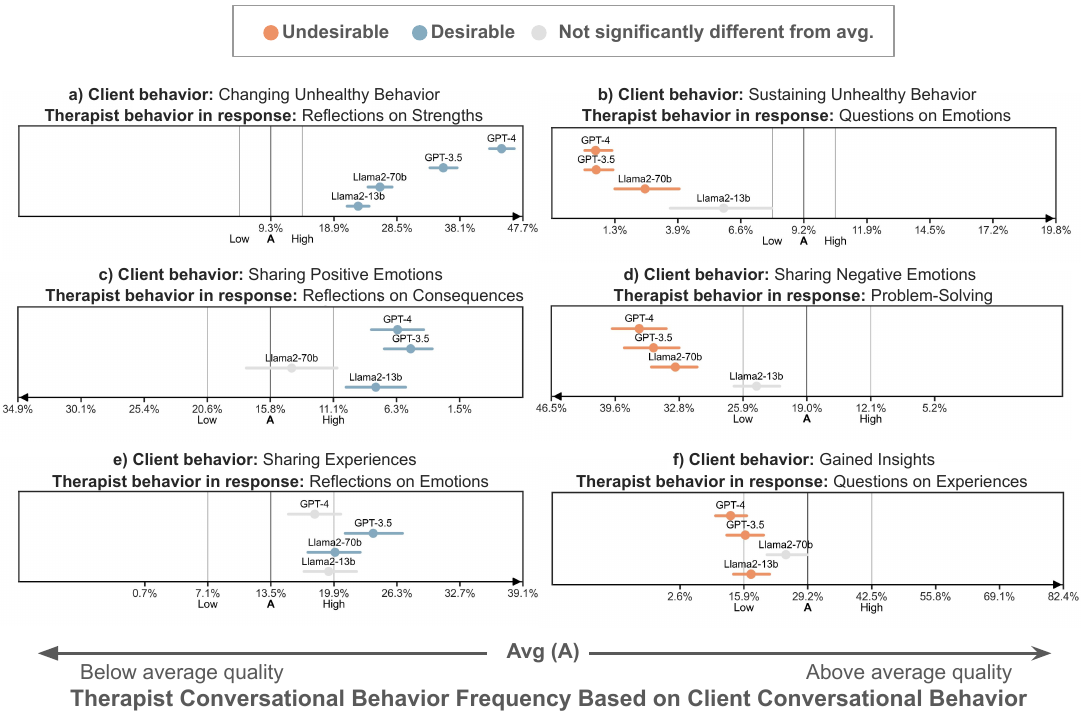}
\centering
\caption{\showedits{Difference between the frequency of conversational behaviors observed in LLM therapists (GPT-4, GPT-3.5-turbo, Llama2-70b, Llama2-13b) or low-quality human therapy \textit{in response to specific client behaviors} (\textit{Adaptability}), relative to average-, low-, and high-quality human therapy. \textbf{A}: average-quality, Low: low-quality, and High: high-quality therapy. The direction of the arrow on the x-axis indicates the direction in which the frequency is increasing (we flip the axis if low-quality is more frequent than high-quality, such that low-quality is visualized below the average quality marker). Values colored in \textcolor{blueplot}{blue} indicate desirable behaviors (significantly closer to high-quality than low-quality) whereas values colored in \textcolor{orangeplot}{orange} indicate undesirable behaviors (significantly closer to low-quality than high-quality). Values in \textcolor{gray}{gray} are not statistically significantly different from average-quality at $p = \frac{0.05}{m}$ using Two-sided Student’s t-test, following Bonferroni correction (m: number of (client, therapist) intents tested = 13*6 = 78). Error bars indicate 95\% bootstrapped confidence intervals. Here, a key finding is that LLMs respond with significantly lower \textsc{Questions on Emotions} when clients express \textsc{Sustaining Unhealthy Behavior} (subfigure (b)), similar to low-quality human therapy.}}
\label{fig:clientvst-avg-new}
\end{figure*}

\showedits{\xhdr{Summary} \editsnew{LLM therapists providing MI support respond to some client behaviors similar to low-quality human MI therapy (Figure \ref{fig:clientvst-avg-new}; Supplementary Tables~\ref{tab:adapt_new_hl_p1}-\ref{tab:adapt_new_hl_p2}}). For example, both respond with more \textsc{Problem-Solving} when clients \textsc{Share Negative Emotions}. However, unlike low-quality human therapy, LLM therapists respond with more \textsc{Reflections on Strengths} when clients express \textsc{Changing Unhealthy Behavior}, which may be more desirable due to their empathetic and validating nature.}



\showedits{\xhdr{If clients express \textsc{Changing Unhealthy Behavior}, LLM therapists respond with significantly higher \textsc{Reflections on Strengths}, \textit{unlike} low-quality human therapy} We observe that GPT-4, GPT-3.5, and Llama2-70b are more frequent in expressing \textsc{Reflections on Strengths} by 6.5\%, 2.9\%, and 2.4\% respectively  when clients express \textsc{Changing Unhealthy
Behavior}, compared to average-quality human therapy (Figure~\ref{fig:clientvst-avg-new}a). This suggests that LLM therapists tend to reinforce the strengths of clients towards changing unhealthy behaviors when clients express a desire or take action to do so. This more closely resembles the higher degrees of \textsc{Reflections on Strengths} behavior observed in high-quality therapy than low-quality therapy, suggesting that such behavior may be desirable.}

\showedits{\xhdr{If clients express \textsc{Sustaining Unhealthy Behavior}, LLM therapists respond with significantly {fewer} \textsc{Questions on Emotions}, \textit{similar to} low-quality human therapy} When clients express \textsc{Sustaining Unhealthy Behavior}, we find that GPT-4, GPT-3.5, and Llama2-70b are {less} frequent in exhibiting \textsc{Questions on Emotions} by 8.8\%, 8.7\%, and 6.7\% respectively, compared to average-quality human therapy (Figure~\ref{fig:clientvst-avg-new}b). This more closely resembles low-quality therapy, suggesting that such behavior may be undesirable, potentially hindering the deeper understanding of the sustenance of unhealthy behavior. }


\showedits{\xhdr{If clients \textsc{Share Negative Emotions}, LLM therapists respond with higher \textsc{Problem-Solving}, \textit{similar to} low-quality human therapy} We find that GPT-4, GPT-3.5, and Llama2-70b are more frequent in expressing \textsc{Problem-Solving} by 18.1\%, 16.6\%, and 14.2\% respectively  when clients \textsc{Share Negative Emotions}, compared to average-quality human therapy (Figure~\ref{fig:clientvst-avg-new}d). This more closely resembles low-quality therapy than high-quality suggesting that such a behavior may be undesirable.}

\showedits{\xhdr{If clients share \textsc{Gained Insights}, all LLM therapists respond with fewer \textsc{Questions on Experiences}, \textit{similar to} low-quality human therapy} When clients express \textsc{Questions on Experiences}, we find that GPT-4, GPT-3.5, and Llama2-13b express fewer \textsc{Questions on Experiences} by 16.1\%, 13.1\%, and 11.8\% respectively, compared to average-quality human therapy (Figure~\ref{fig:clientvst-avg-new}f). Asking fewer \textsc{Questions on Experiences} could indicate lesser engagement with client insights.  This more closely resembles low-quality therapy than high-quality therapy suggesting that such a behavior may be undesirable.}

\subsection*{Can LLM Behavior be Modulated Towards High-Quality Psychotherapy?}
\label{sec:prompt-variation}
\begin{figure*}[t]
\includegraphics[width=\textwidth]{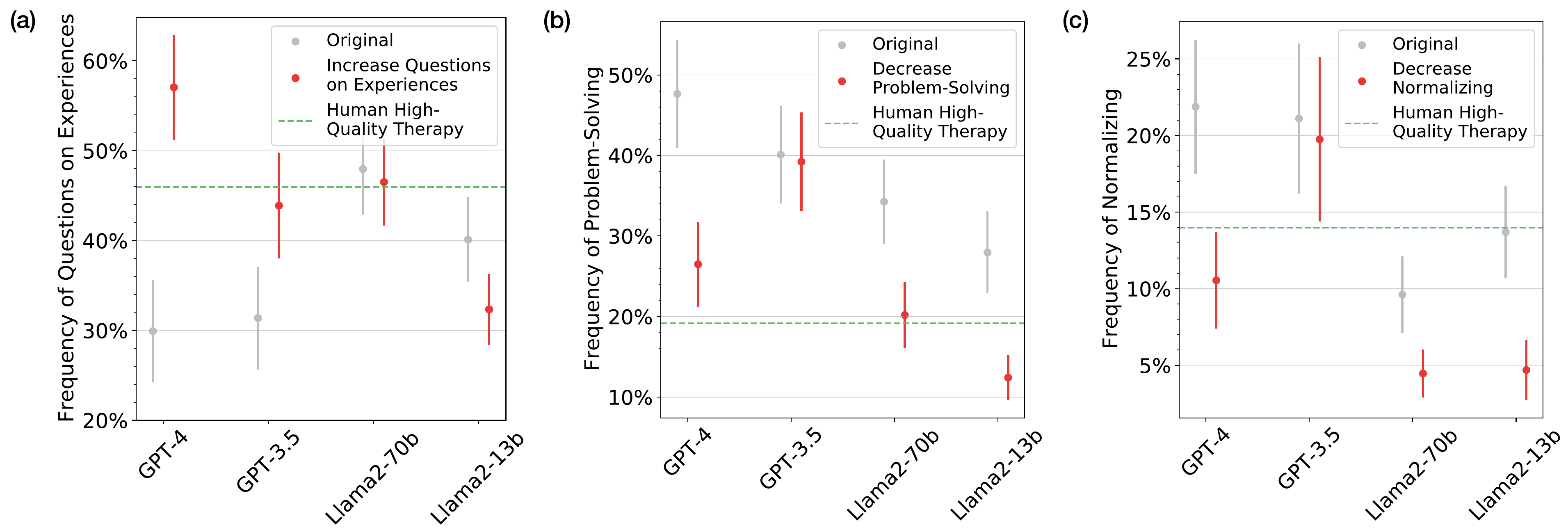}
\centering
\vspace{-15pt}
\caption{We incorporate simple prompts that aim to calibrate LLM therapists, specifically (a) increase \textsc{Questions on Experiences}, (b) decrease \textsc{Problem-Solving}, and (c) decrease \textsc{Normalizing}. 
Subfigures show changes in the frequency of conversational behaviors based on changes in prompts to different LLM therapists (GPT-4, GPT-3.5 turbo, Llama2-70b, Llama2-13b). 
Changes to individual LLMs are shown in pairs (\textcolor{gray}{left} -- Original prompt; \textcolor{redplot}{right} -- modulated prompt). The corresponding high-quality human therapy behavior frequency is shown as green dashed lines. For instance, GPT-4 increases the frequency of \textsc{Questions on Experiences} from 29.9\% to 57.0\% with the modulated prompt. In general, we find that only GPT-4 is able to modulate behavior frequency to a statistically and practically significant amount into the desired direction always, whereas the modulation is inconsistent for other models. Error bars indicate 95\% bootstrapped confidence intervals. }

\label{fig:control_propmt_variation}
\end{figure*}


LLMs are being used to support mental health \cite{hamilton-linkedin,fastcompany-therapy,stade2023artificial,mental-health-gpts,sharma2023naturemi,sharma-etal-2023-cognitive,openai-employee,reddit-examples,twitter-examples}, but the behavior of LLM therapists often differs significantly from high-quality human therapy. Therefore, it is crucial to develop methods that make them more consistent with high-quality care.

Here, we perform one specific experiment to study whether variations in prompts may help in modulating their behavior toward high-quality therapy. We noticed that LLM therapists respond less with \textsc{Questions on Experiences}, more on \textsc{Problem-Solving}, and more on \textsc{Normalizing} than human high-quality therapy (Figure \ref{fig:control_propmt_variation}, Results). Here, we vary our original system prompt stated in Methods to: (a) increase \textsc{Questions on Experiences}, (b) decrease \textsc{Problem-Solving}, and (c) decrease \textsc{Normalizing} (see Methods section).


\xhdr{Only GPT-4 is able to modulate these behaviors successfully} We find that only GPT-4 is able to successfully modulate the three behaviors in the intended direction, whereas the behavior modulation of GPT-3.5, Llama2-70b, and Llama2-13b is inconsistent (Figure \ref{fig:control_propmt_variation}). Specifically, GPT-4 increases the occurrence of \textsc{Questions on Experiences} from 29.9\% to 57.0\%, decreases the \textsc{Problem-Solving} behavior from 47.6\% to 26.5\%, and decreases \textsc{Normalizing} behavior from 21.9\% to 10.5\%.

\xhdr{GPT-3.5 is able to increase behavior frequency successfully, but does not decrease behavior frequency} Certain behaviors (e.g., \textsc{Questions}), might necessitate an increase in frequency, while others may call for less frequent occurrences (e.g., \textsc{Solutions}) depending on what is preferred in high-quality care. We observe that GPT-3.5 has better modulation performance when increasing \textsc{Questions on Experiences} -- it increases such questions from 31.4\% to 43.9\%. However, it attains unsatisfactory performance when decreasing \textsc{Problem-Solving} (a reduction in frequency from 40.1\% to 39.2\%) and decreasing \textsc{Normalizing} (a reduction in frequency from 21.9\% to 20.5\%).

\xhdr{Llama2 variants perform better with the modulation of decreasing a behavior rather than increasing} Llama2-70b and Llama2-13b, in contrast to GPT-3.5, are able to decrease the frequency of behaviors but not increase them when required for modulation. Specifically, for \textsc{Problem-Solving}, Llama2-70b decreases the behavior frequency from 34.3\% to 20.2\%, and Llama2-13b decreases it from 28.0\% to 12.4\%. For \textsc{Normalizing}, they decrease the frequency from 9.6\% to 4.5\%, and from 13.7\% to 4.7\% respectively. However, they fail to successfully increase the frequency of \textsc{Questions on Experiences}. Llama2-70b slightly decreases the behavior frequency from 48.0\% to 46.5\%, and Llama2-13b decreases the behavior frequency from 40.1\% to 32.3\%.

Overall, our findings indicate that the ability to modulate the LLM behavior to a statistically and practically significant amount is inconsistent across LLMs. Currently, people using LLMs as therapists may struggle to modulate LLM behaviors to a desired direction always, potentially compromising the quality of care. Even if that was possible, most users do not have the expertise to design effective LLM prompts \cite{zamfirescu2023johnny}, and may not possess a clear understanding of what constitutes quality psychotherapy. Future LLMs with improved prompt tuning may enable better control over behaviors.

\section*{Discussion}
\label{sec:discussion}
This section discusses our findings' implications for using LLMs for psychotherapy.

\subsection*{More research is needed to ensure quality care}

\begin{table*}
\small
\centering
\resizebox{0.95\textwidth}{!}{%
\def\arraystretch{1.15}
\centering
\setlength{\tabcolsep}{4pt}
\begin{tabular}{c|ccc|ccc|ccc|ccc}
\toprule
\multirow{2}{*}{\textbf{Model}} & \multicolumn{3}{c}{\textbf{Frequency}} & \multicolumn{3}{c}{\textbf{Order}} & \multicolumn{3}{c}{\textbf{Adaptability}} & \multicolumn{3}{c}{\textbf{Aggregate (Average)}} \\
\cmidrule(lr){2-4} \cmidrule(lr){5-7} \cmidrule(lr){8-10} \cmidrule(lr){11-13}  
& \textbf{\ding{51}} & \textbf{\ding{55}} & \textbf{?} & \textbf{\ding{51}} & \textbf{\ding{55}} & \textbf{?} & \textbf{\ding{51}} & \textbf{\ding{55}} & \textbf{?} & \textbf{\ding{51}} & \textbf{\ding{55}} & \textbf{?} \\
 \midrule
GPT-4 & 30.8\% & 46.2\% & 23.0\% &  46.2\% & 7.7\% & 46.1\% & 24.3\% & 42.3\% & 33.4\% & 33.7\% & 32.1\% & 34.2\% \\
GPT-3.5 & 38.5\% & 61.5\% & 0.0\% & 23.1\% & 15.4\% & 61.5\% & 24.4\% & 35.9\% & 39.7\% & 28.7\% & 37.6\% & 33.7\% \\
Llama2-70b & 38.5\% & 30.8\% & 30.7\% & 46.2\% & 15.4\% & 38.4\% & 30.8\% & 28.2\% & 41.0\% & 38.4\% & 24.8\% & 36.7\% \\
Llama2-13b & 30.8\% & 30.8\% & 38.4\% & 30.8\% & 23.1\% & 46.1\% & 23.1\% & 20.5\% & 56.4\% & 28.2\% & 24.8\% & 47.0\%  \\
\bottomrule 
\end{tabular}
}
\caption{\showedits{Percentage of analysis in which the models were found to be desirable (\textbf{\ding{51}}), undesirable (\textit{\ding{55}}), or uncertain (\textbf{?}). On one hand, in some cases, LLMs demonstrate desirable behaviors resembling high-quality therapy, which could be promising. However, the prevalence of undesirable behaviors resembling low-quality therapy is concerning. These behaviors deviate from established therapeutic guidelines and could affect the overall quality of care.}
}
\label{tab:optimism-pessimism}
\end{table*}

LLM technologies hold great promise to improve the access to mental health support through scalable interventions that can reach a large population \cite{choudhury2023llms,Sharma2023FacilitatingSM}. To illustrate this potential, developers and end users have shared anecdotal examples on social media and other platforms that demonstrate striking similarities between LLMs like ChatGPT and human therapists (e.g., being able to problem-solve \cite{twitter-examples} or being heard \cite{reddit-examples,reddit-examples-2}). This is prompting developers and users alike to substitute human therapists and established, evidence-based techniques with them. Many people are now choosing LLM therapists to work through their mental health issues\cite{openai-employee,reddit-examples,twitter-examples}. 


However, our research suggests that LLM therapists currently {often} resemble the behavior of therapists in low-quality therapy rather than high-quality therapy sessions. \showedits{Table~\ref{tab:optimism-pessimism} summarizes the number of instances in which the LLM therapist behavior was found to be desirable (\ding{51}), undesirable (\ding{55}), or uncertain (\textbf{?}). 
We can interpret these reuslts from multiple perspectives. 
On one hand, some behaviors show promise aligning more closely high-quality therapy than low-quality therapy. Conversely, the prevalence of undesirable behaviors closer to low-quality therapy could be a cause for concern. These behaviors deviate from established treatment methods and are discouraged according to psychotherapy recommendations. This may result in inferior and possibly harmful interactions, or unhealthy relationships between clients and LLMs, affecting overall therapeutic alliance and quality of care \cite{horvath2001alliance,fairburn2011therapist}. Our results suggest that more research is needed to ensure high-quality care, consistent with concerns raised by experts about the lack of evidence supporting the efficacy of LLMs and the risks associated with their use for mental health support \cite{stade2023artificial,choudhury2023llms,li2020developing,tate2023chatgpt}.}




\subsection*{LLM training prioritizes certain behaviors, but psychotherapy applications exhibit conflicting needs}
Our research identifies behaviors that LLMs currently overemphasize (e.g., \textsc{Problem-Solving}) or underemphasize (e.g., asking \textsc{Questions}), which are actually more closely associated with low-quality therapy sessions. The overemphasized/underemphasized behaviors could have been learned during RLHF which LLM-training organizations use to align LLM behaviors with human preferences and improve model performance \cite{bai2022training,ouyang2022training}. However, such behaviors may not be preferable across all settings, including not in mental health. Current RLHF alignment focuses on short-term outcomes like the helpfulness of the immediate next response\cite{bai2022training,ouyang2022training}. However, mental health prioritizes longer-term outcomes like assessing behavioral change across multiple sessions\cite{lambert2013bergin}. Such a notion may affect which behaviors are emphasized at different stages of a conversation.

\showedits{Recent studies advocate for a pluralistic alignment of LLMs to accommodate diverse values and perspectives\cite{sorensen2024roadmap}. In particular, RLHF optimizing for a reward from a single user is insufficient in representing diverse human preferences \cite{Chakraborty2024MaxMinRLHFTE} and is insufficient for tailoring LLMs to specific contexts like therapy. This reward is currently based on a ``third person without relevant expertise,'' which is unlikely to train LLMs to exhibit the behaviors in therapy. This highlights the need for a human-centered approach in building these models that considers the unique requirements of different applications, such as the quality of care in healthcare.}


\showedits{\subsection*{Need for public, transparent evaluation methods to inform development, deployment, and adoption of LLM-supported mental health applications}}

As LLMs are integrated with new and existing mental health interventions in both commercial \cite{hamilton-linkedin,fastcompany-therapy,mental-health-gpts} and academic settings \cite{stade2023artificial,sharma2023naturemi,sharma-etal-2023-cognitive,hsu2023helping,xu2023leveraging}, it is important to have public, transparent, trusted means of evaluating LLMs. Our framework serves as an initial step towards this goal. More work is needed to comprehensively assess LLM use in mental health, particularly in terms of quality, safety, and ethical considerations \cite{choudhury2023llms,pendse2022treatment,huang2023survey,duffourc2023generative,de2022chatbots, weidinger2021ethical}.  This important line of research could build upon the qualitative perspectives offered by mental health experts and researchers \cite{choudhury2023llms,chung2023challenges,stade2023artificial} and quality, safety, and ethical considerations set up in psychotherapy research \cite{miller2012motivational,lambert2013bergin,fairburn2011therapist,barlow1999dissemination,waltz1993testing}.

\clearpage
\section*{Methods}

\subsection*{\framework: Framework for Assessing Conversational Behavior of LLM Therapists}
\label{sec:framework}

We develop \framework, a novel computational framework to systematically assess the \textbf{B}ehavior \textbf{O}f \textbf{L}LM \textbf{T}herapists and compare them against human therapists. See Figure~\ref{fig:intro-figure} for an overview. Given an LLM (e.g., GPT-4) that is being used as a therapist, we aim to assess the ways in which it responds when supporting clients with mental health issues and whether its response behavior aligns with or deviates from the response behavior of high- and low-quality therapy conversations.

To achieve this, \framework~simulates therapy conversations between simulated clients and the LLM. Next, it leverages psychotherapeutic approaches as defined in psychotherapy research \cite{lee-etal-2019-identifying,cao2019observing} to identify the behavior of the LLM therapist and compares them against high- and low-quality human therapy conversations, as identified by experts in previous studies \cite{perez-rosas-etal-2019-makes}. 

\subsubsection*{LLM Therapists}
\label{sec:llm-therapists}
A common way in which current LLMs are being used as therapists is through custom ``system prompts'' that instruct them to function as therapists. \framework~uses a simple ``system prompt'' which was designed based on commonly reported example prompts by users on mental health subreddits \cite{reddit-examples,reddit-examples-2}:
\begin{quote}
\textit{Act as if you're a professional therapist. You provide evidence-based therapy to help clients seeking help with mental health challenges. You should maintain your therapist persona while responding. Communicate in a conversational style, mirroring the style of previous therapist responses.}
\end{quote}

We make subtle variations to this prompt to adapt it to different therapy types, such as Motivational Interviewing (Appendix Figure~\ref{fig:prompts_for_simulation}). We study the effects of variations in ``system prompts'' (Results).

\subsubsection*{Datasets of Therapy Conversations}
\label{sec:dataset-description}

Acquiring datasets of therapy conversations poses a significant challenge as they are typically private and rarely shared. 
Moreover, potential privacy issues may arise when exposing therapy datasets to public LLM APIs as they may contain sensitive client information. To circumvent these potential concerns, \framework~leverages publicly available datasets of therapy conversations.

Publicly available therapy conversation datasets are limited. Here, we use two datasets that carefully preprocess publicly available YouTube videos on therapy. This ensures high-quality transcripts while maintaining the confidentiality of sensitive personal information. We use both datasets with appropriate license and consent as provided by the authors within their terms of use.

\xhdr{(1) High-and-Low-Quality Therapy Conversation Dataset (High-Low Quality)} The first dataset, created by Perez-Rosas et al. \cite{perez-rosas-etal-2019-makes}, comprises 259 therapy conversations, primarily focusing on the evidence-based motivational interviewing (MI) therapy. Based on the MI psychotherapy principles, they evaluate the quality of conversations in the dataset to have 155 high-quality therapy transcripts and 104 low-quality transcripts. We use both high-quality and low-quality therapy conversations by human therapists to understand desirable and undesirable conversational behaviors.

\xhdr{(2) HOPE Dataset} The second dataset from Malhotra et al. \cite{malhotra2022speaker} was used to study dialogue acts in therapy. This dataset contains 212 therapy transcripts and includes conversations employing different types of therapy techniques (e.g., MI, Cognitive Behavioral Therapy). 

Both of these datasets (available in CSV formats with speaker and utterance columns) were originally sourced from public video-sharing platforms like YouTube. We use these datasets to simulate conversations between LLM therapists and simulated clients. Moreover, we use them to study the behavior of LLM therapists in varied mental health contexts (Results). 

\editsnew{\framework~is general and could be instantiated to assess different LLM therapists based on varied therapeutic frameworks. Our main analysis and findings are focused on the High-Low Quality dataset, which facilitates assessment within a specific therapeutic approach of Motivational Interviewing and {enables} the interpretation of findings {within a specific therapeutic framework}.  To test the generalizability of our results, we apply \framework~on the HOPE dataset that includes more general therapy. We find highly similar results for all four of our analyses including the frequency (Supplementary Table~\ref{tab:dist_new_hope}), temporal order (Supplementary Table~\ref{tab:order_new_hope}), and adaptability (Supplementary Tables~\ref{tab:adapt_new_hope_p1}-\ref{tab:adapt_new_hope_p2}). This suggests that LLM therapists might be exhibiting similar behavior more broadly. Future work is needed to assess their behavior on more datasets and more types of therapeutic frameworks.} 


\subsubsection*{Simulating Conversations between LLM Therapists and Clients} 
\label{subsec:simulating}

\label{sec:simulation-methods}
To analyze the behavior of LLM therapists, we need a dataset that contains conversations between LLM therapists and clients. Recruiting clients struggling with mental health issues and asking them to interact with LLM therapists presents significant ethical challenges and risks. Such an approach may expose vulnerable populations to potentially harmful LLM generations. 

Instead, \framework~uses the public therapy datasets to simulate conversations between LLM therapists and simulated clients. We develop two simulation strategies that represent two common use cases of LLM therapists -- (1) LLM Single Response Simulation: using LLMs to write a single response within an ongoing conversation session between a human therapist and a client, and (2) LLM Full Conversation Simulation: facilitating a {full, multi-turn} conversation session between LLMs and clients.

\xhdr{(1) LLM Single Response Simulation} \label{sec:single-response-sim}
Let $C_{i}$ be an ongoing conversation between a human therapist and client with utterances $u_{i,1}, u_{i,2}, ..., u_{i,j}$ ending with a client utterance $u_{i,j}$. We aim to use an LLM $\mathcal{L}$ to generate a single response $u_{i,j+1}^{(\mathcal{L})}$ to the client utterance. To simulate such single responses, we use the therapy datasets to create all possible sequences of $u_{i,1}, u_{i,2}, ..., u_{i,j}$ ending with a client utterance. For each of these sequences, we use the ``system prompt''  to generate the next LLM therapist response $u_{i,j+1}^{(\mathcal{L})}$. Check Appendix Tables of the first part \ref{tab:llm_example_single_response} and second part \ref{tab:llm_example_single_response_continue} for an illustrative example.

\xhdr{(2) LLM Full Conversation Simulation} 
In addition to single-response generation, we also simulate the full conversations between LLM therapists and clients. Here, we leverage the utterances from clients in the therapy datasets to simulate artificial clients using GPT-4. For each human-human conversation between a client and a therapist, we provide it as input to GPT-4 and ask it to simulate the same client talking to a different therapist in a parallel universe, ignoring that this conversation ever happened (to avoid GPT-4 continuing from the provided human-human conversation rather than simulating a new similar conversation), but keeping the same conversational style, the manner of addressing certain topics or concerns, and the life events and emotions being discussed (also see Appendix Figure~\ref{fig:prompts_for_simulation}). Then, we generate a conversation between this artificial client and the LLM therapist. We start the generation {by randomly picking between the client or therapist to go first}. We let them talk for up to 20 turns (which roughly corresponds to the mean conversation length (20.8 turns) in the two datasets used) or until one of them outputs an end token. See Appendix Tables~\ref{tab:llm_example_simulated} and \ref{tab:llm_example_simulated_hope} for illustrative examples.

\xhdr{Evaluating Simulated Conversations} \editsnew{We evaluate the consistency of the simulated conversations. We define \textit{consistency} as generating an utterance that is {logically and semantically} consistent with the previous utterances without conflicting with them. We use a GPT-4-based few-shot prompting method\cite{ziems2023can} to measure the consistency and coverage of the simulated conversations (this measure achieves an accuracy of 95.83\% and macro-f1 of 82.22\% in our human evaluation). We find that the simulated conversations have a {near-perfect} mean consistency of 2.95 (out of 3; std = 0.13). In addition, we qualitatively assess the conversations and find that simulated conversations follow the source conversational snippets in their overall theme and structure.}



\subsubsection*{Behavioral Techniques in Psychotherapy}
\label{subsec:behavioral-codes}
To assess the conversational behavior exhibited by therapists and clients during therapy sessions, we characterize the psychotherapeutic approaches underlying utterances. We draw upon the clinically relevant dialogue acts of therapist and client utterances (intentions of the speaker in the utterance \cite{stolcke-etal-2000-dialogue}) established by experts in psychotherapy in two previous studies \cite{lee-etal-2019-identifying, cao2019observing}.

We discuss the conversational behaviors in the following. Then we describe our method for automatically identifying these behaviors.

\paragraph*{Conversational Behavior of Therapists} 
\label{sec:therapist-conversational-behavior}

For characterizing the conversational behavior of therapists, we focus on 13 major psychotherapeutic approaches based on techniques identified in Lee et al. \cite{lee-etal-2019-identifying} and Cao et al. \cite{cao2019observing} organized into five semantically related categories:

\xhdr{(a) \textsc{Reflections}} A foundational tool for any therapist is to understand, restate, and emphasize the client's feelings and experiences (e.g., ``\textit{You are feeling overwhelmed and struggling to find a balance between work and family}'') \cite{arnold2014behind}. Such reflective listening has been shown to help clients gain a deeper understanding of their feelings and experiences and also improve the therapeutic relationship between the therapist and client through increased empathy \cite{braillon2020practicing, rautalinko2007reflective}. Here, we focus on six different ways of reflecting {differentiating between} (1) \textsc{Needs}, (2) \textsc{Emotions}, (3) \textsc{Values}, (4) \textsc{Consequences}, (5) \textsc{Conflicts}, and (6) \textsc{Strengths} (Appendix Table \ref{tab:therapist_intent_def})

\xhdr{(b) \textsc{Questions}} Open-ended questions constitute a fundamental element of therapy, helping clients express emotions and share thoughts and experiences they might not explore otherwise (e.g., ``\textit{How were you feeling in that moment?}'') \cite{lundh2019three,rautalinko2007reflective}. Also, expressing an active interest in clients' emotions and experiences enhances relationship forming with clients~\cite{miller2003manual,robert2011empathy}. Here, we focus on open-ended questions {differentiating between} (7) \textsc{Experiences}, (8) \textsc{Perspectives}, and (9) \textsc{Emotions} (Appendix Table \ref{tab:therapist_intent_def}).

\xhdr{(c) \textsc{Solutions}} To achieve the therapeutic goal effectively, therapists often offer solutions to client's problems by discussing concrete problem-solving approaches and constructing a specific plan of action \cite{nelson2013practical,solomonov2019multitheoretical,fawcett2014learning}. Here, we focus on these two types of solutions {differentiating between} (10) \textsc{Problem-Solving}, and (11) \textsc{Planning} (Appendix Table~\ref{tab:therapist_intent_def}).

\xhdr{(d) \textsc{Normalizing}} Being warm, friendly, and respectful has been found to contribute positively to the relationship (or alliance) between the client and therapist, which can be achieved through acknowledging and validating clients's feelings and experiences, empathizing, and reassuring (e.g., \textit{I hear you, it's perfectly normal to feel overwhelmed...}). Hence, we combine these related concepts into (12) \textsc{Normalizing}.

\xhdr{(e) \textsc{Psychoeducation}} Therapists commonly educate clients about coping strategies, treatment approaches, symptoms, or diagnosis (e.g., ``\textit{Cognitive Behavioral Therapy aims to...}'') \cite{lundh2019three}. Here, we broadly call this aspect as (13) \textsc{Psychoeducation}.

Refer to Appendix Table \ref{tab:therapist_intent_def} for definitions and examples of all 13 therapist conversational behaviors.

\paragraph*{Conversational Behavior of Clients} 
\label{sec:client-conversational-behavior}

For a therapist, knowing when to express specific behaviors based on different client behaviors (e.g., when to reflect vs. when to utilize problem-solving) is an important skill. Here, to more thoroughly assess therapist behavior based on how they respond and adapt to clients, we characterize the conversational behavior of clients. We use six types of expressions from clients based on the existing works \cite{lee-etal-2019-identifying, cao2019observing} organized into three semantically related categories. We focus on a combination of client behaviors that include the changes in client behaviors, the emotions and experiences they share, and the insights they gain from therapy \cite{carey2007psychological, tulver2023restructuring}:

\xhdr{(a) \textsc{Behavior Change}} Some clients may struggle with unhealthy behaviors or symptoms (E.g., ``\textit{I'm having trouble sleeping}''). Here, we differentiate between clients expressing {(1) \textsc{Changing Unhealthy Behavior}} and (2) \textsc{Sustaining Unhealthy Behavior} (Appendix Table~\ref{tab:client_intent_def}).
 
\xhdr{(b) \textsc{Self-Disclosure of Affect or Experiences}} During the conversation, clients may share their feelings or related experiences (e.g., ``\textit{I lost my job due to the pandemic}''). Here, we focus on clients' expression of (3) \textsc{Sharing Positive Emotions}, (4) \textsc{Sharing Negative Emotions}, and (5) \textsc{Sharing Experiences} (Appendix Table~\ref{tab:client_intent_def}).

\xhdr{(c) \textsc{Gaining Insights}} Clients often report learning something new about themselves or their situation during therapy (e.g., ``\textit{I hadn't considered how much I avoid confrontations}'').  Here, we detect such (6) \textsc{Gained Insights} by clients.

Refer to Appendix Table \ref{tab:client_intent_def} for definitions and examples of all six client conversational behaviors.

\subsection*{Identifying Conversational Behavior in Psychotherapy Conversations}

\framework~automatically identifies the conversational behavior of LLM therapists based on the simulated conversations. We annotate the therapy datasets with conversational behavior and use it to train and evaluate GPT-based methods.  

\subsubsection*{Classification Setup}
\label{sec:classification-setup}
We frame the identification of conversational behaviors as either a single multi-label classification task (identify all possible behaviors in a given utterance at once simultaneously) or a multiple binary-label classification task (identify independently whether a particular conversation behavior is exhibited by an utterance).

Let $C_{i}$ be a conversation between a therapist and a client with utterances $u_{i,1}, u_{i,2}, ...$.

\xhdr{Multi-label classification} For each therapist utterance, $u^{(t)} \in \{u_{i,t_1}, u_{i,t_2}, ...\}$, we aim to identify all possible behaviors exhibited by $u^{(t)}$ among the 13 different therapist codes. Similarly, for each client utterance, $u^{(c)} \in \{u_{i,c_1}, u_{i,c_2}, ...\}$, we aim to identify all possible behaviors exhibited by $u^{(c)}$ among the 6 different client codes.

\xhdr{Binary-label classification} For each therapist utterance, $u^{(t)} \in \{u_{i,t_1}, u_{i,t_2}, ...\}$ and for each therapist behavioral code in the 13 possible therapist codes, we aim to identify whether $u^{(t)}$ exhibits that behavior. Similarly, for each client utterance, $u^{(c)} \in \{u_{i,c_1}, u_{i,c_2}, ...\}$ and for each client behavioral code in the 6 possible client codes, we aim to identify whether $u^{(c)}$ exhibits that behavior.

\subsubsection*{Annotations of Conversational Behavior}
\label{sec:self-annotated-dataset}
The broad range of therapist and client conversational behavior and their complexity make crowdwork data annotation process challenging. Here, two co-authors of this paper with significant expertise in NLP and mental health research, and who were actively engaged in defining the conversational behaviors of therapists and clients, annotated 283 therapist utterances and 91 client utterances (inter-rater agreement, krippendorff's alpha = 0.7834). Our annotated dataset includes conversations from the human therapy datasets as well as the simulated conversations based on LLM therapists.

\subsubsection*{Models}
\label{sec:classifcation-methods}

We compare three models to identify conversational behaviors in therapist and client utterances. Note that GPT-4-based methods are the state-of-the-art for this task.

\xhdr{Finetuning (multi-label)} The first method involved fine-tuning GPT-3 series models (GPT-3: davinci-002, GPT-3.5: gpt-3.5-turbo-0613) \cite{brown2020language} with our annotated dataset described above (Note that GPT-4 did not allow finetuning at the time of the writing of this paper. Finetuning for classifying individual conversational behavior for the binary classification method was prohibitively expensive. Therefore, we instead finetune two multi-label classification models for the conversation behaviors of therapists and clients, respectively.)

\xhdr{Prompting / In-context Learning (multi-label)} We facilitate in-context learning by prompting GPT-3 \cite{brown2020language} and GPT-4 \cite{OpenAI2023GPT4TR} models to identify all possible multiple intents in a single utterance. We explored two variations of prompting -- (1) We prompt the models with definitions of different conversational behaviors sourced from prior work\cite{lee-etal-2019-identifying, cao2019observing} (Appendix Tables~\ref{tab:therapist_intent_def} and \ref{tab:client_intent_def}); (2) We prompt the models with definitions as well as in-context examples from our annotated dataset(We tried different numbers of in-context examples and found three examples to work the best). We use the following base instruction prompt for both variations -- ``\textit{What are all possible conversational behaviors of this utterance}.'' Check Appendix Figure \ref{fig:prompts_variations_classification} for all employed prompts.

\xhdr{Prompting / In-context Learning (binary-label)} We create binary questions focusing on one intent at a time with its definition and in-context examples (e.g., ``\textit{Classify if the utterance contains \{conversational behavior\}. Answer in Yes or No}''). Check Appendix Figure \ref{fig:prompts_variations_classification} for all employed prompts.

\xhdr{Experimental Setup} We create five random train-test splits of our annotated dataset, each with a 60:40 split. We use the training split to finetune GPT-3 models and as demonstration examples for in-context learning. For all models, we report the mean and standard deviation of Macro-Precision, Macro-Recall, and Macro-F1 scores for the 13-class therapist behavior classification task and the 6-class client behavior classification task. Given the limited dataset size, averaging across five random splits ensures a more robust reflection of classification performance. As a baseline for comparison, we assign random conversational behavior to utterances.

\showedits{\subsection*{Behavior Modulation Experiments}
We noticed that LLM therapists respond less with \textsc{Questions on Experiences}, more on \textsc{Problem-Solving}, and more on \textsc{Normalizing} than human high-quality therapy (Figure \ref{fig:control_propmt_variation}, Results). Here, we vary our original system prompt stated in Methods to: (a) increase \textsc{Questions on Experiences}, by adding an instruction that asks LLMs to ``\textit{focus more on asking questions to allow client to express their experiences}'', (b) decrease \textsc{Problem-Solving}, by adding an instruction that asks LLMs to ``\textit{focus less on offering possible solutions to
client’s problem}'', and (c) decrease \textsc{Normalizing}, by adding an instruction that asks LLMs to ``\textit{focus less on validating client’s experiences or feelings as normal,
on sympathizing with their challenges, and on providing reassurance}''. }

\showedits{\subsection*{Limitations}
We utilize validated yet automated and imperfect measures for the inference of behaviors. Our assessment is solely based on automated metrics and does not include a human evaluation. However, this was a deliberate choice to mitigate potential harm from untested interventions as well as to enable automated, and cost-effective assessment. Our analysis framework depends on the high- and low-quality labels from a previous study. However, it is adaptable to any other definition of quality (e.g., based on patient outcomes). We leverage a limited set of psychotherapeutic approaches and several other possible approaches exist based on specific framework.}

\section*{Inclusion and Ethics}
This study does not involve interaction or intervention with human subjects or access to private identifiable information and was deemed exempt from IRB review. Both the High-Low Quality and the HOPE datasets were used with appropriate license and consent as provided by the authors within their terms of use. These datasets leverage publicly available YouTube videos on therapy. This ensures high-quality transcripts while maintaining the confidentiality of sensitive personal information. Our work does not make any treatment recommendations or diagnostic claims.

\section*{Data availability}


Simulated conversation data on behavioral assessment of LLM therapists (\textsc{Bolt}) is available at \\ \href{https://github.com/behavioral-data/BOLT}{github.com/behavioral-data/BOLT}. The High-Quality dataset conversations are publicly available at \\ \href{https://lit.eecs.umich.edu/downloads.html}{https://lit.eecs.umich.edu/downloads.html} \cite{perez-rosas-etal-2019-makes}. The HOPE dataset conversations are publicly available at  \href{https://github.com/LCS2-IIITD/SPARTA_WSDM2022}{https://github.com/LCS2-IIITD/SPARTA\_WSDM2022} \cite{malhotra2022speaker}.

\section*{Code availability}
\rone{Source code of the behavioral assessment of LLM therapists (\textsc{Bolt}) is available at \\ \href{https://github.com/behavioral-data/BOLT}{github.com/behavioral-data/BOLT}.}

\section*{Acknowledgements}
We thank Derrick Hull for providing us with access to the definitions and examples of behavioral techniques described in Lee et al., 2019\cite{lee-etal-2019-identifying}. We are grateful to Perez-Rosas et al., 2019 \cite{perez-rosas-etal-2019-makes} and Malholtra et al., 2022 \cite{malhotra2022speaker} for the public access of their datasets. We thank Zac Imel, Jina Suh, and the UW Behavioral Data Science Group members for their suggestions and feedback. T.A., A.S., and I.W.L. were supported in part by NSF grant IIS-1901386, NSF CAREER IIS-2142794, NSF grant CNS-2025022, NIH grant R01MH125179, Bill \& Melinda Gates Foundation (INV-004841), the Office of Naval Research (\#N00014-21-1-2154), a Microsoft AI for Accessibility grant, a Garvey Institute Innovation grant, and UW Azure Cloud Computing Credits.

\section*{Author Contributions}
Y.Y.C., A.S., I.W.L., and T.A. were involved with the conceptualization of 
\framework~and related analyses. Y.Y.C. and A.S. trained the classification models and conducted analyses of the data. Y.Y.C., A.S., I.W.L., and T.A. interpreted the data, drafted the manuscript, and made significant intellectual contributions to the manuscript.
\section*{Figure Captions}

\xhdr{Figure 1} Overview of \framework, a computational framework that enables systematic assessment of the behavior of LLM therapists and compares them to high- and low-quality human therapy.

\showedits{\xhdr{Figure 2} Difference in the frequency of conversational behaviors exhibited by LLM therapists (GPT-4, GPT-3.5-turbo, Llama2-70b, Llama2-13b), relative to average-, low-, and high-quality human therapy. \textbf{A}: average-quality, Low: low-quality, and High: high-quality therapy. The direction of the arrow on the x-axis indicates the direction in which the frequency is increasing (we flip the axis if low-quality is more frequent than high-quality, such that low-quality is visualized below the average quality marker). Values colored in \textcolor{blueplot}{blue} indicate desirable behaviors (significantly closer to high-quality than low-quality) whereas values colored in \textcolor{orangeplot}{orange} indicate undesirable behaviors (significantly closer to low-quality than high-quality). Values in \textcolor{gray}{gray} are not statistically significantly different from average-quality at p = 0.05 at $p = \frac{0.05}{m}$ using Two-sided Student’s t-test, following Bonferroni correction (m: number of intents tested = 13). Error bars indicate 95\% bootstrapped confidence intervals. A key insight we find is that LLMs respond with significantly higher \textsc{Problem-Solving} (subfigure (a)), similar to low-quality human therapy. On the other hand, LLMs respond with significantly higher \textsc{Reflections on Strengths} (subfigure (l)), similar to high-quality therapy, but with a frequency that significantly exceeds high-quality therapy.}

\showedits{\xhdr{Figure 3} Difference in the \textit{temporal order} of conversational behaviors, operationalized as the turn numbers in which behaviors are first exhibited in a conversation by LLM therapists (GPT-4, GPT-3.5-turbo, Llama2-70b, Llama2-13b), relative to average-, low-, and high-quality human therapy. \textbf{A}: average-quality, Low: low-quality, and High: high-quality therapy. The direction of the arrow on the x-axis indicates the direction in which the order is increasing (we flip the axis if low-quality is exhibited later than high-quality, such that low-quality is visualized below the average quality marker). Values colored in \textcolor{blueplot}{blue} indicate desirable behaviors (significantly closer to high-quality than low-quality) whereas values colored in \textcolor{orangeplot}{orange} indicate undesirable behaviors (significantly closer to low-quality than high-quality). Values in \textcolor{gray}{gray} are not statistically significantly different from average-quality at $p = \frac{0.05}{m}$ using Two-sided Student’s t-test, following Bonferroni correction (m: number of intents tested = 13). Error bars indicate 95\% bootstrapped confidence intervals. Most LLM therapists start providing \textsc{Planning} (subfigure (b)) earlier in the conversations but provide \textsc{Normalizing} (subfigure (j)) later against common recommendations \cite{cochran2015heart}.}

\showedits{\xhdr{Figure 4} Difference between the frequency of conversational behaviors observed in LLM therapists (GPT-4, GPT-3.5-turbo, Llama2-70b, Llama2-13b) or low-quality human therapy \textit{in response to specific client behaviors} (\textit{Adaptability}), relative to average-, low-, and high-quality human therapy. \textbf{A}: average-quality, Low: low-quality, and High: high-quality therapy. The direction of the arrow on the x-axis indicates the direction in which the frequency is increasing (we flip the axis if low-quality is more frequent than high-quality, such that low-quality is visualized below the average quality marker). Values colored in \textcolor{blueplot}{blue} indicate desirable behaviors (significantly closer to high-quality than low-quality) whereas values colored in \textcolor{orangeplot}{orange} indicate undesirable behaviors (significantly closer to low-quality than high-quality). Values in \textcolor{gray}{gray} are not statistically significantly different from average-quality at $p = \frac{0.05}{m}$ using Two-sided Student’s t-test, following Bonferroni correction (m: number of (client, therapist) intents tested = 13*6 = 78). Error bars indicate 95\% bootstrapped confidence intervals. Here, a key finding is that LLMs respond with significantly lower \textsc{Questions on Emotions} when clients express \textsc{Sustaining Unhealthy Behavior} (subfigure (b)), similar to low-quality human therapy. }

\showedits{\xhdr{Figure 5} We incorporate simple prompts that aim to calibrate LLM therapists, specifically (a) increase \textsc{Questions on Experiences}, (b) decrease \textsc{Problem-Solving}, and (c) decrease \textsc{Normalizing}. 
Subfigures show changes in the frequency of conversational behaviors based on changes in prompts to different LLM therapists (GPT-4, GPT-3.5 turbo, Llama2-70b, Llama2-13b). 
Changes to individual LLMs are shown in pairs (\textcolor{gray}{left} -- Original prompt; \textcolor{redplot}{right} -- modulated prompt). The corresponding high-quality human therapy behavior frequency is shown as green dashed lines. For instance, GPT-4 increases the frequency of \textsc{Questions on Experiences} from 29.9\% to 57.0\% with the modulated prompt. In general, we find that only GPT-4 is able to modulate behavior frequency to a statistically and practically significant amount into the desired direction always, whereas the modulation is inconsistent for other models. Error bars indicate 95\% bootstrapped confidence intervals.}

\section*{Table Captions}
\showedits{\xhdr{Table 1} Percentage of analysis in which the models were found to be desirable (\textbf{\ding{51}}), undesirable (\textit{\ding{55}}), or uncertain (\textbf{?}). On one hand, in some cases, LLMs demonstrate desirable behaviors resembling high-quality therapy, which could be promising. However, the prevalence of undesirable behaviors resembling low-quality therapy is concerning. These behaviors deviate from established therapeutic guidelines and could affect the overall quality of care.}

\bibliography{_references}

\begin{thebibliography}{10}
\bibitem{fastcompany-therapy}
\bibinfo{author}{Broderick, R.}
\newblock \emph{\bibinfo{title}{People are using AI for therapy, whether the tech is ready for it or not, accessed 2023}} (\bibinfo{year}{2023}).
\end{thebibliography}

\appendix
\newpage

\setcounter{table}{0}
\renewcommand{\thetable}{S\arabic{table}}
\setcounter{figure}{0}
\renewcommand{\thefigure}{S\arabic{figure}}

\section*{Supplementary Materials}
\subsection*{List of supplementary materials}
Table S1-S29\\
Figures S1 to S2\\



\newpage
\begin{table*}[t]
    \centering
    \small
    \caption{\showedits{Frequency of conversational behaviors exhibited by LLM therapists (GPT-4, GPT-3.5-turbo, Llama2-70b, Llama2-13b) in High-low quality dataset \cite{perez-rosas-etal-2019-makes}. Values in \textcolor{gray}{gray} are not statistically significantly different from human therapist average at $p = \frac{0.05}{m}$ using Two-sided Student’s t-test after Bonferroni correction, where $m$ is the number of t-test per one model ($m = 13$). $d$ is the cohen's d when compared with human therapist average. For instance, GPT-4 responds with \textsc{Reflections on Needs} by 40.0\% while human therapists responds by 23.4\% on average ($P = 8.0 \times 10^{-73}$, Cohen's $d = \num{0.4}$, two-sided Student’s $t$-test).}}
    \resizebox{1\textwidth}{!}{%
    \def\arraystretch{1.15}
    \setlength{\tabcolsep}{2pt}
    \begin{tabular}{l|ccc|ccc|ccc|ccc|ccc|ccc}
         \toprule 
         \multirow{4}{*}{\textbf{Behavior}} & \multicolumn{12}{c}{\textbf{LLM Therapists (Single Response \& Full Conversation Simulations)}} & \multicolumn{3}{c}{\textbf{Human Therapists}} \\
         \cmidrule(lr){2-13} \cmidrule(lr){14-16}
         & \multicolumn{3}{c}{\textbf{GPT-4}} &\multicolumn{3}{c}{\textbf{GPT-3.5}}&\multicolumn{3}{c}{\textbf{Llama2-70b}}&\multicolumn{3}{c}{\textbf{Llama2-13b}}& \textbf{Average}&\textbf{High-quality}&\textbf{Low-quality}\\
& \multicolumn{3}{c}{\textbf{$(n = 4893)$}}& \multicolumn{3}{c}{\textbf{$(n = 4746)$}}& \multicolumn{3}{c}{\textbf{$(n = 4970)$}}& \multicolumn{3}{c}{\textbf{$(n = 4820)$}}&\textbf{$(n = 5446)$}&\textbf{$(n = 3907)$}&\textbf{$(n = 1539)$}\\
         \cmidrule(lr){2-4} \cmidrule(lr){5-7} \cmidrule(lr){8-10} \cmidrule(lr){11-13} \cmidrule(lr){14-14} \cmidrule(lr){15-15}  \cmidrule(lr){16-16} 
        & freq. & $p$ & $d$ & freq. & $p$ & $d$ & freq. & $p$& $d$ & freq. & $p$ & $d$ & freq. & freq. & freq. \\
         \midrule
Reflections on Needs&$\phantom{-}40.0_{49.0}$&$8.0 \times 10^{-73}$&$\phantom{-}0.4$&$\phantom{-}32.3_{46.8}$&$2.3 \times 10^{-22}$&$\phantom{-}0.2$&$\phantom{-}36.0_{48.0}$&$6.7 \times 10^{-44}$&$\phantom{-}0.3$&$\phantom{-}34.5_{47.5}$&$4.6 \times 10^{-34}$&$\phantom{-}0.2$&$\phantom{-}23.4_{43.1}$&$\phantom{-}26.2_{44.0}$&$\phantom{-}20.6_{40.5}$\\

Reflections on Emotions&$\phantom{-}13.8_{34.5}$&$4.5 \times 10^{-6}$&$\phantom{-}0.1$&$\phantom{-}19.4_{39.6}$&$2.3 \times 10^{-34}$&$\phantom{-}0.2$&$\phantom{-}14.9_{35.6}$&$6.3 \times 10^{-10}$&$\phantom{-}0.1$&$\phantom{-}14.4_{35.1}$&$4.8 \times 10^{-8}$&$\phantom{-}0.1$&$\phantom{-}10.5_{32.8}$&$\phantom{-}14.6_{35.3}$&$\phantom{-}6.3_{24.3}$\\

Reflections on Values&\textcolor{gray}{$\phantom{-}5.5_{22.7}$}&$1.0 \times 10^{0}$&$\phantom{-}0.0$&$\phantom{-}6.2_{24.1}$&$5.0 \times 10^{-2}$&$\phantom{-}0.1$&\textcolor{gray}{$\phantom{-}4.8_{21.4}$}&$1.0 \times 10^{0}$&$\phantom{-}0.0$&\textcolor{gray}{$\phantom{-}5.7_{23.1}$}&$8.6 \times 10^{-1}$&$\phantom{-}0.0$&$\phantom{-}4.8_{22.8}$&$\phantom{-}6.3_{24.4}$&$\phantom{-}3.3_{17.9}$\\

Reflections on Consequences&\textcolor{gray}{$\phantom{-}23.4_{42.4}$}&$5.4 \times 10^{-1}$&$\phantom{-}0.0$&$\phantom{-}14.5_{35.2}$&$5.1 \times 10^{-21}$&$-0.2$&\textcolor{gray}{$\phantom{-}22.4_{41.7}$}&$1.0 \times 10^{0}$&$\phantom{-}0.0$&\textcolor{gray}{$\phantom{-}24.0_{42.7}$}&$8.2 \times 10^{-2}$&$\phantom{-}0.1$&$\phantom{-}21.8_{40.1}$&$\phantom{-}17.9_{38.4}$&$\phantom{-}25.7_{43.7}$\\

Reflections on Conflicts&\textcolor{gray}{$\phantom{-}5.9_{23.5}$}&$1.0 \times 10^{0}$&$\phantom{-}0.0$&$\phantom{-}4.8_{21.4}$&$1.5 \times 10^{-3}$&$-0.1$&\textcolor{gray}{$\phantom{-}8.1_{27.3}$}&$6.9 \times 10^{-2}$&$\phantom{-}0.1$&\textcolor{gray}{$\phantom{-}7.8_{26.8}$}&$3.6 \times 10^{-1}$&$\phantom{-}0.0$&$\phantom{-}6.6_{26.1}$&$\phantom{-}8.3_{27.6}$&$\phantom{-}5.0_{21.8}$\\

Reflections on Strengths&$\phantom{-}39.1_{48.8}$&$1.9 \times 10^{-281}$&$\phantom{-}0.7$&$\phantom{-}32.1_{46.7}$&$3.8 \times 10^{-176}$&$\phantom{-}0.6$&$\phantom{-}24.0_{42.7}$&$8.0 \times 10^{-85}$&$\phantom{-}0.4$&$\phantom{-}21.9_{41.3}$&$7.0 \times 10^{-64}$&$\phantom{-}0.3$&$\phantom{-}9.5_{31.3}$&$\phantom{-}12.8_{33.5}$&$\phantom{-}6.2_{24.2}$\\

Questions on Experiences&$\phantom{-}29.8_{45.7}$&$2.1 \times 10^{-49}$&$-0.3$&$\phantom{-}29.9_{45.8}$&$7.4 \times 10^{-48}$&$-0.3$&$\phantom{-}48.2_{50.0}$&$2.4 \times 10^{-4}$&$\phantom{-}0.1$&$\phantom{-}40.7_{49.1}$&$1.2 \times 10^{-2}$&$-0.1$&$\phantom{-}44.0_{49.9}$&$\phantom{-}48.9_{50.0}$&$\phantom{-}39.1_{48.8}$\\

Questions on Perspectives&$\phantom{-}5.9_{23.5}$&$2.9 \times 10^{-7}$&$\phantom{-}0.1$&$\phantom{-}2.4_{15.4}$&$4.9 \times 10^{-2}$&$-0.1$&$\phantom{-}8.4_{27.7}$&$3.2 \times 10^{-24}$&$\phantom{-}0.2$&$\phantom{-}7.3_{26.0}$&$1.0 \times 10^{-15}$&$\phantom{-}0.2$&$\phantom{-}3.5_{19.9}$&$\phantom{-}4.9_{21.7}$&$\phantom{-}2.0_{14.1}$\\

Questions on Emotions&$\phantom{-}2.2_{14.7}$&$7.9 \times 10^{-75}$&$-0.4$&$\phantom{-}1.7_{13.0}$&$8.4 \times 10^{-83}$&$-0.4$&$\phantom{-}8.9_{28.5}$&$4.8 \times 10^{-5}$&$-0.1$&\textcolor{gray}{$\phantom{-}11.7_{32.2}$}&$1.0 \times 10^{0}$&$\phantom{-}0.0$&$\phantom{-}11.7_{33.0}$&$\phantom{-}13.3_{34.0}$&$\phantom{-}10.1_{30.2}$\\

Problem-Solving&$\phantom{-}57.8_{49.4}$&$3.5 \times 10^{-256}$&$\phantom{-}0.7$&$\phantom{-}49.2_{50.0}$&$7.2 \times 10^{-137}$&$\phantom{-}0.5$&$\phantom{-}45.8_{49.8}$&$6.1 \times 10^{-104}$&$\phantom{-}0.4$&$\phantom{-}34.4_{47.5}$&$1.3 \times 10^{-20}$&$\phantom{-}0.2$&$\phantom{-}25.9_{42.6}$&$\phantom{-}21.2_{40.9}$&$\phantom{-}30.5_{46.0}$\\

Planning&$\phantom{-}57.1_{49.5}$&$0.0 \times 10^{0}$&$\phantom{-}0.9$&$\phantom{-}42.6_{49.5}$&$1.2 \times 10^{-182}$&$\phantom{-}0.6$&$\phantom{-}44.6_{49.7}$&$1.6 \times 10^{-213}$&$\phantom{-}0.6$&$\phantom{-}36.2_{48.1}$&$7.1 \times 10^{-109}$&$\phantom{-}0.4$&$\phantom{-}17.1_{37.6}$&$\phantom{-}16.9_{37.5}$&$\phantom{-}17.2_{37.8}$\\

Normalizing&$\phantom{-}45.5_{49.8}$&$2.9 \times 10^{-169}$&$\phantom{-}0.6$&$\phantom{-}36.9_{48.3}$&$1.3 \times 10^{-76}$&$\phantom{-}0.4$&\textcolor{gray}{$\phantom{-}19.3_{39.5}$}&$1.0 \times 10^{0}$&$\phantom{-}0.0$&$\phantom{-}24.9_{43.3}$&$8.8 \times 10^{-7}$&$\phantom{-}0.1$&$\phantom{-}20.5_{39.5}$&$\phantom{-}17.8_{38.3}$&$\phantom{-}23.2_{42.2}$\\

Psychoeducation&$\phantom{-}16.4_{37.0}$&$2.5 \times 10^{-45}$&$\phantom{-}0.3$&$\phantom{-}24.5_{43.0}$&$1.0 \times 10^{-130}$&$\phantom{-}0.5$&$\phantom{-}12.6_{33.2}$&$2.2 \times 10^{-17}$&$\phantom{-}0.2$&\textcolor{gray}{$\phantom{-}8.9_{28.4}$}&$1.7 \times 10^{-1}$&$\phantom{-}0.0$&$\phantom{-}7.6_{24.8}$&$\phantom{-}5.2_{22.3}$&$\phantom{-}9.9_{29.8}$\\

    \bottomrule
    \end{tabular}
    }
    \label{tab:dist_new_hl}
\end{table*}
\begin{table*}[t]
    \centering
    \small
    \caption{\showedits{Frequency of conversational behaviors exhibited by LLM
therapists (GPT-4, GPT-3.5-turbo, Llama2-70b, Llama2-13b) in High-low quality dataset \cite{perez-rosas-etal-2019-makes} for two different types of simulations separately (Single response and Full conversation). Values in \textcolor{gray}{gray} are not statistically significantly different from human therapist average at $p = \frac{0.05}{m}$ using Two-sided Student’s t-test after Bonferroni correction, where $m$ is the number of t-test per one model ($m = 13$). $d$ is the cohen's d when compared with human therapist average. For instance, GPT-4 with single response simulation responds with \textsc{Reflections on Needs} by 31.1\% and GPT-4 with full conversation simulation responds by 42.5\% when compared with the human therapists responds by 23.4\% on average from Table \ref{tab:dist_new_hl}. (Single: $P = 2.3 \times 10^{-6}$, Cohen's $d = \num{0.2}$; two-sided student t-test) (Full: $P = 6.4 \times 10^{-84}$, Cohen's $d = \num{0.4}$; two-sided student t-test)}}
    \resizebox{1.05\textwidth}{!}{%
    \def\arraystretch{1.15}
    \setlength{\tabcolsep}{2pt}
    \begin{tabular}{l|ccc|ccc|ccc|ccc|ccc|ccc|ccc|ccc|ccc|ccc}
         \toprule 
         \multirow{3}{*}{\textbf{Behavior}} & \multicolumn{12}{c}{\textbf{LLM Therapists (Single Response Simulations)}} & \multicolumn{12}{c}{\textbf{LLM Therapists (Full Conversation Simulations)}} \\
         \cmidrule(lr){2-13}  \cmidrule(lr){14-25} 
         & \multicolumn{3}{c}{\textbf{GPT-4}} &\multicolumn{3}{c}{\textbf{GPT-3.5}}&\multicolumn{3}{c}{\textbf{Llama2-70b}}&\multicolumn{3}{c}{\textbf{Llama2-13b}}& \multicolumn{3}{c}{\textbf{GPT-4}} &\multicolumn{3}{c}{\textbf{GPT-3.5}}&\multicolumn{3}{c}{\textbf{Llama2-70b}}&\multicolumn{3}{c}{\textbf{Llama2-13b}}\\
& \multicolumn{3}{c}{\textbf{$(n = 3837)$}}& \multicolumn{3}{c}{\textbf{$(n = 3907)$}}& \multicolumn{3}{c}{\textbf{$(n = 3906)$}}& \multicolumn{3}{c}{\textbf{$(n = 3895)$}}& \multicolumn{3}{c}{\textbf{$(n = 1056)$}}& \multicolumn{3}{c}{\textbf{$(n = 839)$}}& \multicolumn{3}{c}{\textbf{$(n = 1064)$}}& \multicolumn{3}{c}{\textbf{$(n = 925)$}}\\
         \cmidrule(lr){2-4} \cmidrule(lr){5-7} \cmidrule(lr){8-10} \cmidrule(lr){11-13} \cmidrule(lr){14-16} \cmidrule(lr){17-19} \cmidrule(lr){20-22}  \cmidrule(lr){23-25}
        & freq. & $p$ & $d$ & freq. & $p$ & $d$ & freq. & $p$& $d$ & freq. & $p$ & $d$ & freq. & $p$& $d$ &  freq. & $p$ & $d$ & freq. & $p$ & $d$ & freq. & $p$& $d$ \\
         \midrule
Reflections on Needs&$\phantom{-}31.1_{46.3}$&$2.3 \times 10^{-6}$&$\phantom{-}0.2$&$\phantom{-}35.4_{47.8}$&$2.0 \times 10^{-12}$&$\phantom{-}0.3$&\textcolor{gray}{$\phantom{-}20.9_{40.7}$}&$1.0 \times 10^{0}$&$-0.1$&\textcolor{gray}{$\phantom{-}19.9_{39.9}$}&$2.7 \times 10^{-1}$&$-0.1$&$\phantom{-}42.5_{49.4}$&$6.4 \times 10^{-84}$&$\phantom{-}0.4$&$\phantom{-}31.6_{46.5}$&$1.6 \times 10^{-17}$&$\phantom{-}0.2$&$\phantom{-}40.1_{49.0}$&$2.2 \times 10^{-66}$&$\phantom{-}0.4$&$\phantom{-}38.0_{48.5}$&$4.7 \times 10^{-51}$&$\phantom{-}0.3$\\

Reflections on Emotions&\textcolor{gray}{$\phantom{-}8.4_{27.8}$}&$7.7 \times 10^{-1}$&$-0.1$&\textcolor{gray}{$\phantom{-}8.5_{27.8}$}&$1.0 \times 10^{0}$&$-0.1$&\textcolor{gray}{$\phantom{-}9.4_{29.2}$}&$1.0 \times 10^{0}$&$\phantom{-}0.0$&\textcolor{gray}{$\phantom{-}9.8_{29.8}$}&$1.0 \times 10^{0}$&$\phantom{-}0.0$&$\phantom{-}15.3_{36.0}$&$2.0 \times 10^{-10}$&$\phantom{-}0.1$&$\phantom{-}21.8_{41.3}$&$1.9 \times 10^{-47}$&$\phantom{-}0.3$&$\phantom{-}16.4_{37.0}$&$6.3 \times 10^{-15}$&$\phantom{-}0.2$&$\phantom{-}15.5_{36.2}$&$3.1 \times 10^{-11}$&$\phantom{-}0.1$\\

Reflections on Values&\textcolor{gray}{$\phantom{-}4.5_{20.6}$}&$1.0 \times 10^{0}$&$\phantom{-}0.0$&\textcolor{gray}{$\phantom{-}5.1_{22.1}$}&$1.0 \times 10^{0}$&$\phantom{-}0.0$&\textcolor{gray}{$\phantom{-}3.1_{17.3}$}&$2.5 \times 10^{-1}$&$-0.1$&\textcolor{gray}{$\phantom{-}3.5_{18.3}$}&$1.0 \times 10^{0}$&$-0.1$&\textcolor{gray}{$\phantom{-}5.7_{23.3}$}&$8.1 \times 10^{-1}$&$\phantom{-}0.0$&$\phantom{-}6.4_{24.5}$&$1.9 \times 10^{-2}$&$\phantom{-}0.1$&\textcolor{gray}{$\phantom{-}5.3_{22.4}$}&$1.0 \times 10^{0}$&$\phantom{-}0.0$&\textcolor{gray}{$\phantom{-}6.2_{24.1}$}&$7.3 \times 10^{-2}$&$\phantom{-}0.1$\\

Reflections on Values&\textcolor{gray}{$\phantom{-}4.5_{20.6}$}&$6.1 \times 10^{-1}$&$-0.0$&\textcolor{gray}{$\phantom{-}5.1_{22.1}$}&$7.3 \times 10^{-1}$&$0.0$&$\phantom{-}3.1_{17.3}$&$1.9 \times 10^{-2}$&$-0.1$&\textcolor{gray}{$\phantom{-}3.5_{18.3}$}&$8.2 \times 10^{-2}$&$-0.1$&\textcolor{gray}{$\phantom{-}5.7_{23.3}$}&$6.2 \times 10^{-2}$&$0.0$&$\phantom{-}6.4_{24.5}$&$1.5 \times 10^{-3}$&$\phantom{-}0.1$&\textcolor{gray}{$\phantom{-}5.3_{22.4}$}&$3.5 \times 10^{-1}$&$0.0$&$\phantom{-}6.2_{24.1}$&$5.6 \times 10^{-3}$&$\phantom{-}0.1$\\

Reflections on Conflicts&$\phantom{-}3.0_{17.2}$&$2.0 \times 10^{-4}$&$-0.1$&$\phantom{-}2.0_{14.1}$&$7.1 \times 10^{-6}$&$-0.2$&$\phantom{-}3.2_{17.6}$&$4.7 \times 10^{-4}$&$-0.1$&$\phantom{-}3.6_{18.6}$&$7.5 \times 10^{-3}$&$-0.1$&\textcolor{gray}{$\phantom{-}6.7_{25.0}$}&$1.0 \times 10^{0}$&$\phantom{-}0.0$&\textcolor{gray}{$\phantom{-}5.4_{22.6}$}&$2.1 \times 10^{-1}$&$-0.1$&$\phantom{-}9.4_{29.3}$&$1.6 \times 10^{-5}$&$\phantom{-}0.1$&$\phantom{-}8.8_{28.3}$&$1.9 \times 10^{-3}$&$\phantom{-}0.1$\\

Reflections on Strengths&$\phantom{-}68.3_{46.6}$&$0.0 \times 10^{0}$&$\phantom{-}1.7$&$\phantom{-}54.6_{49.8}$&$4.0 \times 10^{-249}$&$\phantom{-}1.3$&$\phantom{-}35.4_{47.9}$&$9.5 \times 10^{-106}$&$\phantom{-}0.7$&$\phantom{-}37.7_{48.5}$&$2.4 \times 10^{-112}$&$\phantom{-}0.8$&$\phantom{-}31.1_{46.3}$&$3.0 \times 10^{-151}$&$\phantom{-}0.6$&$\phantom{-}27.3_{44.5}$&$4.8 \times 10^{-109}$&$\phantom{-}0.5$&$\phantom{-}20.9_{40.7}$&$9.2 \times 10^{-51}$&$\phantom{-}0.3$&$\phantom{-}18.1_{38.5}$&$6.1 \times 10^{-31}$&$\phantom{-}0.2$\\

Questions on Experiences&$\phantom{-}29.4_{45.6}$&$1.4 \times 10^{-17}$&$-0.3$&$\phantom{-}20.9_{40.7}$&$6.0 \times 10^{-36}$&$-0.5$&$\phantom{-}32.7_{46.9}$&$1.3 \times 10^{-10}$&$-0.2$&$\phantom{-}35.5_{47.9}$&$1.7 \times 10^{-5}$&$-0.2$&$\phantom{-}29.9_{45.8}$&$4.5 \times 10^{-42}$&$-0.3$&$\phantom{-}31.8_{46.6}$&$1.5 \times 10^{-31}$&$-0.3$&$\phantom{-}52.4_{49.9}$&$1.3 \times 10^{-14}$&$\phantom{-}0.2$&\textcolor{gray}{$\phantom{-}42.0_{49.4}$}&$7.3 \times 10^{-1}$&$\phantom{-}0.0$\\

Questions on Perspectives&\textcolor{gray}{$\phantom{-}3.2_{17.7}$}&$1.0 \times 10^{0}$&$\phantom{-}0.0$&\textcolor{gray}{$\phantom{-}2.0_{14.1}$}&$5.4 \times 10^{-1}$&$-0.1$&$\phantom{-}6.9_{25.3}$&$1.7 \times 10^{-5}$&$\phantom{-}0.2$&\textcolor{gray}{$\phantom{-}5.3_{22.4}$}&$1.5 \times 10^{-1}$&$\phantom{-}0.1$&$\phantom{-}6.6_{24.8}$&$2.7 \times 10^{-10}$&$\phantom{-}0.1$&\textcolor{gray}{$\phantom{-}2.5_{15.7}$}&$1.8 \times 10^{-1}$&$-0.1$&$\phantom{-}8.8_{28.3}$&$3.2 \times 10^{-25}$&$\phantom{-}0.2$&$\phantom{-}7.7_{26.7}$&$1.7 \times 10^{-17}$&$\phantom{-}0.2$\\

Questions on Emotions&$\phantom{-}1.3_{11.4}$&$8.1 \times 10^{-23}$&$-0.3$&$\phantom{-}2.0_{14.1}$&$7.1 \times 10^{-16}$&$-0.3$&$\phantom{-}3.7_{18.8}$&$1.6 \times 10^{-13}$&$-0.3$&$\phantom{-}5.4_{22.6}$&$2.8 \times 10^{-7}$&$-0.2$&$\phantom{-}2.4_{15.5}$&$3.6 \times 10^{-57}$&$-0.3$&$\phantom{-}1.7_{12.8}$&$5.4 \times 10^{-71}$&$-0.4$&\textcolor{gray}{$\phantom{-}10.3_{30.5}$}&$5.1 \times 10^{-1}$&$\phantom{-}0.0$&\textcolor{gray}{$\phantom{-}13.2_{33.9}$}&$3.8 \times 10^{-1}$&$\phantom{-}0.0$\\

Problem-Solving&$\phantom{-}44.2_{49.7}$&$4.2 \times 10^{-34}$&$\phantom{-}0.4$&$\phantom{-}55.2_{49.8}$&$1.9 \times 10^{-70}$&$\phantom{-}0.7$&$\phantom{-}54.1_{49.9}$&$4.9 \times 10^{-79}$&$\phantom{-}0.6$&$\phantom{-}34.7_{47.6}$&$1.3 \times 10^{-7}$&$\phantom{-}0.2$&$\phantom{-}61.5_{48.7}$&$3.2 \times 10^{-284}$&$\phantom{-}0.8$&$\phantom{-}47.9_{50.0}$&$7.5 \times 10^{-112}$&$\phantom{-}0.5$&$\phantom{-}43.5_{49.6}$&$8.9 \times 10^{-74}$&$\phantom{-}0.4$&$\phantom{-}34.3_{47.5}$&$2.8 \times 10^{-18}$&$\phantom{-}0.2$\\

Planning&$\phantom{-}66.8_{47.1}$&$4.7 \times 10^{-279}$&$\phantom{-}1.3$&$\phantom{-}70.2_{45.8}$&$2.4 \times 10^{-269}$&$\phantom{-}1.4$&$\phantom{-}64.1_{48.0}$&$7.9 \times 10^{-252}$&$\phantom{-}1.2$&$\phantom{-}44.8_{49.8}$&$1.8 \times 10^{-82}$&$\phantom{-}0.7$&$\phantom{-}54.4_{49.8}$&$0.0 \times 10^{0}$&$\phantom{-}0.9$&$\phantom{-}36.7_{48.2}$&$6.2 \times 10^{-104}$&$\phantom{-}0.5$&$\phantom{-}39.3_{48.8}$&$1.4 \times 10^{-130}$&$\phantom{-}0.5$&$\phantom{-}34.1_{47.4}$&$6.5 \times 10^{-81}$&$\phantom{-}0.4$\\

Normalizing&$\phantom{-}65.7_{47.5}$&$2.5 \times 10^{-218}$&$\phantom{-}1.1$&$\phantom{-}70.8_{45.5}$&$9.2 \times 10^{-227}$&$\phantom{-}1.2$&\textcolor{gray}{$\phantom{-}22.5_{41.8}$}&$1.0 \times 10^{0}$&$\phantom{-}0.0$&$\phantom{-}28.6_{45.2}$&$1.9 \times 10^{-7}$&$\phantom{-}0.2$&$\phantom{-}39.9_{49.0}$&$5.4 \times 10^{-95}$&$\phantom{-}0.4$&$\phantom{-}29.6_{45.6}$&$1.7 \times 10^{-23}$&$\phantom{-}0.2$&\textcolor{gray}{$\phantom{-}18.4_{38.8}$}&$1.5 \times 10^{-1}$&$-0.1$&$\phantom{-}24.0_{42.7}$&$5.2 \times 10^{-4}$&$\phantom{-}0.1$\\

Psychoeducation&$\phantom{-}20.5_{40.4}$&$7.3 \times 10^{-42}$&$\phantom{-}0.5$&$\phantom{-}25.3_{43.5}$&$1.0 \times 10^{-62}$&$\phantom{-}0.6$&$\phantom{-}26.9_{44.4}$&$6.0 \times 10^{-85}$&$\phantom{-}0.7$&$\phantom{-}18.1_{38.5}$&$4.2 \times 10^{-26}$&$\phantom{-}0.4$&$\phantom{-}15.2_{36.0}$&$7.7 \times 10^{-33}$&$\phantom{-}0.3$&$\phantom{-}24.4_{43.0}$&$1.5 \times 10^{-121}$&$\phantom{-}0.5$&\textcolor{gray}{$\phantom{-}8.7_{28.2}$}&$5.5 \times 10^{-1}$&$\phantom{-}0.0$&\textcolor{gray}{$\phantom{-}6.7_{25.0}$}&$1.0 \times 10^{0}$&$\phantom{-}0.0$\\

    \bottomrule
    \end{tabular}
    }
    \label{tab:dist_new_single_full_hl}
\end{table*}
\begin{table*}[t]
    \centering
    \small
    \caption{\showedits{Temporal Order (First Occurrence) of conversational behaviors during a conversation by LLM
therapists (GPT-4, GPT-3.5-turbo, Llama2-70b, Llama2-13b) in High-low quality dataset \cite{perez-rosas-etal-2019-makes}. Values in \textcolor{gray}{gray} are not statistically significantly different from human therapist average at $p = \frac{0.05}{m}$ using Two-sided Student’s t-test after Bonferroni correction, where $m$ is the number of t-test per one model ($m = 13$). $d$ is the cohen's $d$ when compared with human therapist average. For instance, GPT-4 responds with \textsc{Reflections on Needs} at 3.1 turns earlier than average human therapists responds.  (GPT-4: 3.1 turn, Average human therapists: 6.2 turn; $P = 4.3 \times 10^{-15}$, Cohen's $d = 0.7$, two-sided Student’s $t$-test).}}
    \resizebox{1\textwidth}{!}{%
    \def\arraystretch{1.15}
    \setlength{\tabcolsep}{2pt}
    \begin{tabular}{l|ccc|ccc|ccc|ccc|ccc|ccc}
         \toprule 
         \multirow{4}{*}{\textbf{Behavior}} & \multicolumn{12}{c}{\textbf{LLM Therapists (Single Response \& Full Conversation Simulations)}} & \multicolumn{3}{c}{\textbf{Human Therapists}} \\
         \cmidrule(lr){2-13} \cmidrule(lr){14-16}
         & \multicolumn{3}{c}{\textbf{GPT-4}} &\multicolumn{3}{c}{\textbf{GPT-3.5}}&\multicolumn{3}{c}{\textbf{Llama2-70b}}&\multicolumn{3}{c}{\textbf{Llama2-13b}}& \textbf{Average}&\textbf{High-quality}&\textbf{Low-quality}\\
& \multicolumn{3}{c}{\textbf{$(n = 4893)$}}& \multicolumn{3}{c}{\textbf{$(n = 4746)$}}& \multicolumn{3}{c}{\textbf{$(n = 4970)$}}& \multicolumn{3}{c}{\textbf{$(n = 4820)$}}&\textbf{$(n = 5446)$}&\textbf{$(n = 3907)$}&\textbf{$(n = 1539)$}\\
         \cmidrule(lr){2-4} \cmidrule(lr){5-7} \cmidrule(lr){8-10} \cmidrule(lr){11-13} \cmidrule(lr){14-14} \cmidrule(lr){15-15}  \cmidrule(lr){16-16} 
        & turn & $p$ & $d$ & turn & $p$ & $d$ & turn & $p$& $d$ & turn & $p$ & $d$ & turn & turn & turn \\
         \midrule
         Reflections on Needs&$\phantom{-}3.1_{2.9}$&$4.3 \times 10^{-15}$&$\phantom{-}0.7$&\textcolor{gray}{$\phantom{-}6.4_{5.4}$}&$1.0 \times 10^{0}$&$\phantom{-}0.0$&\textcolor{gray}{$\phantom{-}5.1_{5.0}$}&$2.5 \times 10^{-1}$&$\phantom{-}0.2$&\textcolor{gray}{$\phantom{-}5.3_{5.5}$}&$9.1 \times 10^{-1}$&$\phantom{-}0.2$&$\phantom{-}6.2_{5.5}$&$\phantom{-}4.4_{4.5}$&$\phantom{-}8.1_{6.2}$\\
Reflections on Emotions&\textcolor{gray}{$\phantom{-}6.1_{5.0}$}&$1.0 \times 10^{0}$&$\phantom{-}0.0$&\textcolor{gray}{$\phantom{-}5.7_{4.8}$}&$1.0 \times 10^{0}$&$\phantom{-}0.1$&\textcolor{gray}{$\phantom{-}7.2_{5.6}$}&$1.0 \times 10^{0}$&$-0.2$&\textcolor{gray}{$\phantom{-}6.1_{5.4}$}&$1.0 \times 10^{0}$&$\phantom{-}0.0$&$\phantom{-}6.3_{4.9}$&$\phantom{-}5.2_{4.8}$&$\phantom{-}7.4_{5.0}$\\
Reflections on Values&\textcolor{gray}{$\phantom{-}8.3_{5.1}$}&$3.6 \times 10^{-1}$&$\phantom{-}0.3$&\textcolor{gray}{$\phantom{-}9.1_{5.5}$}&$1.0 \times 10^{0}$&$\phantom{-}0.1$&\textcolor{gray}{$\phantom{-}10.3_{5.7}$}&$1.0 \times 10^{0}$&$-0.1$&\textcolor{gray}{$\phantom{-}10.1_{5.6}$}&$1.0 \times 10^{0}$&$\phantom{-}0.0$&$\phantom{-}9.9_{5.5}$&$\phantom{-}8.8_{5.3}$&$\phantom{-}11.0_{5.8}$\\

Reflections on Consequences&\textcolor{gray}{$\phantom{-}6.5_{4.2}$}&$1.0 \times 10^{0}$&$\phantom{-}0.1$&\textcolor{gray}{$\phantom{-}7.6_{4.9}$}&$1.0 \times 10^{0}$&$-0.1$&\textcolor{gray}{$\phantom{-}6.5_{4.5}$}&$1.0 \times 10^{0}$&$\phantom{-}0.1$&\textcolor{gray}{$\phantom{-}7.0_{5.4}$}&$1.0 \times 10^{0}$&$\phantom{-}0.0$&$\phantom{-}7.0_{5.0}$&$\phantom{-}6.2_{4.6}$&$\phantom{-}7.8_{5.4}$\\
Reflections on Conflicts&$\phantom{-}7.2_{4.6}$&$3.9 \times 10^{-4}$&$\phantom{-}0.6$&\textcolor{gray}{$\phantom{-}8.1_{4.6}$}&$1.3 \times 10^{-1}$&$\phantom{-}0.4$&$\phantom{-}7.8_{4.9}$&$7.9 \times 10^{-3}$&$\phantom{-}0.4$&$\phantom{-}7.8_{5.3}$&$9.1 \times 10^{-3}$&$\phantom{-}0.4$&$\phantom{-}10.0_{5.2}$&$\phantom{-}8.2_{4.7}$&$\phantom{-}11.8_{5.6}$\\
Reflections on Strengths&$\phantom{-}7.9_{5.4}$&$5.9 \times 10^{-4}$&$\phantom{-}0.4$&\textcolor{gray}{$\phantom{-}9.1_{6.3}$}&$1.0 \times 10^{0}$&$\phantom{-}0.2$&\textcolor{gray}{$\phantom{-}8.7_{6.3}$}&$2.1 \times 10^{-1}$&$\phantom{-}0.2$&\textcolor{gray}{$\phantom{-}9.4_{6.5}$}&$1.0 \times 10^{0}$&$\phantom{-}0.1$&$\phantom{-}10.2_{6.1}$&$\phantom{-}8.4_{6.0}$&$\phantom{-}11.9_{5.7}$\\
Questions on Experiences&\textcolor{gray}{$\phantom{-}2.7_{2.5}$}&$1.0 \times 10^{0}$&$\phantom{-}0.1$&\textcolor{gray}{$\phantom{-}3.7_{4.3}$}&$1.5 \times 10^{-1}$&$-0.2$&\textcolor{gray}{$\phantom{-}2.9_{3.0}$}&$1.0 \times 10^{0}$&$\phantom{-}0.0$&$\phantom{-}3.8_{4.0}$&$2.8 \times 10^{-2}$&$-0.3$&$\phantom{-}2.9_{2.5}$&$\phantom{-}2.2_{2.1}$&$\phantom{-}3.6_{2.9}$\\
Questions on Perspectives&$\phantom{-}8.5_{4.7}$&$3.1 \times 10^{-7}$&$\phantom{-}0.8$&$\phantom{-}9.7_{5.3}$&$1.6 \times 10^{-2}$&$\phantom{-}0.6$&$\phantom{-}8.7_{5.7}$&$4.2 \times 10^{-6}$&$\phantom{-}0.7$&$\phantom{-}8.7_{5.4}$&$2.8 \times 10^{-6}$&$\phantom{-}0.7$&$\phantom{-}12.6_{5.1}$&$\phantom{-}10.0_{4.7}$&$\phantom{-}15.3_{4.2}$\\
Questions on Emotions&\textcolor{gray}{$\phantom{-}6.2_{4.9}$}&$1.0 \times 10^{0}$&$\phantom{-}0.1$&\textcolor{gray}{$\phantom{-}6.0_{5.9}$}&$1.0 \times 10^{0}$&$\phantom{-}0.2$&\textcolor{gray}{$\phantom{-}5.3_{5.6}$}&$1.6 \times 10^{-1}$&$\phantom{-}0.3$&\textcolor{gray}{$\phantom{-}5.4_{5.5}$}&$1.6 \times 10^{-1}$&$\phantom{-}0.3$&$\phantom{-}7.0_{6.4}$&$\phantom{-}6.7_{6.3}$&$\phantom{-}7.3_{6.5}$\\
Problem-Solving&\textcolor{gray}{$\phantom{-}7.8_{5.2}$}&$1.0 \times 10^{0}$&$\phantom{-}0.1$&\textcolor{gray}{$\phantom{-}7.1_{5.1}$}&$1.5 \times 10^{-1}$&$\phantom{-}0.2$&\textcolor{gray}{$\phantom{-}7.5_{4.8}$}&$7.1 \times 10^{-1}$&$\phantom{-}0.2$&\textcolor{gray}{$\phantom{-}9.0_{5.8}$}&$1.0 \times 10^{0}$&$-0.1$&$\phantom{-}8.3_{5.4}$&$\phantom{-}8.7_{5.6}$&$\phantom{-}8.0_{5.1}$\\
Planning&$\phantom{-}9.3_{5.4}$&$4.2 \times 10^{-5}$&$\phantom{-}0.4$&$\phantom{-}10.2_{6.1}$&$3.8 \times 10^{-2}$&$\phantom{-}0.3$&$\phantom{-}8.7_{5.3}$&$7.2 \times 10^{-8}$&$\phantom{-}0.5$&\textcolor{gray}{$\phantom{-}10.7_{6.1}$}&$6.0 \times 10^{-1}$&$\phantom{-}0.2$&$\phantom{-}11.9_{6.5}$&$\phantom{-}11.5_{6.8}$&$\phantom{-}12.4_{6.0}$\\
Normalizing&\textcolor{gray}{$\phantom{-}7.9_{6.0}$}&$9.0 \times 10^{-1}$&$-0.2$&\textcolor{gray}{$\phantom{-}7.0_{6.0}$}&$1.0 \times 10^{0}$&$\phantom{-}0.0$&$\phantom{-}9.3_{6.8}$&$2.8 \times 10^{-3}$&$-0.4$&$\phantom{-}9.0_{6.3}$&$5.3 \times 10^{-3}$&$-0.3$&$\phantom{-}6.9_{6.2}$&$\phantom{-}6.2_{6.3}$&$\phantom{-}7.7_{6.0}$\\
Psychoeducation&$\phantom{-}11.1_{5.5}$&$2.5 \times 10^{-2}$&$-0.4$&\textcolor{gray}{$\phantom{-}8.2_{5.7}$}&$1.0 \times 10^{0}$&$\phantom{-}0.1$&$\phantom{-}11.8_{5.7}$&$2.3 \times 10^{-4}$&$-0.5$&$\phantom{-}11.5_{5.8}$&$5.6 \times 10^{-3}$&$-0.5$&$\phantom{-}8.9_{5.5}$&$\phantom{-}8.5_{5.6}$&$\phantom{-}9.2_{5.5}$\\

    \bottomrule
    \end{tabular}
    }
    \label{tab:order_new_hl}
\end{table*}
\begin{table*}[t]
    \centering
    \small
    \caption{\showedits{Temporal Order (First Occurrence) of conversational behaviors during a conversation by LLM
therapists (GPT-4, GPT-3.5-turbo, Llama2-70b, Llama2-13b) in HOPE dataset \cite{malhotra2022speaker} for two different types of simulations separately (Single response and Full conversation). Values in \textcolor{gray}{gray} are not statistically significantly different from human therapist average at $p = \frac{0.05}{m}$ using Two-sided Student’s t-test after Bonferroni correction, where $m$ is the number of t-test per one model ($m = 13$). $d$ is the cohen's $d$ when compared with human therapist average. For instance, GPT-4 with single response simulation responds with \textsc{Reflections on Needs} at 4.3 turn which is 1.9 turns earlier than average human therapists. Similarly, GPT-4 with full conversation simulation responds at 2.0 turn which is 4.2 turns earlier than average human therapists.  (Single: 4.3 turn, Average human therapists: 6.2 turn from Table \ref{tab:order_new_hl}; $P = 1.1 \times 10^{-3}$, Cohen's $d = 0.4$, two-sided Student’s $t$-test) (Full: 2.0 turn, Average human therapists: 6.2 turn from Table \ref{tab:order_new_hl}; $P = 9.2 \times 10^{-16}$, Cohen's $d = 0.7$).}}
    \resizebox{1.05\textwidth}{!}{%
    \def\arraystretch{1.15}
    \setlength{\tabcolsep}{2pt}
    \begin{tabular}{l|ccc|ccc|ccc|ccc|ccc|ccc|ccc|ccc|ccc|ccc}
         \toprule 
         \multirow{3}{*}{\textbf{Behavior}} & \multicolumn{12}{c}{\textbf{LLM Therapists (Single Response Simulations)}} & \multicolumn{12}{c}{\textbf{LLM Therapists (Full Conversation Simulations)}} \\
         \cmidrule(lr){2-13}  \cmidrule(lr){14-25} 
         & \multicolumn{3}{c}{\textbf{GPT-4}} &\multicolumn{3}{c}{\textbf{GPT-3.5}}&\multicolumn{3}{c}{\textbf{Llama2-70b}}&\multicolumn{3}{c}{\textbf{Llama2-13b}}& \multicolumn{3}{c}{\textbf{GPT-4}} &\multicolumn{3}{c}{\textbf{GPT-3.5}}&\multicolumn{3}{c}{\textbf{Llama2-70b}}&\multicolumn{3}{c}{\textbf{Llama2-13b}}\\
& \multicolumn{3}{c}{\textbf{$(n = 3837)$}}& \multicolumn{3}{c}{\textbf{$(n = 3907)$}}& \multicolumn{3}{c}{\textbf{$(n = 3906)$}}& \multicolumn{3}{c}{\textbf{$(n = 3895)$}}& \multicolumn{3}{c}{\textbf{$(n = 1056)$}}& \multicolumn{3}{c}{\textbf{$(n = 839)$}}& \multicolumn{3}{c}{\textbf{$(n = 1064)$}}& \multicolumn{3}{c}{\textbf{$(n = 925)$}}\\
         \cmidrule(lr){2-4} \cmidrule(lr){5-7} \cmidrule(lr){8-10} \cmidrule(lr){11-13} \cmidrule(lr){14-16} \cmidrule(lr){17-19} \cmidrule(lr){20-22}  \cmidrule(lr){23-25}
        & turn & $p$ & $d$ & turn & $p$ & $d$ & turn & $p$& $d$ & turn & $p$ & $d$ & turn & $p$& $d$ &  turn & $p$ & $d$ & turn & $p$ & $d$ & turn & $p$& $d$ \\
         \midrule
Reflections on Needs&$\phantom{-}4.3_{2.6}$&$1.1 \times 10^{-3}$&$\phantom{-}0.4$&$\phantom{-}8.2_{6.2}$&$1.7 \times 10^{-2}$&$-0.3$&$\phantom{-}8.3_{5.1}$&$1.4 \times 10^{-2}$&$-0.4$&$\phantom{-}9.6_{5.9}$&$6.5 \times 10^{-6}$&$-0.6$&$\phantom{-}2.0_{2.8}$&$9.2 \times 10^{-16}$&$\phantom{-}0.9$&$\phantom{-}4.7_{3.9}$&$4.5 \times 10^{-2}$&$\phantom{-}0.3$&$\phantom{-}2.9_{3.4}$&$1.0 \times 10^{-9}$&$\phantom{-}0.7$&$\phantom{-}2.4_{2.5}$&$6.7 \times 10^{-14}$&$\phantom{-}0.8$\\
Reflections on Emotions&$\phantom{-}8.8_{5.1}$&$9.3 \times 10^{-3}$&$-0.5$&\textcolor{gray}{$\phantom{-}7.8_{5.3}$}&$8.1 \times 10^{-1}$&$-0.3$&$\phantom{-}9.3_{5.2}$&$3.1 \times 10^{-4}$&$-0.6$&$\phantom{-}8.8_{5.3}$&$8.7 \times 10^{-3}$&$-0.5$&$\phantom{-}4.6_{4.2}$&$3.3 \times 10^{-2}$&$\phantom{-}0.4$&\textcolor{gray}{$\phantom{-}4.7_{4.2}$}&$7.1 \times 10^{-2}$&$\phantom{-}0.3$&\textcolor{gray}{$\phantom{-}6.0_{5.4}$}&$1.0 \times 10^{0}$&$\phantom{-}0.1$&\textcolor{gray}{$\phantom{-}4.8_{5.0}$}&$1.4 \times 10^{-1}$&$\phantom{-}0.3$\\
Reflections on Values&\textcolor{gray}{$\phantom{-}8.2_{4.1}$}&$1.0 \times 10^{0}$&$\phantom{-}0.3$&\textcolor{gray}{$\phantom{-}9.5_{6.7}$}&$1.0 \times 10^{0}$&$\phantom{-}0.1$&$\phantom{-}13.6_{5.1}$&$2.7 \times 10^{-2}$&$-0.7$&\textcolor{gray}{$\phantom{-}13.1_{5.3}$}&$1.6 \times 10^{-1}$&$-0.6$&\textcolor{gray}{$\phantom{-}8.4_{5.6}$}&$9.3 \times 10^{-1}$&$\phantom{-}0.3$&\textcolor{gray}{$\phantom{-}8.9_{5.0}$}&$1.0 \times 10^{0}$&$\phantom{-}0.2$&\textcolor{gray}{$\phantom{-}9.2_{5.5}$}&$1.0 \times 10^{0}$&$\phantom{-}0.1$&\textcolor{gray}{$\phantom{-}9.3_{5.5}$}&$1.0 \times 10^{0}$&$\phantom{-}0.1$\\
Reflections on Consequences&\textcolor{gray}{$\phantom{-}7.5_{3.4}$}&$1.0 \times 10^{0}$&$-0.1$&$\phantom{-}9.1_{4.7}$&$3.7 \times 10^{-2}$&$-0.4$&\textcolor{gray}{$\phantom{-}9.0_{4.6}$}&$1.1 \times 10^{-1}$&$-0.4$&$\phantom{-}11.3_{5.3}$&$6.5 \times 10^{-7}$&$-0.8$&\textcolor{gray}{$\phantom{-}6.0_{4.6}$}&$8.3 \times 10^{-1}$&$\phantom{-}0.2$&\textcolor{gray}{$\phantom{-}6.7_{4.8}$}&$1.0 \times 10^{0}$&$\phantom{-}0.1$&\textcolor{gray}{$\phantom{-}5.7_{4.2}$}&$1.0 \times 10^{-1}$&$\phantom{-}0.3$&$\phantom{-}5.5_{4.6}$&$4.5 \times 10^{-2}$&$\phantom{-}0.3$\\
Reflections on Conflicts&\textcolor{gray}{$\phantom{-}7.1_{3.5}$}&$1.8 \times 10^{-1}$&$\phantom{-}0.6$&\textcolor{gray}{$\phantom{-}6.9_{2.3}$}&$3.4 \times 10^{-1}$&$\phantom{-}0.6$&\textcolor{gray}{$\phantom{-}8.0_{3.2}$}&$1.0 \times 10^{0}$&$\phantom{-}0.4$&\textcolor{gray}{$\phantom{-}11.0_{5.5}$}&$1.0 \times 10^{0}$&$-0.2$&$\phantom{-}7.2_{4.9}$&$2.7 \times 10^{-3}$&$\phantom{-}0.6$&\textcolor{gray}{$\phantom{-}8.4_{5.1}$}&$6.7 \times 10^{-1}$&$\phantom{-}0.3$&$\phantom{-}7.8_{5.3}$&$1.7 \times 10^{-2}$&$\phantom{-}0.4$&$\phantom{-}7.1_{5.0}$&$2.1 \times 10^{-4}$&$\phantom{-}0.6$\\
Reflections on Strengths&\textcolor{gray}{$\phantom{-}9.0_{5.2}$}&$1.0 \times 10^{0}$&$\phantom{-}0.2$&\textcolor{gray}{$\phantom{-}11.4_{6.1}$}&$1.0 \times 10^{0}$&$-0.2$&\textcolor{gray}{$\phantom{-}11.2_{6.1}$}&$1.0 \times 10^{0}$&$-0.2$&\textcolor{gray}{$\phantom{-}12.2_{6.6}$}&$1.1 \times 10^{-1}$&$-0.3$&$\phantom{-}6.8_{5.4}$&$5.4 \times 10^{-6}$&$\phantom{-}0.6$&$\phantom{-}7.0_{5.7}$&$7.2 \times 10^{-5}$&$\phantom{-}0.5$&$\phantom{-}6.1_{5.6}$&$1.1 \times 10^{-7}$&$\phantom{-}0.7$&$\phantom{-}7.1_{5.6}$&$1.1 \times 10^{-4}$&$\phantom{-}0.5$\\
Questions on Experiences&$\phantom{-}3.8_{2.1}$&$5.7 \times 10^{-3}$&$-0.4$&$\phantom{-}5.9_{5.2}$&$8.3 \times 10^{-12}$&$-0.9$&$\phantom{-}4.5_{3.5}$&$5.4 \times 10^{-6}$&$-0.5$&$\phantom{-}6.1_{4.5}$&$6.0 \times 10^{-17}$&$-0.9$&$\phantom{-}1.7_{2.3}$&$3.1 \times 10^{-5}$&$\phantom{-}0.5$&$\phantom{-}2.0_{2.5}$&$1.6 \times 10^{-2}$&$\phantom{-}0.3$&$\phantom{-}1.4_{1.2}$&$4.5 \times 10^{-10}$&$\phantom{-}0.7$&$\phantom{-}1.6_{1.3}$&$3.5 \times 10^{-8}$&$\phantom{-}0.6$\\
Questions on Perspectives&$\phantom{-}8.0_{5.1}$&$3.6 \times 10^{-3}$&$\phantom{-}0.9$&$\phantom{-}6.3_{2.0}$&$4.1 \times 10^{-2}$&$\phantom{-}1.3$&\textcolor{gray}{$\phantom{-}11.2_{4.5}$}&$1.0 \times 10^{0}$&$\phantom{-}0.3$&\textcolor{gray}{$\phantom{-}10.4_{4.7}$}&$4.4 \times 10^{-1}$&$\phantom{-}0.5$&$\phantom{-}8.6_{4.6}$&$3.9 \times 10^{-6}$&$\phantom{-}0.8$&\textcolor{gray}{$\phantom{-}10.1_{5.4}$}&$1.0 \times 10^{-1}$&$\phantom{-}0.5$&$\phantom{-}7.6_{5.8}$&$3.4 \times 10^{-8}$&$\phantom{-}0.9$&$\phantom{-}8.2_{5.5}$&$7.4 \times 10^{-7}$&$\phantom{-}0.8$\\
Questions on Emotions&\textcolor{gray}{$\phantom{-}4.9_{3.1}$}&$1.0 \times 10^{0}$&$\phantom{-}0.3$&\textcolor{gray}{$\phantom{-}7.0_{5.9}$}&$1.0 \times 10^{0}$&$\phantom{-}0.0$&\textcolor{gray}{$\phantom{-}6.9_{6.0}$}&$1.0 \times 10^{0}$&$\phantom{-}0.0$&\textcolor{gray}{$\phantom{-}5.7_{4.2}$}&$1.0 \times 10^{0}$&$\phantom{-}0.2$&\textcolor{gray}{$\phantom{-}6.4_{5.1}$}&$1.0 \times 10^{0}$&$\phantom{-}0.1$&\textcolor{gray}{$\phantom{-}5.8_{6.0}$}&$1.0 \times 10^{0}$&$\phantom{-}0.2$&$\phantom{-}4.9_{5.5}$&$4.6 \times 10^{-2}$&$\phantom{-}0.4$&\textcolor{gray}{$\phantom{-}5.3_{5.9}$}&$2.3 \times 10^{-1}$&$\phantom{-}0.3$\\
Problem-Solving&$\phantom{-}10.3_{4.9}$&$3.9 \times 10^{-3}$&$-0.4$&\textcolor{gray}{$\phantom{-}8.9_{4.6}$}&$1.0 \times 10^{0}$&$-0.1$&\textcolor{gray}{$\phantom{-}9.2_{4.5}$}&$1.0 \times 10^{0}$&$-0.2$&$\phantom{-}11.9_{5.8}$&$8.8 \times 10^{-8}$&$-0.7$&$\phantom{-}5.4_{4.3}$&$1.0 \times 10^{-6}$&$\phantom{-}0.6$&$\phantom{-}5.6_{5.0}$&$1.7 \times 10^{-5}$&$\phantom{-}0.5$&$\phantom{-}5.7_{4.5}$&$2.7 \times 10^{-5}$&$\phantom{-}0.5$&$\phantom{-}6.5_{4.4}$&$7.1 \times 10^{-3}$&$\phantom{-}0.4$\\
Planning&\textcolor{gray}{$\phantom{-}11.4_{5.1}$}&$1.0 \times 10^{0}$&$\phantom{-}0.1$&\textcolor{gray}{$\phantom{-}12.2_{5.4}$}&$1.0 \times 10^{0}$&$\phantom{-}0.0$&$\phantom{-}9.8_{5.1}$&$1.6 \times 10^{-2}$&$\phantom{-}0.4$&\textcolor{gray}{$\phantom{-}12.7_{6.5}$}&$1.0 \times 10^{0}$&$-0.1$&$\phantom{-}7.5_{5.0}$&$5.0 \times 10^{-10}$&$\phantom{-}0.7$&$\phantom{-}8.4_{6.2}$&$1.2 \times 10^{-5}$&$\phantom{-}0.5$&$\phantom{-}7.7_{5.3}$&$5.0 \times 10^{-9}$&$\phantom{-}0.7$&$\phantom{-}9.0_{5.2}$&$1.4 \times 10^{-4}$&$\phantom{-}0.5$\\
Normalizing&$\phantom{-}10.3_{6.2}$&$1.6 \times 10^{-5}$&$-0.5$&$\phantom{-}9.6_{6.1}$&$2.4 \times 10^{-3}$&$-0.4$&$\phantom{-}12.4_{6.7}$&$2.0 \times 10^{-10}$&$-0.9$&$\phantom{-}11.2_{6.3}$&$1.3 \times 10^{-6}$&$-0.7$&\textcolor{gray}{$\phantom{-}5.7_{4.9}$}&$5.5 \times 10^{-1}$&$\phantom{-}0.2$&$\phantom{-}4.7_{5.0}$&$6.1 \times 10^{-3}$&$\phantom{-}0.4$&\textcolor{gray}{$\phantom{-}6.7_{5.7}$}&$1.0 \times 10^{0}$&$\phantom{-}0.0$&\textcolor{gray}{$\phantom{-}7.6_{6.0}$}&$1.0 \times 10^{0}$&$-0.1$\\
Psychoeducation&$\phantom{-}12.0_{5.1}$&$7.3 \times 10^{-3}$&$-0.6$&$\phantom{-}12.2_{5.4}$&$3.8 \times 10^{-3}$&$-0.6$&$\phantom{-}13.2_{5.3}$&$3.3 \times 10^{-7}$&$-0.8$&$\phantom{-}13.5_{5.9}$&$1.1 \times 10^{-5}$&$-0.8$&\textcolor{gray}{$\phantom{-}10.4_{5.7}$}&$7.0 \times 10^{-1}$&$-0.3$&$\phantom{-}6.2_{4.7}$&$1.7 \times 10^{-3}$&$\phantom{-}0.5$&\textcolor{gray}{$\phantom{-}10.1_{5.8}$}&$1.0 \times 10^{0}$&$-0.2$&\textcolor{gray}{$\phantom{-}9.6_{5.1}$}&$1.0 \times 10^{0}$&$-0.1$\\

    \bottomrule
    \end{tabular}
    }
    \label{tab:order_new_hope_single_and_full}
\end{table*}
\begin{table*}[t]
    \centering
    \small
    \caption{\showedits{Adaptability (frequency of conversational behaviors exhibited by therapists \textit{in response to specific client behaviors}) by LLM
therapists (GPT-4, GPT-3.5-turbo, Llama2-70b, Llama2-13b) in High-low quality dataset \cite{perez-rosas-etal-2019-makes}. Values in \textcolor{gray}{gray} are not statistically significantly different from human therapist average at $p = \frac{0.05}{m}$ using Two-sided Student’s t-test after Bonferroni correction, where $m$ is the number of t-test per one model ($m = 13 \times 6 = 78$). $d$ is the cohen's d when compared with human therapist average. For instance, GPT-4 exhibits \textsc{Reflections on Consequences} by 19.9\%  and human therapists responds by 27.3\% on average when client expresses \textsc{Changing unhealthy behavior}. ($P = 4.1 \times 10^{-6}$, Cohen's $d = \num{-0.2}$, two-sided Student’s $t$-test).}}
    \resizebox{1\textwidth}{!}{%
    \def\arraystretch{1.15}
    \setlength{\tabcolsep}{2pt}
    \begin{tabular}{l|ccc|ccc|ccc|ccc|ccc|ccc}
         \toprule 
         \multirow{3}{*}{\textbf{Therapist Behavior}} & \multicolumn{12}{c}{\textbf{LLM Therapists (Single Response \& Full Conversation Simulations)}} & \multicolumn{3}{c}{\textbf{Human Therapists}} \\
         \cmidrule(lr){2-13} \cmidrule(lr){14-16}
         & \multicolumn{3}{c}{\textbf{GPT-4}} &\multicolumn{3}{c}{\textbf{GPT-3.5}}&\multicolumn{3}{c}{\textbf{Llama2-70b}}&\multicolumn{3}{c}{\textbf{Llama2-13b}}& \textbf{Average}&\textbf{High-quality}&\textbf{Low-quality}\\
         
         \cmidrule(lr){2-4} \cmidrule(lr){5-7} \cmidrule(lr){8-10} \cmidrule(lr){11-13} \cmidrule(lr){14-14} \cmidrule(lr){15-15}  \cmidrule(lr){16-16} 

        & freq. & $p$ & $d$ & freq. & $p$ & $d$ & freq. & $p$& $d$ & freq. & $p$ & $d$ & freq. & freq. & freq. \\
        \midrule
          \multicolumn{16}{c}{\textbf{Client behavior: Changing unhealthy behavior}} \\
     \midrule
$n$& \multicolumn{3}{c}{\textbf{$2284$}}& \multicolumn{3}{c}{\textbf{$2148$}}& \multicolumn{3}{c}{\textbf{$2211$}}& \multicolumn{3}{c}{\textbf{$2082$}}&\textbf{$1553$}&\textbf{$1198$}&\textbf{$355$}\\
         \midrule
Reflections on Needs&\textcolor{gray}{$\phantom{-}29.3_{45.5}$}&$1.0 \times 10^{0}$&$\phantom{-}0.0$&\textcolor{gray}{$\phantom{-}29.1_{45.4}$}&$1.0 \times 10^{0}$&$\phantom{-}0.0$&\textcolor{gray}{$\phantom{-}26.9_{44.4}$}&$1.0 \times 10^{0}$&$\phantom{-}0.0$&\textcolor{gray}{$\phantom{-}27.8_{44.8}$}&$1.0 \times 10^{0}$&$\phantom{-}0.0$&$\phantom{-}27.7_{45.5}$&$\phantom{-}30.6_{46.1}$&$\phantom{-}24.8_{43.2}$\\
Reflections on Emotions&\textcolor{gray}{$\phantom{-}8.2_{27.4}$}&$1.0 \times 10^{0}$&$-0.1$&\textcolor{gray}{$\phantom{-}11.8_{32.2}$}&$1.0 \times 10^{0}$&$\phantom{-}0.0$&\textcolor{gray}{$\phantom{-}8.7_{28.2}$}&$1.0 \times 10^{0}$&$-0.1$&\textcolor{gray}{$\phantom{-}8.6_{28.0}$}&$1.0 \times 10^{0}$&$-0.1$&$\phantom{-}10.4_{33.3}$&$\phantom{-}14.6_{35.3}$&$\phantom{-}6.2_{24.1}$\\
Reflections on Values&\textcolor{gray}{$\phantom{-}4.1_{19.9}$}&$1.0 \times 10^{0}$&$\phantom{-}0.0$&\textcolor{gray}{$\phantom{-}4.9_{21.7}$}&$1.0 \times 10^{0}$&$\phantom{-}0.0$&\textcolor{gray}{$\phantom{-}3.6_{18.6}$}&$1.0 \times 10^{0}$&$-0.1$&\textcolor{gray}{$\phantom{-}4.5_{20.8}$}&$1.0 \times 10^{0}$&$\phantom{-}0.0$&$\phantom{-}4.8_{23.7}$&$\phantom{-}7.0_{25.5}$&$\phantom{-}2.5_{15.7}$\\

Reflections on Consequences&$\phantom{-}19.9_{39.9}$&$4.1 \times 10^{-6}$&$-0.2$&$\phantom{-}12.3_{32.9}$&$4.9 \times 10^{-31}$&$-0.4$&$\phantom{-}16.6_{37.2}$&$3.1 \times 10^{-14}$&$-0.3$&$\phantom{-}19.6_{39.7}$&$1.8 \times 10^{-6}$&$-0.2$&$\phantom{-}27.3_{42.3}$&$\phantom{-}19.9_{39.9}$&$\phantom{-}34.6_{47.7}$\\
Reflections on Conflicts&$\phantom{-}4.1_{19.9}$&$1.3 \times 10^{-2}$&$-0.1$&$\phantom{-}3.3_{17.9}$&$5.2 \times 10^{-5}$&$-0.2$&\textcolor{gray}{$\phantom{-}4.4_{20.5}$}&$6.7 \times 10^{-2}$&$-0.1$&\textcolor{gray}{$\phantom{-}6.1_{24.0}$}&$1.0 \times 10^{0}$&$\phantom{-}0.0$&$\phantom{-}6.9_{26.6}$&$\phantom{-}8.3_{27.5}$&$\phantom{-}5.6_{23.1}$\\
Reflections on Strengths&$\phantom{-}44.4_{49.7}$&$1.0 \times 10^{-121}$&$\phantom{-}0.8$&$\phantom{-}35.6_{47.9}$&$5.5 \times 10^{-73}$&$\phantom{-}0.6$&$\phantom{-}25.9_{43.8}$&$2.7 \times 10^{-34}$&$\phantom{-}0.4$&$\phantom{-}22.6_{41.8}$&$4.7 \times 10^{-23}$&$\phantom{-}0.3$&$\phantom{-}9.3_{32.4}$&$\phantom{-}14.1_{34.8}$&$\phantom{-}4.5_{20.8}$\\
Questions on Experiences&$\phantom{-}15.2_{35.9}$&$2.6 \times 10^{-47}$&$-0.5$&$\phantom{-}16.3_{36.9}$&$5.1 \times 10^{-40}$&$-0.5$&$\phantom{-}27.7_{44.8}$&$2.0 \times 10^{-5}$&$-0.2$&$\phantom{-}24.3_{42.9}$&$8.5 \times 10^{-12}$&$-0.3$&$\phantom{-}35.6_{48.6}$&$\phantom{-}40.2_{49.1}$&$\phantom{-}31.0_{46.3}$\\

Questions on Perspectives&\textcolor{gray}{$\phantom{-}2.8_{16.4}$}&$1.0 \times 10^{0}$&$\phantom{-}0.0$&$\phantom{-}0.8_{9.1}$&$3.2 \times 10^{-6}$&$-0.2$&\textcolor{gray}{$\phantom{-}3.9_{19.3}$}&$1.0 \times 10^{0}$&$\phantom{-}0.0$&\textcolor{gray}{$\phantom{-}4.8_{21.4}$}&$1.0 \times 10^{0}$&$\phantom{-}0.1$&$\phantom{-}3.5_{20.2}$&$\phantom{-}4.8_{21.5}$&$\phantom{-}2.3_{14.9}$\\
Questions on Emotions&$\phantom{-}0.7_{8.1}$&$2.6 \times 10^{-37}$&$-0.4$&$\phantom{-}0.3_{5.3}$&$2.9 \times 10^{-40}$&$-0.5$&$\phantom{-}3.2_{17.5}$&$4.6 \times 10^{-14}$&$-0.3$&$\phantom{-}4.2_{20.0}$&$1.3 \times 10^{-8}$&$-0.2$&$\phantom{-}9.6_{30.7}$&$\phantom{-}11.3_{31.6}$&$\phantom{-}7.9_{27.0}$\\

Problem-Solving&$\phantom{-}40.7_{49.1}$&$8.0 \times 10^{-14}$&$\phantom{-}0.3$&$\phantom{-}38.1_{48.6}$&$1.9 \times 10^{-8}$&$\phantom{-}0.2$&$\phantom{-}37.3_{48.4}$&$4.5 \times 10^{-7}$&$\phantom{-}0.2$&\textcolor{gray}{$\phantom{-}29.4_{45.6}$}&$1.0 \times 10^{0}$&$\phantom{-}0.0$&$\phantom{-}28.4_{42.7}$&$\phantom{-}20.4_{40.3}$&$\phantom{-}36.3_{48.2}$\\

Planning&$\phantom{-}32.4_{46.8}$&$1.2 \times 10^{-39}$&$\phantom{-}0.4$&$\phantom{-}32.4_{46.8}$&$2.4 \times 10^{-39}$&$\phantom{-}0.5$&$\phantom{-}32.4_{46.8}$&$1.4 \times 10^{-39}$&$\phantom{-}0.5$&$\phantom{-}25.9_{43.8}$&$2.6 \times 10^{-18}$&$\phantom{-}0.3$&$\phantom{-}13.7_{32.6}$&$\phantom{-}10.7_{30.9}$&$\phantom{-}16.6_{37.3}$\\
Normalizing&$\phantom{-}28.0_{44.9}$&$1.1 \times 10^{-23}$&$\phantom{-}0.3$&$\phantom{-}23.4_{42.3}$&$2.7 \times 10^{-11}$&$\phantom{-}0.2$&$\phantom{-}9.9_{29.9}$&$5.6 \times 10^{-3}$&$-0.1$&\textcolor{gray}{$\phantom{-}13.8_{34.5}$}&$1.0 \times 10^{0}$&$\phantom{-}0.0$&$\phantom{-}14.0_{32.6}$&$\phantom{-}10.5_{30.7}$&$\phantom{-}17.5_{38.0}$\\
Psychoeducation&\textcolor{gray}{$\phantom{-}6.1_{24.0}$}&$1.0 \times 10^{0}$&$\phantom{-}0.0$&$\phantom{-}12.0_{32.5}$&$1.7 \times 10^{-10}$&$\phantom{-}0.2$&$\phantom{-}8.5_{27.8}$&$1.9 \times 10^{-2}$&$\phantom{-}0.1$&\textcolor{gray}{$\phantom{-}5.8_{23.4}$}&$1.0 \times 10^{0}$&$\phantom{-}0.0$&$\phantom{-}5.4_{20.5}$&$\phantom{-}3.5_{18.4}$&$\phantom{-}7.3_{26.1}$\\
\midrule
        \multicolumn{16}{c}{\textbf{Client behavior: Sustaining unhealthy behavior}} \\
        \midrule
$n$ & \multicolumn{3}{c}{\textbf{$436$}} & \multicolumn{3}{c}{\textbf{$419$}} & \multicolumn{3}{c}{\textbf{$553$}} & \multicolumn{3}{c}{\textbf{$445$}}&\textbf{$389$}&\textbf{$275$}&\textbf{$114$}\\
\midrule
Reflections on Needs&\textcolor{gray}{$\phantom{-}36.5_{48.2}$}&$5.1 \times 10^{-1}$&$\phantom{-}0.2$&\textcolor{gray}{$\phantom{-}35.8_{48.0}$}&$9.7 \times 10^{-1}$&$\phantom{-}0.2$&\textcolor{gray}{$\phantom{-}31.5_{46.5}$}&$1.0 \times 10^{0}$&$\phantom{-}0.1$&\textcolor{gray}{$\phantom{-}30.1_{45.9}$}&$1.0 \times 10^{0}$&$\phantom{-}0.1$&$\phantom{-}27.6_{45.3}$&$\phantom{-}30.5_{46.1}$&$\phantom{-}24.6_{43.2}$\\
Reflections on Emotions&\textcolor{gray}{$\phantom{-}9.2_{28.9}$}&$1.0 \times 10^{0}$&$-0.1$&\textcolor{gray}{$\phantom{-}14.8_{35.5}$}&$1.0 \times 10^{0}$&$\phantom{-}0.1$&\textcolor{gray}{$\phantom{-}11.0_{31.4}$}&$1.0 \times 10^{0}$&$\phantom{-}0.0$&\textcolor{gray}{$\phantom{-}14.2_{34.9}$}&$1.0 \times 10^{0}$&$\phantom{-}0.1$&$\phantom{-}11.7_{33.2}$&$\phantom{-}13.8_{34.6}$&$\phantom{-}9.6_{29.7}$\\
Reflections on Values&\textcolor{gray}{$\phantom{-}4.1_{19.9}$}&$1.0 \times 10^{0}$&$-0.1$&\textcolor{gray}{$\phantom{-}3.6_{18.6}$}&$1.0 \times 10^{0}$&$-0.1$&\textcolor{gray}{$\phantom{-}4.7_{21.2}$}&$1.0 \times 10^{0}$&$-0.1$&\textcolor{gray}{$\phantom{-}3.8_{19.2}$}&$1.0 \times 10^{0}$&$-0.1$&$\phantom{-}6.8_{25.0}$&$\phantom{-}6.5_{24.8}$&$\phantom{-}7.0_{25.7}$\\
Reflections on Consequences&\textcolor{gray}{$\phantom{-}35.1_{47.8}$}&$1.0 \times 10^{0}$&$\phantom{-}0.0$&\textcolor{gray}{$\phantom{-}27.4_{44.7}$}&$5.8 \times 10^{-1}$&$-0.2$&\textcolor{gray}{$\phantom{-}26.6_{44.2}$}&$1.2 \times 10^{-1}$&$-0.2$&\textcolor{gray}{$\phantom{-}31.5_{46.5}$}&$1.0 \times 10^{0}$&$-0.1$&$\phantom{-}36.1_{46.8}$&$\phantom{-}26.5_{44.2}$&$\phantom{-}45.6_{50.0}$\\
Reflections on Conflicts&\textcolor{gray}{$\phantom{-}11.2_{31.6}$}&$1.0 \times 10^{0}$&$\phantom{-}0.1$&\textcolor{gray}{$\phantom{-}6.9_{25.4}$}&$1.0 \times 10^{0}$&$\phantom{-}0.0$&\textcolor{gray}{$\phantom{-}6.7_{25.0}$}&$1.0 \times 10^{0}$&$-0.1$&\textcolor{gray}{$\phantom{-}8.1_{27.3}$}&$1.0 \times 10^{0}$&$\phantom{-}0.0$&$\phantom{-}8.2_{27.1}$&$\phantom{-}7.6_{26.6}$&$\phantom{-}8.8_{28.4}$\\
Reflections on Strengths&\textcolor{gray}{$\phantom{-}11.7_{32.2}$}&$8.8 \times 10^{-2}$&$\phantom{-}0.2$&\textcolor{gray}{$\phantom{-}8.1_{27.3}$}&$1.0 \times 10^{0}$&$\phantom{-}0.1$&\textcolor{gray}{$\phantom{-}7.6_{26.5}$}&$1.0 \times 10^{0}$&$\phantom{-}0.1$&\textcolor{gray}{$\phantom{-}4.7_{21.2}$}&$1.0 \times 10^{0}$&$\phantom{-}0.0$&$\phantom{-}5.2_{23.6}$&$\phantom{-}6.9_{25.4}$&$\phantom{-}3.5_{18.5}$\\
Questions on Experiences&$\phantom{-}19.5_{39.7}$&$2.3 \times 10^{-8}$&$-0.4$&$\phantom{-}20.5_{40.4}$&$3.5 \times 10^{-7}$&$-0.4$&\textcolor{gray}{$\phantom{-}29.7_{45.7}$}&$1.6 \times 10^{-1}$&$-0.2$&\textcolor{gray}{$\phantom{-}30.6_{46.1}$}&$6.5 \times 10^{-1}$&$-0.2$&$\phantom{-}39.3_{49.4}$&$\phantom{-}46.2_{49.9}$&$\phantom{-}32.5_{47.0}$\\
Questions on Perspectives&\textcolor{gray}{$\phantom{-}7.6_{26.5}$}&$9.9 \times 10^{-1}$&$\phantom{-}0.2$&\textcolor{gray}{$\phantom{-}2.6_{16.0}$}&$1.0 \times 10^{0}$&$\phantom{-}0.0$&\textcolor{gray}{$\phantom{-}6.9_{25.3}$}&$1.0 \times 10^{0}$&$\phantom{-}0.1$&\textcolor{gray}{$\phantom{-}6.1_{23.9}$}&$1.0 \times 10^{0}$&$\phantom{-}0.1$&$\phantom{-}3.5_{19.3}$&$\phantom{-}4.4_{20.5}$&$\phantom{-}2.6_{16.1}$\\
Questions on Emotions&$\phantom{-}0.5_{6.8}$&$2.5 \times 10^{-7}$&$-0.4$&$\phantom{-}0.5_{6.9}$&$5.5 \times 10^{-7}$&$-0.4$&$\phantom{-}2.5_{15.7}$&$6.7 \times 10^{-4}$&$-0.3$&\textcolor{gray}{$\phantom{-}5.8_{23.5}$}&$1.0 \times 10^{0}$&$-0.1$&$\phantom{-}9.2_{29.7}$&$\phantom{-}10.5_{30.8}$&$\phantom{-}7.9_{27.1}$\\
Problem-Solving&\textcolor{gray}{$\phantom{-}31.4_{46.5}$}&$8.2 \times 10^{-1}$&$\phantom{-}0.2$&\textcolor{gray}{$\phantom{-}27.9_{44.9}$}&$1.0 \times 10^{0}$&$\phantom{-}0.1$&$\phantom{-}34.0_{47.4}$&$3.5 \times 10^{-2}$&$\phantom{-}0.2$&\textcolor{gray}{$\phantom{-}21.8_{41.3}$}&$1.0 \times 10^{0}$&$\phantom{-}0.0$&$\phantom{-}23.6_{40.3}$&$\phantom{-}15.6_{36.4}$&$\phantom{-}31.6_{46.7}$\\
Planning&$\phantom{-}12.8_{33.5}$&$4.7 \times 10^{-2}$&$\phantom{-}0.2$&\textcolor{gray}{$\phantom{-}9.3_{29.1}$}&$1.0 \times 10^{0}$&$\phantom{-}0.1$&$\phantom{-}19.9_{40.0}$&$5.1 \times 10^{-8}$&$\phantom{-}0.4$&\textcolor{gray}{$\phantom{-}10.1_{30.2}$}&$1.0 \times 10^{0}$&$\phantom{-}0.2$&$\phantom{-}5.9_{22.1}$&$\phantom{-}4.0_{19.6}$&$\phantom{-}7.9_{27.1}$\\
Normalizing&\textcolor{gray}{$\phantom{-}14.7_{35.4}$}&$1.0 \times 10^{0}$&$\phantom{-}0.0$&\textcolor{gray}{$\phantom{-}15.5_{36.2}$}&$1.0 \times 10^{0}$&$\phantom{-}0.0$&$\phantom{-}6.9_{25.3}$&$1.4 \times 10^{-2}$&$-0.2$&\textcolor{gray}{$\phantom{-}7.4_{26.2}$}&$1.1 \times 10^{-1}$&$-0.2$&$\phantom{-}14.0_{33.2}$&$\phantom{-}10.5_{30.8}$&$\phantom{-}17.5_{38.2}$\\
Psychoeducation&$\phantom{-}2.1_{14.2}$&$1.3 \times 10^{-5}$&$-0.4$&\textcolor{gray}{$\phantom{-}11.9_{32.5}$}&$1.0 \times 10^{0}$&$\phantom{-}0.1$&\textcolor{gray}{$\phantom{-}5.2_{22.3}$}&$3.4 \times 10^{-1}$&$-0.2$&$\phantom{-}2.9_{16.9}$&$5.6 \times 10^{-4}$&$-0.3$&$\phantom{-}9.8_{26.7}$&$\phantom{-}4.7_{21.3}$&$\phantom{-}14.9_{35.8}$\\
        \midrule
          \multicolumn{16}{c}{\textbf{Client behavior: Sharing positive emotions}} \\
          \midrule
          $n$ & \multicolumn{3}{c}{\textbf{$658$}} & \multicolumn{3}{c}{\textbf{$554$}} & \multicolumn{3}{c}{\textbf{$408$}} & \multicolumn{3}{c}{\textbf{$534$}}&\textbf{$294$}&\textbf{$226$}&\textbf{$68$}\\
          \midrule
          Reflections on Needs&\textcolor{gray}{$\phantom{-}12.9_{33.6}$}&$1.0 \times 10^{0}$&$-0.1$&\textcolor{gray}{$\phantom{-}17.9_{38.3}$}&$1.0 \times 10^{0}$&$\phantom{-}0.0$&\textcolor{gray}{$\phantom{-}20.6_{40.5}$}&$1.0 \times 10^{0}$&$\phantom{-}0.1$&\textcolor{gray}{$\phantom{-}16.7_{37.3}$}&$1.0 \times 10^{0}$&$\phantom{-}0.0$&$\phantom{-}17.3_{40.9}$&$\phantom{-}24.3_{43.0}$&$\phantom{-}10.3_{30.6}$\\
          Reflections on Emotions&\textcolor{gray}{$\phantom{-}5.3_{22.5}$}&$1.0 \times 10^{0}$&$-0.1$&\textcolor{gray}{$\phantom{-}8.1_{27.3}$}&$1.0 \times 10^{0}$&$\phantom{-}0.0$&\textcolor{gray}{$\phantom{-}5.6_{23.1}$}&$1.0 \times 10^{0}$&$-0.1$&\textcolor{gray}{$\phantom{-}5.8_{23.4}$}&$1.0 \times 10^{0}$&$-0.1$&$\phantom{-}7.5_{32.0}$&$\phantom{-}15.0_{35.8}$&$\phantom{-}0.0_{0.0}$\\
          Reflections on Values&\textcolor{gray}{$\phantom{-}2.9_{16.8}$}&$1.0 \times 10^{0}$&$\phantom{-}0.0$&\textcolor{gray}{$\phantom{-}4.0_{19.5}$}&$1.0 \times 10^{0}$&$\phantom{-}0.0$&\textcolor{gray}{$\phantom{-}5.1_{22.1}$}&$1.0 \times 10^{0}$&$\phantom{-}0.1$&\textcolor{gray}{$\phantom{-}6.4_{24.4}$}&$1.0 \times 10^{0}$&$\phantom{-}0.1$&$\phantom{-}3.4_{20.6}$&$\phantom{-}5.3_{22.5}$&$\phantom{-}1.5_{12.1}$\\
          Reflections on Consequences&$\phantom{-}6.2_{24.2}$&$6.6 \times 10^{-5}$&$-0.3$&$\phantom{-}5.2_{22.3}$&$5.1 \times 10^{-6}$&$-0.4$&\textcolor{gray}{$\phantom{-}14.2_{35.0}$}&$1.0 \times 10^{0}$&$\phantom{-}0.0$&$\phantom{-}7.9_{26.9}$&$1.8 \times 10^{-2}$&$-0.3$&$\phantom{-}15.8_{34.0}$&$\phantom{-}11.1_{31.4}$&$\phantom{-}20.6_{40.7}$\\
Reflections on Conflicts&\textcolor{gray}{$\phantom{-}1.8_{13.4}$}&$1.0 \times 10^{0}$&$-0.1$&\textcolor{gray}{$\phantom{-}2.0_{14.0}$}&$1.0 \times 10^{0}$&$-0.1$&\textcolor{gray}{$\phantom{-}4.9_{21.6}$}&$1.0 \times 10^{0}$&$\phantom{-}0.1$&\textcolor{gray}{$\phantom{-}1.9_{13.6}$}&$1.0 \times 10^{0}$&$-0.1$&$\phantom{-}3.4_{20.6}$&$\phantom{-}5.3_{22.5}$&$\phantom{-}1.5_{12.1}$\\
Reflections on Strengths&$\phantom{-}66.1_{47.4}$&$3.7 \times 10^{-62}$&$\phantom{-}1.3$&$\phantom{-}52.3_{50.0}$&$2.7 \times 10^{-33}$&$\phantom{-}0.9$&$\phantom{-}29.2_{45.5}$&$1.9 \times 10^{-7}$&$\phantom{-}0.5$&$\phantom{-}39.3_{48.9}$&$5.8 \times 10^{-17}$&$\phantom{-}0.7$&$\phantom{-}10.2_{34.0}$&$\phantom{-}15.9_{36.7}$&$\phantom{-}4.4_{20.7}$\\
Questions on Experiences&$\phantom{-}7.9_{27.0}$&$1.5 \times 10^{-19}$&$-0.7$&$\phantom{-}10.6_{30.9}$&$2.2 \times 10^{-12}$&$-0.6$&\textcolor{gray}{$\phantom{-}24.3_{42.9}$}&$1.0 \times 10^{0}$&$-0.2$&$\phantom{-}15.7_{36.4}$&$8.3 \times 10^{-6}$&$-0.4$&$\phantom{-}31.6_{47.6}$&$\phantom{-}36.7_{48.3}$&$\phantom{-}26.5_{44.4}$\\
Questions on Perspectives&\textcolor{gray}{$\phantom{-}2.1_{14.4}$}&$1.0 \times 10^{0}$&$\phantom{-}0.1$&\textcolor{gray}{$\phantom{-}0.7_{8.5}$}&$1.0 \times 10^{0}$&$-0.1$&\textcolor{gray}{$\phantom{-}3.9_{19.4}$}&$1.0 \times 10^{0}$&$\phantom{-}0.1$&\textcolor{gray}{$\phantom{-}3.4_{18.1}$}&$1.0 \times 10^{0}$&$\phantom{-}0.1$&$\phantom{-}1.3_{14.2}$&$\phantom{-}2.7_{16.1}$&$\phantom{-}0.0_{0.0}$\\
Questions on Emotions&$\phantom{-}0.5_{6.7}$&$3.6 \times 10^{-7}$&$-0.4$&$\phantom{-}0.5_{7.3}$&$8.4 \times 10^{-6}$&$-0.4$&\textcolor{gray}{$\phantom{-}4.9_{21.6}$}&$1.0 \times 10^{0}$&$-0.1$&\textcolor{gray}{$\phantom{-}3.6_{18.5}$}&$1.0 \times 10^{0}$&$-0.2$&$\phantom{-}7.5_{28.9}$&$\phantom{-}10.6_{30.9}$&$\phantom{-}4.4_{20.7}$\\

Problem-Solving&\textcolor{gray}{$\phantom{-}19.8_{39.8}$}&$9.3 \times 10^{-1}$&$\phantom{-}0.2$&$\phantom{-}23.1_{42.2}$&$3.3 \times 10^{-2}$&$\phantom{-}0.3$&$\phantom{-}26.0_{43.9}$&$2.0 \times 10^{-3}$&$\phantom{-}0.3$&\textcolor{gray}{$\phantom{-}14.2_{35.0}$}&$1.0 \times 10^{0}$&$\phantom{-}0.0$&$\phantom{-}13.0_{33.6}$&$\phantom{-}12.8_{33.5}$&$\phantom{-}13.2_{34.1}$\\
Planning&$\phantom{-}47.9_{50.0}$&$7.0 \times 10^{-33}$&$\phantom{-}0.9$&$\phantom{-}39.4_{48.9}$&$5.1 \times 10^{-21}$&$\phantom{-}0.7$&$\phantom{-}32.6_{46.9}$&$3.4 \times 10^{-13}$&$\phantom{-}0.6$&$\phantom{-}25.1_{43.4}$&$1.5 \times 10^{-7}$&$\phantom{-}0.4$&$\phantom{-}8.2_{26.9}$&$\phantom{-}7.5_{26.4}$&$\phantom{-}8.8_{28.6}$\\
Normalizing&$\phantom{-}59.0_{49.2}$&$7.1 \times 10^{-25}$&$\phantom{-}0.8$&$\phantom{-}42.6_{49.5}$&$4.0 \times 10^{-7}$&$\phantom{-}0.4$&$\phantom{-}12.5_{33.1}$&$2.0 \times 10^{-2}$&$-0.3$&\textcolor{gray}{$\phantom{-}29.0_{45.4}$}&$1.0 \times 10^{0}$&$\phantom{-}0.1$&$\phantom{-}22.7_{40.6}$&$\phantom{-}19.0_{39.3}$&$\phantom{-}26.5_{44.4}$\\
Psychoeducation&\textcolor{gray}{$\phantom{-}4.3_{20.2}$}&$1.0 \times 10^{0}$&$\phantom{-}0.1$&\textcolor{gray}{$\phantom{-}7.4_{26.2}$}&$7.5 \times 10^{-1}$&$\phantom{-}0.2$&\textcolor{gray}{$\phantom{-}5.6_{23.1}$}&$1.0 \times 10^{0}$&$\phantom{-}0.1$&\textcolor{gray}{$\phantom{-}4.7_{21.1}$}&$1.0 \times 10^{0}$&$\phantom{-}0.1$&$\phantom{-}3.1_{15.3}$&$\phantom{-}1.8_{13.2}$&$\phantom{-}4.4_{20.7}$\\

    \bottomrule
    \end{tabular}
    }
    \label{tab:adapt_new_hl_p1}
\end{table*}
\begin{table*}[t]
    \centering
    \small
    \caption{\showedits{(Continue Table \ref{tab:adapt_new_hl_p1}) Adaptability (frequency of conversational behaviors exhibited by therapists \textit{in response to specific client behaviors}) by LLM
therapists (GPT-4, GPT-3.5-turbo, Llama2-70b, Llama2-13b) in High-low quality dataset \cite{perez-rosas-etal-2019-makes}. Values in \textcolor{gray}{gray} are not statistically significantly different from human therapist average at $p = \frac{0.05}{m}$ using Two-sided Student’s t-test after Bonferroni correction, where $m$ is the number of t-test per one model ($m = 13 \times 6 = 78$). $d$ is the cohen's d when compared with human therapist average. For instance, GPT-4 exhibits \textsc{Reflections on Needs} by 38.9\%  and human therapists responds by 24.5\% on average when client expresses \textsc{Sharing negative emotions}. ($P = 4.0 \times 10^{-9}$, Cohen's $d = \num{0.3}$, two-sided Student’s $t$-test).}}
    \resizebox{1\textwidth}{!}{%
    \def\arraystretch{1.15}
    \setlength{\tabcolsep}{2pt}
    \begin{tabular}{l|ccc|ccc|ccc|ccc|ccc|ccc}
         \toprule 
         \multirow{3}{*}{\textbf{Therapist Behavior}} & \multicolumn{12}{c}{\textbf{LLM Therapists (Single Response \& Full Conversation Simulations)}} & \multicolumn{3}{c}{\textbf{Human Therapists}} \\
         \cmidrule(lr){2-13} \cmidrule(lr){14-16}
         & \multicolumn{3}{c}{\textbf{GPT-4}} &\multicolumn{3}{c}{\textbf{GPT-3.5}}&\multicolumn{3}{c}{\textbf{Llama2-70b}}&\multicolumn{3}{c}{\textbf{Llama2-13b}}& \textbf{Average}&\textbf{High-quality}&\textbf{Low-quality}\\
         \cmidrule(lr){2-4} \cmidrule(lr){5-7} \cmidrule(lr){8-10} \cmidrule(lr){11-13} \cmidrule(lr){14-14} \cmidrule(lr){15-15}  \cmidrule(lr){16-16} 
        & freq. & $p$ & $d$ & freq. & $p$ & $d$ & freq. & $p$& $d$ & freq. & $p$ & $d$ & freq. & freq. & freq. \\
\midrule

          \multicolumn{16}{c}{\textbf{Client behavior: Sharing negative emotions}} \\
          \midrule
          $n$ & \multicolumn{3}{c}{\textbf{$1063$}} & \multicolumn{3}{c}{\textbf{$1016$}} & \multicolumn{3}{c}{\textbf{$1236$}} & \multicolumn{3}{c}{\textbf{$1114$}}&\textbf{$820$}&\textbf{$646$}&\textbf{$174$}\\
          \midrule
Reflections on Needs&$\phantom{-}38.9_{48.8}$&$4.0 \times 10^{-9}$&$\phantom{-}0.3$&$\phantom{-}32.3_{46.8}$&$2.5 \times 10^{-2}$&$\phantom{-}0.2$&\textcolor{gray}{$\phantom{-}28.9_{45.3}$}&$1.0 \times 10^{0}$&$\phantom{-}0.1$&$\phantom{-}31.9_{46.6}$&$3.9 \times 10^{-2}$&$\phantom{-}0.2$&$\phantom{-}24.5_{44.5}$&$\phantom{-}28.9_{45.4}$&$\phantom{-}20.1_{40.2}$\\
Reflections on Emotions&$\phantom{-}27.4_{44.6}$&$8.0 \times 10^{-3}$&$\phantom{-}0.2$&$\phantom{-}33.9_{47.3}$&$1.4 \times 10^{-9}$&$\phantom{-}0.3$&\textcolor{gray}{$\phantom{-}25.1_{43.4}$}&$2.9 \times 10^{-1}$&$\phantom{-}0.1$&\textcolor{gray}{$\phantom{-}24.0_{42.7}$}&$1.0 \times 10^{0}$&$\phantom{-}0.1$&$\phantom{-}19.4_{42.7}$&$\phantom{-}27.4_{44.6}$&$\phantom{-}11.5_{32.0}$\\
Reflections on Values&\textcolor{gray}{$\phantom{-}4.5_{20.8}$}&$1.0 \times 10^{0}$&$-0.1$&\textcolor{gray}{$\phantom{-}4.9_{21.6}$}&$1.0 \times 10^{0}$&$-0.1$&\textcolor{gray}{$\phantom{-}4.0_{19.5}$}&$1.0 \times 10^{0}$&$-0.1$&\textcolor{gray}{$\phantom{-}5.6_{22.9}$}&$1.0 \times 10^{0}$&$\phantom{-}0.0$&$\phantom{-}6.3_{25.4}$&$\phantom{-}7.4_{26.2}$&$\phantom{-}5.2_{22.2}$\\
Reflections on Consequences&\textcolor{gray}{$\phantom{-}21.3_{40.9}$}&$1.0 \times 10^{0}$&$\phantom{-}0.0$&$\phantom{-}14.0_{34.7}$&$6.1 \times 10^{-4}$&$-0.2$&\textcolor{gray}{$\phantom{-}16.7_{37.4}$}&$2.8 \times 10^{-1}$&$-0.1$&\textcolor{gray}{$\phantom{-}22.2_{41.6}$}&$1.0 \times 10^{0}$&$\phantom{-}0.0$&$\phantom{-}21.8_{39.8}$&$\phantom{-}18.3_{38.7}$&$\phantom{-}25.3_{43.6}$\\
Reflections on Conflicts&\textcolor{gray}{$\phantom{-}7.9_{27.0}$}&$1.0 \times 10^{0}$&$\phantom{-}0.0$&\textcolor{gray}{$\phantom{-}5.3_{22.4}$}&$8.2 \times 10^{-1}$&$-0.1$&\textcolor{gray}{$\phantom{-}8.4_{27.8}$}&$1.0 \times 10^{0}$&$\phantom{-}0.0$&\textcolor{gray}{$\phantom{-}8.6_{28.1}$}&$1.0 \times 10^{0}$&$\phantom{-}0.0$&$\phantom{-}8.4_{29.5}$&$\phantom{-}10.5_{30.7}$&$\phantom{-}6.3_{24.4}$\\
Reflections on Strengths&$\phantom{-}23.1_{42.2}$&$2.0 \times 10^{-20}$&$\phantom{-}0.5$&$\phantom{-}17.2_{37.8}$&$8.9 \times 10^{-10}$&$\phantom{-}0.3$&$\phantom{-}17.8_{38.3}$&$2.8 \times 10^{-11}$&$\phantom{-}0.3$&$\phantom{-}11.8_{32.2}$&$1.3 \times 10^{-2}$&$\phantom{-}0.2$&$\phantom{-}6.6_{26.6}$&$\phantom{-}8.5_{27.9}$&$\phantom{-}4.6_{21.0}$\\
Questions on Experiences&$\phantom{-}22.3_{41.6}$&$2.8 \times 10^{-7}$&$-0.3$&$\phantom{-}18.4_{38.8}$&$2.2 \times 10^{-13}$&$-0.4$&\textcolor{gray}{$\phantom{-}32.9_{47.0}$}&$1.0 \times 10^{0}$&$\phantom{-}0.0$&\textcolor{gray}{$\phantom{-}34.9_{47.7}$}&$1.0 \times 10^{0}$&$\phantom{-}0.0$&$\phantom{-}34.6_{48.6}$&$\phantom{-}40.6_{49.1}$&$\phantom{-}28.7_{45.4}$\\
Questions on Perspectives&\textcolor{gray}{$\phantom{-}3.3_{17.9}$}&$1.0 \times 10^{0}$&$\phantom{-}0.0$&$\phantom{-}0.8_{8.8}$&$3.1 \times 10^{-2}$&$-0.2$&\textcolor{gray}{$\phantom{-}5.6_{23.0}$}&$1.0 \times 10^{0}$&$\phantom{-}0.1$&\textcolor{gray}{$\phantom{-}4.9_{21.7}$}&$1.0 \times 10^{0}$&$\phantom{-}0.1$&$\phantom{-}3.3_{19.9}$&$\phantom{-}4.8_{21.4}$&$\phantom{-}1.7_{13.1}$\\
Questions on Emotions&$\phantom{-}1.6_{12.6}$&$2.8 \times 10^{-12}$&$-0.4$&$\phantom{-}1.1_{10.4}$&$3.6 \times 10^{-14}$&$-0.4$&$\phantom{-}3.2_{17.7}$&$3.1 \times 10^{-7}$&$-0.3$&$\phantom{-}4.5_{20.7}$&$1.1 \times 10^{-3}$&$-0.2$&$\phantom{-}9.6_{31.1}$&$\phantom{-}11.8_{32.2}$&$\phantom{-}7.5_{26.4}$\\
Problem-Solving&$\phantom{-}37.1_{48.3}$&$4.4 \times 10^{-17}$&$\phantom{-}0.4$&$\phantom{-}35.5_{47.9}$&$2.8 \times 10^{-14}$&$\phantom{-}0.4$&$\phantom{-}33.2_{47.1}$&$2.3 \times 10^{-11}$&$\phantom{-}0.3$&\textcolor{gray}{$\phantom{-}24.4_{43.0}$}&$2.5 \times 10^{-1}$&$\phantom{-}0.1$&$\phantom{-}19.0_{35.7}$&$\phantom{-}12.1_{32.6}$&$\phantom{-}25.9_{43.9}$\\
Planning&$\phantom{-}13.9_{34.6}$&$8.8 \times 10^{-4}$&$\phantom{-}0.2$&$\phantom{-}12.8_{33.4}$&$1.9 \times 10^{-2}$&$\phantom{-}0.2$&$\phantom{-}19.3_{39.4}$&$7.0 \times 10^{-12}$&$\phantom{-}0.3$&$\phantom{-}13.5_{34.2}$&$2.7 \times 10^{-3}$&$\phantom{-}0.2$&$\phantom{-}7.6_{25.0}$&$\phantom{-}6.0_{23.8}$&$\phantom{-}9.2_{29.0}$\\
Normalizing&$\phantom{-}23.0_{42.1}$&$3.4 \times 10^{-2}$&$\phantom{-}0.2$&$\phantom{-}30.3_{46.0}$&$2.8 \times 10^{-10}$&$\phantom{-}0.3$&\textcolor{gray}{$\phantom{-}12.9_{33.6}$}&$1.0 \times 10^{0}$&$-0.1$&\textcolor{gray}{$\phantom{-}14.3_{35.0}$}&$1.0 \times 10^{0}$&$-0.1$&$\phantom{-}16.5_{36.2}$&$\phantom{-}14.7_{35.4}$&$\phantom{-}18.4_{38.9}$\\
Psychoeducation&\textcolor{gray}{$\phantom{-}4.5_{20.8}$}&$1.0 \times 10^{0}$&$\phantom{-}0.0$&$\phantom{-}10.3_{30.5}$&$2.5 \times 10^{-4}$&$\phantom{-}0.2$&\textcolor{gray}{$\phantom{-}6.3_{24.3}$}&$1.0 \times 10^{0}$&$\phantom{-}0.1$&\textcolor{gray}{$\phantom{-}3.0_{17.0}$}&$1.0 \times 10^{0}$&$-0.1$&$\phantom{-}4.7_{17.5}$&$\phantom{-}2.0_{14.1}$&$\phantom{-}7.5_{26.4}$\\
        \midrule
          \multicolumn{16}{c}{\textbf{Client behavior: Sharing experiences}} \\
          \midrule
          $n$ & \multicolumn{3}{c}{\textbf{$865$}} & \multicolumn{3}{c}{\textbf{$842$}} & \multicolumn{3}{c}{\textbf{$940$}} & \multicolumn{3}{c}{\textbf{$879$}}&\textbf{$759$}&\textbf{$533$}&\textbf{$226$}\\
          \midrule
Reflections on Needs&$\phantom{-}33.3_{47.2}$&$1.4 \times 10^{-4}$&$\phantom{-}0.2$&\textcolor{gray}{$\phantom{-}28.0_{44.9}$}&$9.6 \times 10^{-1}$&$\phantom{-}0.1$&\textcolor{gray}{$\phantom{-}26.0_{43.9}$}&$1.0 \times 10^{0}$&$\phantom{-}0.1$&\textcolor{gray}{$\phantom{-}27.0_{44.4}$}&$1.0 \times 10^{0}$&$\phantom{-}0.1$&$\phantom{-}22.5_{42.8}$&$\phantom{-}26.5_{44.2}$&$\phantom{-}18.6_{39.0}$\\
Reflections on Emotions&\textcolor{gray}{$\phantom{-}17.9_{38.4}$}&$1.0 \times 10^{0}$&$\phantom{-}0.1$&$\phantom{-}23.9_{42.7}$&$1.8 \times 10^{-5}$&$\phantom{-}0.3$&$\phantom{-}20.0_{40.0}$&$4.3 \times 10^{-2}$&$\phantom{-}0.2$&\textcolor{gray}{$\phantom{-}19.3_{39.5}$}&$1.6 \times 10^{-1}$&$\phantom{-}0.2$&$\phantom{-}13.5_{36.8}$&$\phantom{-}19.9_{40.0}$&$\phantom{-}7.1_{25.7}$\\
Reflections on Values&\textcolor{gray}{$\phantom{-}5.7_{23.1}$}&$1.0 \times 10^{0}$&$\phantom{-}0.0$&\textcolor{gray}{$\phantom{-}5.5_{22.7}$}&$1.0 \times 10^{0}$&$\phantom{-}0.0$&\textcolor{gray}{$\phantom{-}4.5_{20.7}$}&$1.0 \times 10^{0}$&$\phantom{-}0.0$&\textcolor{gray}{$\phantom{-}5.1_{22.1}$}&$1.0 \times 10^{0}$&$\phantom{-}0.0$&$\phantom{-}4.7_{23.1}$&$\phantom{-}7.1_{25.8}$&$\phantom{-}2.2_{14.7}$\\
Reflections on Consequences&\textcolor{gray}{$\phantom{-}19.4_{39.6}$}&$1.0 \times 10^{0}$&$\phantom{-}0.0$&$\phantom{-}10.8_{31.1}$&$1.8 \times 10^{-3}$&$-0.2$&\textcolor{gray}{$\phantom{-}14.1_{34.9}$}&$1.0 \times 10^{0}$&$-0.1$&\textcolor{gray}{$\phantom{-}16.3_{36.9}$}&$1.0 \times 10^{0}$&$-0.1$&$\phantom{-}18.2_{38.5}$&$\phantom{-}17.8_{38.3}$&$\phantom{-}18.6_{39.0}$\\
Reflections on Conflicts&\textcolor{gray}{$\phantom{-}6.4_{24.4}$}&$1.0 \times 10^{0}$&$\phantom{-}0.0$&\textcolor{gray}{$\phantom{-}3.4_{18.2}$}&$1.0 \times 10^{0}$&$-0.1$&\textcolor{gray}{$\phantom{-}7.2_{25.9}$}&$1.0 \times 10^{0}$&$\phantom{-}0.1$&\textcolor{gray}{$\phantom{-}5.5_{22.7}$}&$1.0 \times 10^{0}$&$\phantom{-}0.0$&$\phantom{-}5.8_{25.1}$&$\phantom{-}8.1_{27.3}$&$\phantom{-}3.5_{18.5}$\\
Reflections on Strengths&$\phantom{-}21.5_{41.1}$&$5.4 \times 10^{-11}$&$\phantom{-}0.4$&$\phantom{-}18.3_{38.7}$&$1.4 \times 10^{-6}$&$\phantom{-}0.3$&\textcolor{gray}{$\phantom{-}13.5_{34.2}$}&$1.0 \times 10^{-1}$&$\phantom{-}0.2$&\textcolor{gray}{$\phantom{-}12.5_{33.1}$}&$7.3 \times 10^{-1}$&$\phantom{-}0.1$&$\phantom{-}8.4_{29.9}$&$\phantom{-}12.0_{32.5}$&$\phantom{-}4.9_{21.6}$\\
Questions on Experiences&$\phantom{-}23.5_{42.4}$&$3.6 \times 10^{-18}$&$-0.5$&$\phantom{-}25.3_{43.5}$&$1.3 \times 10^{-14}$&$-0.4$&$\phantom{-}35.1_{47.8}$&$4.1 \times 10^{-3}$&$-0.2$&$\phantom{-}36.4_{48.1}$&$4.7 \times 10^{-2}$&$-0.2$&$\phantom{-}44.7_{49.8}$&$\phantom{-}45.2_{49.8}$&$\phantom{-}44.2_{49.8}$\\
Questions on Perspectives&\textcolor{gray}{$\phantom{-}4.4_{20.5}$}&$7.9 \times 10^{-1}$&$\phantom{-}0.1$&\textcolor{gray}{$\phantom{-}1.4_{11.9}$}&$1.0 \times 10^{0}$&$\phantom{-}0.0$&\textcolor{gray}{$\phantom{-}5.0_{21.8}$}&$1.3 \times 10^{-1}$&$\phantom{-}0.2$&\textcolor{gray}{$\phantom{-}4.6_{20.9}$}&$5.1 \times 10^{-1}$&$\phantom{-}0.1$&$\phantom{-}2.0_{15.6}$&$\phantom{-}3.2_{17.6}$&$\phantom{-}0.9_{9.4}$\\
Questions on Emotions&$\phantom{-}2.2_{14.7}$&$2.4 \times 10^{-7}$&$-0.3$&$\phantom{-}1.8_{13.2}$&$1.5 \times 10^{-8}$&$-0.3$&$\phantom{-}3.9_{19.5}$&$2.6 \times 10^{-3}$&$-0.2$&\textcolor{gray}{$\phantom{-}5.8_{23.4}$}&$1.0 \times 10^{0}$&$-0.1$&$\phantom{-}8.8_{28.6}$&$\phantom{-}9.2_{28.9}$&$\phantom{-}8.4_{27.8}$\\
Problem-Solving&$\phantom{-}29.6_{45.7}$&$7.1 \times 10^{-13}$&$\phantom{-}0.4$&$\phantom{-}23.9_{42.7}$&$2.1 \times 10^{-5}$&$\phantom{-}0.3$&$\phantom{-}21.2_{40.9}$&$6.7 \times 10^{-3}$&$\phantom{-}0.2$&\textcolor{gray}{$\phantom{-}15.1_{35.9}$}&$1.0 \times 10^{0}$&$\phantom{-}0.0$&$\phantom{-}14.0_{32.7}$&$\phantom{-}9.4_{29.2}$&$\phantom{-}18.6_{39.0}$\\

Planning&$\phantom{-}12.9_{33.6}$&$5.9 \times 10^{-3}$&$\phantom{-}0.2$&\textcolor{gray}{$\phantom{-}10.5_{30.6}$}&$1.0 \times 10^{0}$&$\phantom{-}0.1$&$\phantom{-}13.1_{33.7}$&$3.3 \times 10^{-3}$&$\phantom{-}0.2$&\textcolor{gray}{$\phantom{-}8.3_{27.6}$}&$1.0 \times 10^{0}$&$\phantom{-}0.0$&$\phantom{-}7.0_{25.5}$&$\phantom{-}6.9_{25.4}$&$\phantom{-}7.1_{25.7}$\\

Normalizing&\textcolor{gray}{$\phantom{-}16.2_{36.9}$}&$1.0 \times 10^{0}$&$\phantom{-}0.1$&\textcolor{gray}{$\phantom{-}18.2_{38.6}$}&$1.0 \times 10^{0}$&$\phantom{-}0.1$&\textcolor{gray}{$\phantom{-}8.8_{28.4}$}&$5.5 \times 10^{-2}$&$-0.2$&\textcolor{gray}{$\phantom{-}11.0_{31.4}$}&$1.0 \times 10^{0}$&$-0.1$&$\phantom{-}14.0_{34.7}$&$\phantom{-}13.9_{34.6}$&$\phantom{-}14.2_{34.9}$\\
Psychoeducation&\textcolor{gray}{$\phantom{-}3.1_{17.4}$}&$4.5 \times 10^{-1}$&$-0.1$&\textcolor{gray}{$\phantom{-}5.7_{23.2}$}&$1.0 \times 10^{0}$&$\phantom{-}0.0$&\textcolor{gray}{$\phantom{-}2.8_{16.4}$}&$7.7 \times 10^{-2}$&$-0.2$&$\phantom{-}1.1_{10.6}$&$7.4 \times 10^{-7}$&$-0.3$&$\phantom{-}5.7_{20.7}$&$\phantom{-}2.6_{16.0}$&$\phantom{-}8.8_{28.5}$\\
        \midrule
          \multicolumn{16}{c}{\textbf{Client behavior: Gained insights}} \\
          \midrule
          $n$ & \multicolumn{3}{c}{\textbf{$450$}} & \multicolumn{3}{c}{\textbf{$365$}} & \multicolumn{3}{c}{\textbf{$430$}} & \multicolumn{3}{c}{\textbf{$437$}}&\textbf{$258$}&\textbf{$214$}&\textbf{$44$}
          \\
          \midrule
Reflections on Needs&\textcolor{gray}{$\phantom{-}28.0_{44.9}$}&$1.0 \times 10^{0}$&$\phantom{-}0.0$&\textcolor{gray}{$\phantom{-}29.3_{45.6}$}&$1.0 \times 10^{0}$&$\phantom{-}0.1$&\textcolor{gray}{$\phantom{-}27.9_{44.9}$}&$1.0 \times 10^{0}$&$\phantom{-}0.0$&\textcolor{gray}{$\phantom{-}22.9_{42.1}$}&$1.0 \times 10^{0}$&$-0.1$&$\phantom{-}27.0_{45.1}$&$\phantom{-}29.0_{45.5}$&$\phantom{-}25.0_{43.8}$\\
Reflections on Emotions&\textcolor{gray}{$\phantom{-}9.1_{28.8}$}&$1.0 \times 10^{0}$&$-0.1$&\textcolor{gray}{$\phantom{-}11.8_{32.3}$}&$1.0 \times 10^{0}$&$\phantom{-}0.0$&\textcolor{gray}{$\phantom{-}10.5_{30.6}$}&$1.0 \times 10^{0}$&$-0.1$&\textcolor{gray}{$\phantom{-}6.9_{25.3}$}&$1.0 \times 10^{0}$&$-0.2$&$\phantom{-}12.5_{35.5}$&$\phantom{-}15.9_{36.6}$&$\phantom{-}9.1_{29.1}$\\
Reflections on Values&\textcolor{gray}{$\phantom{-}5.6_{22.9}$}&$1.0 \times 10^{0}$&$-0.1$&\textcolor{gray}{$\phantom{-}8.8_{28.3}$}&$1.0 \times 10^{0}$&$\phantom{-}0.0$&\textcolor{gray}{$\phantom{-}6.7_{25.1}$}&$1.0 \times 10^{0}$&$-0.1$&\textcolor{gray}{$\phantom{-}7.8_{26.8}$}&$1.0 \times 10^{0}$&$\phantom{-}0.0$&$\phantom{-}8.1_{28.6}$&$\phantom{-}9.3_{29.2}$&$\phantom{-}6.8_{25.5}$\\
Reflections on Consequences&\textcolor{gray}{$\phantom{-}18.2_{38.6}$}&$1.0 \times 10^{0}$&$\phantom{-}0.0$&\textcolor{gray}{$\phantom{-}12.6_{33.2}$}&$1.0 \times 10^{0}$&$-0.2$&\textcolor{gray}{$\phantom{-}15.8_{36.5}$}&$1.0 \times 10^{0}$&$-0.1$&\textcolor{gray}{$\phantom{-}14.9_{35.6}$}&$1.0 \times 10^{0}$&$-0.1$&$\phantom{-}18.8_{37.0}$&$\phantom{-}15.0_{35.7}$&$\phantom{-}22.7_{42.4}$\\
Reflections on Conflicts&\textcolor{gray}{$\phantom{-}4.2_{20.1}$}&$1.0 \times 10^{0}$&$\phantom{-}0.0$&\textcolor{gray}{$\phantom{-}3.8_{19.2}$}&$1.0 \times 10^{0}$&$\phantom{-}0.0$&\textcolor{gray}{$\phantom{-}6.3_{24.3}$}&$1.0 \times 10^{0}$&$\phantom{-}0.1$&\textcolor{gray}{$\phantom{-}4.8_{21.4}$}&$1.0 \times 10^{0}$&$\phantom{-}0.0$&$\phantom{-}3.9_{21.9}$&$\phantom{-}5.6_{23.1}$&$\phantom{-}2.3_{15.1}$\\
Reflections on Strengths&$\phantom{-}51.3_{50.0}$&$4.0 \times 10^{-24}$&$\phantom{-}0.9$&$\phantom{-}36.4_{48.2}$&$2.1 \times 10^{-9}$&$\phantom{-}0.6$&$\phantom{-}31.4_{46.5}$&$2.1 \times 10^{-6}$&$\phantom{-}0.4$&$\phantom{-}32.7_{47.0}$&$2.8 \times 10^{-7}$&$\phantom{-}0.5$&$\phantom{-}12.5_{35.5}$&$\phantom{-}15.9_{36.6}$&$\phantom{-}9.1_{29.1}$\\
Questions on Experiences&$\phantom{-}13.1_{33.8}$&$2.3 \times 10^{-5}$&$-0.4$&$\phantom{-}16.2_{36.9}$&$1.2 \times 10^{-2}$&$-0.3$&\textcolor{gray}{$\phantom{-}24.7_{43.1}$}&$1.0 \times 10^{0}$&$-0.1$&$\phantom{-}17.4_{37.9}$&$3.0 \times 10^{-2}$&$-0.3$&$\phantom{-}29.2_{48.6}$&$\phantom{-}42.5_{49.6}$&$\phantom{-}15.9_{37.0}$\\
Questions on Perspectives&\textcolor{gray}{$\phantom{-}3.1_{17.4}$}&$1.0 \times 10^{0}$&$-0.1$&\textcolor{gray}{$\phantom{-}2.2_{14.7}$}&$1.0 \times 10^{0}$&$-0.1$&\textcolor{gray}{$\phantom{-}4.4_{20.6}$}&$1.0 \times 10^{0}$&$\phantom{-}0.0$&\textcolor{gray}{$\phantom{-}4.3_{20.4}$}&$1.0 \times 10^{0}$&$\phantom{-}0.0$&$\phantom{-}4.2_{22.7}$&$\phantom{-}6.1_{23.9}$&$\phantom{-}2.3_{15.1}$\\
Questions on Emotions&$\phantom{-}0.7_{8.1}$&$5.2 \times 10^{-9}$&$-0.5$&$\phantom{-}1.9_{13.7}$&$9.7 \times 10^{-5}$&$-0.4$&\textcolor{gray}{$\phantom{-}4.2_{20.1}$}&$5.0 \times 10^{-2}$&$-0.3$&$\phantom{-}4.1_{19.9}$&$4.0 \times 10^{-2}$&$-0.3$&$\phantom{-}10.8_{30.7}$&$\phantom{-}10.3_{30.4}$&$\phantom{-}11.4_{32.1}$\\

Problem-Solving&$\phantom{-}41.1_{49.3}$&$2.0 \times 10^{-2}$&$\phantom{-}0.3$&\textcolor{gray}{$\phantom{-}37.5_{48.5}$}&$7.0 \times 10^{-1}$&$\phantom{-}0.2$&\textcolor{gray}{$\phantom{-}37.0_{48.3}$}&$8.9 \times 10^{-1}$&$\phantom{-}0.2$&\textcolor{gray}{$\phantom{-}25.9_{43.8}$}&$1.0 \times 10^{0}$&$\phantom{-}0.0$&$\phantom{-}27.9_{39.6}$&$\phantom{-}15.0_{35.7}$&$\phantom{-}40.9_{49.7}$\\
Planning&$\phantom{-}34.9_{47.7}$&$4.4 \times 10^{-10}$&$\phantom{-}0.5$&$\phantom{-}28.2_{45.1}$&$3.5 \times 10^{-5}$&$\phantom{-}0.4$&$\phantom{-}38.4_{48.7}$&$9.5 \times 10^{-13}$&$\phantom{-}0.6$&$\phantom{-}25.9_{43.8}$&$4.4 \times 10^{-4}$&$\phantom{-}0.4$&$\phantom{-}11.5_{32.1}$&$\phantom{-}11.7_{32.2}$&$\phantom{-}11.4_{32.1}$\\
Normalizing&$\phantom{-}32.4_{46.9}$&$3.1 \times 10^{-3}$&$\phantom{-}0.3$&\textcolor{gray}{$\phantom{-}30.1_{45.9}$}&$5.6 \times 10^{-2}$&$\phantom{-}0.3$&\textcolor{gray}{$\phantom{-}13.7_{34.4}$}&$1.0 \times 10^{0}$&$-0.1$&\textcolor{gray}{$\phantom{-}21.7_{41.3}$}&$1.0 \times 10^{0}$&$\phantom{-}0.1$&$\phantom{-}18.6_{35.1}$&$\phantom{-}12.1_{32.7}$&$\phantom{-}25.0_{43.8}$\\
Psychoeducation&\textcolor{gray}{$\phantom{-}8.7_{28.2}$}&$1.0 \times 10^{0}$&$\phantom{-}0.0$&\textcolor{gray}{$\phantom{-}12.6_{33.2}$}&$1.0 \times 10^{0}$&$\phantom{-}0.1$&\textcolor{gray}{$\phantom{-}9.5_{29.4}$}&$1.0 \times 10^{0}$&$\phantom{-}0.0$&\textcolor{gray}{$\phantom{-}9.2_{28.9}$}&$1.0 \times 10^{0}$&$\phantom{-}0.0$&$\phantom{-}8.4_{18.4}$&$\phantom{-}0.9_{9.6}$&$\phantom{-}15.9_{37.0}$\\
    \bottomrule
    \end{tabular}
    }
    \label{tab:adapt_new_hl_p2}
\end{table*}
\begin{table*}[t]
    \centering
    \small
    \caption{\showedits{Adaptability (frequency of conversational behaviors exhibited by therapists \textit{in response to specific client behaviors}) by LLM therapists (GPT-4, GPT-3.5-turbo, Llama2-70b, Llama2-13b) with two simulation methods (Single response, Full conversation) in High-low quality dataset \cite{perez-rosas-etal-2019-makes}. Values in \textcolor{gray}{gray} are not statistically significantly different from human therapist average at $p = \frac{0.05}{m}$ using Two-sided Student’s t-test after Bonferroni correction, where $m$ is the number of t-test per one model ($m = 13 \times 6 = 78$). $d$ is the cohen's d when compared with human therapist average. For instance, when client expresses \textsc{changing unhealthy behavior}, GPT-4 with single response simulation exhibits \textsc{Reflections on Consequences} by 38.9\% which is 15.8\% less than average human therapists. (Single: 11.5\%, Average human therapists: 27.3\% from Table \ref{tab:adapt_new_hl_p1}; $P = 5.3 \times 10^{-17}$, Cohen's $d = -0.4$, two-sided Student’s $t$-test)}}
    \resizebox{1.05\textwidth}{!}{%
    \def\arraystretch{1.15}
    \setlength{\tabcolsep}{2pt}
\begin{tabular}{l|ccc|ccc|ccc|ccc|ccc|ccc|ccc|ccc|ccc|ccc}
         \toprule 
         \multirow{3}{*}{\textbf{Therapist Behavior}} & \multicolumn{12}{c}{\textbf{LLM Therapists (Single Response Simulations)}} & \multicolumn{12}{c}{\textbf{LLM Therapists (Full Conversation Simulations)}} \\
         \cmidrule(lr){2-13}  \cmidrule(lr){14-25} 
         & \multicolumn{3}{c}{\textbf{GPT-4}} &\multicolumn{3}{c}{\textbf{GPT-3.5}}&\multicolumn{3}{c}{\textbf{Llama2-70b}}&\multicolumn{3}{c}{\textbf{Llama2-13b}}& \multicolumn{3}{c}{\textbf{GPT-4}} &\multicolumn{3}{c}{\textbf{GPT-3.5}}&\multicolumn{3}{c}{\textbf{Llama2-70b}}&\multicolumn{3}{c}{\textbf{Llama2-13b}}\\
         \cmidrule(lr){2-4} \cmidrule(lr){5-7} \cmidrule(lr){8-10} \cmidrule(lr){11-13} \cmidrule(lr){14-16} \cmidrule(lr){17-19} \cmidrule(lr){20-22}  \cmidrule(lr){23-25}
        & freq. & $p$ & $d$ & freq. & $p$ & $d$ & freq. & $p$& $d$ & freq. & $p$ & $d$ & freq. & $p$& $d$ &  freq. & $p$ & $d$ & freq. & $p$ & $d$ & freq. & $p$& $d$ \\
        \midrule
          \multicolumn{25}{c}{\textbf{Client behavior: Changing unhealthy behavior}} \\
          \midrule
$n$& \multicolumn{3}{c}{\textbf{$1553$}}& \multicolumn{3}{c}{\textbf{$731$}}& \multicolumn{3}{c}{\textbf{$1553$}}& \multicolumn{3}{c}{\textbf{$595$}}& \multicolumn{3}{c}{\textbf{$1553$}}& \multicolumn{3}{c}{\textbf{$658$}}& \multicolumn{3}{c}{\textbf{$1553$}}& \multicolumn{3}{c}{\textbf{$529$}}
\\
\bottomrule
    \end{tabular}
    }
     \label{tab:adapt_single_full_p1}
\end{table*}
\begin{table*}[t]
    \centering
    \small
    \caption{\showedits{(Continue Table \ref{tab:adapt_single_full_p1}) Adaptability (frequency of conversational behaviors exhibited by therapists \textit{in response to specific client behaviors}) by LLM therapists (GPT-4, GPT-3.5-turbo, Llama2-70b, Llama2-13b) with two simulation methods (Single response, Full conversation) in High-low quality dataset \cite{perez-rosas-etal-2019-makes}. Values in \textcolor{gray}{gray} are not statistically significantly different from human therapist average at $p = \frac{0.05}{m}$ using Two-sided Student’s t-test after Bonferroni correction, where $m$ is the number of t-test per one model ($m = 13 \times 6 = 78$). $d$ is the cohen's d when compared with human therapist average. For instance, when client expresses \textsc{changing unhealthy behavior}, GPT-4 with single response simulation exhibits \textsc{Reflections on Needs} by 38.9\% which is 23.6\% more than average human therapists. (Single: 48.1\%, Average human therapists: 24.5\% from Table \ref{tab:adapt_new_hl_p2}; $P = 2.3 \times 10^{-10}$, Cohen's $d = 0.5$, two-sided Student’s $t$-test)}}
    \resizebox{1.05\textwidth}{!}{%
    \def\arraystretch{1.15}
    \setlength{\tabcolsep}{2pt}
\begin{tabular}{l|ccc|ccc|ccc|ccc|ccc|ccc|ccc|ccc|ccc|ccc}
         \toprule 
         \multirow{3}{*}{\textbf{Therapist Behavior}} & \multicolumn{12}{c}{\textbf{LLM Therapists (Single Response Simulations)}} & \multicolumn{12}{c}{\textbf{LLM Therapists (Full Conversation Simulations)}} \\
         \cmidrule(lr){2-13}  \cmidrule(lr){14-25} 
         & \multicolumn{3}{c}{\textbf{GPT-4}} &\multicolumn{3}{c}{\textbf{GPT-3.5}}&\multicolumn{3}{c}{\textbf{Llama2-70b}}&\multicolumn{3}{c}{\textbf{Llama2-13b}}& \multicolumn{3}{c}{\textbf{GPT-4}} &\multicolumn{3}{c}{\textbf{GPT-3.5}}&\multicolumn{3}{c}{\textbf{Llama2-70b}}&\multicolumn{3}{c}{\textbf{Llama2-13b}}\\
         \cmidrule(lr){2-4} \cmidrule(lr){5-7} \cmidrule(lr){8-10} \cmidrule(lr){11-13} \cmidrule(lr){14-16} \cmidrule(lr){17-19} \cmidrule(lr){20-22}  \cmidrule(lr){23-25}
        & freq. & $p$ & $d$ & freq. & $p$ & $d$ & freq. & $p$& $d$ & freq. & $p$ & $d$ & freq. & $p$& $d$ &  freq. & $p$ & $d$ & freq. & $p$ & $d$ & freq. & $p$& $d$ \\
        \midrule
          \multicolumn{25}{c}{\textbf{Client behavior: Sharing negative emotions}} \\
          \midrule
          $n$& \multicolumn{3}{c}{\textbf{$820$}}& \multicolumn{3}{c}{\textbf{$243$}}& \multicolumn{3}{c}{\textbf{$820$}}& \multicolumn{3}{c}{\textbf{$196$}}& \multicolumn{3}{c}{\textbf{$820$}}& \multicolumn{3}{c}{\textbf{$416$}}& \multicolumn{3}{c}{\textbf{$820$}}& \multicolumn{3}{c}{\textbf{$294$}}\\
\bottomrule
    \end{tabular}
    }
    \label{tab:adapt_single_full_p2}
\end{table*}
\begin{table*}[t]
    \centering
    \small
    \caption{\showedits{Frequency of linguistic attributes (from LIWC analysis) by LLM therapists (GPT-4, GPT-3.5-turbo, Llama2-70b, Llama2-13b) in High-low quality dataset \cite{perez-rosas-etal-2019-makes}. Values in \textcolor{gray}{gray} are not statistically significantly different from human therapist average at $p = \frac{0.05}{m}$ using Two-sided Student’s t-test after Bonferroni correction, where $m$ is the number of t-test per one model ($m = 14$). $d$ is the cohen's d when compared with human therapist average. For instance, GPT-4 exhibits \textsc{Positive Emotions} by 1.5\% while human therapists responds by 1.0\% on average ($P = 8.4 \times 10^{-13}$, Cohen's $d = \num{0.1}$, two-sided Student’s $t$-test).}}
    \resizebox{1\textwidth}{!}{%
    \def\arraystretch{1.15}
    \setlength{\tabcolsep}{2pt}
    \begin{tabular}{l|ccc|ccc|ccc|ccc|ccc|ccc}
         \toprule 
         \multirow{4}{*}{\textbf{Linguistic Attributes}} & \multicolumn{12}{c}{\textbf{LLM Therapists (Single Response \& Full Conversation Simulations)}} & \multicolumn{3}{c}{\textbf{Human Therapists}} \\

         \cmidrule(lr){2-13} \cmidrule(lr){14-16}
         & \multicolumn{3}{c}{\textbf{GPT-4}} &\multicolumn{3}{c}{\textbf{GPT-3.5}}&\multicolumn{3}{c}{\textbf{Llama2-70b}}&\multicolumn{3}{c}{\textbf{Llama2-13b}}& \textbf{Average}&\textbf{High-quality}&\textbf{Low-quality}\\
& \multicolumn{3}{c}{\textbf{$(n = 4893)$}}& \multicolumn{3}{c}{\textbf{$(n = 4746)$}}& \multicolumn{3}{c}{\textbf{$(n = 4970)$}}& \multicolumn{3}{c}{\textbf{$(n = 4820)$}}&\textbf{$(n = 5446)$}&\textbf{$(n = 3907)$}&\textbf{$(n = 1539)$}\\
         \cmidrule(lr){2-4} \cmidrule(lr){5-7} \cmidrule(lr){8-10} \cmidrule(lr){11-13} \cmidrule(lr){14-14} \cmidrule(lr){15-15}  \cmidrule(lr){16-16} 
        & freq. & $p$ & $d$ & freq. & $p$ & $d$ & freq. & $p$& $d$ & freq. & $p$ & $d$ & freq. & freq. & freq. \\
         \midrule
Positive Emotions&$\phantom{-}1.5_{2.1}$&$8.4 \times 10^{-13}$&$\phantom{-}0.1$&$\phantom{-}1.2_{1.8}$&$9.2 \times 10^{-5}$&$\phantom{-}0.1$&\textcolor{gray}{$\phantom{-}1.1_{3.1}$}&$1.0 \times 10^{0}$&$\phantom{-}0.0$&\textcolor{gray}{$\phantom{-}1.1_{3.3}$}&$2.9 \times 10^{-1}$&$\phantom{-}0.0$&$\phantom{-}1.0_{4.2}$&$\phantom{-}1.2_{4.3}$&$\phantom{-}0.7_{3.6}$\\
Negative Emotions&\textcolor{gray}{$\phantom{-}0.8_{1.5}$}&$1.0 \times 10^{0}$&$\phantom{-}0.0$&$\phantom{-}0.9_{1.6}$&$9.3 \times 10^{-6}$&$\phantom{-}0.1$&\textcolor{gray}{$\phantom{-}0.8_{1.9}$}&$6.4 \times 10^{-2}$&$\phantom{-}0.1$&\textcolor{gray}{$\phantom{-}0.6_{1.8}$}&$1.0 \times 10^{0}$&$\phantom{-}0.0$&$\phantom{-}0.7_{3.6}$&$\phantom{-}0.7_{3.4}$&$\phantom{-}0.6_{4.3}$\\
BigWords&$\phantom{-}24.5_{7.1}$&$0.0 \times 10^{0}$&$\phantom{-}1.6$&$\phantom{-}23.1_{7.7}$&$0.0 \times 10^{0}$&$\phantom{-}1.4$&$\phantom{-}23.7_{11.0}$&$0.0 \times 10^{0}$&$\phantom{-}1.2$&$\phantom{-}19.8_{12.6}$&$0.0 \times 10^{0}$&$\phantom{-}0.8$&$\phantom{-}10.7_{9.9}$&$\phantom{-}11.0_{9.9}$&$\phantom{-}10.4_{10.1}$\\
Self&$\phantom{-}1.5_{2.3}$&$1.1 \times 10^{-84}$&$-0.4$&$\phantom{-}1.8_{2.6}$&$4.3 \times 10^{-68}$&$-0.3$&$\phantom{-}1.3_{2.4}$&$7.9 \times 10^{-134}$&$-0.5$&$\phantom{-}1.9_{3.0}$&$9.3 \times 10^{-61}$&$-0.3$&$\phantom{-}3.1_{5.1}$&$\phantom{-}2.5_{4.5}$&$\phantom{-}3.8_{6.1}$\\
Other&\textcolor{gray}{$\phantom{-}8.7_{3.3}$}&$8.2 \times 10^{-1}$&$\phantom{-}0.0$&$\phantom{-}8.2_{3.3}$&$8.7 \times 10^{-12}$&$-0.1$&$\phantom{-}7.5_{4.8}$&$2.4 \times 10^{-38}$&$-0.2$&$\phantom{-}6.7_{4.9}$&$2.0 \times 10^{-85}$&$-0.4$&$\phantom{-}8.9_{7.1}$&$\phantom{-}8.7_{6.9}$&$\phantom{-}9.2_{7.4}$\\
Health&$\phantom{-}2.7_{2.8}$&$1.1 \times 10^{-30}$&$\phantom{-}0.2$&$\phantom{-}3.0_{3.3}$&$1.6 \times 10^{-60}$&$\phantom{-}0.3$&$\phantom{-}2.0_{2.9}$&$4.8 \times 10^{-2}$&$\phantom{-}0.1$&\textcolor{gray}{$\phantom{-}1.7_{3.9}$}&$6.1 \times 10^{-1}$&$\phantom{-}0.0$&$\phantom{-}1.8_{4.3}$&$\phantom{-}1.2_{3.2}$&$\phantom{-}2.5_{6.1}$\\
Wellness&$\phantom{-}0.9_{1.6}$&$8.8 \times 10^{-62}$&$\phantom{-}0.3$&$\phantom{-}0.8_{1.6}$&$2.9 \times 10^{-51}$&$\phantom{-}0.3$&$\phantom{-}0.6_{1.3}$&$1.1 \times 10^{-14}$&$\phantom{-}0.1$&$\phantom{-}0.5_{1.9}$&$1.2 \times 10^{-3}$&$\phantom{-}0.1$&$\phantom{-}0.3_{2.1}$&$\phantom{-}0.2_{1.3}$&$\phantom{-}0.4_{3.5}$\\
Behavioral Activation&$\phantom{-}6.8_{4.0}$&$7.8 \times 10^{-260}$&$\phantom{-}0.7$&$\phantom{-}5.9_{4.3}$&$4.5 \times 10^{-175}$&$\phantom{-}0.5$&$\phantom{-}5.7_{5.2}$&$7.6 \times 10^{-129}$&$\phantom{-}0.5$&$\phantom{-}5.1_{5.7}$&$4.6 \times 10^{-69}$&$\phantom{-}0.3$&$\phantom{-}3.2_{5.9}$&$\phantom{-}3.5_{5.9}$&$\phantom{-}2.9_{5.9}$\\
Meaning&$\phantom{-}3.5_{2.3}$&$1.4 \times 10^{-253}$&$\phantom{-}0.7$&$\phantom{-}3.1_{2.2}$&$1.1 \times 10^{-209}$&$\phantom{-}0.6$&$\phantom{-}2.1_{2.4}$&$4.7 \times 10^{-38}$&$\phantom{-}0.2$&$\phantom{-}1.9_{2.7}$&$6.5 \times 10^{-15}$&$\phantom{-}0.2$&$\phantom{-}1.4_{3.5}$&$\phantom{-}1.4_{3.4}$&$\phantom{-}1.4_{3.7}$\\
Purpose&$\phantom{-}3.8_{2.6}$&$1.3 \times 10^{-285}$&$\phantom{-}0.7$&$\phantom{-}3.5_{2.5}$&$1.8 \times 10^{-257}$&$\phantom{-}0.7$&$\phantom{-}2.5_{2.6}$&$3.5 \times 10^{-63}$&$\phantom{-}0.3$&$\phantom{-}2.2_{3.0}$&$2.4 \times 10^{-27}$&$\phantom{-}0.2$&$\phantom{-}1.5_{3.6}$&$\phantom{-}1.5_{3.5}$&$\phantom{-}1.5_{3.8}$\\
Motivation&$\phantom{-}3.4_{2.4}$&$6.8 \times 10^{-264}$&$\phantom{-}0.7$&$\phantom{-}3.1_{2.3}$&$9.2 \times 10^{-248}$&$\phantom{-}0.6$&$\phantom{-}2.1_{2.3}$&$7.1 \times 10^{-56}$&$\phantom{-}0.3$&$\phantom{-}1.9_{2.8}$&$4.8 \times 10^{-26}$&$\phantom{-}0.2$&$\phantom{-}1.3_{3.3}$&$\phantom{-}1.3_{3.3}$&$\phantom{-}1.3_{3.4}$\\
Sadness&$\phantom{-}0.7_{1.3}$&$1.4 \times 10^{-37}$&$\phantom{-}0.3$&$\phantom{-}0.8_{1.4}$&$3.5 \times 10^{-67}$&$\phantom{-}0.3$&$\phantom{-}0.6_{1.5}$&$1.5 \times 10^{-33}$&$\phantom{-}0.2$&$\phantom{-}0.5_{1.6}$&$9.3 \times 10^{-13}$&$\phantom{-}0.1$&$\phantom{-}0.3_{1.5}$&$\phantom{-}0.3_{1.5}$&$\phantom{-}0.2_{1.6}$\\
Sympathy&$\phantom{-}0.3_{0.7}$&$4.1 \times 10^{-11}$&$\phantom{-}0.1$&$\phantom{-}0.4_{0.9}$&$8.5 \times 10^{-41}$&$\phantom{-}0.3$&$\phantom{-}0.3_{1.1}$&$6.7 \times 10^{-18}$&$\phantom{-}0.2$&$\phantom{-}0.3_{1.1}$&$1.4 \times 10^{-9}$&$\phantom{-}0.1$&$\phantom{-}0.1_{1.0}$&$\phantom{-}0.2_{1.1}$&$\phantom{-}0.1_{0.8}$\\
Stress&\textcolor{gray}{$\phantom{-}1.1_{1.8}$}&$1.0 \times 10^{0}$&$\phantom{-}0.0$&$\phantom{-}1.4_{1.9}$&$2.7 \times 10^{-7}$&$\phantom{-}0.1$&$\phantom{-}1.4_{2.2}$&$3.6 \times 10^{-8}$&$\phantom{-}0.1$&\textcolor{gray}{$\phantom{-}1.1_{2.2}$}&$1.0 \times 10^{0}$&$\phantom{-}0.0$&$\phantom{-}1.1_{3.5}$&$\phantom{-}1.1_{3.2}$&$\phantom{-}1.1_{4.2}$\\

    \bottomrule
    \end{tabular}
    }
    \label{tab:liwc_new_hl}
\end{table*}
\begin{table*}[t]
    \centering
    \small
    \caption{\showedits{Frequency of linguistic attributes (from LIWC analysis) by LLM therapists (GPT-4, GPT-3.5-turbo, Llama2-70b, Llama2-13b) in High-low quality dataset \cite{perez-rosas-etal-2019-makes} for two different types of simulations separately (Single response and Full conversation). Values in \textcolor{gray}{gray} are not statistically significantly different from human therapist average at $p = \frac{0.05}{m}$ using Two-sided Student’s t-test after Bonferroni correction, where $m$ is the number of t-test per one model ($m = 13$). $d$ is the cohen's d when compared with human therapist average. For instance, GPT-4 with single response simulation responds with \textsc{Positive Emotions} by 2.0\% and GPT-4 with full conversation simulation responds by 1.3\% when compared with the human therapists responds by 1.0\% on average from Table \ref{tab:liwc_new_hl}. (Single: $P = 2.0 \times 10^{-14}$, Cohen's $d = \num{0.3}$; two-sided student t-test) (Full: $P = 3.7 \times 10^{-5}$, Cohen's $d = \num{0.1}$; two-sided student t-test).
}}
    
    \resizebox{1.05\textwidth}{!}{%
    \def\arraystretch{1.15}
    \setlength{\tabcolsep}{2pt}
    \begin{tabular}{l|ccc|ccc|ccc|ccc|ccc|ccc|ccc|ccc|ccc|ccc}
         \toprule 
         \multirow{3}{*}{\textbf{Behavior}} & \multicolumn{12}{c}{\textbf{LLM Therapists (Single Response Simulations)}} & \multicolumn{12}{c}{\textbf{LLM Therapists (Full Conversation Simulations)}} \\
         \cmidrule(lr){2-13}  \cmidrule(lr){14-25} 
         & \multicolumn{3}{c}{\textbf{GPT-4}} &\multicolumn{3}{c}{\textbf{GPT-3.5}}&\multicolumn{3}{c}{\textbf{Llama2-70b}}&\multicolumn{3}{c}{\textbf{Llama2-13b}}& \multicolumn{3}{c}{\textbf{GPT-4}} &\multicolumn{3}{c}{\textbf{GPT-3.5}}&\multicolumn{3}{c}{\textbf{Llama2-70b}}&\multicolumn{3}{c}{\textbf{Llama2-13b}}\\
& \multicolumn{3}{c}{\textbf{$(n = 3837)$}}& \multicolumn{3}{c}{\textbf{$(n = 3907)$}}& \multicolumn{3}{c}{\textbf{$(n = 3906)$}}& \multicolumn{3}{c}{\textbf{$(n = 3895)$}}& \multicolumn{3}{c}{\textbf{$(n = 1056)$}}& \multicolumn{3}{c}{\textbf{$(n = 839)$}}& \multicolumn{3}{c}{\textbf{$(n = 1064)$}}& \multicolumn{3}{c}{\textbf{$(n = 925)$}}\\
         \cmidrule(lr){2-4} \cmidrule(lr){5-7} \cmidrule(lr){8-10} \cmidrule(lr){11-13} \cmidrule(lr){14-16} \cmidrule(lr){17-19} \cmidrule(lr){20-22}  \cmidrule(lr){23-25}
        & freq. & $p$ & $d$ & freq. & $p$ & $d$ & freq. & $p$& $d$ & freq. & $p$ & $d$ & freq. & $p$& $d$ &  freq. & $p$ & $d$ & freq. & $p$ & $d$ & freq. & $p$& $d$ \\
         \midrule
Positive Emotions&$\phantom{-}2.0_{1.7}$&$2.0 \times 10^{-14}$&$\phantom{-}0.3$&$\phantom{-}1.7_{1.7}$&$1.5 \times 10^{-5}$&$\phantom{-}0.2$&\textcolor{gray}{$\phantom{-}1.3_{1.4}$}&$5.2 \times 10^{-2}$&$\phantom{-}0.1$&$\phantom{-}1.7_{2.1}$&$3.2 \times 10^{-6}$&$\phantom{-}0.2$&$\phantom{-}1.3_{2.2}$&$3.7 \times 10^{-5}$&$\phantom{-}0.1$&$\phantom{-}1.2_{1.8}$&$2.5 \times 10^{-2}$&$\phantom{-}0.1$&\textcolor{gray}{$\phantom{-}1.0_{3.3}$}&$1.0 \times 10^{0}$&$\phantom{-}0.0$&\textcolor{gray}{$\phantom{-}1.0_{3.5}$}&$1.0 \times 10^{0}$&$\phantom{-}0.0$\\
Negative Emotions&\textcolor{gray}{$\phantom{-}0.5_{1.0}$}&$1.0 \times 10^{0}$&$-0.1$&\textcolor{gray}{$\phantom{-}0.6_{1.1}$}&$1.0 \times 10^{0}$&$\phantom{-}0.0$&\textcolor{gray}{$\phantom{-}0.7_{1.1}$}&$1.0 \times 10^{0}$&$\phantom{-}0.0$&\textcolor{gray}{$\phantom{-}0.6_{1.1}$}&$1.0 \times 10^{0}$&$\phantom{-}0.0$&\textcolor{gray}{$\phantom{-}0.8_{1.6}$}&$2.7 \times 10^{-1}$&$\phantom{-}0.0$&$\phantom{-}1.0_{1.7}$&$1.7 \times 10^{-7}$&$\phantom{-}0.1$&$\phantom{-}0.9_{2.0}$&$2.5 \times 10^{-2}$&$\phantom{-}0.1$&\textcolor{gray}{$\phantom{-}0.6_{1.9}$}&$1.0 \times 10^{0}$&$\phantom{-}0.0$\\
BigWords&$\phantom{-}28.2_{4.8}$&$0.0 \times 10^{0}$&$\phantom{-}1.9$&$\phantom{-}25.1_{6.8}$&$0.0 \times 10^{0}$&$\phantom{-}1.5$&$\phantom{-}32.5_{11.1}$&$0.0 \times 10^{0}$&$\phantom{-}2.2$&$\phantom{-}32.0_{16.0}$&$0.0 \times 10^{0}$&$\phantom{-}1.9$&$\phantom{-}23.5_{7.3}$&$0.0 \times 10^{0}$&$\phantom{-}1.4$&$\phantom{-}22.8_{7.7}$&$0.0 \times 10^{0}$&$\phantom{-}1.4$&$\phantom{-}22.0_{10.2}$&$0.0 \times 10^{0}$&$\phantom{-}1.1$&$\phantom{-}17.8_{10.6}$&$3.1 \times 10^{-268}$&$\phantom{-}0.7$\\
Self&$\phantom{-}2.2_{1.9}$&$3.1 \times 10^{-8}$&$-0.2$&$\phantom{-}2.4_{2.3}$&$1.1 \times 10^{-3}$&$-0.1$&$\phantom{-}1.0_{1.6}$&$7.0 \times 10^{-39}$&$-0.4$&$\phantom{-}1.3_{1.8}$&$8.9 \times 10^{-25}$&$-0.4$&$\phantom{-}1.4_{2.4}$&$7.5 \times 10^{-84}$&$-0.4$&$\phantom{-}1.7_{2.6}$&$2.1 \times 10^{-69}$&$-0.3$&$\phantom{-}1.4_{2.5}$&$2.1 \times 10^{-106}$&$-0.4$&$\phantom{-}1.9_{3.2}$&$4.8 \times 10^{-46}$&$-0.3$\\
Other&$\phantom{-}9.6_{2.3}$&$2.9 \times 10^{-2}$&$\phantom{-}0.1$&\textcolor{gray}{$\phantom{-}8.4_{3.0}$}&$3.4 \times 10^{-1}$&$-0.1$&$\phantom{-}5.8_{3.4}$&$2.1 \times 10^{-43}$&$-0.5$&$\phantom{-}5.6_{4.8}$&$1.3 \times 10^{-41}$&$-0.5$&$\phantom{-}8.5_{3.5}$&$3.2 \times 10^{-3}$&$-0.1$&$\phantom{-}8.2_{3.4}$&$3.3 \times 10^{-11}$&$-0.1$&$\phantom{-}7.8_{5.0}$&$3.5 \times 10^{-20}$&$-0.2$&$\phantom{-}6.9_{4.9}$&$1.0 \times 10^{-64}$&$-0.3$\\
Health&$\phantom{-}2.8_{2.3}$&$3.8 \times 10^{-12}$&$\phantom{-}0.2$&$\phantom{-}3.2_{3.1}$&$4.0 \times 10^{-18}$&$\phantom{-}0.3$&\textcolor{gray}{$\phantom{-}1.9_{2.0}$}&$1.0 \times 10^{0}$&$\phantom{-}0.0$&\textcolor{gray}{$\phantom{-}1.6_{2.3}$}&$1.0 \times 10^{0}$&$-0.1$&$\phantom{-}2.7_{2.9}$&$6.7 \times 10^{-23}$&$\phantom{-}0.2$&$\phantom{-}3.0_{3.4}$&$3.3 \times 10^{-52}$&$\phantom{-}0.3$&$\phantom{-}2.1_{3.0}$&$2.3 \times 10^{-2}$&$\phantom{-}0.1$&\textcolor{gray}{$\phantom{-}1.7_{4.1}$}&$1.0 \times 10^{0}$&$\phantom{-}0.0$\\
Wellness&$\phantom{-}1.1_{1.5}$&$1.7 \times 10^{-31}$&$\phantom{-}0.4$&$\phantom{-}1.2_{1.8}$&$1.4 \times 10^{-31}$&$\phantom{-}0.4$&$\phantom{-}0.7_{1.2}$&$5.6 \times 10^{-8}$&$\phantom{-}0.2$&$\phantom{-}0.8_{1.5}$&$1.6 \times 10^{-8}$&$\phantom{-}0.2$&$\phantom{-}0.9_{1.6}$&$2.0 \times 10^{-43}$&$\phantom{-}0.3$&$\phantom{-}0.8_{1.6}$&$1.3 \times 10^{-37}$&$\phantom{-}0.3$&$\phantom{-}0.5_{1.4}$&$2.2 \times 10^{-10}$&$\phantom{-}0.1$&\textcolor{gray}{$\phantom{-}0.4_{2.0}$}&$2.0 \times 10^{-1}$&$\phantom{-}0.0$\\
Behavioral Activation&$\phantom{-}7.8_{3.0}$&$4.0 \times 10^{-128}$&$\phantom{-}0.8$&$\phantom{-}8.9_{4.5}$&$4.0 \times 10^{-147}$&$\phantom{-}1.0$&$\phantom{-}6.9_{3.8}$&$1.1 \times 10^{-84}$&$\phantom{-}0.7$&$\phantom{-}6.4_{4.2}$&$1.4 \times 10^{-54}$&$\phantom{-}0.6$&$\phantom{-}6.5_{4.2}$&$2.6 \times 10^{-182}$&$\phantom{-}0.6$&$\phantom{-}5.4_{4.1}$&$3.5 \times 10^{-116}$&$\phantom{-}0.4$&$\phantom{-}5.4_{5.5}$&$6.3 \times 10^{-93}$&$\phantom{-}0.4$&$\phantom{-}4.9_{5.9}$&$6.4 \times 10^{-49}$&$\phantom{-}0.3$\\
Meaning&$\phantom{-}3.6_{1.8}$&$5.0 \times 10^{-84}$&$\phantom{-}0.7$&$\phantom{-}3.4_{2.1}$&$7.6 \times 10^{-60}$&$\phantom{-}0.6$&$\phantom{-}2.3_{1.7}$&$1.6 \times 10^{-16}$&$\phantom{-}0.3$&$\phantom{-}2.3_{1.8}$&$1.9 \times 10^{-12}$&$\phantom{-}0.3$&$\phantom{-}3.4_{2.4}$&$1.3 \times 10^{-201}$&$\phantom{-}0.7$&$\phantom{-}3.0_{2.1}$&$6.8 \times 10^{-177}$&$\phantom{-}0.6$&$\phantom{-}2.1_{2.5}$&$5.9 \times 10^{-29}$&$\phantom{-}0.2$&$\phantom{-}1.8_{2.8}$&$1.0 \times 10^{-9}$&$\phantom{-}0.1$\\
Purpose&$\phantom{-}4.3_{1.9}$&$1.1 \times 10^{-124}$&$\phantom{-}0.8$&$\phantom{-}4.7_{2.7}$&$9.1 \times 10^{-126}$&$\phantom{-}0.9$&$\phantom{-}2.8_{2.0}$&$1.3 \times 10^{-29}$&$\phantom{-}0.4$&$\phantom{-}2.7_{2.0}$&$1.3 \times 10^{-20}$&$\phantom{-}0.3$&$\phantom{-}3.7_{2.7}$&$2.3 \times 10^{-211}$&$\phantom{-}0.7$&$\phantom{-}3.3_{2.4}$&$2.4 \times 10^{-199}$&$\phantom{-}0.6$&$\phantom{-}2.4_{2.7}$&$8.3 \times 10^{-48}$&$\phantom{-}0.3$&$\phantom{-}2.1_{3.1}$&$9.6 \times 10^{-19}$&$\phantom{-}0.2$\\
Motivation&$\phantom{-}3.9_{1.8}$&$6.7 \times 10^{-124}$&$\phantom{-}0.8$&$\phantom{-}4.3_{2.6}$&$5.4 \times 10^{-128}$&$\phantom{-}0.9$&$\phantom{-}2.4_{1.8}$&$1.7 \times 10^{-23}$&$\phantom{-}0.3$&$\phantom{-}2.3_{1.9}$&$9.7 \times 10^{-17}$&$\phantom{-}0.3$&$\phantom{-}3.2_{2.5}$&$1.8 \times 10^{-190}$&$\phantom{-}0.6$&$\phantom{-}2.9_{2.2}$&$1.8 \times 10^{-190}$&$\phantom{-}0.6$&$\phantom{-}2.1_{2.4}$&$2.9 \times 10^{-43}$&$\phantom{-}0.3$&$\phantom{-}1.8_{2.9}$&$9.6 \times 10^{-19}$&$\phantom{-}0.2$\\
Sadness&$\phantom{-}0.5_{0.9}$&$6.2 \times 10^{-5}$&$\phantom{-}0.2$&$\phantom{-}0.5_{0.9}$&$2.6 \times 10^{-2}$&$\phantom{-}0.1$&$\phantom{-}0.6_{1.0}$&$7.5 \times 10^{-10}$&$\phantom{-}0.2$&$\phantom{-}0.7_{1.4}$&$5.8 \times 10^{-12}$&$\phantom{-}0.3$&$\phantom{-}0.7_{1.4}$&$8.1 \times 10^{-38}$&$\phantom{-}0.3$&$\phantom{-}0.8_{1.5}$&$3.0 \times 10^{-72}$&$\phantom{-}0.3$&$\phantom{-}0.6_{1.6}$&$6.7 \times 10^{-30}$&$\phantom{-}0.2$&$\phantom{-}0.5_{1.7}$&$7.1 \times 10^{-9}$&$\phantom{-}0.1$\\
Sympathy&\textcolor{gray}{$\phantom{-}0.2_{0.5}$}&$2.9 \times 10^{-1}$&$\phantom{-}0.1$&\textcolor{gray}{$\phantom{-}0.2_{0.6}$}&$5.9 \times 10^{-1}$&$\phantom{-}0.1$&$\phantom{-}0.3_{0.7}$&$1.5 \times 10^{-6}$&$\phantom{-}0.2$&$\phantom{-}0.4_{1.3}$&$7.2 \times 10^{-12}$&$\phantom{-}0.3$&$\phantom{-}0.3_{0.8}$&$4.4 \times 10^{-11}$&$\phantom{-}0.1$&$\phantom{-}0.4_{1.0}$&$9.8 \times 10^{-45}$&$\phantom{-}0.3$&$\phantom{-}0.3_{1.2}$&$8.8 \times 10^{-16}$&$\phantom{-}0.2$&$\phantom{-}0.2_{1.0}$&$4.2 \times 10^{-6}$&$\phantom{-}0.1$\\
Stress&$\phantom{-}0.7_{1.2}$&$1.2 \times 10^{-2}$&$-0.1$&\textcolor{gray}{$\phantom{-}0.9_{1.4}$}&$6.5 \times 10^{-1}$&$-0.1$&\textcolor{gray}{$\phantom{-}1.2_{1.5}$}&$1.0 \times 10^{0}$&$\phantom{-}0.0$&\textcolor{gray}{$\phantom{-}1.0_{1.5}$}&$1.0 \times 10^{0}$&$\phantom{-}0.0$&\textcolor{gray}{$\phantom{-}1.2_{1.9}$}&$2.3 \times 10^{-1}$&$\phantom{-}0.1$&$\phantom{-}1.5_{1.9}$&$1.5 \times 10^{-10}$&$\phantom{-}0.1$&$\phantom{-}1.4_{2.3}$&$1.9 \times 10^{-8}$&$\phantom{-}0.1$&\textcolor{gray}{$\phantom{-}1.1_{2.3}$}&$1.0 \times 10^{0}$&$\phantom{-}0.0$\\

    \bottomrule
    \end{tabular}
    }
    \label{tab:liwc_new_single_full_hl}
\end{table*}

\begin{table*}[t]
    \centering
    \small
    \caption{\showedits{Frequency of conversational behaviors exhibited by LLM therapists (GPT-4, GPT-3.5-turbo, Llama2-70b, Llama2-13b) in HOPE dataset \cite{malhotra2022speaker}. Values in \textcolor{gray}{gray} are not statistically significantly different from human therapist average at $p = \frac{0.05}{m}$ using Two-sided Student’s t-test after Bonferroni correction, where $m$ is the number of t-test per one model ($m = 13$). $d$ is the cohen's d when compared with human therapist average. For instance, GPT-4 responds with \textsc{Reflections on Needs} by 40.2\% while human therapists responds by 23.4\% on average ($P = 6.9 \times 10^{-90}$, Cohen's $d = \num{0.4}$, two-sided Student’s $t$-test).}}
    \resizebox{1\textwidth}{!}{%
    \def\arraystretch{1.15}
    \setlength{\tabcolsep}{2pt}
    \begin{tabular}{l|ccc|ccc|ccc|ccc|ccc|ccc}
         \toprule 
         \multirow{3}{*}{\textbf{Behavior}} & \multicolumn{12}{c}{\textbf{LLM Therapists (Single Response \& Full Conversation Simulations)}} & \multicolumn{3}{c}{\textbf{Human Therapists}} \\
         \cmidrule(lr){2-13} \cmidrule(lr){14-16}
         & \multicolumn{3}{c}{\textbf{GPT-4}} &\multicolumn{3}{c}{\textbf{GPT-3.5}}&\multicolumn{3}{c}{\textbf{Llama2-70b}}&\multicolumn{3}{c}{\textbf{Llama2-13b}}& \textbf{Average}&\textbf{High-quality}&\textbf{Low-quality}\\
& \multicolumn{3}{c}{\textbf{$(n = 7705)$}}& \multicolumn{3}{c}{\textbf{$(n = 7135)$}}& \multicolumn{3}{c}{\textbf{$(n = 7461)$}}& \multicolumn{3}{c}{\textbf{$(n = 7464)$}}&\textbf{$(n = 5446)$}&\textbf{$(n = 3907)$}&\textbf{$(n = 1539)$}\\
         \cmidrule(lr){2-4} \cmidrule(lr){5-7} \cmidrule(lr){8-10} \cmidrule(lr){11-13} \cmidrule(lr){14-14} \cmidrule(lr){15-15}  \cmidrule(lr){16-16} 
        & freq. & $p$ & $d$ & freq. & $p$ & $d$ & freq. & $p$& $d$ & freq. & $p$ & $d$ & freq. & freq. & freq. \\
         \midrule
Reflections on Needs&$\phantom{-}40.2_{49.0}$&$6.9 \times 10^{-90}$&$\phantom{-}0.4$&$\phantom{-}35.0_{47.7}$&$7.2 \times 10^{-44}$&$\phantom{-}0.3$&$\phantom{-}29.0_{45.4}$&$2.6 \times 10^{-11}$&$\phantom{-}0.1$&$\phantom{-}28.8_{45.3}$&$9.1 \times 10^{-11}$&$\phantom{-}0.1$&$\phantom{-}23.4_{43.1}$&$\phantom{-}26.2_{44.0}$&$\phantom{-}20.6_{40.5}$\\

Reflections on Emotions&$\phantom{-}23.1_{42.2}$&$1.2 \times 10^{-74}$&$\phantom{-}0.3$&$\phantom{-}25.2_{43.4}$&$8.4 \times 10^{-95}$&$\phantom{-}0.4$&$\phantom{-}17.1_{37.6}$&$3.1 \times 10^{-24}$&$\phantom{-}0.2$&$\phantom{-}21.3_{41.0}$&$4.5 \times 10^{-57}$&$\phantom{-}0.3$&$\phantom{-}10.5_{32.8}$&$\phantom{-}14.6_{35.3}$&$\phantom{-}6.3_{24.3}$\\

Reflections on Values&\textcolor{gray}{$\phantom{-}4.4_{20.6}$}&$1.0 \times 10^{0}$&$\phantom{-}0.0$&$\phantom{-}3.5_{18.4}$&$3.9 \times 10^{-3}$&$-0.1$&$\phantom{-}3.7_{19.0}$&$4.0 \times 10^{-2}$&$-0.1$&\textcolor{gray}{$\phantom{-}4.1_{19.8}$}&$6.2 \times 10^{-1}$&$\phantom{-}0.0$&$\phantom{-}4.8_{22.8}$&$\phantom{-}6.3_{24.4}$&$\phantom{-}3.3_{17.9}$\\

Reflections on Consequences&\textcolor{gray}{$\phantom{-}22.3_{41.7}$}&$1.0 \times 10^{0}$&$\phantom{-}0.0$&\textcolor{gray}{$\phantom{-}21.4_{41.0}$}&$1.0 \times 10^{0}$&$\phantom{-}0.0$&$\phantom{-}19.6_{39.7}$&$3.2 \times 10^{-2}$&$-0.1$&\textcolor{gray}{$\phantom{-}21.2_{40.9}$}&$1.0 \times 10^{0}$&$\phantom{-}0.0$&$\phantom{-}21.8_{40.1}$&$\phantom{-}17.9_{38.4}$&$\phantom{-}25.7_{43.7}$\\

Reflections on Conflicts&$\phantom{-}4.1_{19.8}$&$2.9 \times 10^{-9}$&$-0.1$&$\phantom{-}3.5_{18.5}$&$9.1 \times 10^{-14}$&$-0.1$&\textcolor{gray}{$\phantom{-}6.3_{24.2}$}&$1.0 \times 10^{0}$&$\phantom{-}0.0$&\textcolor{gray}{$\phantom{-}6.6_{24.8}$}&$1.0 \times 10^{0}$&$\phantom{-}0.0$&$\phantom{-}6.6_{26.1}$&$\phantom{-}8.3_{27.6}$&$\phantom{-}5.0_{21.8}$\\

Reflections on Strengths&$\phantom{-}33.5_{47.2}$&$4.2 \times 10^{-225}$&$\phantom{-}0.6$&$\phantom{-}27.4_{44.6}$&$1.4 \times 10^{-135}$&$\phantom{-}0.5$&$\phantom{-}16.8_{37.4}$&$1.9 \times 10^{-30}$&$\phantom{-}0.2$&$\phantom{-}15.6_{36.3}$&$5.8 \times 10^{-22}$&$\phantom{-}0.2$&$\phantom{-}9.5_{31.3}$&$\phantom{-}12.8_{33.5}$&$\phantom{-}6.2_{24.2}$\\

Questions on Experiences&$\phantom{-}36.7_{48.2}$&$6.0 \times 10^{-16}$&$-0.1$&$\phantom{-}29.5_{45.6}$&$6.2 \times 10^{-63}$&$-0.3$&$\phantom{-}57.6_{49.4}$&$6.7 \times 10^{-52}$&$\phantom{-}0.3$&$\phantom{-}56.6_{49.6}$&$1.5 \times 10^{-44}$&$\phantom{-}0.3$&$\phantom{-}44.0_{49.9}$&$\phantom{-}48.9_{50.0}$&$\phantom{-}39.1_{48.8}$\\

Questions on Perspectives&$\phantom{-}7.6_{26.6}$&$2.3 \times 10^{-21}$&$\phantom{-}0.2$&\textcolor{gray}{$\phantom{-}3.8_{19.1}$}&$1.0 \times 10^{0}$&$\phantom{-}0.0$&$\phantom{-}12.1_{32.6}$&$1.4 \times 10^{-65}$&$\phantom{-}0.3$&$\phantom{-}11.3_{31.7}$&$2.2 \times 10^{-56}$&$\phantom{-}0.3$&$\phantom{-}3.5_{19.9}$&$\phantom{-}4.9_{21.7}$&$\phantom{-}2.0_{14.1}$\\

Questions on Emotions&$\phantom{-}3.0_{17.1}$&$4.7 \times 10^{-84}$&$-0.3$&$\phantom{-}6.1_{24.0}$&$3.7 \times 10^{-27}$&$-0.2$&\textcolor{gray}{$\phantom{-}12.5_{33.0}$}&$1.0 \times 10^{0}$&$\phantom{-}0.0$&$\phantom{-}18.1_{38.5}$&$7.7 \times 10^{-22}$&$\phantom{-}0.2$&$\phantom{-}11.7_{33.0}$&$\phantom{-}13.3_{34.0}$&$\phantom{-}10.1_{30.2}$\\

Problem-Solving&$\phantom{-}53.4_{49.9}$&$1.7 \times 10^{-229}$&$\phantom{-}0.6$&$\phantom{-}60.1_{49.0}$&$0.0 \times 10^{0}$&$\phantom{-}0.7$&$\phantom{-}33.7_{47.3}$&$6.9 \times 10^{-21}$&$\phantom{-}0.2$&\textcolor{gray}{$\phantom{-}24.3_{42.9}$}&$5.4 \times 10^{-1}$&$\phantom{-}0.0$&$\phantom{-}25.9_{42.6}$&$\phantom{-}21.2_{40.9}$&$\phantom{-}30.5_{46.0}$\\

Planning&$\phantom{-}47.9_{50.0}$&$3.6 \times 10^{-306}$&$\phantom{-}0.7$&$\phantom{-}53.3_{49.9}$&$0.0 \times 10^{0}$&$\phantom{-}0.8$&$\phantom{-}32.5_{46.8}$&$7.2 \times 10^{-87}$&$\phantom{-}0.4$&$\phantom{-}26.8_{44.3}$&$5.8 \times 10^{-38}$&$\phantom{-}0.2$&$\phantom{-}17.1_{37.6}$&$\phantom{-}16.9_{37.5}$&$\phantom{-}17.2_{37.8}$\\

Normalizing&$\phantom{-}43.5_{49.6}$&$1.9 \times 10^{-171}$&$\phantom{-}0.5$&$\phantom{-}47.6_{49.9}$&$8.9 \times 10^{-228}$&$\phantom{-}0.6$&$\phantom{-}16.6_{37.2}$&$1.4 \times 10^{-7}$&$-0.1$&\textcolor{gray}{$\phantom{-}19.4_{39.5}$}&$1.0 \times 10^{0}$&$\phantom{-}0.0$&$\phantom{-}20.5_{39.5}$&$\phantom{-}17.8_{38.3}$&$\phantom{-}23.2_{42.2}$\\

Psychoeducation&$\phantom{-}19.1_{39.3}$&$1.6 \times 10^{-79}$&$\phantom{-}0.3$&$\phantom{-}18.7_{39.0}$&$9.5 \times 10^{-74}$&$\phantom{-}0.3$&$\phantom{-}12.8_{33.5}$&$1.2 \times 10^{-21}$&$\phantom{-}0.2$&\textcolor{gray}{$\phantom{-}7.4_{26.3}$}&$1.0 \times 10^{0}$&$\phantom{-}0.0$&$\phantom{-}7.6_{24.8}$&$\phantom{-}5.2_{22.3}$&$\phantom{-}9.9_{29.8}$\\

    \bottomrule
    \end{tabular}
    }
    \label{tab:dist_new_hope}
\end{table*}
\begin{table*}[t]
    \centering
    \small
    \caption{\showedits{Frequency of conversational behaviors exhibited by LLM
therapists (GPT-4, GPT-3.5-turbo, Llama2-70b, Llama2-13b) in HOPE dataset \cite{malhotra2022speaker} for two different types of simulations separately (Single response and Full conversation). Values in \textcolor{gray}{gray} are not statistically significantly different from human therapist average at $p = \frac{0.05}{m}$ using Two-sided Student’s t-test after Bonferroni correction, where $m$ is the number of t-test per one model ($m = 13$). $d$ is the cohen's d when compared with human therapist average. For instance, GPT-4 with single response simulation responds with \textsc{Reflections on Needs} by 33.5\% and GPT-4 with full conversation simulation responds by 41.9\% when compared with the human therapists responds by 23.4\% on average from Table \ref{tab:dist_new_hope}. (Single: $P = 2.6 \times 10^{-14}$, Cohen's $d = \num{0.2}$; two-sided student t-test) (Full: $P = 8.1 \times 10^{-99}$, Cohen's $d = \num{0.4}$; two-sided student t-test)}}
    \resizebox{1.05\textwidth}{!}{%
    \def\arraystretch{1.15}
    \setlength{\tabcolsep}{2pt}
    \begin{tabular}{l|ccc|ccc|ccc|ccc|ccc|ccc|ccc|ccc|ccc|ccc}
         \toprule 
         \multirow{3}{*}{\textbf{Behavior}} & \multicolumn{12}{c}{\textbf{LLM Therapists (Single Response Simulations)}} & \multicolumn{12}{c}{\textbf{LLM Therapists (Full Conversation Simulations)}} \\
         \cmidrule(lr){2-13}  \cmidrule(lr){14-25} 
         & \multicolumn{3}{c}{\textbf{GPT-4}} &\multicolumn{3}{c}{\textbf{GPT-3.5}}&\multicolumn{3}{c}{\textbf{Llama2-70b}}&\multicolumn{3}{c}{\textbf{Llama2-13b}}& \multicolumn{3}{c}{\textbf{GPT-4}} &\multicolumn{3}{c}{\textbf{GPT-3.5}}&\multicolumn{3}{c}{\textbf{Llama2-70b}}&\multicolumn{3}{c}{\textbf{Llama2-13b}}\\
& \multicolumn{3}{c}{\textbf{$(n = 6158)$}}& \multicolumn{3}{c}{\textbf{$(n = 6144)$}}& \multicolumn{3}{c}{\textbf{$(n = 6174)$}}& \multicolumn{3}{c}{\textbf{$(n = 6282)$}}& \multicolumn{3}{c}{\textbf{$(n = 1547)$}}& \multicolumn{3}{c}{\textbf{$(n = 991)$}}& \multicolumn{3}{c}{\textbf{$(n = 1287)$}}& \multicolumn{3}{c}{\textbf{$(n = 1182)$}}\\
         \cmidrule(lr){2-4} \cmidrule(lr){5-7} \cmidrule(lr){8-10} \cmidrule(lr){11-13} \cmidrule(lr){14-16} \cmidrule(lr){17-19} \cmidrule(lr){20-22}  \cmidrule(lr){23-25}
        & freq. & $p$ & $d$ & freq. & $p$ & $d$ & freq. & $p$& $d$ & freq. & $p$ & $d$ & freq. & $p$& $d$ &  freq. & $p$ & $d$ & freq. & $p$ & $d$ & freq. & $p$& $d$ \\
         \midrule
Reflections on Needs&$\phantom{-}33.5_{47.2}$&$2.6 \times 10^{-14}$&$\phantom{-}0.2$&$\phantom{-}39.6_{48.9}$&$4.5 \times 10^{-25}$&$\phantom{-}0.4$&$\phantom{-}19.6_{39.7}$&$4.9 \times 10^{-2}$&$-0.1$&\textcolor{gray}{$\phantom{-}19.5_{39.7}$}&$6.2 \times 10^{-2}$&$-0.1$&$\phantom{-}41.9_{49.3}$&$8.1 \times 10^{-99}$&$\phantom{-}0.4$&$\phantom{-}34.3_{47.5}$&$1.3 \times 10^{-36}$&$\phantom{-}0.2$&$\phantom{-}30.9_{46.2}$&$2.2 \times 10^{-18}$&$\phantom{-}0.2$&$\phantom{-}30.6_{46.1}$&$6.6 \times 10^{-17}$&$\phantom{-}0.2$\\

Reflections on Emotions&\textcolor{gray}{$\phantom{-}10.0_{29.9}$}&$1.0 \times 10^{0}$&$\phantom{-}0.0$&\textcolor{gray}{$\phantom{-}13.5_{34.2}$}&$9.4 \times 10^{-2}$&$\phantom{-}0.1$&\textcolor{gray}{$\phantom{-}9.9_{29.9}$}&$1.0 \times 10^{0}$&$\phantom{-}0.0$&$\phantom{-}16.1_{36.7}$&$2.4 \times 10^{-6}$&$\phantom{-}0.2$&$\phantom{-}26.4_{44.1}$&$5.5 \times 10^{-103}$&$\phantom{-}0.4$&$\phantom{-}27.1_{44.5}$&$8.8 \times 10^{-111}$&$\phantom{-}0.4$&$\phantom{-}18.6_{38.9}$&$3.0 \times 10^{-32}$&$\phantom{-}0.2$&$\phantom{-}22.3_{41.6}$&$1.5 \times 10^{-62}$&$\phantom{-}0.3$\\

Reflections on Values&\textcolor{gray}{$\phantom{-}5.1_{22.0}$}&$1.0 \times 10^{0}$&$\phantom{-}0.0$&\textcolor{gray}{$\phantom{-}4.3_{20.4}$}&$1.0 \times 10^{0}$&$\phantom{-}0.0$&$\phantom{-}2.7_{16.3}$&$2.2 \times 10^{-2}$&$-0.1$&\textcolor{gray}{$\phantom{-}3.1_{17.4}$}&$2.0 \times 10^{-1}$&$-0.1$&\textcolor{gray}{$\phantom{-}4.3_{20.2}$}&$1.0 \times 10^{0}$&$\phantom{-}0.0$&$\phantom{-}3.4_{18.0}$&$1.6 \times 10^{-3}$&$-0.1$&\textcolor{gray}{$\phantom{-}4.0_{19.5}$}&$3.3 \times 10^{-1}$&$\phantom{-}0.0$&\textcolor{gray}{$\phantom{-}4.3_{20.2}$}&$1.0 \times 10^{0}$&$\phantom{-}0.0$\\

Reflections on Consequences&$\phantom{-}10.9_{31.2}$&$1.3 \times 10^{-21}$&$-0.3$&$\phantom{-}12.0_{32.5}$&$5.6 \times 10^{-12}$&$-0.3$&$\phantom{-}5.9_{23.6}$&$7.3 \times 10^{-41}$&$-0.4$&$\phantom{-}8.5_{28.0}$&$7.1 \times 10^{-26}$&$-0.3$&$\phantom{-}25.2_{43.4}$&$1.6 \times 10^{-4}$&$\phantom{-}0.1$&\textcolor{gray}{$\phantom{-}22.9_{42.0}$}&$1.0 \times 10^{0}$&$\phantom{-}0.0$&\textcolor{gray}{$\phantom{-}22.5_{41.8}$}&$1.0 \times 10^{0}$&$\phantom{-}0.0$&\textcolor{gray}{$\phantom{-}23.6_{42.5}$}&$2.6 \times 10^{-1}$&$\phantom{-}0.0$\\

Reflections on Conflicts&$\phantom{-}2.0_{14.0}$&$2.5 \times 10^{-10}$&$-0.2$&$\phantom{-}1.8_{13.4}$&$1.7 \times 10^{-7}$&$-0.2$&$\phantom{-}2.3_{15.1}$&$1.5 \times 10^{-7}$&$-0.2$&$\phantom{-}3.9_{19.3}$&$7.9 \times 10^{-3}$&$-0.1$&$\phantom{-}4.6_{21.0}$&$5.3 \times 10^{-5}$&$-0.1$&$\phantom{-}3.8_{19.2}$&$3.2 \times 10^{-10}$&$-0.1$&\textcolor{gray}{$\phantom{-}7.1_{25.6}$}&$1.0 \times 10^{0}$&$\phantom{-}0.0$&\textcolor{gray}{$\phantom{-}7.1_{25.7}$}&$1.0 \times 10^{0}$&$\phantom{-}0.0$\\

Reflections on Strengths&$\phantom{-}70.4_{45.7}$&$0.0 \times 10^{0}$&$\phantom{-}1.7$&$\phantom{-}56.5_{49.6}$&$5.5 \times 10^{-300}$&$\phantom{-}1.4$&$\phantom{-}35.3_{47.8}$&$3.4 \times 10^{-118}$&$\phantom{-}0.7$&$\phantom{-}31.6_{46.5}$&$4.3 \times 10^{-85}$&$\phantom{-}0.6$&$\phantom{-}24.3_{42.9}$&$4.5 \times 10^{-94}$&$\phantom{-}0.4$&$\phantom{-}22.7_{41.9}$&$6.2 \times 10^{-78}$&$\phantom{-}0.4$&$\phantom{-}13.0_{33.6}$&$1.4 \times 10^{-7}$&$\phantom{-}0.1$&$\phantom{-}12.6_{33.2}$&$5.2 \times 10^{-6}$&$\phantom{-}0.1$\\

Questions on Experiences&$\phantom{-}31.5_{46.5}$&$2.3 \times 10^{-17}$&$-0.3$&$\phantom{-}21.6_{41.2}$&$5.9 \times 10^{-39}$&$-0.5$&$\phantom{-}38.3_{48.6}$&$2.8 \times 10^{-3}$&$-0.1$&\textcolor{gray}{$\phantom{-}41.1_{49.2}$}&$9.3 \times 10^{-1}$&$-0.1$&$\phantom{-}38.0_{48.5}$&$8.5 \times 10^{-10}$&$-0.1$&$\phantom{-}30.7_{46.1}$&$2.4 \times 10^{-48}$&$-0.3$&$\phantom{-}61.6_{48.6}$&$2.0 \times 10^{-80}$&$\phantom{-}0.4$&$\phantom{-}59.5_{49.1}$&$1.3 \times 10^{-62}$&$\phantom{-}0.3$\\

Questions on Perspectives&\textcolor{gray}{$\phantom{-}4.5_{20.8}$}&$9.1 \times 10^{-1}$&$\phantom{-}0.1$&\textcolor{gray}{$\phantom{-}2.7_{16.3}$}&$1.0 \times 10^{0}$&$\phantom{-}0.0$&$\phantom{-}5.5_{22.8}$&$1.7 \times 10^{-2}$&$\phantom{-}0.1$&$\phantom{-}5.4_{22.6}$&$4.0 \times 10^{-2}$&$\phantom{-}0.1$&$\phantom{-}8.4_{27.8}$&$2.4 \times 10^{-26}$&$\phantom{-}0.2$&\textcolor{gray}{$\phantom{-}4.0_{19.5}$}&$1.0 \times 10^{0}$&$\phantom{-}0.0$&$\phantom{-}13.5_{34.2}$&$2.9 \times 10^{-78}$&$\phantom{-}0.4$&$\phantom{-}12.4_{33.0}$&$5.0 \times 10^{-66}$&$\phantom{-}0.3$\\

Questions on Emotions&$\phantom{-}1.5_{12.1}$&$1.2 \times 10^{-31}$&$-0.3$&$\phantom{-}3.0_{17.1}$&$8.2 \times 10^{-15}$&$-0.3$&$\phantom{-}3.1_{17.4}$&$1.8 \times 10^{-18}$&$-0.3$&$\phantom{-}8.5_{28.0}$&$2.7 \times 10^{-2}$&$-0.1$&$\phantom{-}3.4_{18.2}$&$1.9 \times 10^{-63}$&$-0.3$&$\phantom{-}6.6_{24.8}$&$3.5 \times 10^{-20}$&$-0.2$&$\phantom{-}14.4_{35.1}$&$2.9 \times 10^{-4}$&$\phantom{-}0.1$&$\phantom{-}19.9_{39.9}$&$7.3 \times 10^{-32}$&$\phantom{-}0.2$\\

Problem-Solving&$\phantom{-}40.5_{49.1}$&$2.1 \times 10^{-29}$&$\phantom{-}0.3$&$\phantom{-}58.5_{49.3}$&$3.6 \times 10^{-99}$&$\phantom{-}0.7$&$\phantom{-}51.4_{50.0}$&$1.4 \times 10^{-74}$&$\phantom{-}0.6$&$\phantom{-}34.6_{47.6}$&$5.4 \times 10^{-9}$&$\phantom{-}0.2$&$\phantom{-}56.6_{49.6}$&$1.1 \times 10^{-262}$&$\phantom{-}0.7$&$\phantom{-}60.4_{48.9}$&$0.0 \times 10^{0}$&$\phantom{-}0.7$&$\phantom{-}30.0_{45.8}$&$7.6 \times 10^{-6}$&$\phantom{-}0.1$&$\phantom{-}22.4_{41.7}$&$9.8 \times 10^{-5}$&$-0.1$\\

Planning&$\phantom{-}66.3_{47.3}$&$0.0 \times 10^{0}$&$\phantom{-}1.2$&$\phantom{-}73.2_{44.3}$&$0.0 \times 10^{0}$&$\phantom{-}1.4$&$\phantom{-}61.6_{48.7}$&$2.6 \times 10^{-258}$&$\phantom{-}1.1$&$\phantom{-}42.0_{49.4}$&$4.2 \times 10^{-81}$&$\phantom{-}0.6$&$\phantom{-}43.3_{49.6}$&$1.8 \times 10^{-211}$&$\phantom{-}0.6$&$\phantom{-}50.1_{50.0}$&$0.0 \times 10^{0}$&$\phantom{-}0.7$&$\phantom{-}26.4_{44.1}$&$5.3 \times 10^{-33}$&$\phantom{-}0.2$&$\phantom{-}23.9_{42.7}$&$7.0 \times 10^{-19}$&$\phantom{-}0.2$\\

Normalizing&$\phantom{-}61.9_{48.6}$&$7.1 \times 10^{-240}$&$\phantom{-}1.0$&$\phantom{-}71.2_{45.3}$&$6.9 \times 10^{-262}$&$\phantom{-}1.3$&\textcolor{gray}{$\phantom{-}23.9_{42.7}$}&$7.7 \times 10^{-2}$&$\phantom{-}0.1$&\textcolor{gray}{$\phantom{-}23.8_{42.6}$}&$1.4 \times 10^{-1}$&$\phantom{-}0.1$&$\phantom{-}38.9_{48.7}$&$6.0 \times 10^{-105}$&$\phantom{-}0.4$&$\phantom{-}43.8_{49.6}$&$3.8 \times 10^{-163}$&$\phantom{-}0.5$&$\phantom{-}15.1_{35.8}$&$1.1 \times 10^{-13}$&$-0.1$&\textcolor{gray}{$\phantom{-}18.5_{38.9}$}&$8.9 \times 10^{-2}$&$-0.1$\\

Psychoeducation&$\phantom{-}23.4_{42.4}$&$2.8 \times 10^{-74}$&$\phantom{-}0.5$&$\phantom{-}33.4_{47.2}$&$1.4 \times 10^{-135}$&$\phantom{-}0.9$&$\phantom{-}29.4_{45.6}$&$7.5 \times 10^{-117}$&$\phantom{-}0.7$&$\phantom{-}17.8_{38.2}$&$3.1 \times 10^{-29}$&$\phantom{-}0.4$&$\phantom{-}18.0_{38.4}$&$3.0 \times 10^{-64}$&$\phantom{-}0.3$&$\phantom{-}16.3_{37.0}$&$5.3 \times 10^{-48}$&$\phantom{-}0.3$&$\phantom{-}9.4_{29.2}$&$3.8 \times 10^{-3}$&$\phantom{-}0.1$&$\phantom{-}5.5_{22.8}$&$3.9 \times 10^{-5}$&$-0.1$\\

    \bottomrule
    \end{tabular}
    }
    \label{tab:dist_new_single_full_hope}
\end{table*}
\begin{table*}[t]
    \centering
    \small
    \caption{\showedits{Temporal Order (First Occurrence) of conversational behaviors during a conversation by LLM
therapists (GPT-4, GPT-3.5-turbo, Llama2-70b, Llama2-13b) in HOPE dataset \cite{malhotra2022speaker}. Values in \textcolor{gray}{gray} are not statistically significantly different from human therapist average at $p = \frac{0.05}{m}$ using Two-sided Student’s t-test after Bonferroni correction, where $m$ is the number of t-test per one model ($m = 13$). $d$ is the cohen's $d$ when compared with human therapist average. For instance, GPT-4 responds with \textsc{Reflections on Needs} at 3.3 turns earlier than average human therapists responds.  (GPT-4: 2.9 turn, Average human therapists: 6.2 turn; $P = 1.0 \times 10^{-19}$, Cohen's $d = 0.8$, two-sided Student’s $t$-test).}}
    \resizebox{1\textwidth}{!}{%
    \def\arraystretch{1.15}
    \setlength{\tabcolsep}{2pt}
    \begin{tabular}{l|ccc|ccc|ccc|ccc|ccc|ccc}
         \toprule 
         \multirow{3}{*}{\textbf{Behavior}} & \multicolumn{12}{c}{\textbf{LLM Therapists (Single Response \& Full Conversation Simulations)}} & \multicolumn{3}{c}{\textbf{Human Therapists}} \\
         \cmidrule(lr){2-13} \cmidrule(lr){14-16}
         & \multicolumn{3}{c}{\textbf{GPT-4}} &\multicolumn{3}{c}{\textbf{GPT-3.5}}&\multicolumn{3}{c}{\textbf{Llama2-70b}}&\multicolumn{3}{c}{\textbf{Llama2-13b}}& \textbf{Average}&\textbf{High-quality}&\textbf{Low-quality}\\
& \multicolumn{3}{c}{\textbf{$(n = 7705)$}}& \multicolumn{3}{c}{\textbf{$(n = 7135)$}}& \multicolumn{3}{c}{\textbf{$(n = 7461)$}}& \multicolumn{3}{c}{\textbf{$(n = 7464)$}}&\textbf{$(n = 5446)$}&\textbf{$(n = 3907)$}&\textbf{$(n = 1539)$}\\
         \cmidrule(lr){2-4} \cmidrule(lr){5-7} \cmidrule(lr){8-10} \cmidrule(lr){11-13} \cmidrule(lr){14-14} \cmidrule(lr){15-15}  \cmidrule(lr){16-16} 
        & turn & $p$ & $d$ & turn & $p$ & $d$ & turn & $p$& $d$ & turn & $p$ & $d$ & turn & turn & turn \\
         \midrule
Reflections on Needs&$\phantom{-}2.9_{3.3}$&$1.0 \times 10^{-19}$&$\phantom{-}0.8$&\textcolor{gray}{$\phantom{-}6.1_{6.7}$}&$1.0 \times 10^{0}$&$\phantom{-}0.0$&\textcolor{gray}{$\phantom{-}5.3_{5.5}$}&$4.8 \times 10^{-1}$&$\phantom{-}0.2$&\textcolor{gray}{$\phantom{-}6.1_{6.4}$}&$1.0 \times 10^{0}$&$\phantom{-}0.0$&$\phantom{-}6.2_{5.5}$&$\phantom{-}4.4_{4.5}$&$\phantom{-}8.1_{6.2}$\\

Reflections on Emotions&\textcolor{gray}{$\phantom{-}5.8_{4.9}$}&$1.0 \times 10^{0}$&$\phantom{-}0.1$&\textcolor{gray}{$\phantom{-}6.0_{5.3}$}&$1.0 \times 10^{0}$&$\phantom{-}0.1$&\textcolor{gray}{$\phantom{-}6.4_{5.1}$}&$1.0 \times 10^{0}$&$\phantom{-}0.0$&\textcolor{gray}{$\phantom{-}7.0_{5.8}$}&$1.0 \times 10^{0}$&$-0.1$&$\phantom{-}6.3_{4.9}$&$\phantom{-}5.2_{4.8}$&$\phantom{-}7.4_{5.0}$\\

Reflections on Values&\textcolor{gray}{$\phantom{-}9.0_{5.9}$}&$1.0 \times 10^{0}$&$\phantom{-}0.2$&\textcolor{gray}{$\phantom{-}9.3_{6.4}$}&$1.0 \times 10^{0}$&$\phantom{-}0.1$&\textcolor{gray}{$\phantom{-}10.3_{6.0}$}&$1.0 \times 10^{0}$&$-0.1$&\textcolor{gray}{$\phantom{-}10.3_{5.8}$}&$1.0 \times 10^{0}$&$-0.1$&$\phantom{-}9.9_{5.5}$&$\phantom{-}8.8_{5.3}$&$\phantom{-}11.0_{5.8}$\\

Reflections on Consequences&\textcolor{gray}{$\phantom{-}7.0_{4.8}$}&$1.0 \times 10^{0}$&$\phantom{-}0.0$&\textcolor{gray}{$\phantom{-}7.1_{5.7}$}&$1.0 \times 10^{0}$&$\phantom{-}0.0$&\textcolor{gray}{$\phantom{-}6.2_{4.9}$}&$1.0 \times 10^{0}$&$\phantom{-}0.2$&\textcolor{gray}{$\phantom{-}7.3_{5.7}$}&$1.0 \times 10^{0}$&$-0.1$&$\phantom{-}7.0_{5.0}$&$\phantom{-}6.2_{4.6}$&$\phantom{-}7.8_{5.4}$\\

Reflections on Conflicts&\textcolor{gray}{$\phantom{-}8.2_{4.6}$}&$5.8 \times 10^{-2}$&$\phantom{-}0.4$&\textcolor{gray}{$\phantom{-}8.8_{5.3}$}&$1.0 \times 10^{0}$&$\phantom{-}0.2$&\textcolor{gray}{$\phantom{-}8.7_{5.3}$}&$4.2 \times 10^{-1}$&$\phantom{-}0.2$&$\phantom{-}8.1_{5.3}$&$2.3 \times 10^{-2}$&$\phantom{-}0.4$&$\phantom{-}10.0_{5.2}$&$\phantom{-}8.2_{4.7}$&$\phantom{-}11.8_{5.6}$\\

Reflections on Strengths&$\phantom{-}7.8_{5.1}$&$4.6 \times 10^{-5}$&$\phantom{-}0.4$&\textcolor{gray}{$\phantom{-}8.9_{6.2}$}&$3.9 \times 10^{-1}$&$\phantom{-}0.2$&\textcolor{gray}{$\phantom{-}9.1_{6.3}$}&$1.0 \times 10^{0}$&$\phantom{-}0.2$&\textcolor{gray}{$\phantom{-}10.1_{6.8}$}&$1.0 \times 10^{0}$&$\phantom{-}0.0$&$\phantom{-}10.2_{6.1}$&$\phantom{-}8.4_{6.0}$&$\phantom{-}11.9_{5.7}$\\

Questions on Experiences&\textcolor{gray}{$\phantom{-}2.7_{2.8}$}&$1.0 \times 10^{0}$&$\phantom{-}0.1$&$\phantom{-}4.1_{5.7}$&$2.9 \times 10^{-2}$&$-0.3$&\textcolor{gray}{$\phantom{-}3.1_{3.1}$}&$1.0 \times 10^{0}$&$-0.1$&$\phantom{-}4.0_{4.1}$&$1.8 \times 10^{-3}$&$-0.3$&$\phantom{-}2.9_{2.5}$&$\phantom{-}2.2_{2.1}$&$\phantom{-}3.6_{2.9}$\\

Questions on Perspectives&$\phantom{-}7.7_{4.1}$&$1.4 \times 10^{-13}$&$\phantom{-}1.1$&$\phantom{-}7.8_{4.5}$&$2.9 \times 10^{-9}$&$\phantom{-}1.0$&$\phantom{-}8.9_{5.4}$&$8.1 \times 10^{-7}$&$\phantom{-}0.7$&$\phantom{-}8.4_{5.2}$&$6.1 \times 10^{-9}$&$\phantom{-}0.8$&$\phantom{-}12.6_{5.1}$&$\phantom{-}10.0_{4.7}$&$\phantom{-}15.3_{4.2}$\\

Questions on Emotions&\textcolor{gray}{$\phantom{-}6.4_{5.6}$}&$1.0 \times 10^{0}$&$\phantom{-}0.1$&\textcolor{gray}{$\phantom{-}7.7_{6.0}$}&$1.0 \times 10^{0}$&$-0.1$&$\phantom{-}5.0_{5.2}$&$1.0 \times 10^{-2}$&$\phantom{-}0.3$&$\phantom{-}4.4_{4.4}$&$2.9 \times 10^{-6}$&$\phantom{-}0.5$&$\phantom{-}7.0_{6.4}$&$\phantom{-}6.7_{6.3}$&$\phantom{-}7.3_{6.5}$\\

Problem-Solving&\textcolor{gray}{$\phantom{-}7.9_{5.2}$}&$1.0 \times 10^{0}$&$\phantom{-}0.1$&$\phantom{-}7.0_{5.6}$&$4.7 \times 10^{-2}$&$\phantom{-}0.2$&\textcolor{gray}{$\phantom{-}8.7_{5.4}$}&$1.0 \times 10^{0}$&$-0.1$&\textcolor{gray}{$\phantom{-}9.5_{6.2}$}&$1.9 \times 10^{-1}$&$-0.2$&$\phantom{-}8.3_{5.4}$&$\phantom{-}8.7_{5.6}$&$\phantom{-}8.0_{5.1}$\\

Planning&$\phantom{-}9.7_{5.7}$&$2.7 \times 10^{-4}$&$\phantom{-}0.4$&$\phantom{-}9.3_{5.6}$&$8.6 \times 10^{-6}$&$\phantom{-}0.4$&$\phantom{-}9.7_{5.7}$&$2.7 \times 10^{-4}$&$\phantom{-}0.4$&\textcolor{gray}{$\phantom{-}10.5_{6.1}$}&$1.3 \times 10^{-1}$&$\phantom{-}0.2$&$\phantom{-}11.9_{6.5}$&$\phantom{-}11.5_{6.8}$&$\phantom{-}12.4_{6.0}$\\

Normalizing&\textcolor{gray}{$\phantom{-}8.0_{5.8}$}&$4.5 \times 10^{-1}$&$-0.2$&\textcolor{gray}{$\phantom{-}7.3_{6.1}$}&$1.0 \times 10^{0}$&$-0.1$&$\phantom{-}9.5_{6.6}$&$7.5 \times 10^{-5}$&$-0.4$&$\phantom{-}8.8_{6.6}$&$1.4 \times 10^{-2}$&$-0.3$&$\phantom{-}6.9_{6.2}$&$\phantom{-}6.2_{6.3}$&$\phantom{-}7.7_{6.0}$\\

Psychoeducation&$\phantom{-}11.0_{5.9}$&$2.1 \times 10^{-2}$&$-0.4$&\textcolor{gray}{$\phantom{-}10.1_{5.9}$}&$9.8 \times 10^{-1}$&$-0.2$&$\phantom{-}12.1_{5.7}$&$9.3 \times 10^{-6}$&$-0.6$&$\phantom{-}12.8_{6.2}$&$2.2 \times 10^{-6}$&$-0.7$&$\phantom{-}8.9_{5.5}$&$\phantom{-}8.5_{5.6}$&$\phantom{-}9.2_{5.5}$\\

    \bottomrule
    \end{tabular}
    }
    \label{tab:order_new_hope}
\end{table*}
\begin{table*}[t]
    \centering
    \small
    \caption{\showedits{Temporal Order (First Occurrence) of conversational behaviors during a conversation by LLM
therapists (GPT-4, GPT-3.5-turbo, Llama2-70b, Llama2-13b) in HOPE dataset \cite{malhotra2022speaker} for two different types of simulations separately (Single response and Full conversation). Values in \textcolor{gray}{gray} are not statistically significantly different from human therapist average at $p = \frac{0.05}{m}$ using Two-sided Student’s t-test after Bonferroni correction, where $m$ is the number of t-test per one model ($m = 13$). $d$ is the cohen's $d$ when compared with human therapist average. For instance, GPT-4 with single response simulation responds with \textsc{Reflections on Needs} at 4.0 turn which is 2.2 turns earlier than average human therapists. Similarly, GPT-4 with full conversation simulation responds at 1.8 turn which is 4.4 turns earlier than average human therapists. (Single: 4.0 turn, Average human therapists: 6.2 turn from Table \ref{tab:order_new_hope}; $P = 6.4 \times 10^{-6}$, Cohen's $d = 0.5$, two-sided Student’s $t$-test) (Full: 1.8 turn, Average human therapists: 6.2 turn from Table \ref{tab:order_new_hope}; $P = 4.7 \times 10^{-21}$, Cohen's $d = 1.0$).}}
    \resizebox{1.05\textwidth}{!}{%
    \def\arraystretch{1.15}
    \setlength{\tabcolsep}{2pt}
    \begin{tabular}{l|ccc|ccc|ccc|ccc|ccc|ccc|ccc|ccc|ccc|ccc}
         \toprule 
         \multirow{3}{*}{\textbf{Behavior}} & \multicolumn{12}{c}{\textbf{LLM Therapists (Single Response Simulations)}} & \multicolumn{12}{c}{\textbf{LLM Therapists (Full Conversation Simulations)}} \\
         \cmidrule(lr){2-13}  \cmidrule(lr){14-25} 
         & \multicolumn{3}{c}{\textbf{GPT-4}} &\multicolumn{3}{c}{\textbf{GPT-3.5}}&\multicolumn{3}{c}{\textbf{Llama2-70b}}&\multicolumn{3}{c}{\textbf{Llama2-13b}}& \multicolumn{3}{c}{\textbf{GPT-4}} &\multicolumn{3}{c}{\textbf{GPT-3.5}}&\multicolumn{3}{c}{\textbf{Llama2-70b}}&\multicolumn{3}{c}{\textbf{Llama2-13b}}\\
& \multicolumn{3}{c}{\textbf{$(n = 6158)$}}& \multicolumn{3}{c}{\textbf{$(n = 6144)$}}& \multicolumn{3}{c}{\textbf{$(n = 6174)$}}& \multicolumn{3}{c}{\textbf{$(n = 6282)$}}& \multicolumn{3}{c}{\textbf{$(n = 1547)$}}& \multicolumn{3}{c}{\textbf{$(n = 991)$}}& \multicolumn{3}{c}{\textbf{$(n = 1287)$}}& \multicolumn{3}{c}{\textbf{$(n = 1182)$}}\\
         \cmidrule(lr){2-4} \cmidrule(lr){5-7} \cmidrule(lr){8-10} \cmidrule(lr){11-13} \cmidrule(lr){14-16} \cmidrule(lr){17-19} \cmidrule(lr){20-22}  \cmidrule(lr){23-25}
        & turn & $p$ & $d$ & turn & $p$ & $d$ & turn & $p$& $d$ & turn & $p$ & $d$ & turn & $p$& $d$ &  turn & $p$ & $d$ & turn & $p$ & $d$ & turn & $p$& $d$ \\
         \midrule
Reflections on Needs&$\phantom{-}4.0_{3.0}$&$6.4 \times 10^{-6}$&$\phantom{-}0.5$&$\phantom{-}9.3_{7.4}$&$1.3 \times 10^{-5}$&$-0.5$&$\phantom{-}9.3_{5.3}$&$1.3 \times 10^{-6}$&$-0.6$&$\phantom{-}11.5_{6.3}$&$2.1 \times 10^{-15}$&$-0.9$&$\phantom{-}1.8_{3.2}$&$4.7 \times 10^{-21}$&$\phantom{-}1.0$&$\phantom{-}2.8_{3.8}$&$7.5 \times 10^{-12}$&$\phantom{-}0.7$&$\phantom{-}2.4_{3.4}$&$5.3 \times 10^{-16}$&$\phantom{-}0.8$&$\phantom{-}2.6_{3.1}$&$1.5 \times 10^{-15}$&$\phantom{-}0.8$\\

Reflections on Emotions&$\phantom{-}8.2_{5.0}$&$3.5 \times 10^{-2}$&$-0.4$&$\phantom{-}9.5_{5.8}$&$6.9 \times 10^{-5}$&$-0.6$&$\phantom{-}9.5_{4.7}$&$2.4 \times 10^{-5}$&$-0.7$&$\phantom{-}11.7_{5.2}$&$3.6 \times 10^{-16}$&$-1.1$&$\phantom{-}4.5_{4.3}$&$2.9 \times 10^{-3}$&$\phantom{-}0.4$&$\phantom{-}4.3_{4.0}$&$3.8 \times 10^{-4}$&$\phantom{-}0.4$&\textcolor{gray}{$\phantom{-}5.0_{4.7}$}&$1.7 \times 10^{-1}$&$\phantom{-}0.3$&$\phantom{-}4.2_{4.0}$&$9.6 \times 10^{-5}$&$\phantom{-}0.5$\\

Reflections on Values&\textcolor{gray}{$\phantom{-}8.4_{5.3}$}&$1.0 \times 10^{0}$&$\phantom{-}0.3$&\textcolor{gray}{$\phantom{-}8.9_{7.2}$}&$1.0 \times 10^{0}$&$\phantom{-}0.2$&$\phantom{-}14.2_{4.9}$&$4.4 \times 10^{-3}$&$-0.8$&$\phantom{-}14.4_{5.2}$&$1.1 \times 10^{-3}$&$-0.8$&\textcolor{gray}{$\phantom{-}9.4_{6.2}$}&$1.0 \times 10^{0}$&$\phantom{-}0.1$&\textcolor{gray}{$\phantom{-}9.6_{5.8}$}&$1.0 \times 10^{0}$&$\phantom{-}0.1$&\textcolor{gray}{$\phantom{-}9.2_{5.9}$}&$1.0 \times 10^{0}$&$\phantom{-}0.1$&\textcolor{gray}{$\phantom{-}9.2_{5.5}$}&$1.0 \times 10^{0}$&$\phantom{-}0.1$\\

Reflections on Consequences&$\phantom{-}8.6_{4.4}$&$4.4 \times 10^{-2}$&$-0.3$&$\phantom{-}9.9_{6.4}$&$3.7 \times 10^{-4}$&$-0.5$&$\phantom{-}10.1_{4.7}$&$6.0 \times 10^{-4}$&$-0.6$&$\phantom{-}13.0_{5.3}$&$2.4 \times 10^{-14}$&$-1.2$&\textcolor{gray}{$\phantom{-}6.0_{4.8}$}&$6.0 \times 10^{-1}$&$\phantom{-}0.2$&\textcolor{gray}{$\phantom{-}5.8_{4.9}$}&$2.0 \times 10^{-1}$&$\phantom{-}0.2$&$\phantom{-}5.1_{4.3}$&$5.1 \times 10^{-4}$&$\phantom{-}0.4$&$\phantom{-}5.5_{4.4}$&$1.3 \times 10^{-2}$&$\phantom{-}0.3$\\

Reflections on Conflicts&\textcolor{gray}{$\phantom{-}9.2_{3.3}$}&$1.0 \times 10^{0}$&$\phantom{-}0.2$&\textcolor{gray}{$\phantom{-}8.8_{5.2}$}&$1.0 \times 10^{0}$&$\phantom{-}0.2$&\textcolor{gray}{$\phantom{-}10.3_{5.0}$}&$1.0 \times 10^{0}$&$-0.1$&\textcolor{gray}{$\phantom{-}11.8_{4.9}$}&$1.0 \times 10^{0}$&$-0.3$&$\phantom{-}8.0_{4.8}$&$4.1 \times 10^{-2}$&$\phantom{-}0.4$&\textcolor{gray}{$\phantom{-}8.8_{5.3}$}&$1.0 \times 10^{0}$&$\phantom{-}0.2$&\textcolor{gray}{$\phantom{-}8.4_{5.3}$}&$1.6 \times 10^{-1}$&$\phantom{-}0.3$&$\phantom{-}7.2_{5.1}$&$1.7 \times 10^{-4}$&$\phantom{-}0.5$\\

Reflections on Strengths&$\phantom{-}8.4_{4.8}$&$2.1 \times 10^{-2}$&$\phantom{-}0.3$&\textcolor{gray}{$\phantom{-}11.5_{6.2}$}&$5.3 \times 10^{-1}$&$-0.2$&\textcolor{gray}{$\phantom{-}11.2_{5.8}$}&$1.0 \times 10^{0}$&$-0.2$&$\phantom{-}12.4_{6.3}$&$2.0 \times 10^{-2}$&$-0.4$&$\phantom{-}7.3_{5.3}$&$3.0 \times 10^{-5}$&$\phantom{-}0.5$&$\phantom{-}6.6_{5.2}$&$5.5 \times 10^{-8}$&$\phantom{-}0.6$&$\phantom{-}7.0_{6.1}$&$2.2 \times 10^{-5}$&$\phantom{-}0.5$&\textcolor{gray}{$\phantom{-}8.3_{6.6}$}&$9.8 \times 10^{-2}$&$\phantom{-}0.3$\\

Questions on Experiences&$\phantom{-}3.9_{2.6}$&$7.9 \times 10^{-4}$&$-0.4$&$\phantom{-}8.1_{7.1}$&$3.2 \times 10^{-21}$&$-1.1$&$\phantom{-}4.9_{3.1}$&$3.9 \times 10^{-12}$&$-0.7$&$\phantom{-}6.5_{4.2}$&$1.5 \times 10^{-25}$&$-1.1$&$\phantom{-}1.5_{2.5}$&$3.9 \times 10^{-8}$&$\phantom{-}0.6$&$\phantom{-}1.6_{2.4}$&$1.7 \times 10^{-7}$&$\phantom{-}0.5$&$\phantom{-}1.4_{1.9}$&$1.9 \times 10^{-11}$&$\phantom{-}0.7$&$\phantom{-}1.5_{1.9}$&$1.8 \times 10^{-9}$&$\phantom{-}0.6$\\

Questions on Perspectives&$\phantom{-}8.3_{3.9}$&$3.5 \times 10^{-4}$&$\phantom{-}0.9$&\textcolor{gray}{$\phantom{-}8.0_{7.9}$}&$5.2 \times 10^{-1}$&$\phantom{-}0.9$&\textcolor{gray}{$\phantom{-}13.0_{4.3}$}&$1.0 \times 10^{0}$&$-0.1$&\textcolor{gray}{$\phantom{-}13.5_{4.8}$}&$1.0 \times 10^{0}$&$-0.2$&$\phantom{-}7.5_{4.1}$&$1.3 \times 10^{-12}$&$\phantom{-}1.1$&$\phantom{-}7.8_{4.2}$&$2.3 \times 10^{-9}$&$\phantom{-}1.0$&$\phantom{-}7.7_{5.1}$&$4.1 \times 10^{-11}$&$\phantom{-}1.0$&$\phantom{-}7.4_{4.7}$&$3.5 \times 10^{-14}$&$\phantom{-}1.1$\\

Questions on Emotions&\textcolor{gray}{$\phantom{-}6.0_{5.9}$}&$1.0 \times 10^{0}$&$\phantom{-}0.2$&\textcolor{gray}{$\phantom{-}8.1_{6.5}$}&$1.0 \times 10^{0}$&$-0.2$&\textcolor{gray}{$\phantom{-}7.6_{6.1}$}&$1.0 \times 10^{0}$&$-0.1$&\textcolor{gray}{$\phantom{-}5.5_{3.3}$}&$5.3 \times 10^{-1}$&$\phantom{-}0.3$&\textcolor{gray}{$\phantom{-}6.5_{5.5}$}&$1.0 \times 10^{0}$&$\phantom{-}0.1$&\textcolor{gray}{$\phantom{-}7.6_{6.0}$}&$1.0 \times 10^{0}$&$-0.1$&$\phantom{-}4.5_{4.9}$&$5.5 \times 10^{-4}$&$\phantom{-}0.4$&$\phantom{-}3.9_{4.7}$&$1.1 \times 10^{-6}$&$\phantom{-}0.6$\\

Problem-Solving&$\phantom{-}10.6_{4.8}$&$8.3 \times 10^{-5}$&$-0.4$&$\phantom{-}10.2_{5.7}$&$8.3 \times 10^{-3}$&$-0.3$&$\phantom{-}11.2_{4.8}$&$1.0 \times 10^{-7}$&$-0.6$&$\phantom{-}13.1_{5.5}$&$2.0 \times 10^{-15}$&$-0.9$&$\phantom{-}5.3_{4.0}$&$2.6 \times 10^{-9}$&$\phantom{-}0.6$&$\phantom{-}4.2_{3.6}$&$3.4 \times 10^{-17}$&$\phantom{-}0.9$&$\phantom{-}6.1_{4.8}$&$1.9 \times 10^{-4}$&$\phantom{-}0.4$&$\phantom{-}6.6_{5.1}$&$8.9 \times 10^{-3}$&$\phantom{-}0.3$\\

Planning&\textcolor{gray}{$\phantom{-}11.4_{5.5}$}&$1.0 \times 10^{0}$&$\phantom{-}0.1$&\textcolor{gray}{$\phantom{-}12.3_{5.3}$}&$1.0 \times 10^{0}$&$-0.1$&\textcolor{gray}{$\phantom{-}10.9_{5.4}$}&$1.0 \times 10^{0}$&$\phantom{-}0.2$&\textcolor{gray}{$\phantom{-}12.6_{6.0}$}&$1.0 \times 10^{0}$&$-0.1$&$\phantom{-}8.0_{5.4}$&$3.4 \times 10^{-9}$&$\phantom{-}0.7$&$\phantom{-}6.9_{4.6}$&$5.9 \times 10^{-16}$&$\phantom{-}0.9$&$\phantom{-}8.4_{5.8}$&$7.4 \times 10^{-7}$&$\phantom{-}0.6$&$\phantom{-}8.6_{5.7}$&$3.0 \times 10^{-6}$&$\phantom{-}0.5$\\

Normalizing&$\phantom{-}10.6_{5.4}$&$8.6 \times 10^{-9}$&$-0.6$&$\phantom{-}10.9_{5.8}$&$1.4 \times 10^{-8}$&$-0.7$&$\phantom{-}12.9_{6.2}$&$2.1 \times 10^{-15}$&$-1.0$&$\phantom{-}12.3_{6.5}$&$1.1 \times 10^{-10}$&$-0.9$&\textcolor{gray}{$\phantom{-}5.5_{5.1}$}&$1.5 \times 10^{-1}$&$\phantom{-}0.2$&$\phantom{-}4.6_{4.8}$&$2.8 \times 10^{-4}$&$\phantom{-}0.4$&\textcolor{gray}{$\phantom{-}7.1_{5.8}$}&$1.0 \times 10^{0}$&$\phantom{-}0.0$&\textcolor{gray}{$\phantom{-}7.0_{5.8}$}&$1.0 \times 10^{0}$&$\phantom{-}0.0$\\

Psychoeducation&$\phantom{-}13.1_{5.2}$&$2.3 \times 10^{-6}$&$-0.8$&$\phantom{-}11.7_{5.8}$&$1.2 \times 10^{-2}$&$-0.5$&$\phantom{-}13.3_{5.3}$&$1.0 \times 10^{-8}$&$-0.8$&$\phantom{-}15.6_{5.4}$&$8.1 \times 10^{-14}$&$-1.2$&\textcolor{gray}{$\phantom{-}9.6_{5.9}$}&$1.0 \times 10^{0}$&$-0.1$&\textcolor{gray}{$\phantom{-}9.2_{5.8}$}&$1.0 \times 10^{0}$&$-0.1$&\textcolor{gray}{$\phantom{-}10.7_{5.7}$}&$2.0 \times 10^{-1}$&$-0.3$&\textcolor{gray}{$\phantom{-}10.0_{5.6}$}&$1.0 \times 10^{0}$&$-0.2$\\

    \bottomrule
    \end{tabular}
    }
    \label{tab:order_new_single_full_hope}
\end{table*}
\begin{table*}[t]
    \centering
    \small
    \caption{\showedits{Adaptability (frequency of conversational behaviors exhibited by therapists \textit{in response to specific client behaviors}) by LLM
therapists (GPT-4, GPT-3.5-turbo, Llama2-70b, Llama2-13b) in HOPE dataset \cite{malhotra2022speaker}. Values in \textcolor{gray}{gray} are not statistically significantly different from human therapist average at $p = \frac{0.05}{m}$ using Two-sided Student’s t-test after Bonferroni correction, where $m$ is the number of t-test per one model ($m = 13 \times 6 = 78$). $d$ is the cohen's d when compared with human therapist average. For instance, GPT-4 exhibits \textsc{Reflections on Needs} by 10.2\%  and human therapists responds by 27.7\% on average when client expresses \textsc{Changing unhealthy behavior}. ($P = 1.9 \times 10^{-46}$, Cohen's $d = \num{-0.5}$, two-sided Student’s $t$-test).}}
    \resizebox{1\textwidth}{!}{%
    \def\arraystretch{1.15}
    \setlength{\tabcolsep}{2pt}
    \begin{tabular}{l|ccc|ccc|ccc|ccc|ccc|ccc}
         \toprule 
         \multirow{3}{*}{\textbf{Therapist Behavior}} & \multicolumn{12}{c}{\textbf{LLM Therapists (Single Response \& Full Conversation Simulations)}} & \multicolumn{3}{c}{\textbf{Human Therapists}} \\
         \cmidrule(lr){2-13} \cmidrule(lr){14-16}
         & \multicolumn{3}{c}{\textbf{GPT-4}} &\multicolumn{3}{c}{\textbf{GPT-3.5}}&\multicolumn{3}{c}{\textbf{Llama2-70b}}&\multicolumn{3}{c}{\textbf{Llama2-13b}}& \textbf{Average}&\textbf{High-quality}&\textbf{Low-quality}\\
         \cmidrule(lr){2-4} \cmidrule(lr){5-7} \cmidrule(lr){8-10} \cmidrule(lr){11-13} \cmidrule(lr){14-14} \cmidrule(lr){15-15}  \cmidrule(lr){16-16} 
        & freq. & $p$ & $d$ & freq. & $p$ & $d$ & freq. & $p$& $d$ & freq. & $p$ & $d$ & freq. & freq. & freq. \\
        \midrule
          \multicolumn{16}{c}{\textbf{Client behavior: Changing unhealthy behavior}} \\
          \midrule
          $n$ & \multicolumn{3}{c}{\textbf{$2552$}} & \multicolumn{3}{c}{\textbf{$2193$}} & \multicolumn{3}{c}{\textbf{$2349$}} & \multicolumn{3}{c}{\textbf{$2109$}}&\textbf{$1553$}&\textbf{$1198$}&\textbf{$355$}\\
         \midrule
Reflections on Needs&$\phantom{-}10.2_{30.3}$&$1.9 \times 10^{-46}$&$-0.5$&$\phantom{-}8.2_{27.4}$&$2.9 \times 10^{-56}$&$-0.5$&$\phantom{-}6.0_{23.8}$&$3.6 \times 10^{-78}$&$-0.6$&$\phantom{-}5.3_{22.4}$&$1.7 \times 10^{-79}$&$-0.7$&$\phantom{-}27.7_{45.5}$&$\phantom{-}30.6_{46.1}$&$\phantom{-}24.8_{43.2}$\\

Reflections on Emotions&$\phantom{-}2.7_{16.2}$&$5.4 \times 10^{-21}$&$-0.3$&$\phantom{-}2.8_{16.6}$&$5.9 \times 10^{-18}$&$-0.3$&$\phantom{-}2.1_{14.4}$&$3.9 \times 10^{-24}$&$-0.3$&$\phantom{-}3.7_{18.8}$&$7.0 \times 10^{-13}$&$-0.3$&$\phantom{-}10.4_{33.3}$&$\phantom{-}14.6_{35.3}$&$\phantom{-}6.2_{24.1}$\\

Reflections on Values&$\phantom{-}2.0_{14.1}$&$3.0 \times 10^{-4}$&$-0.1$&$\phantom{-}0.1_{3.0}$&$8.2 \times 10^{-18}$&$-0.3$&$\phantom{-}1.0_{9.8}$&$4.7 \times 10^{-10}$&$-0.2$&$\phantom{-}0.9_{9.2}$&$4.4 \times 10^{-10}$&$-0.2$&$\phantom{-}4.8_{23.7}$&$\phantom{-}7.0_{25.5}$&$\phantom{-}2.5_{15.7}$\\

Reflections on Consequences&$\phantom{-}4.5_{20.8}$&$6.6 \times 10^{-108}$&$-0.7$&$\phantom{-}2.1_{14.5}$&$3.9 \times 10^{-133}$&$-0.9$&$\phantom{-}1.8_{13.3}$&$5.2 \times 10^{-148}$&$-0.9$&$\phantom{-}2.8_{16.4}$&$5.6 \times 10^{-119}$&$-0.8$&$\phantom{-}27.3_{42.3}$&$\phantom{-}19.9_{39.9}$&$\phantom{-}34.6_{47.7}$\\

Reflections on Conflicts&$\phantom{-}0.5_{7.4}$&$5.3 \times 10^{-28}$&$-0.4$&$\phantom{-}0.3_{5.6}$&$2.6 \times 10^{-27}$&$-0.4$&$\phantom{-}0.6_{7.7}$&$2.2 \times 10^{-25}$&$-0.4$&$\phantom{-}0.9_{9.5}$&$7.3 \times 10^{-20}$&$-0.3$&$\phantom{-}6.9_{26.6}$&$\phantom{-}8.3_{27.5}$&$\phantom{-}5.6_{23.1}$\\

Reflections on Strengths&$\phantom{-}31.6_{46.5}$&$5.7 \times 10^{-58}$&$\phantom{-}0.5$&$\phantom{-}20.1_{40.1}$&$1.9 \times 10^{-16}$&$\phantom{-}0.3$&$\phantom{-}14.8_{35.5}$&$7.5 \times 10^{-5}$&$\phantom{-}0.2$&\textcolor{gray}{$\phantom{-}12.0_{32.6}$}&$9.2 \times 10^{-1}$&$\phantom{-}0.1$&$\phantom{-}9.3_{32.4}$&$\phantom{-}14.1_{34.8}$&$\phantom{-}4.5_{20.8}$\\

Questions on Experiences&$\phantom{-}6.5_{24.7}$&$1.4 \times 10^{-129}$&$-0.8$&$\phantom{-}1.2_{11.0}$&$6.8 \times 10^{-197}$&$-1.1$&$\phantom{-}6.3_{24.2}$&$8.0 \times 10^{-126}$&$-0.8$&$\phantom{-}6.2_{24.1}$&$7.0 \times 10^{-117}$&$-0.8$&$\phantom{-}35.6_{48.6}$&$\phantom{-}40.2_{49.1}$&$\phantom{-}31.0_{46.3}$\\

Questions on Perspectives&$\phantom{-}1.1_{10.4}$&$2.6 \times 10^{-5}$&$-0.2$&$\phantom{-}0.0_{2.1}$&$7.7 \times 10^{-14}$&$-0.3$&$\phantom{-}1.7_{12.8}$&$2.8 \times 10^{-2}$&$-0.1$&$\phantom{-}1.0_{9.9}$&$3.9 \times 10^{-5}$&$-0.2$&$\phantom{-}3.5_{20.2}$&$\phantom{-}4.8_{21.5}$&$\phantom{-}2.3_{14.9}$\\

Questions on Emotions&$\phantom{-}0.2_{4.4}$&$6.3 \times 10^{-49}$&$-0.5$&$\phantom{-}0.1_{3.7}$&$3.6 \times 10^{-43}$&$-0.5$&$\phantom{-}0.4_{6.5}$&$1.4 \times 10^{-41}$&$-0.5$&$\phantom{-}0.2_{4.4}$&$8.3 \times 10^{-41}$&$-0.5$&$\phantom{-}9.6_{30.7}$&$\phantom{-}11.3_{31.6}$&$\phantom{-}7.9_{27.0}$\\

Problem-Solving&$\phantom{-}14.0_{34.7}$&$1.5 \times 10^{-29}$&$-0.4$&$\phantom{-}14.4_{35.1}$&$1.0 \times 10^{-25}$&$-0.4$&$\phantom{-}20.3_{40.2}$&$2.1 \times 10^{-7}$&$-0.2$&$\phantom{-}10.3_{30.4}$&$3.5 \times 10^{-47}$&$-0.5$&$\phantom{-}28.4_{42.7}$&$\phantom{-}20.4_{40.3}$&$\phantom{-}36.3_{48.2}$\\

Planning&$\phantom{-}19.5_{39.6}$&$8.7 \times 10^{-5}$&$\phantom{-}0.2$&\textcolor{gray}{$\phantom{-}15.3_{36.0}$}&$1.0 \times 10^{0}$&$\phantom{-}0.0$&$\phantom{-}22.9_{42.0}$&$2.5 \times 10^{-11}$&$\phantom{-}0.2$&\textcolor{gray}{$\phantom{-}11.9_{32.4}$}&$1.0 \times 10^{0}$&$-0.1$&$\phantom{-}13.7_{32.6}$&$\phantom{-}10.7_{30.9}$&$\phantom{-}16.6_{37.3}$\\

Normalizing&$\phantom{-}19.3_{39.5}$&$6.3 \times 10^{-4}$&$\phantom{-}0.1$&\textcolor{gray}{$\phantom{-}15.8_{36.5}$}&$1.0 \times 10^{0}$&$\phantom{-}0.1$&$\phantom{-}6.5_{24.7}$&$4.4 \times 10^{-14}$&$-0.3$&$\phantom{-}5.1_{22.0}$&$7.8 \times 10^{-21}$&$-0.3$&$\phantom{-}14.0_{32.6}$&$\phantom{-}10.5_{30.7}$&$\phantom{-}17.5_{38.0}$\\

Psychoeducation&\textcolor{gray}{$\phantom{-}3.4_{18.0}$}&$6.5 \times 10^{-2}$&$-0.1$&$\phantom{-}3.0_{17.0}$&$5.1 \times 10^{-3}$&$-0.1$&$\phantom{-}8.3_{27.5}$&$3.9 \times 10^{-2}$&$\phantom{-}0.1$&\textcolor{gray}{$\phantom{-}4.1_{19.9}$}&$1.0 \times 10^{0}$&$-0.1$&$\phantom{-}5.4_{20.5}$&$\phantom{-}3.5_{18.4}$&$\phantom{-}7.3_{26.1}$\\

\midrule
        \multicolumn{16}{c}{\textbf{Client behavior: Sustaining unhealthy behavior}} \\
        
\midrule
$n$ & \multicolumn{3}{c}{\textbf{$428$}} & \multicolumn{3}{c}{\textbf{$402$}} & \multicolumn{3}{c}{\textbf{$485$}} & \multicolumn{3}{c}{\textbf{$462$}}&\textbf{$389$}&\textbf{$275$}&\textbf{$114$}\\
\midrule
Reflections on Needs&$\phantom{-}5.4_{22.6}$&$1.5 \times 10^{-16}$&$-0.6$&$\phantom{-}2.2_{14.8}$&$7.6 \times 10^{-23}$&$-0.8$&$\phantom{-}7.0_{25.6}$&$9.8 \times 10^{-15}$&$-0.6$&$\phantom{-}6.1_{23.9}$&$4.4 \times 10^{-16}$&$-0.6$&$\phantom{-}27.6_{45.3}$&$\phantom{-}30.5_{46.1}$&$\phantom{-}24.6_{43.2}$\\

Reflections on Emotions&$\phantom{-}0.5_{6.8}$&$1.1 \times 10^{-9}$&$-0.5$&$\phantom{-}0.2_{5.0}$&$1.1 \times 10^{-9}$&$-0.5$&$\phantom{-}2.5_{15.5}$&$5.4 \times 10^{-6}$&$-0.4$&$\phantom{-}3.7_{18.8}$&$8.2 \times 10^{-4}$&$-0.3$&$\phantom{-}11.7_{33.2}$&$\phantom{-}13.8_{34.6}$&$\phantom{-}9.6_{29.7}$\\

Reflections on Values&$\phantom{-}1.2_{10.8}$&$2.0 \times 10^{-3}$&$-0.3$&$\phantom{-}0.0_{0.0}$&$5.6 \times 10^{-6}$&$-0.4$&$\phantom{-}0.6_{7.8}$&$2.9 \times 10^{-5}$&$-0.3$&$\phantom{-}0.9_{9.3}$&$2.2 \times 10^{-4}$&$-0.3$&$\phantom{-}6.8_{25.0}$&$\phantom{-}6.5_{24.8}$&$\phantom{-}7.0_{25.7}$\\

Reflections on Consequences&$\phantom{-}4.7_{21.1}$&$2.7 \times 10^{-31}$&$-0.9$&$\phantom{-}1.0_{9.9}$&$1.5 \times 10^{-41}$&$-1.0$&$\phantom{-}4.5_{20.8}$&$5.3 \times 10^{-35}$&$-0.9$&$\phantom{-}5.4_{22.6}$&$5.4 \times 10^{-31}$&$-0.9$&$\phantom{-}36.1_{46.8}$&$\phantom{-}26.5_{44.2}$&$\phantom{-}45.6_{50.0}$\\

Reflections on Conflicts&$\phantom{-}0.9_{9.6}$&$2.0 \times 10^{-5}$&$-0.4$&$\phantom{-}1.2_{11.1}$&$1.9 \times 10^{-4}$&$-0.3$&$\phantom{-}1.9_{13.5}$&$5.9 \times 10^{-4}$&$-0.3$&$\phantom{-}0.9_{9.3}$&$5.1 \times 10^{-6}$&$-0.4$&$\phantom{-}8.2_{27.1}$&$\phantom{-}7.6_{26.6}$&$\phantom{-}8.8_{28.4}$\\

Reflections on Strengths&\textcolor{gray}{$\phantom{-}3.5_{18.4}$}&$1.0 \times 10^{0}$&$-0.1$&\textcolor{gray}{$\phantom{-}1.2_{11.1}$}&$1.9 \times 10^{-1}$&$-0.2$&\textcolor{gray}{$\phantom{-}4.3_{20.4}$}&$1.0 \times 10^{0}$&$\phantom{-}0.0$&\textcolor{gray}{$\phantom{-}3.7_{18.8}$}&$1.0 \times 10^{0}$&$-0.1$&$\phantom{-}5.2_{23.6}$&$\phantom{-}6.9_{25.4}$&$\phantom{-}3.5_{18.5}$\\

Questions on Experiences&$\phantom{-}3.0_{17.2}$&$1.8 \times 10^{-39}$&$-1.0$&$\phantom{-}0.5_{7.0}$&$5.0 \times 10^{-46}$&$-1.1$&$\phantom{-}6.0_{23.7}$&$5.6 \times 10^{-34}$&$-0.9$&$\phantom{-}5.4_{22.6}$&$2.2 \times 10^{-34}$&$-0.9$&$\phantom{-}39.3_{49.4}$&$\phantom{-}46.2_{49.9}$&$\phantom{-}32.5_{47.0}$\\

Questions on Perspectives&\textcolor{gray}{$\phantom{-}0.7_{8.4}$}&$5.0 \times 10^{-1}$&$-0.2$&$\phantom{-}0.0_{0.0}$&$2.3 \times 10^{-2}$&$-0.3$&\textcolor{gray}{$\phantom{-}1.4_{11.9}$}&$1.0 \times 10^{0}$&$-0.1$&\textcolor{gray}{$\phantom{-}1.7_{13.1}$}&$1.0 \times 10^{0}$&$-0.1$&$\phantom{-}3.5_{19.3}$&$\phantom{-}4.4_{20.5}$&$\phantom{-}2.6_{16.1}$\\

Questions on Emotions&$\phantom{-}0.0_{0.0}$&$1.8 \times 10^{-8}$&$-0.4$&$\phantom{-}0.0_{0.0}$&$6.3 \times 10^{-8}$&$-0.4$&$\phantom{-}0.0_{0.0}$&$1.2 \times 10^{-9}$&$-0.5$&$\phantom{-}0.2_{4.7}$&$1.8 \times 10^{-8}$&$-0.4$&$\phantom{-}9.2_{29.7}$&$\phantom{-}10.5_{30.8}$&$\phantom{-}7.9_{27.1}$\\

Problem-Solving&$\phantom{-}2.8_{16.5}$&$1.1 \times 10^{-19}$&$-0.7$&$\phantom{-}1.5_{12.1}$&$1.9 \times 10^{-22}$&$-0.7$&$\phantom{-}12.2_{32.7}$&$3.2 \times 10^{-4}$&$-0.3$&$\phantom{-}6.9_{25.4}$&$4.0 \times 10^{-11}$&$-0.5$&$\phantom{-}23.6_{40.3}$&$\phantom{-}15.6_{36.4}$&$\phantom{-}31.6_{46.7}$\\

Planning&\textcolor{gray}{$\phantom{-}1.9_{13.6}$}&$1.1 \times 10^{-1}$&$-0.2$&$\phantom{-}0.5_{7.0}$&$2.4 \times 10^{-4}$&$-0.3$&\textcolor{gray}{$\phantom{-}8.9_{28.5}$}&$1.0 \times 10^{0}$&$\phantom{-}0.1$&\textcolor{gray}{$\phantom{-}4.1_{19.9}$}&$1.0 \times 10^{0}$&$-0.1$&$\phantom{-}5.9_{22.1}$&$\phantom{-}4.0_{19.6}$&$\phantom{-}7.9_{27.1}$\\

Normalizing&$\phantom{-}2.1_{14.4}$&$1.9 \times 10^{-9}$&$-0.5$&$\phantom{-}0.7_{8.6}$&$2.0 \times 10^{-12}$&$-0.6$&$\phantom{-}2.3_{14.9}$&$4.5 \times 10^{-10}$&$-0.5$&$\phantom{-}1.7_{13.1}$&$4.4 \times 10^{-11}$&$-0.5$&$\phantom{-}14.0_{33.2}$&$\phantom{-}10.5_{30.8}$&$\phantom{-}17.5_{38.2}$\\

Psychoeducation&$\phantom{-}0.2_{4.8}$&$5.6 \times 10^{-11}$&$-0.5$&$\phantom{-}0.5_{7.0}$&$2.1 \times 10^{-9}$&$-0.5$&$\phantom{-}3.3_{17.9}$&$1.4 \times 10^{-3}$&$-0.3$&$\phantom{-}1.5_{12.2}$&$2.6 \times 10^{-7}$&$-0.4$&$\phantom{-}9.8_{26.7}$&$\phantom{-}4.7_{21.3}$&$\phantom{-}14.9_{35.8}$\\

        \midrule
          \multicolumn{16}{c}{\textbf{Client behavior: Sharing positive emotions}} \\
          \midrule
          $n$ & \multicolumn{3}{c}{\textbf{$815$}} & \multicolumn{3}{c}{\textbf{$596$}} & \multicolumn{3}{c}{\textbf{$389$}} & \multicolumn{3}{c}{\textbf{$437$}}&\textbf{$294$}&\textbf{$226$}&\textbf{$68$}\\
          \midrule
Reflections on Needs&$\phantom{-}4.0_{19.7}$&$7.2 \times 10^{-11}$&$-0.5$&\textcolor{gray}{$\phantom{-}10.7_{31.0}$}&$6.0 \times 10^{-1}$&$-0.2$&$\phantom{-}2.8_{16.6}$&$3.3 \times 10^{-8}$&$-0.5$&$\phantom{-}3.0_{17.0}$&$8.8 \times 10^{-9}$&$-0.5$&$\phantom{-}17.3_{40.9}$&$\phantom{-}24.3_{43.0}$&$\phantom{-}10.3_{30.6}$\\

Reflections on Emotions&$\phantom{-}2.3_{15.1}$&$2.2 \times 10^{-2}$&$-0.2$&\textcolor{gray}{$\phantom{-}3.4_{18.0}$}&$1.0 \times 10^{0}$&$-0.2$&$\phantom{-}0.3_{5.1}$&$9.7 \times 10^{-4}$&$-0.3$&$\phantom{-}0.9_{9.5}$&$4.4 \times 10^{-3}$&$-0.3$&$\phantom{-}7.5_{32.0}$&$\phantom{-}15.0_{35.8}$&$\phantom{-}0.0_{0.0}$\\

Reflections on Values&\textcolor{gray}{$\phantom{-}0.9_{9.2}$}&$3.9 \times 10^{-1}$&$-0.2$&$\phantom{-}0.2_{4.1}$&$1.9 \times 10^{-2}$&$-0.3$&\textcolor{gray}{$\phantom{-}0.8_{8.8}$}&$1.0 \times 10^{0}$&$-0.2$&\textcolor{gray}{$\phantom{-}0.2_{4.8}$}&$1.6 \times 10^{-1}$&$-0.2$&$\phantom{-}3.4_{20.6}$&$\phantom{-}5.3_{22.5}$&$\phantom{-}1.5_{12.1}$\\

Reflections on Consequences&$\phantom{-}0.6_{7.8}$&$4.0 \times 10^{-29}$&$-0.8$&$\phantom{-}0.7_{8.2}$&$1.0 \times 10^{-21}$&$-0.7$&$\phantom{-}0.5_{7.2}$&$3.1 \times 10^{-15}$&$-0.7$&$\phantom{-}0.9_{9.5}$&$2.0 \times 10^{-15}$&$-0.7$&$\phantom{-}15.8_{34.0}$&$\phantom{-}11.1_{31.4}$&$\phantom{-}20.6_{40.7}$\\

Reflections on Conflicts&$\phantom{-}0.0_{0.0}$&$2.2 \times 10^{-4}$&$-0.3$&$\phantom{-}0.0_{0.0}$&$4.9 \times 10^{-3}$&$-0.3$&\textcolor{gray}{$\phantom{-}0.0_{0.0}$}&$9.5 \times 10^{-2}$&$-0.3$&\textcolor{gray}{$\phantom{-}0.7_{8.3}$}&$1.0 \times 10^{0}$&$-0.2$&$\phantom{-}3.4_{20.6}$&$\phantom{-}5.3_{22.5}$&$\phantom{-}1.5_{12.1}$\\

Reflections on Strengths&$\phantom{-}62.2_{48.5}$&$1.4 \times 10^{-55}$&$\phantom{-}1.2$&$\phantom{-}42.8_{49.5}$&$3.6 \times 10^{-21}$&$\phantom{-}0.7$&\textcolor{gray}{$\phantom{-}14.4_{35.2}$}&$1.0 \times 10^{0}$&$\phantom{-}0.1$&\textcolor{gray}{$\phantom{-}17.2_{37.7}$}&$8.4 \times 10^{-1}$&$\phantom{-}0.2$&$\phantom{-}10.2_{34.0}$&$\phantom{-}15.9_{36.7}$&$\phantom{-}4.4_{20.7}$\\

Questions on Experiences&$\phantom{-}0.9_{9.2}$&$3.1 \times 10^{-59}$&$-1.2$&$\phantom{-}0.2_{4.1}$&$4.9 \times 10^{-49}$&$-1.1$&$\phantom{-}4.1_{19.9}$&$3.2 \times 10^{-21}$&$-0.8$&$\phantom{-}6.6_{24.9}$&$2.0 \times 10^{-17}$&$-0.7$&$\phantom{-}31.6_{47.6}$&$\phantom{-}36.7_{48.3}$&$\phantom{-}26.5_{44.4}$\\

Questions on Perspectives&\textcolor{gray}{$\phantom{-}0.1_{3.5}$}&$1.0 \times 10^{0}$&$-0.2$&\textcolor{gray}{$\phantom{-}0.2_{4.1}$}&$1.0 \times 10^{0}$&$-0.1$&\textcolor{gray}{$\phantom{-}0.3_{5.1}$}&$1.0 \times 10^{0}$&$-0.1$&\textcolor{gray}{$\phantom{-}0.5_{6.8}$}&$1.0 \times 10^{0}$&$-0.1$&$\phantom{-}1.3_{14.2}$&$\phantom{-}2.7_{16.1}$&$\phantom{-}0.0_{0.0}$\\

Questions on Emotions&$\phantom{-}0.0_{0.0}$&$1.8 \times 10^{-11}$&$-0.5$&$\phantom{-}0.3_{5.8}$&$6.0 \times 10^{-7}$&$-0.4$&$\phantom{-}0.0_{0.0}$&$3.0 \times 10^{-5}$&$-0.4$&$\phantom{-}0.2_{4.8}$&$2.4 \times 10^{-5}$&$-0.4$&$\phantom{-}7.5_{28.9}$&$\phantom{-}10.6_{30.9}$&$\phantom{-}4.4_{20.7}$\\

Problem-Solving&\textcolor{gray}{$\phantom{-}7.7_{26.7}$}&$5.2 \times 10^{-1}$&$-0.2$&\textcolor{gray}{$\phantom{-}10.6_{30.8}$}&$1.0 \times 10^{0}$&$-0.1$&\textcolor{gray}{$\phantom{-}7.2_{25.9}$}&$8.2 \times 10^{-1}$&$-0.2$&$\phantom{-}3.4_{18.2}$&$6.1 \times 10^{-5}$&$-0.4$&$\phantom{-}13.0_{33.6}$&$\phantom{-}12.8_{33.5}$&$\phantom{-}13.2_{34.1}$\\

Planning&$\phantom{-}43.6_{49.6}$&$1.1 \times 10^{-27}$&$\phantom{-}0.8$&$\phantom{-}31.0_{46.3}$&$1.0 \times 10^{-12}$&$\phantom{-}0.6$&$\phantom{-}18.5_{38.9}$&$8.1 \times 10^{-3}$&$\phantom{-}0.3$&\textcolor{gray}{$\phantom{-}12.4_{32.9}$}&$1.0 \times 10^{0}$&$\phantom{-}0.1$&$\phantom{-}8.2_{26.9}$&$\phantom{-}7.5_{26.4}$&$\phantom{-}8.8_{28.6}$\\

Normalizing&$\phantom{-}58.5_{49.3}$&$1.5 \times 10^{-25}$&$\phantom{-}0.8$&$\phantom{-}38.9_{48.8}$&$8.6 \times 10^{-5}$&$\phantom{-}0.3$&$\phantom{-}8.2_{27.5}$&$3.0 \times 10^{-6}$&$-0.4$&$\phantom{-}11.4_{31.9}$&$2.3 \times 10^{-3}$&$-0.3$&$\phantom{-}22.7_{40.6}$&$\phantom{-}19.0_{39.3}$&$\phantom{-}26.5_{44.4}$\\

Psychoeducation&\textcolor{gray}{$\phantom{-}3.4_{18.2}$}&$1.0 \times 10^{0}$&$\phantom{-}0.0$&\textcolor{gray}{$\phantom{-}1.8_{13.5}$}&$1.0 \times 10^{0}$&$-0.1$&\textcolor{gray}{$\phantom{-}5.1_{22.1}$}&$1.0 \times 10^{0}$&$\phantom{-}0.1$&\textcolor{gray}{$\phantom{-}3.0_{17.0}$}&$1.0 \times 10^{0}$&$\phantom{-}0.0$&$\phantom{-}3.1_{15.3}$&$\phantom{-}1.8_{13.2}$&$\phantom{-}4.4_{20.7}$\\

    \bottomrule
    \end{tabular}
    }
    \label{tab:adapt_new_hope_p1}
\end{table*}
\begin{table*}[t]
    \centering
    \small
    \caption{\showedits{(Continue Table \ref{tab:adapt_new_hope_p1}) Adaptability (frequency of conversational behaviors exhibited by therapists \textit{in response to specific client behaviors}) by LLM
therapists (GPT-4, GPT-3.5-turbo, Llama2-70b, Llama2-13b) in HOPE dataset \cite{malhotra2022speaker}. Values in \textcolor{gray}{gray} are not statistically significantly different from human therapist average at $p = \frac{0.05}{m}$ using Two-sided Student’s t-test after Bonferroni correction, where $m$ is the number of t-test per one model ($m = 13 \times 6 = 78$). $d$ is the cohen's d when compared with human therapist average. For instance, GPT-4 exhibits \textsc{Reflections on Needs} by 14.0\%  and human therapists responds by 24.5\% on average when client expresses \textsc{Sharing negative emotions}. ($P = 3.2 \times 10^{-7}$, Cohen's $d = \num{-0.3}$, two-sided Student’s $t$-test).}}
    \resizebox{1\textwidth}{!}{%
    \def\arraystretch{1.15}
    \setlength{\tabcolsep}{2pt}
    \begin{tabular}{l|ccc|ccc|ccc|ccc|ccc|ccc}
         \toprule 
         \multirow{3}{*}{\textbf{Therapist Behavior}} & \multicolumn{12}{c}{\textbf{LLM Therapists (Single Response \& Full Conversation Simulations)}} & \multicolumn{3}{c}{\textbf{Human Therapists}} \\
         \cmidrule(lr){2-13} \cmidrule(lr){14-16}
         & \multicolumn{3}{c}{\textbf{GPT-4}} &\multicolumn{3}{c}{\textbf{GPT-3.5}}&\multicolumn{3}{c}{\textbf{Llama2-70b}}&\multicolumn{3}{c}{\textbf{Llama2-13b}}& \textbf{Average}&\textbf{High-quality}&\textbf{Low-quality}\\
         \cmidrule(lr){2-4} \cmidrule(lr){5-7} \cmidrule(lr){8-10} \cmidrule(lr){11-13} \cmidrule(lr){14-14} \cmidrule(lr){15-15}  \cmidrule(lr){16-16} 
        & freq. & $p$ & $d$ & freq. & $p$ & $d$ & freq. & $p$& $d$ & freq. & $p$ & $d$ & freq. & freq. & freq. \\
\midrule

          \multicolumn{16}{c}{\textbf{Client behavior: Sharing negative emotions}} \\
          \midrule
          $n$ & \multicolumn{3}{c}{\textbf{$1175$}} & \multicolumn{3}{c}{\textbf{$1052$}} & \multicolumn{3}{c}{\textbf{$1354$}} & \multicolumn{3}{c}{\textbf{$1348$}}&\textbf{$820$}&\textbf{$646$}&\textbf{$174$}\\
          \midrule
Reflections on Needs&$\phantom{-}14.0_{34.8}$&$3.2 \times 10^{-7}$&$-0.3$&$\phantom{-}7.2_{25.9}$&$2.3 \times 10^{-23}$&$-0.5$&$\phantom{-}9.5_{29.4}$&$5.6 \times 10^{-19}$&$-0.4$&$\phantom{-}9.2_{28.9}$&$5.1 \times 10^{-20}$&$-0.4$&$\phantom{-}24.5_{44.5}$&$\phantom{-}28.9_{45.4}$&$\phantom{-}20.1_{40.2}$\\

Reflections on Emotions&$\phantom{-}9.0_{28.7}$&$6.9 \times 10^{-9}$&$-0.3$&$\phantom{-}7.2_{25.9}$&$2.6 \times 10^{-12}$&$-0.4$&$\phantom{-}8.1_{27.3}$&$5.7 \times 10^{-12}$&$-0.3$&$\phantom{-}12.2_{32.8}$&$8.3 \times 10^{-4}$&$-0.2$&$\phantom{-}19.4_{42.7}$&$\phantom{-}27.4_{44.6}$&$\phantom{-}11.5_{32.0}$\\

Reflections on Values&$\phantom{-}2.5_{15.5}$&$2.5 \times 10^{-3}$&$-0.2$&$\phantom{-}0.2_{4.4}$&$2.5 \times 10^{-12}$&$-0.4$&$\phantom{-}0.8_{9.0}$&$5.2 \times 10^{-11}$&$-0.3$&$\phantom{-}1.3_{11.2}$&$2.2 \times 10^{-8}$&$-0.3$&$\phantom{-}6.3_{25.4}$&$\phantom{-}7.4_{26.2}$&$\phantom{-}5.2_{22.2}$\\

Reflections on Consequences&$\phantom{-}7.1_{25.8}$&$6.6 \times 10^{-21}$&$-0.5$&$\phantom{-}3.6_{18.7}$&$1.7 \times 10^{-35}$&$-0.6$&$\phantom{-}3.5_{18.3}$&$5.4 \times 10^{-44}$&$-0.6$&$\phantom{-}4.4_{20.5}$&$1.8 \times 10^{-37}$&$-0.6$&$\phantom{-}21.8_{39.8}$&$\phantom{-}18.3_{38.7}$&$\phantom{-}25.3_{43.6}$\\

Reflections on Conflicts&$\phantom{-}1.4_{11.6}$&$1.3 \times 10^{-11}$&$-0.3$&$\phantom{-}0.6_{7.5}$&$1.7 \times 10^{-14}$&$-0.4$&$\phantom{-}1.1_{10.5}$&$1.4 \times 10^{-14}$&$-0.4$&$\phantom{-}2.3_{15.0}$&$1.7 \times 10^{-8}$&$-0.3$&$\phantom{-}8.4_{29.5}$&$\phantom{-}10.5_{30.7}$&$\phantom{-}6.3_{24.4}$\\

Reflections on Strengths&$\phantom{-}13.5_{34.2}$&$8.4 \times 10^{-5}$&$\phantom{-}0.2$&\textcolor{gray}{$\phantom{-}7.0_{25.6}$}&$1.0 \times 10^{0}$&$\phantom{-}0.0$&$\phantom{-}12.6_{33.1}$&$9.1 \times 10^{-4}$&$\phantom{-}0.2$&\textcolor{gray}{$\phantom{-}10.4_{30.5}$}&$2.3 \times 10^{-1}$&$\phantom{-}0.1$&$\phantom{-}6.6_{26.6}$&$\phantom{-}8.5_{27.9}$&$\phantom{-}4.6_{21.0}$\\

Questions on Experiences&$\phantom{-}11.6_{32.0}$&$4.0 \times 10^{-34}$&$-0.6$&$\phantom{-}3.5_{18.4}$&$1.6 \times 10^{-72}$&$-0.9$&$\phantom{-}13.4_{34.0}$&$3.6 \times 10^{-30}$&$-0.5$&$\phantom{-}15.7_{36.4}$&$1.8 \times 10^{-22}$&$-0.5$&$\phantom{-}34.6_{48.6}$&$\phantom{-}40.6_{49.1}$&$\phantom{-}28.7_{45.4}$\\

Questions on Perspectives&\textcolor{gray}{$\phantom{-}1.1_{10.5}$}&$1.4 \times 10^{-1}$&$-0.1$&$\phantom{-}0.2_{4.4}$&$1.1 \times 10^{-4}$&$-0.2$&\textcolor{gray}{$\phantom{-}2.7_{16.1}$}&$1.0 \times 10^{0}$&$\phantom{-}0.0$&\textcolor{gray}{$\phantom{-}2.2_{14.5}$}&$1.0 \times 10^{0}$&$-0.1$&$\phantom{-}3.3_{19.9}$&$\phantom{-}4.8_{21.4}$&$\phantom{-}1.7_{13.1}$\\

Questions on Emotions&$\phantom{-}0.2_{4.1}$&$2.7 \times 10^{-22}$&$-0.5$&$\phantom{-}0.4_{6.2}$&$1.3 \times 10^{-18}$&$-0.4$&$\phantom{-}0.5_{7.2}$&$1.7 \times 10^{-22}$&$-0.5$&$\phantom{-}0.4_{6.1}$&$1.2 \times 10^{-23}$&$-0.5$&$\phantom{-}9.6_{31.1}$&$\phantom{-}11.8_{32.2}$&$\phantom{-}7.5_{26.4}$\\

Problem-Solving&$\phantom{-}11.1_{31.5}$&$2.1 \times 10^{-5}$&$-0.2$&\textcolor{gray}{$\phantom{-}14.2_{34.9}$}&$2.7 \times 10^{-1}$&$-0.1$&\textcolor{gray}{$\phantom{-}22.6_{41.8}$}&$1.0 \times 10^{0}$&$\phantom{-}0.1$&\textcolor{gray}{$\phantom{-}15.4_{36.1}$}&$1.0 \times 10^{0}$&$-0.1$&$\phantom{-}19.0_{35.7}$&$\phantom{-}12.1_{32.6}$&$\phantom{-}25.9_{43.9}$\\

Planning&\textcolor{gray}{$\phantom{-}5.3_{22.4}$}&$1.0 \times 10^{0}$&$-0.1$&\textcolor{gray}{$\phantom{-}6.0_{23.7}$}&$1.0 \times 10^{0}$&$-0.1$&$\phantom{-}17.6_{38.1}$&$2.5 \times 10^{-9}$&$\phantom{-}0.3$&\textcolor{gray}{$\phantom{-}10.4_{30.5}$}&$1.0 \times 10^{0}$&$\phantom{-}0.1$&$\phantom{-}7.6_{25.0}$&$\phantom{-}6.0_{23.8}$&$\phantom{-}9.2_{29.0}$\\

Normalizing&$\phantom{-}9.8_{29.7}$&$4.1 \times 10^{-4}$&$-0.2$&$\phantom{-}10.6_{30.7}$&$8.7 \times 10^{-3}$&$-0.2$&$\phantom{-}6.7_{25.0}$&$9.3 \times 10^{-12}$&$-0.3$&$\phantom{-}6.7_{25.0}$&$7.2 \times 10^{-12}$&$-0.3$&$\phantom{-}16.5_{36.2}$&$\phantom{-}14.7_{35.4}$&$\phantom{-}18.4_{38.9}$\\

Psychoeducation&\textcolor{gray}{$\phantom{-}2.6_{15.8}$}&$2.8 \times 10^{-1}$&$-0.1$&\textcolor{gray}{$\phantom{-}2.9_{16.9}$}&$1.0 \times 10^{0}$&$-0.1$&\textcolor{gray}{$\phantom{-}7.6_{26.5}$}&$4.7 \times 10^{-1}$&$\phantom{-}0.1$&\textcolor{gray}{$\phantom{-}4.3_{20.3}$}&$1.0 \times 10^{0}$&$\phantom{-}0.0$&$\phantom{-}4.7_{17.5}$&$\phantom{-}2.0_{14.1}$&$\phantom{-}7.5_{26.4}$\\

        \midrule
          \multicolumn{16}{c}{\textbf{Client behavior: Sharing experiences}} \\
          \midrule
          $n$ & \multicolumn{3}{c}{\textbf{$874$}} & \multicolumn{3}{c}{\textbf{$812$}} & \multicolumn{3}{c}{\textbf{$908$}} & \multicolumn{3}{c}{\textbf{$926$}}&\textbf{$759$}&\textbf{$533$}&\textbf{$226$}\\
          \midrule
Reflections on Needs&$\phantom{-}6.5_{24.7}$&$1.5 \times 10^{-18}$&$-0.5$&$\phantom{-}1.4_{11.6}$&$7.5 \times 10^{-38}$&$-0.7$&$\phantom{-}3.9_{19.3}$&$5.6 \times 10^{-29}$&$-0.6$&$\phantom{-}5.0_{21.7}$&$8.2 \times 10^{-25}$&$-0.5$&$\phantom{-}22.5_{42.8}$&$\phantom{-}26.5_{44.2}$&$\phantom{-}18.6_{39.0}$\\

Reflections on Emotions&$\phantom{-}5.8_{23.5}$&$3.3 \times 10^{-5}$&$-0.3$&$\phantom{-}3.0_{16.9}$&$2.2 \times 10^{-11}$&$-0.4$&$\phantom{-}4.4_{20.5}$&$2.1 \times 10^{-8}$&$-0.3$&$\phantom{-}6.0_{23.8}$&$4.8 \times 10^{-5}$&$-0.2$&$\phantom{-}13.5_{36.8}$&$\phantom{-}19.9_{40.0}$&$\phantom{-}7.1_{25.7}$\\

Reflections on Values&$\phantom{-}0.6_{7.5}$&$6.6 \times 10^{-5}$&$-0.2$&$\phantom{-}0.0_{0.0}$&$8.2 \times 10^{-7}$&$-0.3$&$\phantom{-}0.8_{8.8}$&$2.3 \times 10^{-4}$&$-0.2$&$\phantom{-}1.3_{11.3}$&$7.6 \times 10^{-3}$&$-0.2$&$\phantom{-}4.7_{23.1}$&$\phantom{-}7.1_{25.8}$&$\phantom{-}2.2_{14.7}$\\

Reflections on Consequences&$\phantom{-}2.4_{15.3}$&$4.9 \times 10^{-26}$&$-0.6$&$\phantom{-}1.0_{9.9}$&$1.8 \times 10^{-31}$&$-0.6$&$\phantom{-}1.9_{13.6}$&$1.2 \times 10^{-29}$&$-0.6$&$\phantom{-}1.7_{13.0}$&$5.5 \times 10^{-31}$&$-0.6$&$\phantom{-}18.2_{38.5}$&$\phantom{-}17.8_{38.3}$&$\phantom{-}18.6_{39.0}$\\

Reflections on Conflicts&$\phantom{-}1.0_{10.1}$&$2.0 \times 10^{-5}$&$-0.3$&$\phantom{-}0.1_{3.5}$&$1.7 \times 10^{-8}$&$-0.3$&$\phantom{-}0.9_{9.4}$&$3.8 \times 10^{-6}$&$-0.3$&$\phantom{-}1.1_{10.3}$&$1.6 \times 10^{-5}$&$-0.3$&$\phantom{-}5.8_{25.1}$&$\phantom{-}8.1_{27.3}$&$\phantom{-}3.5_{18.5}$\\

Reflections on Strengths&\textcolor{gray}{$\phantom{-}4.5_{20.7}$}&$1.3 \times 10^{-1}$&$-0.2$&$\phantom{-}1.8_{13.5}$&$1.1 \times 10^{-6}$&$-0.3$&$\phantom{-}4.1_{19.8}$&$2.9 \times 10^{-2}$&$-0.2$&$\phantom{-}3.7_{18.8}$&$5.5 \times 10^{-3}$&$-0.2$&$\phantom{-}8.4_{29.9}$&$\phantom{-}12.0_{32.5}$&$\phantom{-}4.9_{21.6}$\\

Questions on Experiences&$\phantom{-}5.8_{23.5}$&$3.2 \times 10^{-82}$&$-1.0$&$\phantom{-}2.3_{15.1}$&$2.3 \times 10^{-100}$&$-1.2$&$\phantom{-}8.6_{28.0}$&$1.3 \times 10^{-68}$&$-0.9$&$\phantom{-}9.1_{28.7}$&$5.0 \times 10^{-67}$&$-0.9$&$\phantom{-}44.7_{49.8}$&$\phantom{-}45.2_{49.8}$&$\phantom{-}44.2_{49.8}$\\

Questions on Perspectives&$\phantom{-}0.1_{3.4}$&$3.2 \times 10^{-2}$&$-0.2$&$\phantom{-}0.0_{0.0}$&$1.6 \times 10^{-2}$&$-0.2$&\textcolor{gray}{$\phantom{-}1.1_{10.4}$}&$1.0 \times 10^{0}$&$-0.1$&\textcolor{gray}{$\phantom{-}1.0_{9.8}$}&$1.0 \times 10^{0}$&$-0.1$&$\phantom{-}2.0_{15.6}$&$\phantom{-}3.2_{17.6}$&$\phantom{-}0.9_{9.4}$\\

Questions on Emotions&$\phantom{-}0.1_{3.4}$&$1.0 \times 10^{-16}$&$-0.4$&$\phantom{-}0.4_{6.1}$&$3.5 \times 10^{-14}$&$-0.4$&$\phantom{-}0.2_{4.7}$&$1.1 \times 10^{-16}$&$-0.4$&$\phantom{-}0.9_{9.3}$&$2.5 \times 10^{-13}$&$-0.4$&$\phantom{-}8.8_{28.6}$&$\phantom{-}9.2_{28.9}$&$\phantom{-}8.4_{27.8}$\\

Problem-Solving&$\phantom{-}4.7_{21.2}$&$5.5 \times 10^{-10}$&$-0.3$&$\phantom{-}3.2_{17.6}$&$3.4 \times 10^{-14}$&$-0.4$&$\phantom{-}5.9_{23.7}$&$5.8 \times 10^{-7}$&$-0.3$&$\phantom{-}6.2_{24.0}$&$1.4 \times 10^{-6}$&$-0.3$&$\phantom{-}14.0_{32.7}$&$\phantom{-}9.4_{29.2}$&$\phantom{-}18.6_{39.0}$\\

Planning&$\phantom{-}2.3_{15.0}$&$3.0 \times 10^{-4}$&$-0.2$&$\phantom{-}1.7_{13.0}$&$1.6 \times 10^{-5}$&$-0.3$&\textcolor{gray}{$\phantom{-}5.3_{22.4}$}&$1.0 \times 10^{0}$&$-0.1$&\textcolor{gray}{$\phantom{-}4.1_{19.8}$}&$6.7 \times 10^{-1}$&$-0.1$&$\phantom{-}7.0_{25.5}$&$\phantom{-}6.9_{25.4}$&$\phantom{-}7.1_{25.7}$\\

Normalizing&$\phantom{-}4.0_{19.6}$&$3.5 \times 10^{-11}$&$-0.4$&$\phantom{-}3.2_{17.6}$&$5.2 \times 10^{-13}$&$-0.4$&$\phantom{-}2.5_{15.7}$&$7.7 \times 10^{-17}$&$-0.4$&$\phantom{-}2.6_{15.9}$&$7.4 \times 10^{-17}$&$-0.4$&$\phantom{-}14.0_{34.7}$&$\phantom{-}13.9_{34.6}$&$\phantom{-}14.2_{34.9}$\\

Psychoeducation&$\phantom{-}0.7_{8.3}$&$3.5 \times 10^{-9}$&$-0.3$&$\phantom{-}0.2_{5.0}$&$2.7 \times 10^{-11}$&$-0.4$&$\phantom{-}1.5_{12.3}$&$2.7 \times 10^{-5}$&$-0.3$&$\phantom{-}0.5_{7.3}$&$1.3 \times 10^{-10}$&$-0.3$&$\phantom{-}5.7_{20.7}$&$\phantom{-}2.6_{16.0}$&$\phantom{-}8.8_{28.5}$\\

        \midrule
          \multicolumn{16}{c}{\textbf{Client behavior: Gained insights}} \\
          \midrule
          $n$ & \multicolumn{3}{c}{\textbf{$561$}} & \multicolumn{3}{c}{\textbf{$382$}} & \multicolumn{3}{c}{\textbf{$490$}} & \multicolumn{3}{c}{\textbf{$425$}}&\textbf{$258$}&\textbf{$214$}&\textbf{$44$}\\
          \midrule
Reflections on Needs&$\phantom{-}10.0_{30.0}$&$2.4 \times 10^{-8}$&$-0.5$&$\phantom{-}7.3_{26.1}$&$6.5 \times 10^{-10}$&$-0.6$&$\phantom{-}6.1_{24.0}$&$5.3 \times 10^{-14}$&$-0.6$&$\phantom{-}7.8_{26.8}$&$5.3 \times 10^{-10}$&$-0.6$&$\phantom{-}27.0_{45.1}$&$\phantom{-}29.0_{45.5}$&$\phantom{-}25.0_{43.8}$\\

Reflections on Emotions&$\phantom{-}3.4_{18.1}$&$1.1 \times 10^{-4}$&$-0.4$&$\phantom{-}2.9_{16.7}$&$4.2 \times 10^{-4}$&$-0.4$&$\phantom{-}3.1_{17.2}$&$9.8 \times 10^{-5}$&$-0.4$&$\phantom{-}3.5_{18.5}$&$1.4 \times 10^{-3}$&$-0.3$&$\phantom{-}12.5_{35.5}$&$\phantom{-}15.9_{36.6}$&$\phantom{-}9.1_{29.1}$\\

Reflections on Values&$\phantom{-}1.8_{13.2}$&$1.4 \times 10^{-3}$&$-0.3$&$\phantom{-}0.3_{5.1}$&$1.8 \times 10^{-5}$&$-0.4$&$\phantom{-}2.2_{14.8}$&$1.9 \times 10^{-2}$&$-0.3$&$\phantom{-}1.4_{11.8}$&$1.9 \times 10^{-3}$&$-0.3$&$\phantom{-}8.1_{28.6}$&$\phantom{-}9.3_{29.2}$&$\phantom{-}6.8_{25.5}$\\

Reflections on Consequences&$\phantom{-}4.6_{21.0}$&$5.1 \times 10^{-10}$&$-0.5$&$\phantom{-}2.4_{15.2}$&$2.1 \times 10^{-12}$&$-0.6$&$\phantom{-}1.0_{10.1}$&$3.0 \times 10^{-20}$&$-0.8$&$\phantom{-}1.4_{11.8}$&$1.9 \times 10^{-16}$&$-0.7$&$\phantom{-}18.8_{37.0}$&$\phantom{-}15.0_{35.7}$&$\phantom{-}22.7_{42.4}$\\

Reflections on Conflicts&\textcolor{gray}{$\phantom{-}0.7_{8.4}$}&$1.9 \times 10^{-1}$&$-0.2$&\textcolor{gray}{$\phantom{-}0.3_{5.1}$}&$1.3 \times 10^{-1}$&$-0.3$&\textcolor{gray}{$\phantom{-}0.8_{9.0}$}&$4.8 \times 10^{-1}$&$-0.2$&\textcolor{gray}{$\phantom{-}1.4_{11.8}$}&$1.0 \times 10^{0}$&$-0.2$&$\phantom{-}3.9_{21.9}$&$\phantom{-}5.6_{23.1}$&$\phantom{-}2.3_{15.1}$\\

Reflections on Strengths&$\phantom{-}47.4_{50.0}$&$7.8 \times 10^{-21}$&$\phantom{-}0.8$&\textcolor{gray}{$\phantom{-}23.3_{42.3}$}&$6.1 \times 10^{-2}$&$\phantom{-}0.3$&\textcolor{gray}{$\phantom{-}21.4_{41.1}$}&$2.5 \times 10^{-1}$&$\phantom{-}0.2$&\textcolor{gray}{$\phantom{-}17.9_{38.4}$}&$1.0 \times 10^{0}$&$\phantom{-}0.1$&$\phantom{-}12.5_{35.5}$&$\phantom{-}15.9_{36.6}$&$\phantom{-}9.1_{29.1}$\\

Questions on Experiences&$\phantom{-}4.5_{20.7}$&$2.8 \times 10^{-21}$&$-0.8$&$\phantom{-}0.8_{8.8}$&$1.2 \times 10^{-24}$&$-0.9$&$\phantom{-}6.7_{25.1}$&$2.7 \times 10^{-14}$&$-0.6$&$\phantom{-}6.6_{24.8}$&$3.5 \times 10^{-13}$&$-0.6$&$\phantom{-}29.2_{48.6}$&$\phantom{-}42.5_{49.6}$&$\phantom{-}15.9_{37.0}$\\

Questions on Perspectives&\textcolor{gray}{$\phantom{-}2.1_{14.5}$}&$1.0 \times 10^{0}$&$-0.1$&\textcolor{gray}{$\phantom{-}0.3_{5.1}$}&$9.5 \times 10^{-2}$&$-0.3$&\textcolor{gray}{$\phantom{-}2.0_{14.2}$}&$1.0 \times 10^{0}$&$-0.1$&\textcolor{gray}{$\phantom{-}2.6_{15.9}$}&$1.0 \times 10^{0}$&$-0.1$&$\phantom{-}4.2_{22.7}$&$\phantom{-}6.1_{23.9}$&$\phantom{-}2.3_{15.1}$\\

Questions on Emotions&$\phantom{-}0.0_{0.0}$&$2.0 \times 10^{-14}$&$-0.6$&$\phantom{-}0.3_{5.1}$&$6.9 \times 10^{-9}$&$-0.5$&$\phantom{-}0.4_{6.4}$&$9.4 \times 10^{-11}$&$-0.6$&$\phantom{-}0.0_{0.0}$&$7.3 \times 10^{-11}$&$-0.6$&$\phantom{-}10.8_{30.7}$&$\phantom{-}10.3_{30.4}$&$\phantom{-}11.4_{32.1}$\\

Problem-Solving&\textcolor{gray}{$\phantom{-}18.0_{38.5}$}&$5.5 \times 10^{-2}$&$-0.3$&\textcolor{gray}{$\phantom{-}17.3_{37.9}$}&$5.1 \times 10^{-2}$&$-0.3$&\textcolor{gray}{$\phantom{-}24.9_{43.3}$}&$1.0 \times 10^{0}$&$-0.1$&$\phantom{-}11.1_{31.4}$&$9.9 \times 10^{-8}$&$-0.5$&$\phantom{-}27.9_{39.6}$&$\phantom{-}15.0_{35.7}$&$\phantom{-}40.9_{49.7}$\\

Planning&$\phantom{-}25.8_{43.8}$&$2.4 \times 10^{-4}$&$\phantom{-}0.4$&\textcolor{gray}{$\phantom{-}16.2_{36.9}$}&$1.0 \times 10^{0}$&$\phantom{-}0.1$&$\phantom{-}33.7_{47.3}$&$2.4 \times 10^{-9}$&$\phantom{-}0.5$&\textcolor{gray}{$\phantom{-}16.7_{37.3}$}&$1.0 \times 10^{0}$&$\phantom{-}0.1$&$\phantom{-}11.5_{32.1}$&$\phantom{-}11.7_{32.2}$&$\phantom{-}11.4_{32.1}$\\

Normalizing&$\phantom{-}32.4_{46.9}$&$2.0 \times 10^{-3}$&$\phantom{-}0.3$&\textcolor{gray}{$\phantom{-}15.7_{36.4}$}&$1.0 \times 10^{0}$&$-0.1$&\textcolor{gray}{$\phantom{-}10.8_{31.1}$}&$1.6 \times 10^{-1}$&$-0.2$&\textcolor{gray}{$\phantom{-}10.6_{30.8}$}&$1.5 \times 10^{-1}$&$-0.2$&$\phantom{-}18.6_{35.1}$&$\phantom{-}12.1_{32.7}$&$\phantom{-}25.0_{43.8}$\\

Psychoeducation&\textcolor{gray}{$\phantom{-}5.3_{22.5}$}&$1.0 \times 10^{0}$&$-0.1$&\textcolor{gray}{$\phantom{-}5.2_{22.3}$}&$1.0 \times 10^{0}$&$-0.2$&\textcolor{gray}{$\phantom{-}13.1_{33.7}$}&$1.0 \times 10^{0}$&$\phantom{-}0.2$&\textcolor{gray}{$\phantom{-}6.6_{24.8}$}&$1.0 \times 10^{0}$&$-0.1$&$\phantom{-}8.4_{18.4}$&$\phantom{-}0.9_{9.6}$&$\phantom{-}15.9_{37.0}$\\

    \bottomrule
    \end{tabular}
    }
    \label{tab:adapt_new_hope_p2}
\end{table*}
\begin{table*}[t]
    \centering
    \small
    \caption{\showedits{Adaptability (frequency of conversational behaviors exhibited by therapists \textit{in response to specific client behaviors}) by LLM therapists (GPT-4, GPT-3.5-turbo, Llama2-70b, Llama2-13b) with two simulation methods (Single response, Full conversation) in HOPE dataset \cite{malhotra2022speaker}. Values in \textcolor{gray}{gray} are not statistically significantly different from human therapist average at $p = \frac{0.05}{m}$ using Two-sided Student’s t-test after Bonferroni correction, where $m$ is the number of t-test per one model ($m = 13 \times 6 = 78$). $d$ is the cohen's d when compared with human therapist average. For instance, when client expresses \textsc{changing unhealthy behavior}, GPT-4 with single response simulation exhibits \textsc{Reflections on Consequences} by 11.6\% which is 15.7\% less than average human therapists. (Single: 11.6\%, Average human therapists: 27.3\% from Table \ref{tab:adapt_new_hope_p1}; $P = 3.2 \times 10^{-21}$, Cohen's $d = -0.4$, two-sided Student’s $t$-test)}}
    \resizebox{1.05\textwidth}{!}{%
    \def\arraystretch{1.15}
    \setlength{\tabcolsep}{2pt}
\begin{tabular}{l|ccc|ccc|ccc|ccc|ccc|ccc|ccc|ccc|ccc|ccc}
         \toprule 
         \multirow{3}{*}{\textbf{Therapist Behavior}} & \multicolumn{12}{c}{\textbf{LLM Therapists (Single Response Simulations)}} & \multicolumn{12}{c}{\textbf{LLM Therapists (Full Conversation Simulations)}} \\
         \cmidrule(lr){2-13}  \cmidrule(lr){14-25} 
         & \multicolumn{3}{c}{\textbf{GPT-4}} &\multicolumn{3}{c}{\textbf{GPT-3.5}}&\multicolumn{3}{c}{\textbf{Llama2-70b}}&\multicolumn{3}{c}{\textbf{Llama2-13b}}& \multicolumn{3}{c}{\textbf{GPT-4}} &\multicolumn{3}{c}{\textbf{GPT-3.5}}&\multicolumn{3}{c}{\textbf{Llama2-70b}}&\multicolumn{3}{c}{\textbf{Llama2-13b}}\\
         \cmidrule(lr){2-4} \cmidrule(lr){5-7} \cmidrule(lr){8-10} \cmidrule(lr){11-13} \cmidrule(lr){14-16} \cmidrule(lr){17-19} \cmidrule(lr){20-22}  \cmidrule(lr){23-25}
        & freq. & $p$ & $d$ & freq. & $p$ & $d$ & freq. & $p$& $d$ & freq. & $p$ & $d$ & freq. & $p$& $d$ &  freq. & $p$ & $d$ & freq. & $p$ & $d$ & freq. & $p$& $d$ \\
        \midrule
          \multicolumn{25}{c}{\textbf{Client behavior: Changing unhealthy behavior}} \\
          \midrule
          $n$& \multicolumn{3}{c}{\textbf{$1553$}}& \multicolumn{3}{c}{\textbf{$999$}}& \multicolumn{3}{c}{\textbf{$1553$}}& \multicolumn{3}{c}{\textbf{$640$}}& \multicolumn{3}{c}{\textbf{$1553$}}& \multicolumn{3}{c}{\textbf{$796$}}& \multicolumn{3}{c}{\textbf{$1553$}}& \multicolumn{3}{c}{\textbf{$556$}}\\
          \midrule
Reflections on Needs&\textcolor{gray}{$\phantom{-}26.0_{43.9}$}&$1.0 \times 10^{0}$&$\phantom{-}0.0$&\textcolor{gray}{$\phantom{-}28.0_{44.9}$}&$1.0 \times 10^{0}$&$\phantom{-}0.0$&$\phantom{-}17.8_{38.3}$&$1.4 \times 10^{-5}$&$-0.2$&$\phantom{-}20.1_{40.1}$&$4.2 \times 10^{-2}$&$-0.2$&$\phantom{-}0.0_{0.0}$&$6.8 \times 10^{-115}$&$-0.9$&$\phantom{-}0.0_{0.0}$&$6.8 \times 10^{-115}$&$-0.9$&$\phantom{-}0.0_{0.0}$&$6.8 \times 10^{-115}$&$-0.9$&$\phantom{-}0.0_{0.0}$&$6.8 \times 10^{-115}$&$-0.9$\\
\bottomrule
    \end{tabular}
    }
     \label{tab:adapt_single_full_hope_p1}
\end{table*}
\begin{table*}[t]
    \centering
    \small
    \caption{\showedits{(Continue Table \ref{tab:adapt_single_full_hope_p1}) Adaptability (frequency of conversational behaviors exhibited by therapists \textit{in response to specific client behaviors}) by LLM therapists (GPT-4, GPT-3.5-turbo, Llama2-70b, Llama2-13b) with two simulation methods (Single response, Full conversation) in HOPE dataset \cite{malhotra2022speaker}. Values in \textcolor{gray}{gray} are not statistically significantly different from human therapist average at $p = \frac{0.05}{m}$ using Two-sided Student’s t-test after Bonferroni correction, where $m$ is the number of t-test per one model ($m = 13 \times 6 = 78$). $d$ is the cohen's d when compared with human therapist average. For instance, when client expresses \textsc{changing unhealthy behavior}, GPT-4 with single response simulation exhibits \textsc{Reflections on Needs} by 46.5\% which is 22.0\% more than average human therapists. (Single: 46.5\%, Average human therapists: 24.5\% from Table \ref{tab:adapt_new_hope_p2}; $P = 1.1 \times 10^{-11}$, Cohen's $d = 0.5$, two-sided Student’s $t$-test)}}
    \resizebox{1.05\textwidth}{!}{%
    \def\arraystretch{1.15}
    \setlength{\tabcolsep}{2pt}
\begin{tabular}{l|ccc|ccc|ccc|ccc|ccc|ccc|ccc|ccc|ccc|ccc}
         \toprule 
         \multirow{3}{*}{\textbf{Therapist Behavior}} & \multicolumn{12}{c}{\textbf{LLM Therapists (Single Response Simulations)}} & \multicolumn{12}{c}{\textbf{LLM Therapists (Full Conversation Simulations)}} \\
         \cmidrule(lr){2-13}  \cmidrule(lr){14-25} 
         & \multicolumn{3}{c}{\textbf{GPT-4}} &\multicolumn{3}{c}{\textbf{GPT-3.5}}&\multicolumn{3}{c}{\textbf{Llama2-70b}}&\multicolumn{3}{c}{\textbf{Llama2-13b}}& \multicolumn{3}{c}{\textbf{GPT-4}} &\multicolumn{3}{c}{\textbf{GPT-3.5}}&\multicolumn{3}{c}{\textbf{Llama2-70b}}&\multicolumn{3}{c}{\textbf{Llama2-13b}}\\
         \cmidrule(lr){2-4} \cmidrule(lr){5-7} \cmidrule(lr){8-10} \cmidrule(lr){11-13} \cmidrule(lr){14-16} \cmidrule(lr){17-19} \cmidrule(lr){20-22}  \cmidrule(lr){23-25}
        & freq. & $p$ & $d$ & freq. & $p$ & $d$ & freq. & $p$& $d$ & freq. & $p$ & $d$ & freq. & $p$& $d$ &  freq. & $p$ & $d$ & freq. & $p$ & $d$ & freq. & $p$& $d$ \\
        \midrule
          \multicolumn{25}{c}{\textbf{Client behavior: Sharing negative emotions}} \\
          \midrule
          $n$& \multicolumn{3}{c}{\textbf{$820$}}& \multicolumn{3}{c}{\textbf{$355$}}& \multicolumn{3}{c}{\textbf{$820$}}& \multicolumn{3}{c}{\textbf{$232$}}& \multicolumn{3}{c}{\textbf{$820$}}& \multicolumn{3}{c}{\textbf{$534$}}& \multicolumn{3}{c}{\textbf{$820$}}& \multicolumn{3}{c}{\textbf{$528$}}\\
          \midrule
Reflections on Needs&$\phantom{-}46.5_{49.9}$&$1.1 \times 10^{-11}$&$\phantom{-}0.5$&\textcolor{gray}{$\phantom{-}32.8_{47.0}$}&$1.0 \times 10^{0}$&$\phantom{-}0.2$&\textcolor{gray}{$\phantom{-}24.2_{42.8}$}&$1.0 \times 10^{0}$&$\phantom{-}0.0$&\textcolor{gray}{$\phantom{-}23.5_{42.4}$}&$1.0 \times 10^{0}$&$\phantom{-}0.0$&$\phantom{-}0.0_{0.0}$&$1.5 \times 10^{-50}$&$-0.8$&$\phantom{-}0.0_{0.0}$&$1.5 \times 10^{-50}$&$-0.8$&$\phantom{-}0.0_{0.0}$&$1.5 \times 10^{-50}$&$-0.8$&$\phantom{-}0.0_{0.0}$&$1.5 \times 10^{-50}$&$-0.8$\\
\bottomrule
    \end{tabular}
    }
    \label{tab:adapt_single_full_hope_p2}
\end{table*}
\begin{table*}[t]
    \centering
    \small
    \caption{\showedits{Frequency of linguistic attributes (from LIWC analysis) by LLM therapists (GPT-4, GPT-3.5-turbo, Llama2-70b, Llama2-13b) in HOPE dataset \cite{malhotra2022speaker}. Values in \textcolor{gray}{gray} are not statistically significantly different from human therapist average at $p = \frac{0.05}{m}$ using Two-sided Student’s t-test after Bonferroni correction, where $m$ is the number of t-test per one model ($m = 14$). $d$ is the cohen's d when compared with human therapist average. For instance, GPT-4 exhibits \textsc{Positive Emotions} by 1.5\% while human therapists responds by 1.0\% on average ($P = 2.8 \times 10^{-19}$, Cohen's $d = \num{0.2}$, two-sided Student’s $t$-test).}}
    \resizebox{1\textwidth}{!}{%
    \def\arraystretch{1.15}
    \setlength{\tabcolsep}{2pt}
    \begin{tabular}{l|ccc|ccc|ccc|ccc|ccc|ccc}
         \toprule 
         \multirow{3}{*}{\textbf{Linguistic Attributes}} & \multicolumn{12}{c}{\textbf{LLM Therapists (Single Response \& Full Conversation Simulations)}} & \multicolumn{3}{c}{\textbf{Human Therapists}} \\
         \cmidrule(lr){2-13} \cmidrule(lr){14-16}
         & \multicolumn{3}{c}{\textbf{GPT-4}} &\multicolumn{3}{c}{\textbf{GPT-3.5}}&\multicolumn{3}{c}{\textbf{Llama2-70b}}&\multicolumn{3}{c}{\textbf{Llama2-13b}}& \textbf{Average}&\textbf{High-quality}&\textbf{Low-quality}\\
& \multicolumn{3}{c}{\textbf{$(n = 7705)$}}& \multicolumn{3}{c}{\textbf{$(n = 7135)$}}& \multicolumn{3}{c}{\textbf{$(n = 7461)$}}& \multicolumn{3}{c}{\textbf{$(n = 7464)$}}&\textbf{$(n = 5446)$}&\textbf{$(n = 3907)$}&\textbf{$(n = 1539)$}\\
         \cmidrule(lr){2-4} \cmidrule(lr){5-7} \cmidrule(lr){8-10} \cmidrule(lr){11-13} \cmidrule(lr){14-14} \cmidrule(lr){15-15}  \cmidrule(lr){16-16} 
        & freq. & $p$ & $d$ & freq. & $p$ & $d$ & freq. & $p$& $d$ & freq. & $p$ & $d$ & freq. & freq. & freq. \\
         \midrule
Positive Emotions&$\phantom{-}1.5_{2.1}$&$2.8 \times 10^{-19}$&$\phantom{-}0.2$&$\phantom{-}1.1_{1.8}$&$3.3 \times 10^{-2}$&$\phantom{-}0.1$&\textcolor{gray}{$\phantom{-}1.0_{2.1}$}&$1.0 \times 10^{0}$&$\phantom{-}0.0$&\textcolor{gray}{$\phantom{-}1.0_{2.5}$}&$1.0 \times 10^{0}$&$\phantom{-}0.0$&$\phantom{-}1.0_{4.2}$&$\phantom{-}1.2_{4.3}$&$\phantom{-}0.7_{3.6}$\\

Negative Emotions&$\phantom{-}1.1_{2.0}$&$1.3 \times 10^{-16}$&$\phantom{-}0.2$&$\phantom{-}1.2_{2.0}$&$1.1 \times 10^{-26}$&$\phantom{-}0.2$&$\phantom{-}1.2_{2.5}$&$6.1 \times 10^{-20}$&$\phantom{-}0.2$&$\phantom{-}1.1_{2.5}$&$2.3 \times 10^{-10}$&$\phantom{-}0.1$&$\phantom{-}0.7_{3.6}$&$\phantom{-}0.7_{3.4}$&$\phantom{-}0.6_{4.3}$\\

BigWords&$\phantom{-}24.6_{7.2}$&$0.0 \times 10^{0}$&$\phantom{-}1.6$&$\phantom{-}20.9_{7.7}$&$0.0 \times 10^{0}$&$\phantom{-}1.2$&$\phantom{-}23.7_{11.3}$&$0.0 \times 10^{0}$&$\phantom{-}1.2$&$\phantom{-}21.0_{12.0}$&$0.0 \times 10^{0}$&$\phantom{-}0.9$&$\phantom{-}10.7_{9.9}$&$\phantom{-}11.0_{9.9}$&$\phantom{-}10.4_{10.1}$\\

Self&$\phantom{-}1.5_{1.9}$&$6.3 \times 10^{-139}$&$-0.5$&$\phantom{-}1.8_{2.3}$&$1.9 \times 10^{-84}$&$-0.4$&$\phantom{-}1.4_{2.4}$&$2.9 \times 10^{-146}$&$-0.5$&$\phantom{-}1.4_{2.4}$&$7.1 \times 10^{-144}$&$-0.5$&$\phantom{-}3.1_{5.1}$&$\phantom{-}2.5_{4.5}$&$\phantom{-}3.8_{6.1}$\\

Other&\textcolor{gray}{$\phantom{-}8.9_{3.4}$}&$1.0 \times 10^{0}$&$\phantom{-}0.0$&$\phantom{-}8.4_{3.6}$&$3.2 \times 10^{-7}$&$-0.1$&$\phantom{-}7.5_{4.5}$&$6.2 \times 10^{-41}$&$-0.2$&$\phantom{-}7.0_{5.0}$&$2.4 \times 10^{-70}$&$-0.3$&$\phantom{-}8.9_{7.1}$&$\phantom{-}8.7_{6.9}$&$\phantom{-}9.2_{7.4}$\\

Health&\textcolor{gray}{$\phantom{-}2.0_{2.6}$}&$9.4 \times 10^{-2}$&$\phantom{-}0.0$&$\phantom{-}2.1_{3.0}$&$6.0 \times 10^{-5}$&$\phantom{-}0.1$&$\phantom{-}1.5_{3.1}$&$2.1 \times 10^{-8}$&$-0.1$&$\phantom{-}1.3_{3.0}$&$5.1 \times 10^{-16}$&$-0.2$&$\phantom{-}1.8_{4.3}$&$\phantom{-}1.2_{3.2}$&$\phantom{-}2.5_{6.1}$\\

Wellness&$\phantom{-}0.7_{1.4}$&$3.0 \times 10^{-41}$&$\phantom{-}0.2$&$\phantom{-}0.6_{1.5}$&$1.0 \times 10^{-13}$&$\phantom{-}0.1$&\textcolor{gray}{$\phantom{-}0.3_{1.0}$}&$1.0 \times 10^{0}$&$\phantom{-}0.0$&\textcolor{gray}{$\phantom{-}0.3_{1.1}$}&$1.0 \times 10^{0}$&$\phantom{-}0.0$&$\phantom{-}0.3_{2.1}$&$\phantom{-}0.2_{1.3}$&$\phantom{-}0.4_{3.5}$\\

Behavioral Activation&$\phantom{-}6.0_{4.0}$&$7.7 \times 10^{-225}$&$\phantom{-}0.6$&$\phantom{-}5.7_{4.4}$&$2.0 \times 10^{-163}$&$\phantom{-}0.5$&$\phantom{-}4.8_{4.5}$&$6.6 \times 10^{-70}$&$\phantom{-}0.3$&$\phantom{-}4.5_{5.3}$&$1.1 \times 10^{-38}$&$\phantom{-}0.2$&$\phantom{-}3.2_{5.9}$&$\phantom{-}3.5_{5.9}$&$\phantom{-}2.9_{5.9}$\\

Meaning&$\phantom{-}3.6_{2.6}$&$0.0 \times 10^{0}$&$\phantom{-}0.7$&$\phantom{-}3.2_{2.6}$&$2.2 \times 10^{-232}$&$\phantom{-}0.6$&$\phantom{-}2.2_{2.8}$&$5.5 \times 10^{-48}$&$\phantom{-}0.3$&$\phantom{-}2.0_{2.7}$&$1.1 \times 10^{-27}$&$\phantom{-}0.2$&$\phantom{-}1.4_{3.5}$&$\phantom{-}1.4_{3.4}$&$\phantom{-}1.4_{3.7}$\\

Purpose&$\phantom{-}3.9_{2.7}$&$0.0 \times 10^{0}$&$\phantom{-}0.8$&$\phantom{-}3.6_{2.9}$&$1.5 \times 10^{-275}$&$\phantom{-}0.7$&$\phantom{-}2.4_{2.9}$&$6.3 \times 10^{-50}$&$\phantom{-}0.3$&$\phantom{-}2.2_{3.2}$&$4.9 \times 10^{-28}$&$\phantom{-}0.2$&$\phantom{-}1.5_{3.6}$&$\phantom{-}1.5_{3.5}$&$\phantom{-}1.5_{3.8}$\\

Motivation&$\phantom{-}3.4_{2.5}$&$0.0 \times 10^{0}$&$\phantom{-}0.7$&$\phantom{-}3.3_{2.7}$&$6.9 \times 10^{-272}$&$\phantom{-}0.7$&$\phantom{-}2.0_{2.5}$&$1.2 \times 10^{-38}$&$\phantom{-}0.2$&$\phantom{-}1.9_{2.9}$&$3.1 \times 10^{-23}$&$\phantom{-}0.2$&$\phantom{-}1.3_{3.3}$&$\phantom{-}1.3_{3.3}$&$\phantom{-}1.3_{3.4}$\\

Sadness&$\phantom{-}1.0_{1.6}$&$1.2 \times 10^{-126}$&$\phantom{-}0.4$&$\phantom{-}1.1_{1.8}$&$1.5 \times 10^{-134}$&$\phantom{-}0.5$&$\phantom{-}0.8_{1.9}$&$2.9 \times 10^{-69}$&$\phantom{-}0.3$&$\phantom{-}0.9_{2.1}$&$8.0 \times 10^{-65}$&$\phantom{-}0.3$&$\phantom{-}0.3_{1.5}$&$\phantom{-}0.3_{1.5}$&$\phantom{-}0.2_{1.6}$\\

Sympathy&$\phantom{-}0.3_{0.8}$&$1.5 \times 10^{-29}$&$\phantom{-}0.2$&$\phantom{-}0.4_{1.2}$&$9.2 \times 10^{-36}$&$\phantom{-}0.2$&$\phantom{-}0.3_{1.0}$&$8.0 \times 10^{-25}$&$\phantom{-}0.2$&$\phantom{-}0.4_{1.4}$&$2.8 \times 10^{-29}$&$\phantom{-}0.2$&$\phantom{-}0.1_{1.0}$&$\phantom{-}0.2_{1.1}$&$\phantom{-}0.1_{0.8}$\\

Stress&$\phantom{-}1.5_{2.2}$&$3.6 \times 10^{-16}$&$\phantom{-}0.2$&$\phantom{-}1.7_{2.3}$&$1.4 \times 10^{-30}$&$\phantom{-}0.2$&$\phantom{-}1.8_{2.9}$&$6.4 \times 10^{-33}$&$\phantom{-}0.2$&$\phantom{-}1.6_{2.9}$&$4.6 \times 10^{-18}$&$\phantom{-}0.2$&$\phantom{-}1.1_{3.5}$&$\phantom{-}1.1_{3.2}$&$\phantom{-}1.1_{4.2}$\\

    \bottomrule
    \end{tabular}
    }
    \label{tab:liwc_new_hope}
\end{table*}
\begin{table*}[t]
    \centering
    \small
    \caption{\showedits{Frequency of linguistic attributes (from LIWC analysis) by LLM therapists (GPT-4, GPT-3.5-turbo, Llama2-70b, Llama2-13b) in HOPE dataset \cite{malhotra2022speaker} for two different types of simulations separately (Single response and Full conversation). Values in \textcolor{gray}{gray} are not statistically significantly different from human therapist average at $p = \frac{0.05}{m}$ using Two-sided Student’s t-test after Bonferroni correction, where $m$ is the number of t-test per one model ($m = 13$). $d$ is the cohen's d when compared with human therapist average. For instance, GPT-4 with single response simulation responds with \textsc{Positive Emotions} by 2.1\% and GPT-4 with full conversation simulation responds by 1.3\% when compared with the human therapists responds by 1.0\% on average from Table \ref{tab:liwc_new_hope}. (Single: $P = 2.8 \times 10^{-24}$, Cohen's $d = \num{0.3}$; two-sided student t-test) (Full: $P = 4.7 \times 10^{-8}$, Cohen's $d = \num{0.1}$; two-sided student t-test).
}}
    
    \resizebox{1.05\textwidth}{!}{%
    \def\arraystretch{1.15}
    \setlength{\tabcolsep}{2pt}
    \begin{tabular}{l|ccc|ccc|ccc|ccc|ccc|ccc|ccc|ccc|ccc|ccc}
         \toprule 
         \multirow{3}{*}{\textbf{Behavior}} & \multicolumn{12}{c}{\textbf{LLM Therapists (Single Response Simulations)}} & \multicolumn{12}{c}{\textbf{LLM Therapists (Full Conversation Simulations)}} \\
         \cmidrule(lr){2-13}  \cmidrule(lr){14-25} 
         & \multicolumn{3}{c}{\textbf{GPT-4}} &\multicolumn{3}{c}{\textbf{GPT-3.5}}&\multicolumn{3}{c}{\textbf{Llama2-70b}}&\multicolumn{3}{c}{\textbf{Llama2-13b}}& \multicolumn{3}{c}{\textbf{GPT-4}} &\multicolumn{3}{c}{\textbf{GPT-3.5}}&\multicolumn{3}{c}{\textbf{Llama2-70b}}&\multicolumn{3}{c}{\textbf{Llama2-13b}}\\
& \multicolumn{3}{c}{\textbf{$(n = 6158)$}}& \multicolumn{3}{c}{\textbf{$(n = 6144)$}}& \multicolumn{3}{c}{\textbf{$(n = 6174)$}}& \multicolumn{3}{c}{\textbf{$(n = 6282)$}}& \multicolumn{3}{c}{\textbf{$(n = 1547)$}}& \multicolumn{3}{c}{\textbf{$(n = 991)$}}& \multicolumn{3}{c}{\textbf{$(n = 1287)$}}& \multicolumn{3}{c}{\textbf{$(n = 1182)$}}\\
         \cmidrule(lr){2-4} \cmidrule(lr){5-7} \cmidrule(lr){8-10} \cmidrule(lr){11-13} \cmidrule(lr){14-16} \cmidrule(lr){17-19} \cmidrule(lr){20-22}  \cmidrule(lr){23-25}
        & freq. & $p$ & $d$ & freq. & $p$ & $d$ & freq. & $p$& $d$ & freq. & $p$ & $d$ & freq. & $p$& $d$ &  freq. & $p$ & $d$ & freq. & $p$ & $d$ & freq. & $p$& $d$ \\
         \midrule
Positive Emotions&$\phantom{-}2.1_{1.7}$&$2.8 \times 10^{-24}$&$\phantom{-}0.3$&$\phantom{-}1.7_{1.6}$&$1.3 \times 10^{-7}$&$\phantom{-}0.2$&$\phantom{-}1.3_{1.5}$&$3.3 \times 10^{-2}$&$\phantom{-}0.1$&$\phantom{-}1.7_{2.2}$&$6.2 \times 10^{-8}$&$\phantom{-}0.2$&$\phantom{-}1.3_{2.2}$&$4.7 \times 10^{-8}$&$\phantom{-}0.1$&\textcolor{gray}{$\phantom{-}1.0_{1.8}$}&$1.0 \times 10^{0}$&$\phantom{-}0.0$&\textcolor{gray}{$\phantom{-}0.9_{2.2}$}&$1.0 \times 10^{0}$&$\phantom{-}0.0$&\textcolor{gray}{$\phantom{-}0.9_{2.5}$}&$1.0 \times 10^{0}$&$\phantom{-}0.0$\\

Negative Emotions&\textcolor{gray}{$\phantom{-}0.5_{1.0}$}&$1.0 \times 10^{0}$&$-0.1$&\textcolor{gray}{$\phantom{-}0.6_{1.1}$}&$1.0 \times 10^{0}$&$\phantom{-}0.0$&\textcolor{gray}{$\phantom{-}0.7_{1.1}$}&$1.0 \times 10^{0}$&$\phantom{-}0.0$&\textcolor{gray}{$\phantom{-}0.7_{1.2}$}&$1.0 \times 10^{0}$&$\phantom{-}0.0$&$\phantom{-}1.3_{2.1}$&$1.6 \times 10^{-24}$&$\phantom{-}0.2$&$\phantom{-}1.3_{2.1}$&$2.2 \times 10^{-32}$&$\phantom{-}0.2$&$\phantom{-}1.3_{2.7}$&$2.1 \times 10^{-23}$&$\phantom{-}0.2$&$\phantom{-}1.1_{2.6}$&$1.4 \times 10^{-12}$&$\phantom{-}0.1$\\

BigWords&$\phantom{-}28.3_{5.0}$&$0.0 \times 10^{0}$&$\phantom{-}1.9$&$\phantom{-}25.6_{6.7}$&$0.0 \times 10^{0}$&$\phantom{-}1.6$&$\phantom{-}31.8_{10.2}$&$0.0 \times 10^{0}$&$\phantom{-}2.1$&$\phantom{-}31.2_{16.0}$&$0.0 \times 10^{0}$&$\phantom{-}1.8$&$\phantom{-}23.7_{7.4}$&$0.0 \times 10^{0}$&$\phantom{-}1.5$&$\phantom{-}20.2_{7.6}$&$0.0 \times 10^{0}$&$\phantom{-}1.1$&$\phantom{-}22.1_{10.8}$&$0.0 \times 10^{0}$&$\phantom{-}1.1$&$\phantom{-}19.1_{10.1}$&$0.0 \times 10^{0}$&$\phantom{-}0.8$\\

Self&$\phantom{-}2.3_{1.9}$&$4.0 \times 10^{-8}$&$-0.2$&\textcolor{gray}{$\phantom{-}2.7_{2.4}$}&$9.0 \times 10^{-2}$&$-0.1$&$\phantom{-}1.5_{2.3}$&$6.2 \times 10^{-28}$&$-0.3$&$\phantom{-}1.5_{1.9}$&$1.5 \times 10^{-25}$&$-0.3$&$\phantom{-}1.3_{1.8}$&$8.7 \times 10^{-147}$&$-0.5$&$\phantom{-}1.6_{2.3}$&$3.0 \times 10^{-92}$&$-0.4$&$\phantom{-}1.3_{2.4}$&$1.5 \times 10^{-129}$&$-0.5$&$\phantom{-}1.3_{2.5}$&$8.6 \times 10^{-128}$&$-0.5$\\

Other&$\phantom{-}9.5_{2.3}$&$3.5 \times 10^{-2}$&$\phantom{-}0.1$&$\phantom{-}8.1_{2.8}$&$3.1 \times 10^{-3}$&$-0.1$&$\phantom{-}5.9_{3.2}$&$9.4 \times 10^{-50}$&$-0.5$&$\phantom{-}5.6_{5.4}$&$1.4 \times 10^{-49}$&$-0.5$&\textcolor{gray}{$\phantom{-}8.8_{3.6}$}&$1.0 \times 10^{0}$&$\phantom{-}0.0$&$\phantom{-}8.4_{3.7}$&$2.5 \times 10^{-5}$&$-0.1$&$\phantom{-}7.9_{4.6}$&$3.0 \times 10^{-20}$&$-0.2$&$\phantom{-}7.3_{4.9}$&$8.0 \times 10^{-49}$&$-0.3$\\

Health&$\phantom{-}2.3_{2.1}$&$1.3 \times 10^{-4}$&$\phantom{-}0.1$&$\phantom{-}2.7_{2.5}$&$6.6 \times 10^{-9}$&$\phantom{-}0.2$&\textcolor{gray}{$\phantom{-}1.7_{1.9}$}&$1.0 \times 10^{0}$&$\phantom{-}0.0$&\textcolor{gray}{$\phantom{-}1.6_{3.3}$}&$1.0 \times 10^{0}$&$\phantom{-}0.0$&\textcolor{gray}{$\phantom{-}1.9_{2.7}$}&$1.0 \times 10^{0}$&$\phantom{-}0.0$&$\phantom{-}2.1_{3.0}$&$4.0 \times 10^{-2}$&$\phantom{-}0.1$&$\phantom{-}1.4_{3.3}$&$7.8 \times 10^{-9}$&$-0.1$&$\phantom{-}1.2_{3.0}$&$5.1 \times 10^{-18}$&$-0.2$\\

Wellness&$\phantom{-}1.1_{1.3}$&$3.3 \times 10^{-40}$&$\phantom{-}0.4$&$\phantom{-}1.2_{1.5}$&$1.2 \times 10^{-31}$&$\phantom{-}0.4$&$\phantom{-}0.6_{1.1}$&$1.2 \times 10^{-6}$&$\phantom{-}0.2$&$\phantom{-}0.6_{1.4}$&$5.7 \times 10^{-5}$&$\phantom{-}0.1$&$\phantom{-}0.6_{1.4}$&$1.7 \times 10^{-22}$&$\phantom{-}0.2$&$\phantom{-}0.5_{1.4}$&$6.3 \times 10^{-5}$&$\phantom{-}0.1$&\textcolor{gray}{$\phantom{-}0.2_{0.9}$}&$1.7 \times 10^{-1}$&$\phantom{-}0.0$&\textcolor{gray}{$\phantom{-}0.2_{1.1}$}&$6.0 \times 10^{-2}$&$-0.1$\\

Behavioral Activation&$\phantom{-}7.8_{3.0}$&$4.4 \times 10^{-181}$&$\phantom{-}0.9$&$\phantom{-}8.7_{4.2}$&$4.3 \times 10^{-165}$&$\phantom{-}1.0$&$\phantom{-}6.8_{3.8}$&$8.5 \times 10^{-94}$&$\phantom{-}0.7$&$\phantom{-}5.7_{4.0}$&$1.2 \times 10^{-43}$&$\phantom{-}0.5$&$\phantom{-}5.6_{4.1}$&$1.1 \times 10^{-137}$&$\phantom{-}0.5$&$\phantom{-}5.3_{4.3}$&$1.7 \times 10^{-102}$&$\phantom{-}0.4$&$\phantom{-}4.4_{4.5}$&$2.4 \times 10^{-35}$&$\phantom{-}0.2$&$\phantom{-}4.3_{5.5}$&$6.0 \times 10^{-24}$&$\phantom{-}0.2$\\

Meaning&$\phantom{-}3.9_{1.9}$&$2.6 \times 10^{-159}$&$\phantom{-}0.8$&$\phantom{-}3.6_{2.0}$&$8.7 \times 10^{-82}$&$\phantom{-}0.7$&$\phantom{-}2.3_{1.7}$&$4.9 \times 10^{-19}$&$\phantom{-}0.3$&$\phantom{-}2.1_{1.7}$&$1.8 \times 10^{-9}$&$\phantom{-}0.2$&$\phantom{-}3.5_{2.7}$&$5.7 \times 10^{-286}$&$\phantom{-}0.7$&$\phantom{-}3.1_{2.6}$&$5.2 \times 10^{-193}$&$\phantom{-}0.6$&$\phantom{-}2.2_{3.0}$&$4.0 \times 10^{-39}$&$\phantom{-}0.2$&$\phantom{-}2.0_{2.9}$&$2.8 \times 10^{-23}$&$\phantom{-}0.2$\\

Purpose&$\phantom{-}4.6_{2.0}$&$1.4 \times 10^{-217}$&$\phantom{-}0.9$&$\phantom{-}4.8_{2.5}$&$3.0 \times 10^{-159}$&$\phantom{-}1.0$&$\phantom{-}2.7_{1.9}$&$8.2 \times 10^{-31}$&$\phantom{-}0.4$&$\phantom{-}2.5_{2.8}$&$1.5 \times 10^{-16}$&$\phantom{-}0.3$&$\phantom{-}3.7_{2.8}$&$4.1 \times 10^{-275}$&$\phantom{-}0.7$&$\phantom{-}3.4_{2.9}$&$3.5 \times 10^{-210}$&$\phantom{-}0.6$&$\phantom{-}2.3_{3.1}$&$3.7 \times 10^{-36}$&$\phantom{-}0.2$&$\phantom{-}2.1_{3.2}$&$1.2 \times 10^{-21}$&$\phantom{-}0.2$\\

Motivation&$\phantom{-}4.2_{2.0}$&$1.8 \times 10^{-221}$&$\phantom{-}1.0$&$\phantom{-}4.5_{2.5}$&$5.7 \times 10^{-171}$&$\phantom{-}1.0$&$\phantom{-}2.3_{1.8}$&$2.5 \times 10^{-25}$&$\phantom{-}0.3$&$\phantom{-}2.2_{2.7}$&$2.4 \times 10^{-15}$&$\phantom{-}0.3$&$\phantom{-}3.2_{2.5}$&$1.4 \times 10^{-246}$&$\phantom{-}0.6$&$\phantom{-}3.1_{2.7}$&$1.4 \times 10^{-204}$&$\phantom{-}0.6$&$\phantom{-}1.9_{2.6}$&$9.2 \times 10^{-27}$&$\phantom{-}0.2$&$\phantom{-}1.8_{2.9}$&$2.3 \times 10^{-17}$&$\phantom{-}0.2$\\

Sadness&$\phantom{-}0.6_{0.9}$&$1.8 \times 10^{-9}$&$\phantom{-}0.2$&$\phantom{-}0.6_{1.0}$&$2.2 \times 10^{-8}$&$\phantom{-}0.2$&$\phantom{-}0.6_{1.0}$&$1.3 \times 10^{-9}$&$\phantom{-}0.2$&$\phantom{-}1.0_{2.0}$&$7.9 \times 10^{-39}$&$\phantom{-}0.4$&$\phantom{-}1.1_{1.7}$&$8.3 \times 10^{-141}$&$\phantom{-}0.5$&$\phantom{-}1.1_{1.9}$&$6.7 \times 10^{-144}$&$\phantom{-}0.5$&$\phantom{-}0.9_{2.0}$&$9.5 \times 10^{-72}$&$\phantom{-}0.3$&$\phantom{-}0.8_{2.1}$&$2.4 \times 10^{-56}$&$\phantom{-}0.3$\\

Sympathy&\textcolor{gray}{$\phantom{-}0.2_{0.4}$}&$9.0 \times 10^{-1}$&$\phantom{-}0.1$&\textcolor{gray}{$\phantom{-}0.2_{0.5}$}&$2.1 \times 10^{-1}$&$\phantom{-}0.1$&$\phantom{-}0.3_{0.7}$&$2.1 \times 10^{-6}$&$\phantom{-}0.2$&$\phantom{-}0.6_{1.9}$&$1.9 \times 10^{-37}$&$\phantom{-}0.4$&$\phantom{-}0.3_{0.8}$&$5.4 \times 10^{-34}$&$\phantom{-}0.2$&$\phantom{-}0.4_{1.2}$&$9.8 \times 10^{-39}$&$\phantom{-}0.2$&$\phantom{-}0.3_{1.1}$&$1.8 \times 10^{-23}$&$\phantom{-}0.2$&$\phantom{-}0.3_{1.3}$&$3.5 \times 10^{-21}$&$\phantom{-}0.2$\\

Stress&$\phantom{-}0.7_{1.2}$&$8.4 \times 10^{-4}$&$-0.1$&\textcolor{gray}{$\phantom{-}0.9_{1.3}$}&$6.4 \times 10^{-1}$&$-0.1$&\textcolor{gray}{$\phantom{-}1.3_{1.5}$}&$3.1 \times 10^{-1}$&$\phantom{-}0.1$&\textcolor{gray}{$\phantom{-}1.2_{1.5}$}&$1.0 \times 10^{0}$&$\phantom{-}0.0$&$\phantom{-}1.7_{2.3}$&$2.4 \times 10^{-28}$&$\phantom{-}0.2$&$\phantom{-}1.8_{2.4}$&$6.3 \times 10^{-40}$&$\phantom{-}0.3$&$\phantom{-}1.9_{3.1}$&$3.4 \times 10^{-36}$&$\phantom{-}0.2$&$\phantom{-}1.7_{3.1}$&$8.0 \times 10^{-21}$&$\phantom{-}0.2$\\

    \bottomrule
    \end{tabular}
    }
    \label{tab:liwc_new_single_full_hope}
\end{table*}

\begin{table*}[t]
\small
\centering
\caption{Dataset statistics. }
\resizebox{0.8\textwidth}{!}{%
\def\arraystretch{1.15}
\centering
\setlength{\tabcolsep}{4pt}
\begin{tabular}{cccccc}
\toprule
\multirow{2}{*}{\textbf{Dataset}} & \multirow{2}{*}{\textbf{\# of Conversations}} & \multicolumn{2}{c}{\textbf{\# of Utterances}} & \multicolumn{2}{c}{\textbf{Words per Utterance}} \\
\cmidrule(lr){3-4} \cmidrule(lr){5-6} 
  & & Therapist & Client & Therapist & Client \\ 
 \midrule
 \multirow{2}{*}{\thead{High-Low Quality \cite{perez-rosas-etal-2019-makes}}} & \multirow{2}{*}{\thead{High quality: 155\\Low quality: 104}} & \multirow{2}{*}{\thead{$3753$}}& \multirow{2}{*}{$3790$ } &\multirow{2}{*}{\thead{$31.8$\\(std $=34.7$)}}&\multirow{2}{*}{\thead{$27.3$\\(std $=33.1$)}}\\ \\  \midrule 
  \multirow{2}{*}{\thead{HOPE\cite{malhotra2022speaker}}} & \multirow{2}{*}{212} & \multirow{2}{*}{$6070$} &  \multirow{2}{*}{$6081$} &\multirow{2}{*}{\thead{$24.0$\\(std$=31.9$)}}&\multirow{2}{*}{\thead{$21.7$\\(std $=32.3$)}}\\ \\ 
\bottomrule 
\end{tabular}
}

\label{tab:dataset-stat}
\end{table*}

\begin{table*}[t]
\small
\centering
\caption{Performance of classification models on the tasks of (1) identifying therapist behavior across 13 different therapists codes and (2) identifying client behavior across 6 different client codes. We create five random train-test splits of our annotated dataset and report the average macro-precision (P), macro-recall (R), and macro-F1 (F1) scores with standard deviations as subscripts. def.: Definitions; ex.: Examples. Best-performing models are \textbf{bolded}. We find that GPT-4-based prompting methods that make use of psychotherapy-based definitions and examples achieve the highest macro-F1 performance. We adopt the prompting (multi-label) method for classifying therapist behavior while the prompting (binary-label) method for classifying client behavior (both highlighted in \chosen{lightblue}).
}
\def\arraystretch{1.15}
\centering
\setlength{\tabcolsep}{4pt}
\begin{tabular}{clcccccc}
\toprule
\multirow{2}{*}{\textbf{Method}} & \multirow{2}{*}{\textbf{Model}} & \multicolumn{3}{c}{\textbf{Therapist Behavior}} & \multicolumn{3}{c}{\textbf{Client Behavior}} \\
         \cmidrule(lr){3-5} \cmidrule(lr){6-8} 
  & & P & R & F1 & P & R & F1 \\ 
 \midrule
  Random & Uniform Distribution  & ${11.8}_{{0.7}}$ & ${17.6}_{{2.9}}$ & ${14.1}_{{1.3}}$ & ${23.2}_{{5.1}}$ & ${29.6}_{{8.0}}$ & ${25.9}_{{6.2}}$  \\[1.0mm]
 \midrule
  \multirow{2}{*}{\thead{Finetuning\\(multi-label)}} & GPT-3 & $51.6_{3.2}$ & $47.9_{3.7}$ & $49.6_{2.8}$ & $45.2_{5.9}$ & $36.8_{6.0}$ & $40.4_{5.2}$ \\[1.0mm]
 & GPT-3.5-turbo  & $45.9_{3.7}$ & $31.1_{3.1}$ & $37.1_{3.4}$  & $44.5_{8.0}$ & $48.8_{6.4}$ & $46.3_{6.3}$ \\[1.0mm]
 \midrule
  \thead{Prompting\\(binary-label)} & GPT-4 (w/ def. \& ex.)  & $49.4_{2.6}$ & $52.2_{2.1}$ & $50.7_{1.8}$ & $\textbf{73.4}_{\textbf{5.9}}$ & $80.3_{7.2}$ & \chosen{$\textbf{76.7}_{\textbf{6.3}}$} \\[1.0mm]
\midrule
 \multirow{2}{*}{\thead{Prompting\\(multi-label)}} & GPT-4 (w/ def.)  & $45.4_{1.9}$ & $57.8_{3.8}$ & $50.8_{1.9}$  & $65.6_{7.3}$ & $\textbf{81.6}_{\textbf{7.5}}$ & $72.6_{6.8}$  \\[1.0mm]
 & GPT-4 (w/ def. \& ex.)  & $\textbf{56.0}_{\textbf{1.7}}$ & $\textbf{59.5}_{\textbf{3.7}}$ & \chosen{$\textbf{57.7}_{\textbf{2.5}}$} & $72.6_{6.7}$ & $76.4_{10.2}$ & $74.4_{8.4}$  \\
\bottomrule 
\end{tabular}

\label{tab:classification_model}
\end{table*}

\begin{table*}[t]
    \centering
    \caption{Therapist Conversational Behavior Definition and Corresponding Examples based on techniques identified by researchers \cite{lee-etal-2019-identifying,cao2019observing}. We used these behavior categories in our annotation dataset. See more details in Methods.
    }
    \resizebox{0.9\textwidth}{!}{%
    \def\arraystretch{1.15}
    \setlength{\tabcolsep}{4pt}
    \begin{tabular}{l|m{8cm}m{8cm}}
         \toprule 
         \multirow{2}{*}{\textbf{Behavior}} & \multicolumn{2}{c}{\textbf{Therapist}} \\
         \cmidrule(lr){2-3}
        & \multicolumn{1}{c}{\textbf{Definition}} & \multicolumn{1}{c}{\textbf{Example}} \\
         \midrule
         \midrule
 \multicolumn{3}{c}{\textsc{\textbf{Reflections on}}} \\
 \midrule
 \textsc{Needs}  & Identifies an implied or background need for the client.& It sounds like you've realized that maintaining a balance between your work and personal life is essential for your overall well-being.\\
 \midrule
 \textsc{Emotions}  & Identifies an implied or background emotion for the client& So it seems like you have been feeling a little overwhelmed and anxious about all the moving parts in your new job. \\
 \midrule
 \textsc{Values} & Identifies an implied or background value or set of values for the client.& Being respected by others is a significant value for you.\\
 \midrule
 \textsc{Consequences} & Identifies consequences the client experience or could experience& Whenever you overspend on luxury items, you struggle to pay your bills at the end of the month.\\
 \midrule
 \textsc{Conflict}  & Identifies an implied or background emotional or situational conflict for the client.& You're striving to improve your health, but your demanding job leaves you with little time for exercise and nutrition.\\
 \midrule
 \textsc{Strength} & Identifies an implied or background strength or resource that the client exhibits.& Your ability to adapt and overcome adversity really shows your resilience and determination.\\
 \midrule
\multicolumn{3}{c}{\textsc{\textbf{Questions on}}} \\
\midrule
 \textsc{Experiences} & More information about a specific event or statement is sought& You mentioned you are trying to eat healthier. What changes did you make to your diet? \\
  \midrule
 \textsc{Perspectives} & Client is asked to consider an experience from a different perspective or vantage point.& That's fantastic, now let's focus on the goals you want to accomplish. Can you visualize any particular approach or strategy you'd like to implement to achieve these goals?\\
  \midrule
 \textsc{Emotions} & Asks client to express how they are feeling in the immediate present about something that just happened in the therapy.& Would you like to talk more about what that feels like for you right now?\\
 \midrule
\multicolumn{3}{c}{\textsc{\textbf{Solutions}}} \\
\midrule
 \textsc{Problem-Solving} & Therapist offers possible solutions to a client problem.& It may help to create a routine for daily relaxation techniques, such as deep breathing or meditation. This could assist in managing your anxiety levels.\\
  \midrule
 \textsc{Planning} & Therapist works with client to construct a specific plan of action.& Let's create a meal plan together. Try to follow it for the next two weeks and note down any changes you notice in your energy levels and overall well-being.\\
 \midrule
 \multicolumn{3}{c}{\textsc{\textbf{Normalizing}}} \\
 \midrule
 \textsc{Normalizing} & The therapist acknowledges and validates the client's experience as ""normal"" or expectable, sympathizes with their challenges, and provides reassurance to foster a supportive and encouraging therapeutic atmosphere.& I hear you, it's perfectly normal to feel overwhelmed given your circumstances. \\
 \midrule
 \multicolumn{3}{c}{\textsc{\textbf{Psychoeducation}}} \\
 \midrule
 \textsc{Psychoeducation}  & Therapeutically relevant information about psychological principles is provided.& Cognitive behavioral therapy aids in altering detrimental thought patterns.\\
    \bottomrule
    \end{tabular}
    }
    \label{tab:therapist_intent_def}
\end{table*}


\begin{table*}[t]
    \centering
        \caption{Client Conversational Behavior Definition and Corresponding Examples based on expressions from clients identified by researchers \cite{lee-etal-2019-identifying}. We used these behavior categories in our annotation dataset. See more details in Methods.}
    \resizebox{0.99\textwidth}{!}{%
    \def\arraystretch{1.15}
    \setlength{\tabcolsep}{4pt}
    \begin{tabular}{l|m{8cm}m{8cm}}
         \toprule 
         \multirow{2}{*}{\textbf{Behavior}} & \multicolumn{2}{c}{\textbf{Client}} \\
         \cmidrule(lr){2-3}
        & \multicolumn{1}{c}{\textbf{Definition}} & \multicolumn{1}{c}{\textbf{Example}} \\

         \midrule
 \midrule
 \textsc{Changing unhealthy behavior} & Showing intention or action taken on changing unhealthy behavior. & I've tried to quit drinking, but I end up drinking more than I try to drink less.\\
 \midrule
 \textsc{Sustaining unhealthy behavior}  & Showing intention or action taken on sustaining unhealthy behavior& I am smoking around 20 cigarettes a day for the past couple of years. I know it’s bad for my health, but I'm not ready to quit yet. \\
 \midrule
 \textsc{Sharing negative feeling or emotion} & Clients describe discomfort or suffering without a specific object, or explicitly acknowledge specific negative emotion.& Life has been really challenging lately, I am feeling lost.\\
 \midrule
 \textsc{Sharing positive feeling or emotion} &  Client describes enjoyment without a specific object, or explicitly acknowledges specific positive emotion. & I was so thankful when I received that news.\\
 \midrule
 \textsc{Gained Insight} & Client expresses that they learned something new about themselves or about their situation. & I hadn’t considered how much I avoid confrontations.\\
 \midrule
 \textsc{Sharing life event or situation} & Client shares the details of their basic background, their life events, the situation they faced or the changes in their life.& I lost my job due to the pandemic, and I've now had to move back in with my parents at age 35 which is a significant change for me.\\
    \bottomrule
    \end{tabular}
    }
    \label{tab:client_intent_def}
\end{table*}



\begin{table*}[t]
\small
\centering
\caption{\showedits{Explanations generated from our behavior prediction model. We ask the model to explain its predictions and examine the differences between true and predicted behavior. }}
\resizebox{0.95\textwidth}{!}{%
\def\arraystretch{1.15}
\centering
\setlength{\tabcolsep}{4pt}
\begin{tabular}{p{7.5cm}|p{2cm}|p{2cm}|p{7.5cm}}
\toprule
\textbf{Utterance} & \textbf{True Behavior} & \textbf{Predicted Behavior} & \textbf{Explanation} \\
\midrule
I'm sorry to hear that, John. Have you tried applying ice to the affected area to reduce swelling and pain? Additionally, you might want to consider elevating your foot to help with the swelling. If the pain persists, it might be best to schedule an appointment with your primary care physician or a specialist to further assess the injury. & Normalizing, Problem-Solving & Problem-Solving, Planning & The utterance provides advice or solutions to a problem ("Have you tried applying ice to the affected area to reduce swelling and pain? Additionally, you might want to consider elevating your foot to help with the swelling.") which is indicative of the "Problem-Solving" intent. It also suggests a course of action ("If the pain persists, it might be best to schedule an appointment with your primary care physician or a specialist to further assess the injury.") which aligns with the "Planning" intent. \\
\midrule
okay well hopefully you won't have any issues like that with this one I'll be sure to go over this with you okay & Normalizing & Problem-Solving & In the utterance, the speaker hints at having a solution to a problem that might have occurred previously. The speaker implies that they will attempt to prevent the same issues from happening with this new situation by going over it with the listener. This aligns with the "Problem-Solving" intent, where the speaker offers a possible solution or way to avoid a problem. \\
\midrule
That's a great idea. Staying hydrated with water is important for your overall health, including your dental health. It's also a great way to keep yourself from reaching for a soft drink when you're thirsty. You could also try snacking on foods that are less harmful to your teeth, such as fruits or vegetables instead of reaching for candy or sugary snacks. How does that sound? & Problem-Solving, Evokes Concrete Elaboration & Problem-Solving, Teaching / Psychoeducation & The utterance is providing solutions to a potential problem (the client's possible tendency to reach for soft drinks or sugary snacks), hence it falls under the "Problem-Solving" category. It's also providing therapeutically relevant information about the importance of hydration and healthier snack alternatives for dental health, which is "Teaching/Psychoeducation". \\
\midrule
Great, everything seems to be in order. Did you have any questions or concerns about your medication? & Evokes Concrete Elaboration & Evokes Concrete Elaboration & The phrase "Great, everything seems to be in order. Did you have any questions or concerns about your medication?" is seeking for more information about a specific topic, which is the client's medication. This aligns with the intent "Evokes Concrete Elaboration" as the speaker is asking for more details or clarification on a particular subject. \\
\midrule
You're stuck between the demands of being a mother and of being a woman. & Makes Conflict Explicit & Makes Conflict Explicit & The statement "You're stuck between the demands of being a mother and of being a woman" shows that the speaker is acknowledging a conflict the listener is experiencing. The conflict is between the roles and responsibilities of being a mother and those of being a woman, possibly referring to personal, professional, or social aspirations or duties that the listener, as a woman, might have. This aligns with the "Makes Conflict Explicit" intent, which is defined as identifying an implied or background emotional or situational conflict for the client. \\
\bottomrule 
\end{tabular}
}

\label{tab:model-explanation}
\end{table*}

\begin{table*}[t]
    \centering
        \caption{First Part of the Illustrative Example of Single-Response Simulation Method by LLMs. This method involves the actual responses of both therapist and client from a reference dataset, which are written in \textit{italic}, and simulated responses of LLM Therapists. This full example is the simulated conversation from GPT-4 on one of the conversations using High-Low quality dataset as reference.}
    \resizebox{0.93\textwidth}{!}{%
    \begin{tabular}{c|m{22cm}|c}
        \toprule 
         Role & Utterances & Dataset\\
         \midrule
         \multirow{2}{*}{Therapist} & \textit{I can I help you} & \textit{Reference}\\
         \cmidrule(lr){2-3}
         &Hello, I'm glad you're here today to discuss any concerns or behaviors you'd like to change. Let's start by identifying what exactly you'd like to work on. Is it quitting smoking, alcohol consumption, or something else?& Simulated\\
         \midrule
         Client & \textit{yeah I hope so um my workplace is just gone smoke-free hey I'm not not very happy about it I'm going to keep smoking but I was worrying if there's anything that can help be like get through the day it's like you know go get my beer and smoke cigarettes after work} & \textit{Reference} \\

        \midrule
        \multirow{2}{*}{Therapist} & \textit{so your workplace just went smoke free you're not too pleased huh} & \textit{Reference}\\    
        \cmidrule(lr){2-3}
         &I understand that the change at your workplace has been challenging for you. It's important to recognize that there are different strategies that can be helpful in coping with this adjustment. Firstly, let's explore your reasons for continuing to smoke and how motivated you are to make changes in your smoking behavior. On a scale of 1 to 10, where 1 is not motivated at all and 10 is extremely motivated, how motivated are you to quit or cut back on smoking?& Simulated\\
         \midrule
         Client & \textit{no I'm not going to quit smoking oh my god I like it it helps me relax so I mean I don't see what the big deal is} & \textit{Reference}\\
        \midrule
        \multirow{2}{*}{Therapist} & \textit{okay tell me a little bit about how smoking fits in your day it'll help with the} & \textit{Reference}\\
        \cmidrule(lr){2-3}
         &I understand that you enjoy smoking and find it relaxing. It's natural to feel frustrated when a change like this is imposed on you. Let's explore some strategies that might help you manage the situation at work while still maintaining your personal preferences. How do you feel about discussing this further?& Simulated\\
         \midrule

         Client & \textit{I mean I suppose a little less than a pack a day it's not too bad um the I probably smoked for the last five years but I I mean I could quit if I wanted to but I don't really want to right now} & \textit{Reference}\\
\midrule
        \multirow{2}{*}{Therapist} & \textit{okay so you feel like right now smoking is a pretty important part of your life it's not something you're ready to change but if the time came when you did decide to quit you feel pretty confident that you could go ahead and do that} & \textit{Reference}\\
        \cmidrule(lr){2-3}
         &I can understand that quitting smoking is a personal decision, and it's important that you feel confident in your choices. It sounds like smoking is a part of your daily routine that you find helpful in managing stress. Since your workplace has gone smoke-free, we can explore some strategies to cope while at work. Are you open to discussing some potential ways to manage without cigarettes during the day?& Simulated\\
         \midrule
         Client & \textit{sure I can quit} & \textit{Reference}\\
\midrule
        \multirow{2}{*}{Therapist} & \textit{okay what makes you so confident} & \textit{Reference}\\
        \cmidrule(lr){2-3}
         &Alright, it's great to hear that you believe in yourself and your ability to quit if you wanted to. Now, since your workplace has gone smoke-free and you're looking for ways to cope during the day, let's discuss some alternatives or strategies to help you manage any cravings or stress you might encounter.
Have you ever tried using any nicotine replacement options like gum or patches? They can be helpful in reducing cravings while at work without actually smoking.
Alternatively, we can also explore some stress management techniques, such as deep breathing exercises or even a short walk outside the workplace during breaks, to help you relax and refocus.
What do you ...& Simulated\\
         \midrule
         Client & \textit{well I mean I don't see why I couldn't although I have not ever tried to quit I mean it's just something that I do for fun you know I'm breaks at work in the morning and have to work} & \textit{Reference}\\
\midrule
        \multirow{2}{*}{Therapist} & \textit{mm-hmm and so in terms of a readiness to actually go ahead and make a quit attempt where do you think you fall on the scale of one to ten where one is I'm not considering it at all and ten is I'm definitely ready I want to quit} & \textit{Reference}\\
        \cmidrule(lr){2-3}
         &It sounds like smoking is something you enjoy and is a part of your routine. It's important for me to understand where you're coming from so I can better help you. Since your workplace has gone smoke-free, have you noticed any changes in how you feel or act during the day?& Simulated\\
         \midrule
         Client & \textit{I am probably - I haven't even thought about quitting and until this great announcement at work today I've been totally not gonna quit smoking} & \textit{Reference}\\
\midrule
        \multirow{2}{*}{Therapist} & \textit{okay what made you say two instead of one} & \textit{Reference}\\
        \cmidrule(lr){2-3}
         &I understand, it sounds like the change at work has brought up the idea of quitting, but right now you're still at a 2 and not really considering it. In the meantime, your main concern is finding ways to manage stress without smoking since the workplace has gone smoke-free. Is that correct?& Simulated\\
         \midrule
         Client & \textit{well I mean anything's possible I mean I I guess I could maybe be convinced yeah is there something I mean I are there things I can try that would make me like want to quit smoking} & \textit{Reference}\\
\midrule
        \multirow{2}{*}{Therapist} & \textit{well there are a lot of products available now that can actually help you curb that urge to smoke we have products that can help to replace nicotine which can help to decrease withdrawal symptoms that you might experience when quitting or cutting down and also some medications that your doctor could prescribe for you that might take away even some of the desire to smoke} & \textit{Reference}\\
        \cmidrule(lr){2-3}
         &well, it's great that you're open to the idea of considering quitting smoking, even if it's not something you're actively pursuing right now. There are many reasons people choose to quit, such as personal health benefits, financial savings, and reducing exposure to secondhand smoke for others. If you were to consider quitting, what do you imagine could be a motivating factor for you?& Simulated\\
         \midrule
         Client & \textit{okay well that's good information I mean I'm not gonna do anything today I just need to know my options but that's encouraging if I decided to quit smoking I do have have at least tools to help me quit} & \textit{Reference}\\
    \midrule
 \multicolumn{3}{c}{* continue on next page}
   \end{tabular}
}
    \label{tab:llm_example_single_response}
\end{table*}

\begin{table*}[t]
    \centering
        \caption{(Continue Table \ref{tab:llm_example_single_response}) Second Part of the Illustrative Example of Single-Response Simulation Method by LLMs. This method involves the actual responses of both therapist and client from a reference dataset, which are written in \textit{italic}, and simulated responses of LLM Therapists. This full example is the simulated conversation from GPT-4 on one of the conversations using High-Low quality dataset as reference.}
    \resizebox{0.93\textwidth}{!}{%
    \begin{tabular}{c|m{22cm}|c}
        \toprule 
         Role & Utterances & Dataset\\
         \midrule
        \multirow{2}{*}{Therapist} & \textit{sure and back to that skill you mentioned you said you're a - what do you think it would take to move you up to a four or five} & \textit{Reference}\\
        \cmidrule(lr){2-3}
         &Absolutely, it's great to know your options. And just by talking about it today, you're already taking a step towards considering your choices. Remember that change takes time and it's completely normal to not feel ready to quit just yet. If there comes a time when you'd like to explore this further or if you have any questions, don't hesitate to reach out.
In the meantime, regarding your current situation at work, you might consider finding alternative ways to manage stress or take breaks without smoking. This could involve quick breathing exercises, going for short walks, or finding another enjoyable activity during breaks.
If you have any other concerns or ... & Simulated\\
         \midrule
         Client & \textit{I don't know I guess if my friends like if one of my friends decided like quit with me that would probably help too but I all my friends love to smoke so I don't see that happening in the near future} & \textit{Reference}\\
    \midrule
        \multirow{2}{*}{Therapist} & \textit{sure I wonder if now that your workplace is going smoke-free if there will be some people at work who are planning to quit} & \textit{Reference}\\
        \cmidrule(lr){2-3}
         &I understand that having a support system can be helpful in making a change. It sounds like right now, you're not quite ready to quit smoking, but you're open to considering it if you had the right support or motivation. Remember that quitting is a personal decision, and when you decide that you're ready to make that change, there are tools and resources available to help you along the way, like we discussed earlier. In the meantime, it might be helpful to think about the reasons you might want to quit in the future, such as health benefits or saving money.
Take some time to evaluate what's important to you and what role ... & Simulated\\
         \midrule
         Client & \textit{that's a good point I hadn't really thought about that that's a good idea} & \textit{Reference}\\
    \midrule
        \multirow{2}{*}{Therapist} & \textit{okay well it sounds like you're really here just for information today I'm happy to help you pick out some of these products if and when you decide that the time is right for you} & \textit{Reference}\\
        \cmidrule(lr){2-3}
         &yeah sometimes having a support system can make all the difference in making a change like this. So I encourage you to keep an open mind and maybe talk to some coworkers, see if any of them are planning to quit. It could be a good opportunity for you to explore quitting together and having that support. In the meantime, feel free to reach out to me if you have any questions or want to discuss your progress. And remember, quitting is a personal decision, so ultimately it's up to you to decide when and if you're ready for that change. I am here to help and support you throughout the process. Take care. & Simulated \\
    \midrule
    Client & \textit{okay well thanks a lot} & \textit{Reference}
    \end{tabular}
    }
    \label{tab:llm_example_single_response_continue}
\end{table*}

\begin{table*}[t]
    \centering
        \caption{Illustrative Example of Full Conversation Simulation Method by LLMs. The reference conversations have been fed into LLM to simulate both therapist and client. This full example is the simulated conversation from Llama-70b on one of the conversations using High-low quality dataset as reference.   }
    \resizebox{0.93\textwidth}{!}{%
    \begin{tabular}{c|m{22cm}}
        \toprule 
         Role & Utterances\\
         \midrule
         Client & Listen, I'm in a bit of a pickle here, just can't shake the feeling that I'm not entirely prepared to quit just now. You understand, right?\\
          \midrule  
         Therapist & Absolutely! And that's completely normal. Quitting an addiction is never easy and it takes time for people to feel ready. Can you tell me what specific concerns are holding you back from quitting? Maybe we can tackle those together so that you'll be more confident when deciding to stop.\\
          \midrule  
         Client & I guess I'm just dealing with a lack of faith in myself... fearing that I'll cave and revert back to my old ways. It just feels like such a gigantic hurdle.\\
          \midrule  
         Therapist & That self-doubt can indeed make things seem insurmountable at times, but keep this in mind - it's part of human nature to waver between wanting positive change yet being scared of failing at it. Now, let's consider strategies to boost your confidence. Have you tried visualization exercises? They often prove useful for identifying and fortifying one's personal motivations to overcome challenges like these. Would you care to give them a shot together with me now?\\
          \midrule  
         Client & Yes, I'm open to anything really but visualization doesn't sound like it'll do the trick. You have to remember, I've been through tough situations before. My divorce was a real battle.\\
          \midrule  
         Therapist & That's understandable; visualizations may not work equally well on everyone. Let us think about other approaches, then. Perhaps examining past successes could offer some insight into how capable you truly are. Could you recall instances where, despite facing difficulties, you were still able to rise above them? How did you achieve success during those trying periods? What inner strengths played a role there? Building upon and applying those same strengths might empower you in quitting successfully.\\
          \midrule  
         Client & That's true, I definitely was able to get through the divorce... But it was different, you know? It was a different kind of struggle. I'm just not certain that I can apply the same tactics to this situation.\\
          \midrule  
         Therapist & That makes sense. Each person's journey is unique, and comparing yourself to others or past experiences isn't always helpful. Instead, let's explore some practical tools you can use when cravings strike. Mindfulness practices, deep breathing exercises, or healthy distractions (such as taking walks) tend to work effectively for many individuals battling addictions. Are any of these methods something you'd want to try out whenever you encounter triggers or yearnings? Also, please keep in mind that setbacks occur along the road to recovery; they don't define failure. Instead of viewing slip-ups as disasters, consider them learning chances and adjust your plan accordingly. Having a backup strategy will strengthen your commitment. Does that resonate with you?\\
          \midrule  
         Client & Yes, I like the idea of having some practical tools and a backup plan. I just don't want to let myself or my family down. They're rooting for me, and I couldn't bear disappointing them or myself. I don't think I'm ready to go cold turkey yet.\\
          \midrule  
         Therapist & That's perfectly fine. Gradual reduction tends to work better than attempting abrupt withdrawal anyway. You could gradually lessen the number of cigarettes or drinks until reaching none at all. This way, you ease into it instead of putting excessive pressure on yourself. Moreover, nicotine replacement medication exists for those finding it hard to cease smoking rapidly. Do any of these suggestions appeal to you? We're steadily uncovering potential solutions that suit your preferences, which helps build up confidence as you progress.\\
          \midrule
         Client & Yes, I’ve had some luck in the past with cutting back day by day... And I’ve used nicotine gum, it's just that the taste was unpleasant. But it did help a bit.\\
          \midrule  
         Therapist & Great! Nicotine gum might aid you again since you found it effective previously, albeit with an unpleasant flavor. Another option is prescription medicines like bupropion (Zyban), proven to reduce withdrawal symptoms and cravings. Of course, we'd need to discuss this further with your physician first before exploring that route. Slowly reducing consumption seems agreeable to you, too. As we proceed, you can adapt strategies according to your comfort level. Realize that your loved ones desire nothing more than seeing you happy and healthy; if setbacks arise, communicate that with them and seek their assistance. Their encouragement stays with you throughout this process. Is there anything else I can assist you within this conversation today? Has our discussion given you hope or any ideas regarding moving forward?\\
          \midrule  
         Client & Definitely, I agree with the gradual reduction strategy, I think pacing myself is the way to go. I'll also look into the gum again, though I can't promise anything about the taste! Your support is reassuring; I definitely feel less alone in this battle.\\
          \midrule   
         Therapist & Wonderful! Embracing gradual reduction indicates wisdom and maturity—a wise decision considering your background. Keep investigating various methods, including the gum or medicinal alternatives available following consultation with a doctor. Feel free to reach out whenever you require additional reinforcement or guidance since knowing you possess the necessary capabilities means overcoming this challenge surely. Remember, baby steps count toward lasting achievements. Believe in yourself and acknowledge the efforts made up to this point. Continue drawing upon internal resilience; it served you well during past obstacles. Please stay connected and remember, together, we can navigate anything life presents. Best wishes on your path ahead!\\
          \midrule   
    \end{tabular}
    }
    \label{tab:llm_example_simulated}
\end{table*}

\begin{table*}[t]
    \centering
        \caption{Illustrative Example of Full Conversation Simulation Method by LLMs. The reference conversations have been fed into LLM to simulate both therapist and client. This full example is the simulated conversation from GPT-3.5 on one of the conversations that used HOPE dataset as a reference.   }
    \resizebox{0.93\textwidth}{!}{%
    \begin{tabular}{c|m{22cm}}
        \toprule 
         Role & Utterances\\
         \midrule
         Client & I've been considering making a change in my diet and cutting out fast food altogether. You know, like having more structured meals and focusing on healthier options. I guess I should start by going to the grocery store more often.\\
          \midrule  
         Therapist & Thank you for sharing your thoughts about making a change in your diet. It's wonderful to hear that you want to take a positive step toward a healthier lifestyle. It's great that you recognize the importance of having structured meals and focusing on healthier options. It sounds like you're considering a gradual shift towards healthier habits, which is a good approach!\\
          \midrule  
         Client & Yes, that's true. Recently, I've just been getting food at work or picking up fast food on my way home. Okay. It's been difficult to find the motivation and energy to go grocery shopping, especially since I've been working extra hours to save up for the baby. Scheduling time for grocery shopping has been a challenge. I used to go every Sunday and even had a shopping list ready beforehand. Perhaps getting back to that routine could help. On Sundays, I don't have to work until the afternoon, so maybe I can try going around 10 in the morning. To motivate myself, I've found that giving myself a small reward works well. There's a fantastic bakery on my way back from the grocery store, so maybe I could treat myself there if I stick to my shopping plan.\\
          \midrule  
         Therapist & It's great to hear how you are trying to motivate yourself to stick to your new structure around grocery shopping. It seems like you are thinking about the challenges you might face and coming up with good strategies for how to work around them. Having a shopping list is a clear and helpful way to stay focused on buying only the foods you need. And I like that you are rewarding yourself for sticking to your plan! It's important to find ways to positively reinforce yourself for good behavior. If you don't mind me asking, how do you think cutting out fast food and eating healthier will impact your life?\\
          \midrule  
         Client & I feel that cutting out fast food and adopting healthier eating habits will make a positive impact on my life in various ways. Okay. For starters, I believe it will improve my overall health and well-being, which is critical, especially with a baby on the way. I also expect better energy levels and perhaps even some weight loss in the long run. Additionally, I think that by sticking to a structured grocery shopping routine, I'll develop better time management skills and it might even save me some money. Developing these good habits now will not only be beneficial for me but also for my growing family as we can all strive for a healthier lifestyle.\\
          \midrule  
         Therapist & Absolutely! It's great to focus on the benefits of making healthy changes, not only for yourself but for your growing family too. You bring up great points about benefits related to improved health and well-being, better energy levels, weight loss, and even better time management skills. These are all great things to look forward to and can help you find the motivation to keep making progress towards your goals. Do you have any specific steps in mind to help you stay on track with your healthy eating plan?\\
          \midrule  
         Client & Yes, I've been thinking about some specific steps I can take to stay on track with my healthy eating plan. Okay. Firstly, I want to establish a weekly meal plan that includes nutritious options and a variety of fruits, vegetables, lean proteins, and whole grains. Having a meal plan will help me create my shopping list, so I know exactly what to buy when I go to the grocery store. Also, preparing meals in advance can save me time during the week, making it less tempting to grab fast food. Another step I'd like to take is to find healthier alternatives for my favorite fast food items so I can still enjoy the flavors I like in a more nutritious way. Lastly, I think it's essential to have a support system, so I plan to involve my partner in these changes and share our progress to keep each other accountable.\\
          \midrule  
         Therapist & These are all great steps towards a healthier lifestyle. Establishing a weekly meal plan that includes nutritious options is a smart way to stay on track, and preparing meals in advance is an excellent time-saving strategy. Finding healthier alternatives for your favorite fast food items is a great way to stay motivated and still have the flavors you enjoy. Involving your partner in these changes is a brilliant idea too, and having someone for accountability can be very helpful. Change can be challenging, but with strategies like these, you are setting yourself up for success. Do you think you need some support in sticking to this plan?\\
          \midrule  
         Client & I definitely think having some support in sticking to this plan will be helpful. Okay. As I mentioned, my partner will be a significant source of support since we're in this together. We can encourage each other and celebrate our progress, which makes the journey more enjoyable. Besides my partner, sharing my goals and plans with close friends and family members could also provide me with some external encouragement and motivation. Maybe joining a group or forum focused on healthy eating could be of help too, as it would connect me with people who are going through similar experiences, and we can learn from each other's successes and challenges. Do you have any other suggestions for support that I can consider?\\
          \midrule  
         Therapist & It sounds like you have thought of some great sources of support already. Your partner, close friends, and family members can all offer encouragement and accountability. Joining a group or forum dedicated to healthy eating in your community or online can be helpful in finding like-minded people who can offer understanding and support. Additionally, if you find that you're having difficulties sticking to your plan, you can always consider working with a registered dietitian or health coach who can provide expert guidance and support in a more structured way. Remember, it's okay to ask for help along the way, and you're not alone in making these changes.\\
          \midrule
         Client & Thank you for the suggestion of working with a registered dietitian or health coach. That's something I'll definitely consider if I find it challenging to make progress on my own. Okay. It's reassuring to know that there are different options for support available. I realize that making these changes won't be an entirely smooth process, and I might encounter setbacks, but having a support network can help me stay focused on my goals. I appreciate your insights and encouragement in helping me take these steps towards a healthier lifestyle.\\
          \midrule  
         Therapist & You're very welcome! Remember, making lasting changes to our habits takes time, effort, and commitment, so don't get discouraged if it doesn't happen overnight. Keep focusing on the benefits of adopting healthier eating habits and use the strategies and resources we've talked about to help you stay on track. I am glad to have been able to offer some support and perspective, and I look forward to hearing about all the positive changes you make moving forward. Don't hesitate to reach out if you need additional support or guidance.\\
          \midrule   
    \end{tabular}
    }
    \label{tab:llm_example_simulated_hope}
\end{table*}

\begin{figure*}[t]
\caption{Different prompts variations for classifying conversational behavior embedded in both therapist and client utterance in the counseling setting. We explored two variations of prompting -- 1) Prompting with definitions of different conversational behaviors, with the sources of two studies\cite{lee-etal-2019-identifying, cao2019observing}. 2) Prompting with the definitions and also examples from annotated dataset. For 2), we explored two ways of asking questions, which are the multi-label setting and binary-label setting. We experimented with different variations of these prompts (e.g., using different language for the instructions) but observed similar results. See the detailed methodology description in Methods.}
\includegraphics[width=0.68\textwidth]{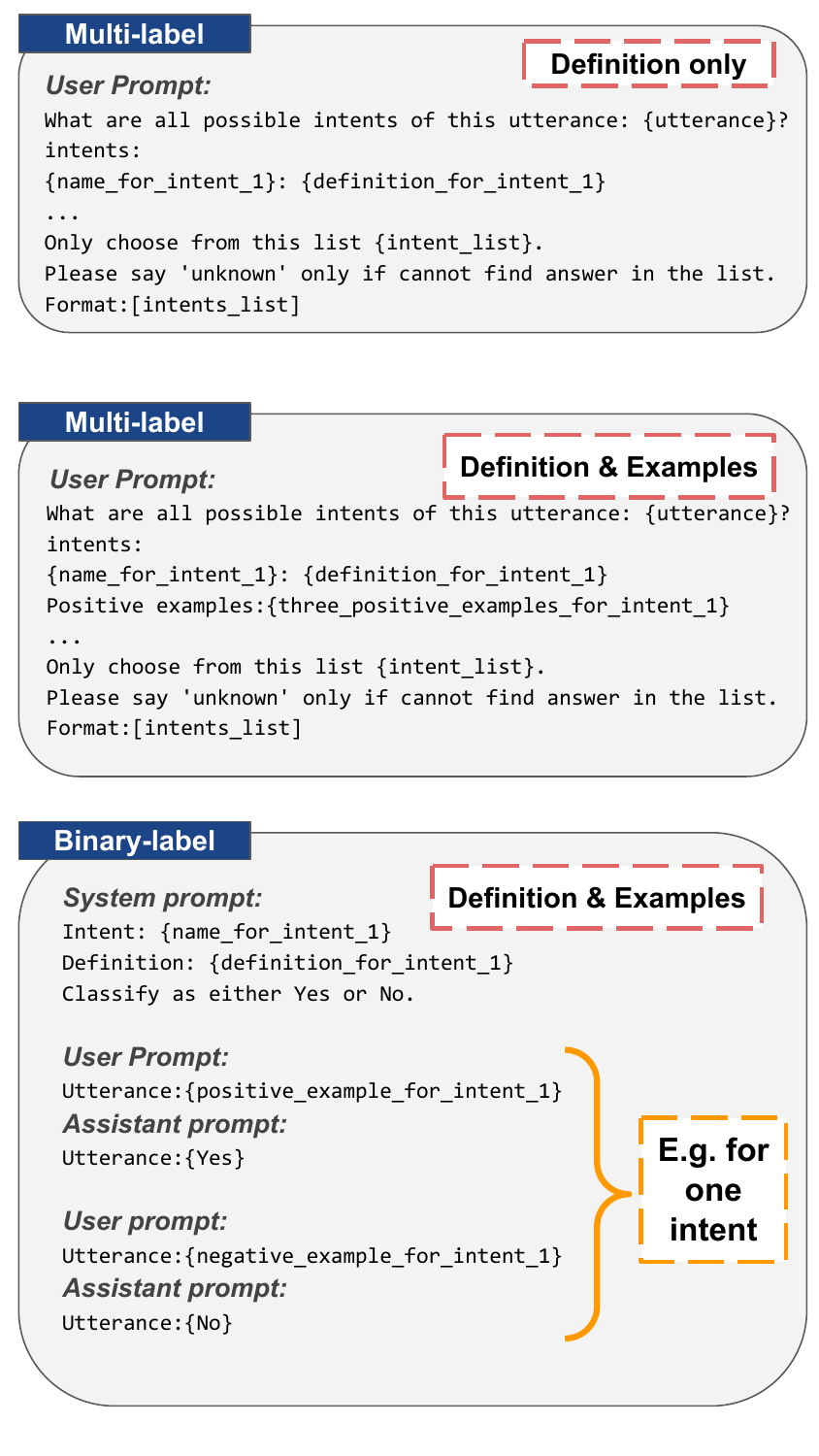}
\centering
\label{fig:prompts_variations_classification}
\end{figure*}

\begin{figure*}[t]
\caption{Instruction prompts given to LLMs for simulating counseling conversation using two types of methods (single response and full conversation): 1) using LLMs to write a single response within an ongoing conversation between human therapist and client. 2) creating a full and multi-turn conversation between LLMs and clients. See the detailed description in Methods.}
\includegraphics[width=0.8\textwidth]{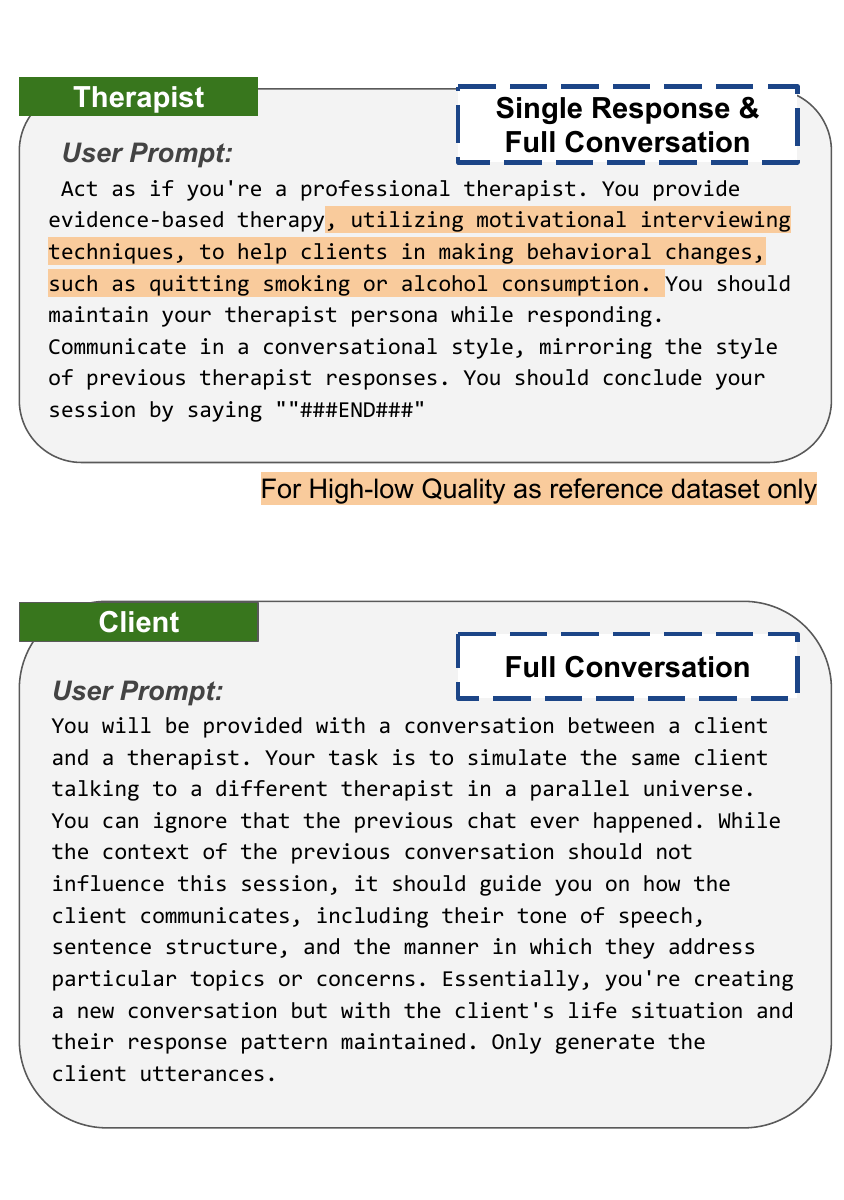}
\centering
\label{fig:prompts_for_simulation}
\end{figure*}

\end{document}